\documentclass{article}
\PassOptionsToPackage{numbers,sort&compress}{natbib}
\usepackage{publicpage}
\usepackage{times}
\usepackage{amsfonts}
\usepackage{flafter}
\usepackage{float}
\usepackage{placeins}
\usepackage{hyperref}
\usepackage[strings]{underscore}
\usepackage{url}
\usepackage{xspace}
\usepackage{amsmath,amsthm,amssymb}
\usepackage{booktabs}
\usepackage{longtable}
\usepackage{graphicx}
\usepackage{array}
\usepackage{caption}
\usepackage{xcolor}
\usepackage{enumitem}
\usepackage[capitalize]{cleveref}
\crefname{section}{Sec.}{Secs.}
\Crefname{section}{Section}{Sections}
\Crefname{table}{Table}{Tables}
\crefname{table}{Tab.}{Tabs.}
\newcommand{\etal}{\emph{et al.}}
\def\publicspread#1to #2\relax{\substack{\textnormal{\fontsize{9}{10}\selectfont [#1 to}\\\textnormal{\fontsize{9}{10}\selectfont #2]}}}
\newcommand{\estimate}[2]{\ensuremath{\textnormal{#1}_{\textnormal{\fontsize{9}{10}\selectfont [#2]}}}}
\newcommand{\Dino}{DINOv2\xspace}
\usepackage{multirow}
\usepackage{float}
\newcommand{\later}{\textsuperscript{\fontsize{7}{8}\selectfont\dag}}
\newcommand{\added}{\later}
\xspaceaddexceptions{\later \added}
\newcommand{\squared}{^{\text{\fontsize{7}{8}\selectfont$2$}}}
\newcommand{\inestimate}[2]{\estimate{#1}{#2}}

\usepackage{colortbl}
\definecolor{methodtint}{HTML}{FBEFE6}
\newcommand{\MarkDictFvuASkFvuTrMeanA}{\textbf{\SkFvuTrMeanA}\xspace}
\newcommand{\MarkDictFvuASkFvuFrMeanA}{\SkFvuFrMeanA\xspace}
\newcommand{\MarkDictLzeroASkLzeroTrMeanA}{\textbf{\SkLzeroTrMeanA}\xspace}
\newcommand{\MarkDictLzeroASkLzeroFrMeanA}{\SkLzeroFrMeanA\xspace}
\newcommand{\MarkCarryCountACertA}{\textbf{\CertA}\xspace}
\newcommand{\MarkCarryCountAFrozenCertA}{\FrozenCertA\xspace}
\newcommand{\MarkCarryShareAShareA}{\textbf{\ShareA}\xspace}
\newcommand{\MarkCarryShareAFrShareA}{\FrShareA\xspace}
\newcommand{\MarkAgreementASelfAgreeA}{\textbf{\SelfAgreeA}\xspace}
\newcommand{\MarkAgreementARuleAgreeA}{\RuleAgreeA\xspace}
\newcommand{\MarkAgreementASiteHalvesA}{\SiteHalvesA\xspace}
\newcommand{\MarkDisagreementADisagreeA}{\DisagreeA\xspace}
\newcommand{\MarkDisagreementASiteDisagreeA}{\textbf{\SiteDisagreeA}\xspace}
\newcommand{\MarkFreshAgreementAFreshSelfA}{\textbf{\FreshSelfA}\xspace}
\newcommand{\MarkFreshAgreementAFreshRuleA}{\FreshRuleA\xspace}
\newcommand{\MarkFreshErrorsAFreshCosErrA}{\FreshCosErrA\xspace}
\newcommand{\MarkFreshErrorsAFreshSiteErrA}{\textbf{\FreshSiteErrA}\xspace}
\newcommand{\MarkFreshErrorsAFittedCountA}{\textbf{\FittedCountA}\xspace}
\newcommand{\MarkWeakAllACertA}{\CertA\xspace}
\newcommand{\MarkWeakAllAClassGroupA}{\ClassGroupA\xspace}
\newcommand{\MarkWeakAllAClassUpperA}{\textbf{\ClassUpperA}\xspace}
\newcommand{\MarkWeakOnlyAWeakMatchA}{\WeakMatchA\xspace}
\newcommand{\MarkWeakOnlyAWeakGroupA}{\WeakGroupA\xspace}
\newcommand{\MarkWeakOnlyAWeakUpperA}{\textbf{\WeakUpperA}\xspace}
\newcommand{\MarkVarianceAOneSvAOneCarried}{\textbf{\SvAOneCarried}\xspace}
\newcommand{\MarkVarianceAOneSvAOneCosine}{\SvAOneCosine\xspace}
\newcommand{\MarkFreshVarianceAOneFreshVarianceAOnePassing}{\textbf{\FreshVarianceAOnePassing}\xspace}
\newcommand{\MarkFreshVarianceAOneFreshVarianceAOneCosine}{\FreshVarianceAOneCosine\xspace}
\newcommand{\MarkVarianceATwoSvATwoCarried}{\textbf{\SvATwoCarried}\xspace}
\newcommand{\MarkVarianceATwoSvATwoCosine}{\textbf{\SvATwoCosine}\xspace}
\newcommand{\MarkFreshVarianceATwoFreshVarianceATwoPassing}{\textbf{\FreshVarianceATwoPassing}\xspace}
\newcommand{\MarkFreshVarianceATwoFreshVarianceATwoCosine}{\textbf{\FreshVarianceATwoCosine}\xspace}
\newcommand{\MarkVarianceAThreeSvAThreeCarried}{\textbf{\SvAThreeCarried}\xspace}
\newcommand{\MarkVarianceAThreeSvAThreeCosine}{\SvAThreeCosine\xspace}
\newcommand{\MarkFreshVarianceAThreeFreshVarianceAThreePassing}{\textbf{\FreshVarianceAThreePassing}\xspace}
\newcommand{\MarkFreshVarianceAThreeFreshVarianceAThreeCosine}{\FreshVarianceAThreeCosine\xspace}
\newcommand{\MarkVarianceAFourSvAFourCarried}{\textbf{\SvAFourCarried}\xspace}
\newcommand{\MarkVarianceAFourSvAFourCosine}{\SvAFourCosine\xspace}
\newcommand{\MarkFreshVarianceAFourFreshVarianceAFourPassing}{\textbf{\FreshVarianceAFourPassing}\xspace}
\newcommand{\MarkFreshVarianceAFourFreshVarianceAFourCosine}{\FreshVarianceAFourCosine\xspace}
\newcommand{\MarkVarianceAFiveSvAFiveCarried}{\textbf{\SvAFiveCarried}\xspace}
\newcommand{\MarkVarianceAFiveSvAFiveCosine}{\SvAFiveCosine\xspace}
\newcommand{\MarkFreshVarianceAFiveFreshVarianceAFivePassing}{\textbf{\FreshVarianceAFivePassing}\xspace}
\newcommand{\MarkFreshVarianceAFiveFreshVarianceAFiveCosine}{\FreshVarianceAFiveCosine\xspace}
\newcommand{\MarkDictFvuBSkFvuTrMeanB}{\textbf{\SkFvuTrMeanB}\xspace}
\newcommand{\MarkDictFvuBSkFvuFrMeanB}{\SkFvuFrMeanB\xspace}
\newcommand{\MarkDictFvuBSkFvuSuMeanB}{\SkFvuSuMeanB\xspace}
\newcommand{\MarkDictLzeroBSkLzeroTrMeanB}{\textbf{\SkLzeroTrMeanB}\xspace}
\newcommand{\MarkDictLzeroBSkLzeroFrMeanB}{\SkLzeroFrMeanB\xspace}
\newcommand{\MarkDictLzeroBSkLzeroSuMeanB}{\SkLzeroSuMeanB\xspace}
\newcommand{\MarkCarryCountBCertB}{\textbf{\CertB}\xspace}
\newcommand{\MarkCarryCountBFrozenCertB}{\FrozenCertB\xspace}
\newcommand{\MarkCarryCountBSurrogateCertB}{\SurrogateCertB\xspace}
\newcommand{\MarkCarryShareBShareB}{\textbf{\ShareB}\xspace}
\newcommand{\MarkCarryShareBFrShareB}{\FrShareB\xspace}
\newcommand{\MarkCarryShareBSuShareB}{\SuShareB\xspace}
\newcommand{\MarkAgreementBSelfAgreeB}{\SelfAgreeB\xspace}
\newcommand{\MarkAgreementBRuleAgreeB}{\RuleAgreeB\xspace}
\newcommand{\MarkAgreementBSiteHalvesB}{\textbf{\SiteHalvesB}\xspace}
\newcommand{\MarkDisagreementBDisagreeB}{\DisagreeB\xspace}
\newcommand{\MarkDisagreementBSiteDisagreeB}{\textbf{\SiteDisagreeB}\xspace}
\newcommand{\MarkFreshAgreementBFreshSelfB}{\textbf{\FreshSelfB}\xspace}
\newcommand{\MarkFreshAgreementBFreshRuleB}{\FreshRuleB\xspace}
\newcommand{\MarkFreshErrorsBFreshCosErrB}{\FreshCosErrB\xspace}
\newcommand{\MarkFreshErrorsBFreshSiteErrB}{\textbf{\FreshSiteErrB}\xspace}
\newcommand{\MarkFreshErrorsBFittedCountB}{\textbf{\FittedCountB}\xspace}
\newcommand{\MarkWeakAllBCertB}{\CertB\xspace}
\newcommand{\MarkWeakAllBClassGroupB}{\ClassGroupB\xspace}
\newcommand{\MarkWeakAllBClassUpperB}{\textbf{\ClassUpperB}\xspace}
\newcommand{\MarkWeakOnlyBWeakMatchB}{\WeakMatchB\xspace}
\newcommand{\MarkWeakOnlyBWeakGroupB}{\WeakGroupB\xspace}
\newcommand{\MarkWeakOnlyBWeakUpperB}{\textbf{\WeakUpperB}\xspace}
\newcommand{\MarkVarianceBOneSvBOneCarried}{\SvBOneCarried\xspace}
\newcommand{\MarkVarianceBOneSvBOneCosine}{\textbf{\SvBOneCosine}\xspace}
\newcommand{\MarkFreshVarianceBOneFreshVarianceBOnePassing}{\FreshVarianceBOnePassing\xspace}
\newcommand{\MarkFreshVarianceBOneFreshVarianceBOneCosine}{\textbf{\FreshVarianceBOneCosine}\xspace}
\newcommand{\MarkVarianceBTwoSvBTwoCarried}{\textbf{\SvBTwoCarried}\xspace}
\newcommand{\MarkVarianceBTwoSvBTwoCosine}{\textbf{\SvBTwoCosine}\xspace}
\newcommand{\MarkFreshVarianceBTwoFreshVarianceBTwoPassing}{\textbf{\FreshVarianceBTwoPassing}\xspace}
\newcommand{\MarkFreshVarianceBTwoFreshVarianceBTwoCosine}{\textbf{\FreshVarianceBTwoCosine}\xspace}
\newcommand{\MarkVarianceBThreeSvBThreeCarried}{\textbf{\SvBThreeCarried}\xspace}
\newcommand{\MarkVarianceBThreeSvBThreeCosine}{\SvBThreeCosine\xspace}
\newcommand{\MarkFreshVarianceBThreeFreshVarianceBThreePassing}{\textbf{\FreshVarianceBThreePassing}\xspace}
\newcommand{\MarkFreshVarianceBThreeFreshVarianceBThreeCosine}{\FreshVarianceBThreeCosine\xspace}
\newcommand{\MarkVarianceBFourSvBFourCarried}{\textbf{\SvBFourCarried}\xspace}
\newcommand{\MarkVarianceBFourSvBFourCosine}{\SvBFourCosine\xspace}
\newcommand{\MarkFreshVarianceBFourFreshVarianceBFourPassing}{\textbf{\FreshVarianceBFourPassing}\xspace}
\newcommand{\MarkFreshVarianceBFourFreshVarianceBFourCosine}{\FreshVarianceBFourCosine\xspace}
\newcommand{\MarkVarianceBFiveSvBFiveCarried}{\textbf{\SvBFiveCarried}\xspace}
\newcommand{\MarkVarianceBFiveSvBFiveCosine}{\SvBFiveCosine\xspace}
\newcommand{\MarkFreshVarianceBFiveFreshVarianceBFivePassing}{\textbf{\FreshVarianceBFivePassing}\xspace}
\newcommand{\MarkFreshVarianceBFiveFreshVarianceBFiveCosine}{\FreshVarianceBFiveCosine\xspace}
\newcommand{\MarkDictFvuCSkFvuTrMeanC}{\textbf{\SkFvuTrMeanC}\xspace}
\newcommand{\MarkDictFvuCSkFvuFrMeanC}{\SkFvuFrMeanC\xspace}
\newcommand{\MarkDictFvuCSkFvuSuMeanC}{\SkFvuSuMeanC\xspace}
\newcommand{\MarkDictLzeroCSkLzeroTrMeanC}{\textbf{\SkLzeroTrMeanC}\xspace}
\newcommand{\MarkDictLzeroCSkLzeroFrMeanC}{\SkLzeroFrMeanC\xspace}
\newcommand{\MarkDictLzeroCSkLzeroSuMeanC}{\SkLzeroSuMeanC\xspace}
\newcommand{\MarkCarryCountCCertC}{\textbf{\CertC}\xspace}
\newcommand{\MarkCarryCountCFrozenCertC}{\FrozenCertC\xspace}
\newcommand{\MarkCarryCountCSurrogateCertC}{\SurrogateCertC\xspace}
\newcommand{\MarkCarryShareCShareC}{\textbf{\ShareC}\xspace}
\newcommand{\MarkCarryShareCFrShareC}{\FrShareC\xspace}
\newcommand{\MarkCarryShareCSuShareC}{\SuShareC\xspace}
\newcommand{\MarkAgreementCSelfAgreeC}{\SelfAgreeC\xspace}
\newcommand{\MarkAgreementCRuleAgreeC}{\RuleAgreeC\xspace}
\newcommand{\MarkAgreementCSiteHalvesC}{\textbf{\SiteHalvesC}\xspace}
\newcommand{\MarkDisagreementCDisagreeC}{\DisagreeC\xspace}
\newcommand{\MarkDisagreementCSiteDisagreeC}{\textbf{\SiteDisagreeC}\xspace}
\newcommand{\MarkFreshAgreementCFreshSelfC}{\textbf{\FreshSelfC}\xspace}
\newcommand{\MarkFreshAgreementCFreshRuleC}{\FreshRuleC\xspace}
\newcommand{\MarkFreshErrorsCFreshCosErrC}{\FreshCosErrC\xspace}
\newcommand{\MarkFreshErrorsCFreshSiteErrC}{\textbf{\FreshSiteErrC}\xspace}
\newcommand{\MarkFreshErrorsCFittedCountC}{\FittedCountC\xspace}
\newcommand{\MarkWeakAllCCertC}{\CertC\xspace}
\newcommand{\MarkWeakAllCClassGroupC}{\ClassGroupC\xspace}
\newcommand{\MarkWeakAllCClassUpperC}{\textbf{\ClassUpperC}\xspace}
\newcommand{\MarkWeakOnlyCWeakMatchC}{\WeakMatchC\xspace}
\newcommand{\MarkWeakOnlyCWeakGroupC}{\WeakGroupC\xspace}
\newcommand{\MarkWeakOnlyCWeakUpperC}{\textbf{\WeakUpperC}\xspace}
\newcommand{\MarkVarianceCOneSvCOneCarried}{\textbf{\SvCOneCarried}\xspace}
\newcommand{\MarkVarianceCOneSvCOneCosine}{\SvCOneCosine\xspace}
\newcommand{\MarkFreshVarianceCOneFreshVarianceCOnePassing}{\FreshVarianceCOnePassing\xspace}
\newcommand{\MarkFreshVarianceCOneFreshVarianceCOneCosine}{\textbf{\FreshVarianceCOneCosine}\xspace}
\newcommand{\MarkVarianceCTwoSvCTwoCarried}{\textbf{\SvCTwoCarried}\xspace}
\newcommand{\MarkVarianceCTwoSvCTwoCosine}{\SvCTwoCosine\xspace}
\newcommand{\MarkFreshVarianceCTwoFreshVarianceCTwoPassing}{\textbf{\FreshVarianceCTwoPassing}\xspace}
\newcommand{\MarkFreshVarianceCTwoFreshVarianceCTwoCosine}{\FreshVarianceCTwoCosine\xspace}
\newcommand{\MarkVarianceCThreeSvCThreeCarried}{\SvCThreeCarried\xspace}
\newcommand{\MarkVarianceCThreeSvCThreeCosine}{\textbf{\SvCThreeCosine}\xspace}
\newcommand{\MarkFreshVarianceCThreeFreshVarianceCThreePassing}{\FreshVarianceCThreePassing\xspace}
\newcommand{\MarkFreshVarianceCThreeFreshVarianceCThreeCosine}{\textbf{\FreshVarianceCThreeCosine}\xspace}
\newcommand{\MarkVarianceCFourSvCFourCarried}{\textbf{\SvCFourCarried}\xspace}
\newcommand{\MarkVarianceCFourSvCFourCosine}{\SvCFourCosine\xspace}
\newcommand{\MarkFreshVarianceCFourFreshVarianceCFourPassing}{\textbf{\FreshVarianceCFourPassing}\xspace}
\newcommand{\MarkFreshVarianceCFourFreshVarianceCFourCosine}{\FreshVarianceCFourCosine\xspace}
\newcommand{\MarkVarianceCFiveSvCFiveCarried}{\textbf{\SvCFiveCarried}\xspace}
\newcommand{\MarkVarianceCFiveSvCFiveCosine}{\SvCFiveCosine\xspace}
\newcommand{\MarkFreshVarianceCFiveFreshVarianceCFivePassing}{\textbf{\FreshVarianceCFivePassing}\xspace}
\newcommand{\MarkFreshVarianceCFiveFreshVarianceCFiveCosine}{\FreshVarianceCFiveCosine\xspace}
\newcommand{\MarkCarryCountDevDevCert}{\textbf{\DevCert}\xspace}
\newcommand{\MarkCarryCountDevDevFrozenCert}{\DevFrozenCert\xspace}
\newcommand{\MarkCarryCountDevDevSurrogateCert}{\DevSurrogateCert\xspace}
\newcommand{\MarkCarryShareDevTabDevTr}{\textbf{\TabDevTr}\xspace}
\newcommand{\MarkCarryShareDevTabDevFr}{\TabDevFr\xspace}
\newcommand{\MarkCarryShareDevTabDevSu}{\TabDevSu\xspace}
\newcommand{\MarkWeakAllDevDevCert}{\DevCert\xspace}
\newcommand{\MarkWeakAllDevDevClassGroup}{\DevClassGroup\xspace}
\newcommand{\MarkWeakAllDevDevClassUpper}{\textbf{\DevClassUpper}\xspace}
\newcommand{\MarkWeakOnlyDevDevWeakMatch}{\DevWeakMatch\xspace}
\newcommand{\MarkWeakOnlyDevDevWeakGroup}{\DevWeakGroup\xspace}
\newcommand{\MarkWeakOnlyDevDevWeakUpper}{\textbf{\DevWeakUpper}\xspace}
\newcommand{\MarkDevelopmentTrSdMatchTr}{\SdMatchTr\xspace}
\newcommand{\MarkDevelopmentTrSdGroupOneTr}{\SdGroupOneTr\xspace}
\newcommand{\MarkDevelopmentTrSdGroupTwoTr}{\SdGroupTwoTr\xspace}
\newcommand{\MarkDevelopmentTrSdGroupFourTr}{\SdGroupFourTr\xspace}
\newcommand{\MarkDevelopmentTrSdGroupEightTr}{\SdGroupEightTr\xspace}
\newcommand{\MarkDevelopmentTrSdUpperTr}{\textbf{\SdUpperTr}\xspace}
\newcommand{\MarkDevelopmentFrSdMatchFr}{\textbf{\SdMatchFr}\xspace}
\newcommand{\MarkDevelopmentFrSdGroupOneFr}{\textbf{\SdGroupOneFr}\xspace}
\newcommand{\MarkDevelopmentFrSdGroupTwoFr}{\textbf{\SdGroupTwoFr}\xspace}
\newcommand{\MarkDevelopmentFrSdGroupFourFr}{\textbf{\SdGroupFourFr}\xspace}
\newcommand{\MarkDevelopmentFrSdGroupEightFr}{\textbf{\SdGroupEightFr}\xspace}
\newcommand{\MarkDevelopmentFrSdUpperFr}{\textbf{\SdUpperFr}\xspace}
\newcommand{\MarkDevelopmentSuSdMatchSu}{\textbf{\SdMatchSu}\xspace}
\newcommand{\MarkDevelopmentSuSdGroupOneSu}{\textbf{\SdGroupOneSu}\xspace}
\newcommand{\MarkDevelopmentSuSdGroupTwoSu}{\textbf{\SdGroupTwoSu}\xspace}
\newcommand{\MarkDevelopmentSuSdGroupFourSu}{\textbf{\SdGroupFourSu}\xspace}
\newcommand{\MarkDevelopmentSuSdGroupEightSu}{\textbf{\SdGroupEightSu}\xspace}
\newcommand{\MarkDevelopmentSuSdUpperSu}{\textbf{\SdUpperSu}\xspace}
\newcommand{\MarkDevelopmentGaSdMatchGa}{\textbf{\SdMatchGa}\xspace}
\newcommand{\MarkDevelopmentGaSdGroupOneGa}{\textbf{\SdGroupOneGa}\xspace}
\newcommand{\MarkDevelopmentGaSdGroupTwoGa}{\textbf{\SdGroupTwoGa}\xspace}
\newcommand{\MarkDevelopmentGaSdGroupFourGa}{\textbf{\SdGroupFourGa}\xspace}
\newcommand{\MarkDevelopmentGaSdGroupEightGa}{\textbf{\SdGroupEightGa}\xspace}
\newcommand{\MarkDevelopmentGaSdUpperGa}{\textbf{\SdUpperGa}\xspace}
\newcommand{\MarkDevelopmentShSdMatchSh}{\textbf{\SdMatchSh}\xspace}
\newcommand{\MarkDevelopmentShSdGroupOneSh}{\textbf{\SdGroupOneSh}\xspace}
\newcommand{\MarkDevelopmentShSdGroupTwoSh}{\textbf{\SdGroupTwoSh}\xspace}
\newcommand{\MarkDevelopmentShSdGroupFourSh}{\textbf{\SdGroupFourSh}\xspace}
\newcommand{\MarkDevelopmentShSdGroupEightSh}{\textbf{\SdGroupEightSh}\xspace}
\newcommand{\MarkDevelopmentShSdUpperSh}{\textbf{\SdUpperSh}\xspace}
\newcommand{\MarkDevelopmentTightSdMatchTight}{\SdMatchTight\xspace}
\newcommand{\MarkDevelopmentTightSdGroupOneTight}{\SdGroupOneTight\xspace}
\newcommand{\MarkDevelopmentTightSdGroupTwoTight}{\SdGroupTwoTight\xspace}
\newcommand{\MarkDevelopmentTightSdGroupFourTight}{\SdGroupFourTight\xspace}
\newcommand{\MarkDevelopmentTightSdGroupEightTight}{\SdGroupEightTight\xspace}
\newcommand{\MarkDevelopmentTightSdUpperTight}{\textbf{\SdUpperTight}\xspace}
\newcommand{\MarkDevelopmentLooseSdMatchLoose}{\SdMatchLoose\xspace}
\newcommand{\MarkDevelopmentLooseSdGroupOneLoose}{\SdGroupOneLoose\xspace}
\newcommand{\MarkDevelopmentLooseSdGroupTwoLoose}{\SdGroupTwoLoose\xspace}
\newcommand{\MarkDevelopmentLooseSdGroupFourLoose}{\SdGroupFourLoose\xspace}
\newcommand{\MarkDevelopmentLooseSdGroupEightLoose}{\SdGroupEightLoose\xspace}
\newcommand{\MarkDevelopmentLooseSdUpperLoose}{\textbf{\SdUpperLoose}\xspace}
\newcommand{\MarkWeakGridFiveSgMatchFive}{\SgMatchFive\xspace}
\newcommand{\MarkWeakGridFiveSgGroupFive}{\SgGroupFive\xspace}
\newcommand{\MarkWeakGridFiveSgUpperFive}{\textbf{\SgUpperFive}\xspace}
\newcommand{\MarkWeakGridSixSgMatchSix}{\SgMatchSix\xspace}
\newcommand{\MarkWeakGridSixSgGroupSix}{\SgGroupSix\xspace}
\newcommand{\MarkWeakGridSixSgUpperSix}{\textbf{\SgUpperSix}\xspace}
\newcommand{\MarkWeakGridSevenSgMatchSeven}{\SgMatchSeven\xspace}
\newcommand{\MarkWeakGridSevenSgGroupSeven}{\SgGroupSeven\xspace}
\newcommand{\MarkWeakGridSevenSgUpperSeven}{\textbf{\SgUpperSeven}\xspace}
\newcommand{\MarkWeakGridEightSgMatchEight}{\SgMatchEight\xspace}
\newcommand{\MarkWeakGridEightSgGroupEight}{\SgGroupEight\xspace}
\newcommand{\MarkWeakGridEightSgUpperEight}{\textbf{\SgUpperEight}\xspace}
\newcommand{\MarkWeakGridNineSgMatchNine}{\SgMatchNine\xspace}
\newcommand{\MarkWeakGridNineSgGroupNine}{\SgGroupNine\xspace}
\newcommand{\MarkWeakGridNineSgUpperNine}{\textbf{\SgUpperNine}\xspace}
\newcommand{\MarkTimingPairATimeSitePairA}{\textbf{\TimeSitePairA}\xspace}
\newcommand{\MarkTimingPairATimeFullPairA}{\TimeFullPairA\xspace}
\newcommand{\MarkTimingPairBTimeSitePairB}{\textbf{\TimeSitePairB}\xspace}
\newcommand{\MarkTimingPairBTimeFullPairB}{\TimeFullPairB\xspace}

\newcommand{\tablenote}[1]{\par\smallskip{\fontsize{9}{10.5}\selectfont #1\par}}
\newcommand{\cfgA}{\Dino layer~\SiteA}
\newcommand{\cfgB}{SigLIP layer~\SiteB}
\newcommand{\cfgC}{\Dino layer~\SiteC at half width\xspace}
\newcommand{\headA}{\shortstack{\Dino\\layer \SiteA}}
\newcommand{\headB}{\shortstack{SigLIP\\layer \SiteB}}
\newcommand{\headC}{\shortstack{\Dino layer \SiteC\\half width}}
\newcommand{\headCnarrow}{\shortstack{\Dino\\layer \SiteC\\half width}}
\newcommand{\headDev}{\shortstack{Development\\\Dino layer \SiteDev}}

\newcommand{\rowAone}{\Dino}
\newcommand{\rowAtwo}{layer \SiteA}
\newcommand{\rowBone}{SigLIP}
\newcommand{\rowBtwo}{layer \SiteB}
\newcommand{\rowCone}{\Dino layer \SiteC}
\newcommand{\rowCtwo}{half width}
\newcommand{\figOverviewCaption}{\textbf{Ablating a feature and ablating its match in another seed's SAE often change the encoder output alike.} Random decoder SAEs keep their directions random. (b) Incorrect predictions of the cosine rule, which predicts a pass from the features' directions. (c) Those avoided by the layer rule\later{}, which predicts one from what the features add at the edited layer. \later{}Post hoc.}
\newcommand{\noteFigOverview}{On the scoring images, each setting has \UnitsTrained matched pairs and the random decoder pairs counted in \cref{tab:carrying}. Half width marks SAEs with half as many features as in the other two settings.}
\newcommand{\capDictionaries}{\textbf{The trained SAEs reconstruct each layer better than the random decoder SAEs with fewer active features.} Each cell gives the mean over the SAEs of one class in one setting, with the range as subscript. Active features count the features with a positive code on a token. Lower is better, bold marks a setting's lowest value and tint the trained SAEs.}
\newcommand{\noteDictionaries}{Both columns use the \ImagesHeld selection and scoring images for trained and random decoder SAEs, and held out Gaussian tokens scaled by surrogate training statistics for surrogate SAEs.}
\newcommand{\capCarrying}{\textbf{Matched pairs pass in every setting and almost no control pair does.} Rows count the pairs that pass the output test on the scoring images, by kind of pair, and give their shares with intervals as subscripts. A strong match has both cosines above \Weak. Up favours passes, bold marks the largest value of a kind and tint the matched pairs. The last block is post hoc\later.}
\newcommand{\noteCarrying}{Greedy selection and ridge regression are the replacements of \cref{sec:test}. A feature passes into all other seeds where it passes with its match in all four other SAEs.}
\newcommand{\figPredictorsCaption}{\textbf{Higher decoder cosines and lower layer errors go with lower output errors.} Each dot is a pair, its output error plotted over (a) its decoder cosine for the cosine rule and (b) its layer error\later{} for the layer rule. Output test only marks passing pairs that the row's rule predicts to fail, and rule only marks the reverse.}
\newcommand{\noteFigPredictors}{The output error on the scoring images and the layer error on the selection images use logarithmic axes, which omit zero. The threshold of the cosine rule on the encoder cosine is not plotted.}
\newcommand{\capCosine}{\textbf{The layer rule agrees with the output test more often than the cosine rule.} The upper block gives agreement with the outcomes on the two halves of the scoring images, and the lower block counts pairs on all scoring images. Arrows give the better direction, bold the better rule and tint the output test. The layer rule's rows are post hoc\later.}
\newcommand{\noteCosine}{Subscripts give intervals over resampled features. The layer error is computed on the selection images.}
\newcommand{\figEveryCaption}{\textbf{The advantage of the features that pass into all other seeds depends on the setting and the SAE.} For each first SAE, panel (a) counts how many of \RandomSets random sets of the same size reconstruct less variance than the features that pass. Panel (b) compares those features with as many features ranked highest by cosine.}
\newcommand{\noteFigEvery}{On the scoring images, panel (b) divides the variance gain by the cosine ranked set's reconstructed variance, and each random set draws from the SAE's \DrawnPerSource sampled features.}
\newcommand{\capRepeat}{\textbf{No permuted pair passes, and the counts repeat on fresh images and on a second sample of features.} Upper rows count the pairs that pass the output test, and lower rows count each rule's incorrect predictions. Bold marks the fewest incorrect predictions of a block and tint the layer rule.}
\newcommand{\noteRepeat}{The fresh images are \FreshImages images of the COCO 2017 training split. The second sample takes \DrawnPerSource further features from each SAE and is scored on the fresh images. The permuted control gives each of \TenthPermSourcesA sampled features a setting the match of another, under \TenthPermutations permutations. The three rows between the cosine rule and the layer rule take one part of \cref{eq:layer} each, with $c$ the decoder cosine, and the layer rule's threshold is \NinthFitJoint.}
\newcommand{\capNames}{\textbf{More than nine sampled features in ten take no name in every setting, so names cannot be compared across seeds.} Upper rows count sampled features by the largest share of their positive codes that one object category covers on the \ImagesSelect selection images\later, and lower rows count agreeing names among named pairs. The counts have no better direction.}
\newcommand{\noteNames}{A feature takes a name where one category's masks cover at least \NameHold of its positive codes on the selection images, and keeps it where the same holds on the scoring images. The median row gives that largest share for the median sampled feature, and a named pair is one whose two features both take a name.}
\newcommand{\capHalves}{\textbf{The layer rule repeats fewer incorrect predictions on both halves of the scoring images than the cosine rule.} Rows give the cosine rule's incorrect predictions on each half and the second comparison, and below them the incorrect predictions that each rule repeats on both halves. The first four rows were fixed in advance and the rest are post hoc\later.}
\newcommand{\noteHalves}{The layer rule's prediction is fixed on the selection images unless a row names the same halves. A row that subtracts disagreeing pairs takes the pairs whose halves disagree from a rule's repeated incorrect predictions. Intervals span the middle \IntervalLevelPct{}\% of \Resamples feature resamples. Image resamples draw within each half. Omission rows drop each seed as source only or from both positions.}
\newcommand{\capFresh}{\textbf{On fresh images the same shares pass, and in no setting does the output test's agreement exceed the cosine rule's by the required difference\later.} Rows compare the fresh images and their halves over the matched pairs of each setting. Up arrows favour agreement and down arrows fewer incorrect predictions, bold marks the better rule and tint the output test.}
\newcommand{\noteFresh}{The difference's subscript gives the middle \IntervalLevelPct{}\% of \Resamples resamples of the sampled features, and the next row their fifth percentile. The resample count draws fresh images within each half at its own required difference. The layer rule uses the selection images and the original output bound unless marked as fitted.}
\newcommand{\capPredictors}{\textbf{No single part of the layer error predicts incorrectly as rarely as the layer rule in every setting\later.} Each rule is fitted on development pairs and scored on fresh matched pairs, with predictors computed on selection images above and on the same fresh images below. Down arrows favour fewer incorrect predictions, bold marks the fewest and tint the layer rule.}
\newcommand{\notePredictors}{Subscripts give the middle \IntervalLevelPct{}\% of \Resamples resamples of the sampled features, for a rule's incorrect predictions less the layer rule's. Here $c$ is the decoder cosine, $\rho$ the code cosine and $r$ the code norm ratio. The cosine rule predicts a pass where both cosines are at least their thresholds, and every other rule where its score is at most its threshold. At a fixed nonnegative product of $c$ and $\rho$, the layer error is smallest where $r$ equals that product. The layer rule's fitted threshold is \FittedSiteExact in full.}
\newcommand{\capEight}{\textbf{On a third image set one of the \EightCount pairs keeps its disagreement under image resampling\later.} Rows give pairs selected for same image disagreement in the same direction on both scoring (S) and both fresh (F) halves. Subscripts give pointwise intervals over \Resamples paired image resamples, including on the third images (U).}
\newcommand{\noteEight}{A pair is named by its setting, its first SAE's seed and feature and its second SAE's seed, with its outcome on the scoring images. The two share columns give the share of resamples in which the two outcomes are opposite and in which they keep their stored direction, and a disagreement persists where the second share is at least \SeventhFamilyShare on U and on both its halves.}
\newcommand{\noteEightContinued}{S, F and U are the scoring, fresh and third images. Subscripts give paired image resampling intervals. Stored direction keeps the layer and output errors on their original, opposite sides of the bound.}
\newcommand{\capEightContinued}{\textbf{One pair's disagreement persists on the third image set\later.} This part continues \cref{tab:supp-six}, whose first part defines the columns, the image sets and the criterion.}
\newcommand{\capContinuation}{\textbf{Each pair's ratio already lies on its final side of \Certify before the final normalization\later.} Rows give the \EightCount pairs of \cref{tab:supp-six} on the \SeventhThirdImages third images, with the outcome of the output test on the scoring images. Columns give two ratios at the edited layer, just before the final normalization and at the output.}
\newcommand{\noteContinuation}{A pair is named by the seed and feature of its first SAE and the seed and feature of its second SAE. The first ratio is that of \cref{eq:error}, which at the edited layer is the layer error.}
\newcommand{\figRuleCaption}{\textbf{The fitted cosine thresholds include most passing development pairs and exclude most failing ones.} Each mark is one of the \DevUnits development pairs that fitted the cosine rule, plotted by its encoder cosine against its decoder cosine. The two lines mark a decoder cosine of \RuleDec and an encoder cosine of \RuleEnc.}
\newcommand{\noteFigRule}{Each axis shows the distance of a cosine from one on a log scale, with ticks at the cosine itself. A pass meets the original output error bound on the scoring images.}
\newcommand{\capDevelopment}{\textbf{Every replacement passes all pairs of rescaled copies and no control pair.} Cells give the share of development pairs that pass by replacement and kind of pair, with feature resampling intervals conditional on the trained SAEs. Up arrows favour passes of matched and rescaled pairs, down arrows fewer passes of the other pairs, bold marks the best value and tint the match.}
\newcommand{\noteDevelopment}{A replacement stands in for the first feature in the second SAE: the match, greedy selection of at most the stated number of features, or ridge regression on all features of the second SAE. The five columns count \DevUnits, \UnitsFrozen, \UnitsSurrogate, \UnitsKnown and \UnitsKnown pairs.}
\newcommand{\capBounds}{\textbf{Each replacement passes a larger share at a larger bound on the output error.} Cells give the shares of the \UnitsTrained matched pairs that pass on the scoring images, with intervals as subscripts. Up arrows mark higher shares, bold the largest development share at each bound and tint the match. The rows of the three settings are post hoc\later.}
\newcommand{\noteBounds}{The upper rows give the development setting. Intervals resample features with the SAEs fixed.}
\newcommand{\capWeakGrid}{\textbf{Only ridge regression passes weak matches at cutoffs through \SgAtSix.} Columns fix the cutoff among the \DevUnits development pairs. Rows count the weak matches, the passes of each replacement and the empty greedy selections. Up arrows favour passes, bold marks the largest count and tint the match.}
\newcommand{\noteWeakGrid}{A greedy selection has at most \Budget features. A pass meets the original output error bound on the scoring images.}
\newcommand{\capWeak}{\textbf{No replacement makes more than a few weak matches pass.} Rows count the pairs that pass among all matched pairs of each setting and among its weak matches, whose decoder or encoder cosine is at most \Weak. Up is better, bold marks the largest count among replacements on the same pairs and tint the match.}
\newcommand{\noteWeak}{Greedy selection takes at most \Budget features, and ridge regression uses all features of the second SAE. Pairs pass at the original output error bound on the scoring images.}
\newcommand{\capVariance}{\textbf{On most SAEs the features that pass into all other seeds reconstruct more variance than as many features ranked highest by cosine\later.} They reconstruct more on \VarAboveA SAEs of \cfgA, \VarAboveB of \cfgB and \VarAboveC of \cfgC. Rows give each first SAE by its training seed. Up is better, bold marks an SAE's better set and tint the features that pass.}
\newcommand{\noteVariance}{Each variance column gives the share of the layer's variance that a set reconstructs on the scoring images, for the features that pass and for as many features ranked highest by their lowest decoder or encoder cosine across the other seeds.}
\newcommand{\capFreshVariance}{\textbf{On fresh images the features that pass into all other seeds keep a positive mean advantage in every setting\later.} Each set keeps the members it had on the scoring images, and the differences take both signs across the SAEs of \cfgB and of \cfgC. Up is better, bold marks an SAE's better set and tint the features that pass.}
\newcommand{\noteFreshVariance}{Each variance column gives the share of the layer's variance that a set reconstructs on the \FreshImages fresh images, for the features that pass and for as many features ranked highest by the lowest decoder or encoder cosine across the other seeds. The difference subtracts the second from the first.}
\newcommand{\capTiming}{\textbf{The layer rule costs less than the output test in both timed settings\later.} Entries give median seconds, with the full range over runs as subscript. Each time per pair covers the stage its rule adds beside the shared passes above it. Down arrows favour less time, bold marks the cheaper rule per pair and tint the output test.}
\newcommand{\noteTiming}{Each time is the median of three runs on the fresh images. The two timed settings have \WidthA features each. The short cosine prediction timing uses stored arrays on the container processor and excludes matching and transfer. The matching and the cosine predictions cover all \EleventhTimedFeatures matched features of two trained SAEs, and their sum is the cost of the cosine rule for one pair of SAEs. Those two steps ran in a container whose log gives \EleventhContainerProcessors processors and \EleventhContainerThreads threads and names no processor model, and the matching alone also ran on an \SeventhProcessor, so each of these rows compares only with rows of the same processor. The image and ablation steps ran on one \TimingDevice in single precision, on batches of \TimingBatch images at \DinoCrop by \DinoCrop pixels, after one batch and one pair of warm up, with the accelerator synchronized before each clock.}
\newcommand{\capSiteBoundsOne}{\textbf{Incorrect predictions of the layer rule stay few at a common bound of \BoundTight\later.} The output test and the layer rule share the bound. The left block fixes the layer rule's prediction on the selection images, and the right block recomputes both on the images a column names.}
\newcommand{\noteSiteBounds}{The scoring and fresh columns count incorrect predictions on a whole set of \ImagesScore images, and each all halves column counts the pairs with one direction of incorrect prediction on all four halves of \ImagesHalf. Each subscript is a pointwise \IntervalLevelPct{}\% interval over \Resamples paired image resamples, and counts cover the \SeedPairs pairs of each first SAE.}
\newcommand{\capSiteBoundsTwo}{\textbf{Incorrect predictions of the layer rule stay few at the common bound \Certify\later.} The left block fixes predictions on the selection images and the right recomputes them on the scored images. Subscripts give paired image resampling intervals.}
\newcommand{\capSiteBoundsThree}{\textbf{Incorrect predictions of the layer rule stay few at the common bound \BoundLoose\later.} The left block fixes predictions on the selection images and the right recomputes them on the scored images. Subscripts give paired image resampling intervals.}
\newcommand{\capFittedSelection}{\textbf{Incorrect predictions remain on fresh images after the layer rule's threshold is fitted\later.} Cells count each direction of incorrect prediction among the \SeedPairs pairs of each first SAE, at the original threshold and the fitted threshold of \NinthFitJoint.}
\newcommand{\noteFittedSelection}{The layer rule's prediction is fixed on the selection images. Incorrect predictions are counted on the \FreshImages fresh images. The threshold was frozen before these counts, after it agreed with the output test on \FittedAgree of \DevUnits development pairs. Subscripts give pointwise \IntervalLevelPct{}\% intervals over \Resamples paired image resamples.}
\newcommand{\capFittedFresh}{\textbf{Incorrect predictions remain when the layer error and the output error use the same fresh images\later.} Cells count each direction of incorrect prediction among the \SeedPairs pairs of each first SAE on the \FreshImages fresh images, with both errors recomputed in each of \Resamples paired image resamples. The thresholds follow \cref{tab:fitted-selection}.}
\newcommand{\noteFittedFresh}{Subscripts give pointwise \IntervalLevelPct{}\% intervals.}
\newcommand{\capCaseVerdict}{\textbf{The pair whose incorrect prediction persists stays an incorrect prediction of the layer rule at its fitted threshold\later.} Rows compare the pair's layer error and its output error with the two bounds on the \ImagesSelect selection, \ImagesScore scoring, \FreshImages fresh and \SeventhThirdImages third images.}
\newcommand{\noteCaseVerdict}{The pair is feature \EightSevenFeature of the SAE of seed \EightSevenSource with its match, feature \CasePartner of the SAE of seed \EightSevenTarget, at \cfgC, whose SAEs have \WidthC features.}
\newcommand{\capCaseRemovals}{\textbf{The output test still fails the pair when the largest image leaves each set\later.} Rows give the pair's output error on each image set, the shares of the ten largest images and the output error after one image is dropped.}
\newcommand{\noteCaseRemovals}{The squared difference is that between the two ablations' effects, and the squared first effect is the squared size of the first ablation's effect, on an image or a token. Each share gives the part of the set's total that falls in its ten largest images. The last two columns drop the image of largest squared first effect or of largest squared difference, which is one image on each set. The pair is that of \cref{tab:case-verdict}.}
\newcommand{\capCaseImages}{\textbf{The ten images of largest squared difference account for \CaseScoreTopShare, \CaseFreshTopShare and \CaseThirdTopShare of the three sets' squared difference\later.} Rows list each set's ten largest images with each one's squared difference, squared first effect and shares of the set's totals.}
\newcommand{\noteCaseImages}{Each share divides an image's squared difference in encoder output effects or squared first effect by its total over the image set. Each image is named by its number in the COCO collection. The pair is that of \cref{tab:case-verdict}.}
\newcommand{\capCaseTokens}{\textbf{Almost all of the squared difference and of the squared first effect lie at patch tokens\later.} Each row splits both totals of an image set between the class token and the patch tokens, so each pair of shares sums to one.}
\newcommand{\noteCaseTokens}{Each entry is the share of the set's total at the named token positions. The pair is that of \cref{tab:case-verdict}.}
\newcommand{\capCaseTopTokens}{\textbf{The ten tokens of largest squared difference are patch tokens on every set\later.} Rows give each token's squared difference and squared first effect with their shares of the set's totals.}
\newcommand{\noteCaseTopTokens}{Patch row and patch column give the position of the patch in the image, counted from zero at the upper left. Each image is named by its number in the COCO collection. The pair is that of \cref{tab:case-verdict}.}
\newcommand{\capCaseGrid}{\textbf{The squared difference at patch tokens is spread across the image\later.} Regions have four by four patches, counted from zero at the upper left. Entries give each region's share of the squared difference or of the squared first effect over all images and tokens of its set.}
\newcommand{\noteCaseGrid}{The pair is that of \cref{tab:case-verdict}.}
\newcommand{\capPermuted}{\textbf{No pair passes when the matches are permuted within the trained SAEs\later.} Cells count the pairs that pass among \TenthPermMatchedPairsA pairs of \TenthPermSourcesA sampled features a setting on the fresh images. A permuted pair gives a feature the match of another sampled feature of its SAE in the same second SAE.}
\newcommand{\notePermuted}{Each of \TenthPermutations permutations leaves no feature its own match. Up arrows favour passes, bold marks the largest count and tint the matched pairs. Subscripts give the middle \IntervalLevelPct{}\% of \Resamples resamples of the \TenthPermSourcesA features.}
\newcommand{\capSecond}{\textbf{On a second sample of features the layer rule again predicts incorrectly on fewer pairs than the cosine rule\later.} The second sample takes \DrawnPerSource features from each SAE, disjoint from the first, and scores their matched pairs on the fresh images with every rule frozen. Down arrows favour fewer incorrect predictions, bold marks the fewest and tint the layer rule.}
\newcommand{\noteSecond}{Every statistic uses the selection images. Codes denotes the layer error with decoder cosine $c$ set to one. The layer rule predicts a pass at or below \NinthFitJoint, the layer error with $c$ set to one at or below \NinthFitCodes and the cosine rule at \RuleDec and \RuleEnc. Subscripts give the middle \IntervalLevelPct{}\% of \Resamples resamples of the sampled features.}
\newcommand{\capOutputs}{\textbf{At the image level output the layer rule still predicts incorrectly on fewer pairs than the cosine rule, and both mispredict more pairs than on all tokens\later.} Rows score the matched pairs of each setting on the fresh images at \Dino's class token and SigLIP's pooled embedding. Down is better, bold marks the fewest and tint the layer rule.}
\newcommand{\noteOutputs}{The image level output is the class token of the encoder output after the final normalization for \Dino and the pooled image embedding for SigLIP, and the output error is that of \cref{eq:error} at that output. Each rule keeps its prediction from the selection images, with the layer rule's fitted threshold at \NinthFitJoint, the layer error with decoder cosine $c$ set to one at \NinthFitCodes, the code cosine alone at \NinthFitOverlap and the code norm ratio alone at \NinthFitMagnitude. Subscripts in the lower block count the incorrect predictions that repeat on both halves of \FreshHalf images, and the last row's give the middle \IntervalLevelPct{}\% of \Resamples resamples of the sampled features.}

\newcommand{\SiteA}{6\xspace}
\newcommand{\WidthA}{6,144\xspace}
\newcommand{\SiteB}{11\xspace}
\newcommand{\WidthB}{6,144\xspace}
\newcommand{\SiteC}{11\xspace}
\newcommand{\WidthC}{3,072\xspace}
\newcommand{\TokensDino}{257\xspace}
\newcommand{\TokensSiglip}{196\xspace}
\newcommand{\SiteDev}{11\xspace}
\newcommand{\WidthDev}{6,144\xspace}
\newcommand{\ImagesAll}{5,000\xspace}
\newcommand{\ImagesTrain}{4,000\xspace}
\newcommand{\ImagesHeld}{1,000\xspace}
\newcommand{\ImagesSelect}{500\xspace}
\newcommand{\ImagesScore}{500\xspace}
\newcommand{\ImagesHalf}{250\xspace}
\newcommand{\Seeds}{5\xspace}
\newcommand{\DrawnPerSource}{128\xspace}
\newcommand{\DrawnAll}{640\xspace}
\newcommand{\UnitsTrained}{2,560\xspace}
\newcommand{\UnitsFrozen}{3,200\xspace}
\newcommand{\UnitsSurrogate}{2,560\xspace}
\newcommand{\UnitsKnown}{640\xspace}
\newcommand{\Coefficient}{0.001\xspace}
\newcommand{\Steps}{20,000\xspace}
\newcommand{\Batch}{4,096\xspace}
\newcommand{\Rate}{0.0003\xspace}
\newcommand{\WarmSteps}{1,000\xspace}
\newcommand{\DecaySteps}{4,000\xspace}

\newcommand{\LzeroA}{552\xspace}

\newcommand{\LzeroC}{76\xspace}
\newcommand{\Certify}{0.25\xspace}
\newcommand{\Weak}{0.7\xspace}
\newcommand{\MinImages}{8\xspace}
\newcommand{\EffectFloor}{$1.0\times 10^{-8}$\xspace}
\newcommand{\Budget}{8\xspace}
\newcommand{\Ridge}{$1.0\times 10^{-6}$\xspace}
\newcommand{\GreedyTol}{$1.0\times 10^{-6}$\xspace}
\newcommand{\NameHold}{0.5\xspace}
\newcommand{\Resamples}{2,000\xspace}

\newcommand{\LogScale}{2\xspace}
\newcommand{\MarginOne}{0.1\xspace}
\newcommand{\DevCert}{1,038\xspace}
\newcommand{\DevUnits}{2,560\xspace}

\newcommand{\DevFrozenCert}{0\xspace}
\newcommand{\DevSurrogateCert}{0\xspace}

\newcommand{\DevEvery}{215\xspace}
\newcommand{\DevWeakUnits}{1,259\xspace}
\newcommand{\DevWeakMatch}{3\xspace}
\newcommand{\DevWeakGroup}{4\xspace}
\newcommand{\DevWeakUpper}{12\xspace}
\newcommand{\DevWeakEmpty}{1,117\xspace}
\newcommand{\TabDevTr}{0.405\xspace}
\newcommand{\TabDevTrLow}{0.370\xspace}
\newcommand{\TabDevTrHigh}{0.441\xspace}
\newcommand{\TabDevFr}{0.000\xspace}
\newcommand{\TabDevFrLow}{0.000\xspace}
\newcommand{\TabDevFrHigh}{0.000\xspace}
\newcommand{\TabDevSu}{0.000\xspace}
\newcommand{\TabDevSuLow}{0.000\xspace}
\newcommand{\TabDevSuHigh}{0.000\xspace}
\newcommand{\RuleDec}{0.96\xspace}
\newcommand{\RuleEnc}{0.71\xspace}
\newcommand{\RuleDevAgree}{2,505\xspace}

\newcommand{\RuleTies}{3\xspace}
\newcommand{\CertA}{304\xspace}
\newcommand{\ShareA}{0.119\xspace}
\newcommand{\SharePctA}{12\xspace}

\newcommand{\FrozenCertA}{0\xspace}
\newcommand{\TrLowA}{0.096\xspace}
\newcommand{\TrHighA}{0.142\xspace}
\newcommand{\FrShareA}{0.000\xspace}
\newcommand{\FrLowA}{0.000\xspace}
\newcommand{\FrHighA}{0.000\xspace}
\newcommand{\GaugeCertA}{640\xspace}

\newcommand{\RulePredA}{255\xspace}
\newcommand{\DisagreeA}{57\xspace}
\newcommand{\RuleFalsePosA}{4\xspace}
\newcommand{\RuleFalseNegA}{53\xspace}
\newcommand{\StrongNotA}{208\xspace}

\newcommand{\SelfAgreeA}{0.9891\xspace}
\newcommand{\RuleAgreeA}{0.9773\xspace}

\newcommand{\SelfLowA}{0.9844\xspace}
\newcommand{\SelfHighA}{0.9934\xspace}
\newcommand{\RuleLowA}{0.9705\xspace}
\newcommand{\RuleHighA}{0.9838\xspace}
\newcommand{\DiffTwoA}{0.0117\xspace}
\newcommand{\MarginTwoA}{0.0113\xspace}
\newcommand{\DiffTwoLowA}{0.0064\xspace}
\newcommand{\DiffTwoHighA}{0.0174\xspace}
\newcommand{\UnnamedA}{637\xspace}
\newcommand{\NamedCertA}{12\xspace}
\newcommand{\AgreeCertA}{12\xspace}
\newcommand{\NamedNotCertA}{0\xspace}
\newcommand{\AgreeNotCertA}{0\xspace}
\newcommand{\NamedRuleA}{12\xspace}
\newcommand{\AgreeRuleA}{12\xspace}
\newcommand{\NamedShufA}{0\xspace}
\newcommand{\AgreeShufA}{0\xspace}
\newcommand{\WeakUnitsA}{2,048\xspace}
\newcommand{\WeakMatchA}{0\xspace}
\newcommand{\WeakGroupA}{0\xspace}
\newcommand{\WeakUpperA}{2\xspace}
\newcommand{\WeakEmptyA}{1,953\xspace}
\newcommand{\EveryA}{50\xspace}

\newcommand{\CertB}{1,057\xspace}
\newcommand{\ShareB}{0.413\xspace}

\newcommand{\FrozenCertB}{0\xspace}
\newcommand{\TrLowB}{0.377\xspace}
\newcommand{\TrHighB}{0.451\xspace}
\newcommand{\FrShareB}{0.000\xspace}
\newcommand{\FrLowB}{0.000\xspace}
\newcommand{\FrHighB}{0.000\xspace}
\newcommand{\SuShareB}{0.000\xspace}
\newcommand{\SuLowB}{0.000\xspace}
\newcommand{\SuHighB}{0.000\xspace}

\newcommand{\RulePredB}{993\xspace}
\newcommand{\DisagreeB}{66\xspace}
\newcommand{\RuleFalsePosB}{1\xspace}
\newcommand{\RuleFalseNegB}{65\xspace}

\newcommand{\SelfAgreeB}{0.9840\xspace}
\newcommand{\RuleAgreeB}{0.9736\xspace}

\newcommand{\SelfLowB}{0.9773\xspace}
\newcommand{\SelfHighB}{0.9898\xspace}
\newcommand{\RuleLowB}{0.9658\xspace}
\newcommand{\RuleHighB}{0.9809\xspace}
\newcommand{\DiffTwoB}{0.0104\xspace}
\newcommand{\MarginTwoB}{0.0132\xspace}
\newcommand{\DiffTwoLowB}{0.0043\xspace}
\newcommand{\DiffTwoHighB}{0.0174\xspace}
\newcommand{\UnnamedB}{629\xspace}
\newcommand{\NamedCertB}{24\xspace}
\newcommand{\AgreeCertB}{24\xspace}
\newcommand{\NamedNotCertB}{12\xspace}
\newcommand{\AgreeNotCertB}{12\xspace}
\newcommand{\NamedRuleB}{21\xspace}
\newcommand{\AgreeRuleB}{21\xspace}
\newcommand{\NamedShufB}{0\xspace}
\newcommand{\AgreeShufB}{0\xspace}
\newcommand{\WeakUnitsB}{1,222\xspace}
\newcommand{\WeakMatchB}{0\xspace}
\newcommand{\WeakGroupB}{0\xspace}
\newcommand{\WeakUpperB}{4\xspace}
\newcommand{\WeakEmptyB}{1,088\xspace}
\newcommand{\EveryB}{211\xspace}

\newcommand{\CertC}{1,790\xspace}
\newcommand{\ShareC}{0.699\xspace}
\newcommand{\SharePctC}{70\xspace}

\newcommand{\FrozenCertC}{0\xspace}
\newcommand{\TrLowC}{0.668\xspace}
\newcommand{\TrHighC}{0.731\xspace}
\newcommand{\FrShareC}{0.000\xspace}
\newcommand{\FrLowC}{0.000\xspace}
\newcommand{\FrHighC}{0.000\xspace}
\newcommand{\SuShareC}{0.000\xspace}
\newcommand{\SuLowC}{0.000\xspace}
\newcommand{\SuHighC}{0.001\xspace}

\newcommand{\RulePredC}{1,734\xspace}
\newcommand{\DisagreeC}{86\xspace}
\newcommand{\RuleFalsePosC}{15\xspace}
\newcommand{\RuleFalseNegC}{71\xspace}

\newcommand{\SelfAgreeC}{0.9820\xspace}
\newcommand{\RuleAgreeC}{0.9648\xspace}

\newcommand{\SelfLowC}{0.9754\xspace}
\newcommand{\SelfHighC}{0.9875\xspace}
\newcommand{\RuleLowC}{0.9562\xspace}
\newcommand{\RuleHighC}{0.9732\xspace}
\newcommand{\DiffTwoC}{0.0172\xspace}
\newcommand{\MarginTwoC}{0.0176\xspace}
\newcommand{\DiffTwoLowC}{0.0092\xspace}
\newcommand{\DiffTwoHighC}{0.0258\xspace}
\newcommand{\UnnamedC}{590\xspace}
\newcommand{\NamedCertC}{165\xspace}
\newcommand{\AgreeCertC}{165\xspace}
\newcommand{\NamedNotCertC}{18\xspace}
\newcommand{\AgreeNotCertC}{17\xspace}
\newcommand{\NamedRuleC}{157\xspace}
\newcommand{\AgreeRuleC}{157\xspace}
\newcommand{\NamedShufC}{2\xspace}
\newcommand{\AgreeShufC}{0\xspace}
\newcommand{\WeakUnitsC}{544\xspace}
\newcommand{\WeakMatchC}{0\xspace}
\newcommand{\WeakGroupC}{5\xspace}
\newcommand{\WeakUpperC}{11\xspace}
\newcommand{\WeakEmptyC}{466\xspace}
\newcommand{\EveryC}{368\xspace}

\newcommand{\SurrogateCertB}{0\xspace}
\newcommand{\SurrogateCertC}{1\xspace}
\newcommand{\RandomSets}{200\xspace}
\newcommand{\ComputeCalls}{36\xspace}
\newcommand{\ComputeSeconds}{42,447\xspace}
\newcommand{\ComputeSources}{31,795\xspace}
\newcommand{\ComputeTrain}{9,698\xspace}
\newcommand{\SdMatchTr}{0.405\xspace}
\newcommand{\SdMatchTrLow}{0.370\xspace}
\newcommand{\SdMatchTrHigh}{0.441\xspace}
\newcommand{\SdMatchFr}{0.000\xspace}
\newcommand{\SdMatchFrLow}{0.000\xspace}
\newcommand{\SdMatchFrHigh}{0.000\xspace}
\newcommand{\SdMatchSu}{0.000\xspace}
\newcommand{\SdMatchSuLow}{0.000\xspace}
\newcommand{\SdMatchSuHigh}{0.000\xspace}
\newcommand{\SdMatchGa}{1.000\xspace}
\newcommand{\SdMatchGaLow}{1.000\xspace}
\newcommand{\SdMatchGaHigh}{1.000\xspace}
\newcommand{\SdMatchSh}{0.000\xspace}
\newcommand{\SdMatchShLow}{0.000\xspace}
\newcommand{\SdMatchShHigh}{0.000\xspace}
\newcommand{\SdMatchTight}{0.337\xspace}
\newcommand{\SdMatchTightLow}{0.302\xspace}
\newcommand{\SdMatchTightHigh}{0.371\xspace}
\newcommand{\SdMatchLoose}{0.463\xspace}
\newcommand{\SdMatchLooseLow}{0.427\xspace}
\newcommand{\SdMatchLooseHigh}{0.499\xspace}
\newcommand{\SdGroupOneTr}{0.405\xspace}
\newcommand{\SdGroupOneTrLow}{0.370\xspace}
\newcommand{\SdGroupOneTrHigh}{0.441\xspace}
\newcommand{\SdGroupOneFr}{0.000\xspace}
\newcommand{\SdGroupOneFrLow}{0.000\xspace}
\newcommand{\SdGroupOneFrHigh}{0.000\xspace}
\newcommand{\SdGroupOneSu}{0.000\xspace}
\newcommand{\SdGroupOneSuLow}{0.000\xspace}
\newcommand{\SdGroupOneSuHigh}{0.000\xspace}
\newcommand{\SdGroupOneGa}{1.000\xspace}
\newcommand{\SdGroupOneGaLow}{1.000\xspace}
\newcommand{\SdGroupOneGaHigh}{1.000\xspace}
\newcommand{\SdGroupOneSh}{0.000\xspace}
\newcommand{\SdGroupOneShLow}{0.000\xspace}
\newcommand{\SdGroupOneShHigh}{0.000\xspace}
\newcommand{\SdGroupOneTight}{0.337\xspace}
\newcommand{\SdGroupOneTightLow}{0.302\xspace}
\newcommand{\SdGroupOneTightHigh}{0.371\xspace}
\newcommand{\SdGroupOneLoose}{0.464\xspace}
\newcommand{\SdGroupOneLooseLow}{0.427\xspace}
\newcommand{\SdGroupOneLooseHigh}{0.500\xspace}
\newcommand{\SdGroupTwoTr}{0.407\xspace}
\newcommand{\SdGroupTwoTrLow}{0.371\xspace}
\newcommand{\SdGroupTwoTrHigh}{0.443\xspace}
\newcommand{\SdGroupTwoFr}{0.000\xspace}
\newcommand{\SdGroupTwoFrLow}{0.000\xspace}
\newcommand{\SdGroupTwoFrHigh}{0.000\xspace}
\newcommand{\SdGroupTwoSu}{0.000\xspace}
\newcommand{\SdGroupTwoSuLow}{0.000\xspace}
\newcommand{\SdGroupTwoSuHigh}{0.000\xspace}
\newcommand{\SdGroupTwoGa}{1.000\xspace}
\newcommand{\SdGroupTwoGaLow}{1.000\xspace}
\newcommand{\SdGroupTwoGaHigh}{1.000\xspace}
\newcommand{\SdGroupTwoSh}{0.000\xspace}
\newcommand{\SdGroupTwoShLow}{0.000\xspace}
\newcommand{\SdGroupTwoShHigh}{0.000\xspace}
\newcommand{\SdGroupTwoTight}{0.337\xspace}
\newcommand{\SdGroupTwoTightLow}{0.302\xspace}
\newcommand{\SdGroupTwoTightHigh}{0.371\xspace}
\newcommand{\SdGroupTwoLoose}{0.466\xspace}
\newcommand{\SdGroupTwoLooseLow}{0.429\xspace}
\newcommand{\SdGroupTwoLooseHigh}{0.502\xspace}
\newcommand{\SdGroupFourTr}{0.407\xspace}
\newcommand{\SdGroupFourTrLow}{0.370\xspace}
\newcommand{\SdGroupFourTrHigh}{0.443\xspace}
\newcommand{\SdGroupFourFr}{0.000\xspace}
\newcommand{\SdGroupFourFrLow}{0.000\xspace}
\newcommand{\SdGroupFourFrHigh}{0.000\xspace}
\newcommand{\SdGroupFourSu}{0.000\xspace}
\newcommand{\SdGroupFourSuLow}{0.000\xspace}
\newcommand{\SdGroupFourSuHigh}{0.000\xspace}
\newcommand{\SdGroupFourGa}{1.000\xspace}
\newcommand{\SdGroupFourGaLow}{1.000\xspace}
\newcommand{\SdGroupFourGaHigh}{1.000\xspace}
\newcommand{\SdGroupFourSh}{0.000\xspace}
\newcommand{\SdGroupFourShLow}{0.000\xspace}
\newcommand{\SdGroupFourShHigh}{0.000\xspace}
\newcommand{\SdGroupFourTight}{0.337\xspace}
\newcommand{\SdGroupFourTightLow}{0.302\xspace}
\newcommand{\SdGroupFourTightHigh}{0.371\xspace}
\newcommand{\SdGroupFourLoose}{0.466\xspace}
\newcommand{\SdGroupFourLooseLow}{0.429\xspace}
\newcommand{\SdGroupFourLooseHigh}{0.502\xspace}
\newcommand{\SdGroupEightTr}{0.407\xspace}
\newcommand{\SdGroupEightTrLow}{0.370\xspace}
\newcommand{\SdGroupEightTrHigh}{0.443\xspace}
\newcommand{\SdGroupEightFr}{0.000\xspace}
\newcommand{\SdGroupEightFrLow}{0.000\xspace}
\newcommand{\SdGroupEightFrHigh}{0.000\xspace}
\newcommand{\SdGroupEightSu}{0.000\xspace}
\newcommand{\SdGroupEightSuLow}{0.000\xspace}
\newcommand{\SdGroupEightSuHigh}{0.000\xspace}
\newcommand{\SdGroupEightGa}{1.000\xspace}
\newcommand{\SdGroupEightGaLow}{1.000\xspace}
\newcommand{\SdGroupEightGaHigh}{1.000\xspace}
\newcommand{\SdGroupEightSh}{0.000\xspace}
\newcommand{\SdGroupEightShLow}{0.000\xspace}
\newcommand{\SdGroupEightShHigh}{0.000\xspace}
\newcommand{\SdGroupEightTight}{0.337\xspace}
\newcommand{\SdGroupEightTightLow}{0.302\xspace}
\newcommand{\SdGroupEightTightHigh}{0.371\xspace}
\newcommand{\SdGroupEightLoose}{0.466\xspace}
\newcommand{\SdGroupEightLooseLow}{0.429\xspace}
\newcommand{\SdGroupEightLooseHigh}{0.502\xspace}
\newcommand{\SdUpperTr}{0.431\xspace}
\newcommand{\SdUpperTrLow}{0.393\xspace}
\newcommand{\SdUpperTrHigh}{0.468\xspace}
\newcommand{\SdUpperFr}{0.000\xspace}
\newcommand{\SdUpperFrLow}{0.000\xspace}
\newcommand{\SdUpperFrHigh}{0.000\xspace}
\newcommand{\SdUpperSu}{0.000\xspace}
\newcommand{\SdUpperSuLow}{0.000\xspace}
\newcommand{\SdUpperSuHigh}{0.000\xspace}
\newcommand{\SdUpperGa}{1.000\xspace}
\newcommand{\SdUpperGaLow}{1.000\xspace}
\newcommand{\SdUpperGaHigh}{1.000\xspace}
\newcommand{\SdUpperSh}{0.000\xspace}
\newcommand{\SdUpperShLow}{0.000\xspace}
\newcommand{\SdUpperShHigh}{0.000\xspace}
\newcommand{\SdUpperTight}{0.350\xspace}
\newcommand{\SdUpperTightLow}{0.313\xspace}
\newcommand{\SdUpperTightHigh}{0.384\xspace}
\newcommand{\SdUpperLoose}{0.504\xspace}
\newcommand{\SdUpperLooseLow}{0.466\xspace}
\newcommand{\SdUpperLooseHigh}{0.541\xspace}
\newcommand{\BoundTight}{0.1\xspace}
\newcommand{\BoundLoose}{0.5\xspace}
\newcommand{\SgAtFive}{0.5\xspace}
\newcommand{\SgUnitsFive}{973\xspace}
\newcommand{\SgMatchFive}{0\xspace}
\newcommand{\SgGroupFive}{0\xspace}
\newcommand{\SgUpperFive}{2\xspace}
\newcommand{\SgEmptyFive}{944\xspace}
\newcommand{\SgAtSix}{0.6\xspace}
\newcommand{\SgUnitsSix}{1,112\xspace}
\newcommand{\SgMatchSix}{0\xspace}
\newcommand{\SgGroupSix}{0\xspace}
\newcommand{\SgUpperSix}{2\xspace}
\newcommand{\SgEmptySix}{1,047\xspace}
\newcommand{\SgAtSeven}{0.7\xspace}
\newcommand{\SgUnitsSeven}{1,259\xspace}
\newcommand{\SgMatchSeven}{3\xspace}
\newcommand{\SgGroupSeven}{4\xspace}
\newcommand{\SgUpperSeven}{12\xspace}
\newcommand{\SgEmptySeven}{1,117\xspace}
\newcommand{\SgAtEight}{0.8\xspace}
\newcommand{\SgUnitsEight}{1,415\xspace}
\newcommand{\SgMatchEight}{13\xspace}
\newcommand{\SgGroupEight}{16\xspace}
\newcommand{\SgUpperEight}{43\xspace}
\newcommand{\SgEmptyEight}{1,137\xspace}
\newcommand{\SgAtNine}{0.9\xspace}
\newcommand{\SgUnitsNine}{1,658\xspace}
\newcommand{\SgMatchNine}{149\xspace}
\newcommand{\SgGroupNine}{153\xspace}
\newcommand{\SgUpperNine}{205\xspace}
\newcommand{\SgEmptyNine}{1,139\xspace}
\newcommand{\SkFvuTrMeanA}{0.105\xspace}
\newcommand{\SkFvuTrMinA}{0.104\xspace}
\newcommand{\SkFvuTrMaxA}{0.105\xspace}
\newcommand{\SkLzeroTrMeanA}{552\xspace}
\newcommand{\SkLzeroTrMinA}{552\xspace}
\newcommand{\SkLzeroTrMaxA}{554\xspace}
\newcommand{\SkFvuFrMeanA}{0.185\xspace}
\newcommand{\SkFvuFrMinA}{0.185\xspace}
\newcommand{\SkFvuFrMaxA}{0.185\xspace}
\newcommand{\SkLzeroFrMeanA}{1,177\xspace}
\newcommand{\SkLzeroFrMinA}{1,177\xspace}
\newcommand{\SkLzeroFrMaxA}{1,178\xspace}
\newcommand{\ShFirstPosA}{3\xspace}
\newcommand{\ShFirstNegA}{60\xspace}
\newcommand{\ShSecondPosA}{3\xspace}
\newcommand{\ShSecondNegA}{50\xspace}
\newcommand{\SvAOneSize}{12\xspace}
\newcommand{\SvAOneBelow}{200\xspace}
\newcommand{\SvAOneCarried}{0.00614\xspace}
\newcommand{\SvAOneCosine}{0.00553\xspace}
\newcommand{\SvATwoSize}{12\xspace}
\newcommand{\SvATwoBelow}{200\xspace}
\newcommand{\SvATwoCarried}{0.00634\xspace}
\newcommand{\SvATwoCosine}{0.00634\xspace}
\newcommand{\SvAThreeSize}{10\xspace}
\newcommand{\SvAThreeBelow}{200\xspace}
\newcommand{\SvAThreeCarried}{0.00461\xspace}
\newcommand{\SvAThreeCosine}{0.00439\xspace}
\newcommand{\SvAFourSize}{9\xspace}
\newcommand{\SvAFourBelow}{200\xspace}
\newcommand{\SvAFourCarried}{0.00354\xspace}
\newcommand{\SvAFourCosine}{0.00339\xspace}
\newcommand{\SvAFiveSize}{7\xspace}
\newcommand{\SvAFiveBelow}{200\xspace}
\newcommand{\SvAFiveCarried}{0.00401\xspace}
\newcommand{\SvAFiveCosine}{0.00346\xspace}

\newcommand{\SkFvuTrMeanB}{0.096\xspace}
\newcommand{\SkFvuTrMinB}{0.096\xspace}
\newcommand{\SkFvuTrMaxB}{0.096\xspace}
\newcommand{\SkLzeroTrMeanB}{223\xspace}
\newcommand{\SkLzeroTrMinB}{221\xspace}
\newcommand{\SkLzeroTrMaxB}{225\xspace}
\newcommand{\SkFvuFrMeanB}{0.140\xspace}
\newcommand{\SkFvuFrMinB}{0.140\xspace}
\newcommand{\SkFvuFrMaxB}{0.140\xspace}
\newcommand{\SkLzeroFrMeanB}{929\xspace}
\newcommand{\SkLzeroFrMinB}{928\xspace}
\newcommand{\SkLzeroFrMaxB}{930\xspace}
\newcommand{\SkFvuSuMeanB}{0.106\xspace}
\newcommand{\SkFvuSuMinB}{0.106\xspace}
\newcommand{\SkFvuSuMaxB}{0.106\xspace}
\newcommand{\SkLzeroSuMeanB}{819\xspace}
\newcommand{\SkLzeroSuMinB}{817\xspace}
\newcommand{\SkLzeroSuMaxB}{822\xspace}
\newcommand{\ShFirstPosB}{5\xspace}
\newcommand{\ShFirstNegB}{61\xspace}
\newcommand{\ShSecondPosB}{4\xspace}
\newcommand{\ShSecondNegB}{65\xspace}
\newcommand{\SvBOneSize}{44\xspace}
\newcommand{\SvBOneBelow}{190\xspace}
\newcommand{\SvBOneCarried}{0.00739\xspace}
\newcommand{\SvBOneCosine}{0.00741\xspace}
\newcommand{\SvBTwoSize}{45\xspace}
\newcommand{\SvBTwoBelow}{200\xspace}
\newcommand{\SvBTwoCarried}{0.00755\xspace}
\newcommand{\SvBTwoCosine}{0.00755\xspace}
\newcommand{\SvBThreeSize}{41\xspace}
\newcommand{\SvBThreeBelow}{200\xspace}
\newcommand{\SvBThreeCarried}{0.00929\xspace}
\newcommand{\SvBThreeCosine}{0.00879\xspace}
\newcommand{\SvBFourSize}{45\xspace}
\newcommand{\SvBFourBelow}{200\xspace}
\newcommand{\SvBFourCarried}{0.00897\xspace}
\newcommand{\SvBFourCosine}{0.00894\xspace}
\newcommand{\SvBFiveSize}{36\xspace}
\newcommand{\SvBFiveBelow}{200\xspace}
\newcommand{\SvBFiveCarried}{0.00638\xspace}
\newcommand{\SvBFiveCosine}{0.00607\xspace}

\newcommand{\SkFvuTrMeanC}{0.045\xspace}
\newcommand{\SkFvuTrMinC}{0.045\xspace}
\newcommand{\SkFvuTrMaxC}{0.045\xspace}
\newcommand{\SkLzeroTrMeanC}{76\xspace}
\newcommand{\SkLzeroTrMinC}{76\xspace}
\newcommand{\SkLzeroTrMaxC}{77\xspace}
\newcommand{\SkFvuFrMeanC}{0.107\xspace}
\newcommand{\SkFvuFrMinC}{0.107\xspace}
\newcommand{\SkFvuFrMaxC}{0.108\xspace}
\newcommand{\SkLzeroFrMeanC}{389\xspace}
\newcommand{\SkLzeroFrMinC}{389\xspace}
\newcommand{\SkLzeroFrMaxC}{389\xspace}
\newcommand{\SkFvuSuMeanC}{0.074\xspace}
\newcommand{\SkFvuSuMinC}{0.074\xspace}
\newcommand{\SkFvuSuMaxC}{0.074\xspace}
\newcommand{\SkLzeroSuMeanC}{376\xspace}
\newcommand{\SkLzeroSuMinC}{374\xspace}
\newcommand{\SkLzeroSuMaxC}{377\xspace}
\newcommand{\ShFirstPosC}{23\xspace}
\newcommand{\ShFirstNegC}{74\xspace}
\newcommand{\ShSecondPosC}{20\xspace}
\newcommand{\ShSecondNegC}{63\xspace}
\newcommand{\SvCOneSize}{70\xspace}
\newcommand{\SvCOneBelow}{190\xspace}
\newcommand{\SvCOneCarried}{0.00566\xspace}
\newcommand{\SvCOneCosine}{0.00565\xspace}
\newcommand{\SvCTwoSize}{79\xspace}
\newcommand{\SvCTwoBelow}{50\xspace}
\newcommand{\SvCTwoCarried}{0.00653\xspace}
\newcommand{\SvCTwoCosine}{0.00640\xspace}
\newcommand{\SvCThreeSize}{75\xspace}
\newcommand{\SvCThreeBelow}{180\xspace}
\newcommand{\SvCThreeCarried}{0.00595\xspace}
\newcommand{\SvCThreeCosine}{0.00599\xspace}
\newcommand{\SvCFourSize}{73\xspace}
\newcommand{\SvCFourBelow}{28\xspace}
\newcommand{\SvCFourCarried}{0.00608\xspace}
\newcommand{\SvCFourCosine}{0.00585\xspace}
\newcommand{\SvCFiveSize}{71\xspace}
\newcommand{\SvCFiveBelow}{200\xspace}
\newcommand{\SvCFiveCarried}{0.00609\xspace}
\newcommand{\SvCFiveCosine}{0.00590\xspace}

\newcommand{\RandomBelowCSecond}{50\xspace}
\newcommand{\RandomBelowCFourth}{28\xspace}
\newcommand{\DeviceName}{NVIDIA A100-SXM4-80GB\xspace}
\newcommand{\DevicePcie}{NVIDIA A100 80GB PCIe\xspace}
\newcommand{\TorchVersion}{2.5.1+cu124\xspace}
\newcommand{\TransformersVersion}{4.46.3\xspace}

\newcommand{\DinoShort}{256\xspace}
\newcommand{\DinoCrop}{224\xspace}
\newcommand{\TowerSize}{224\xspace}
\newcommand{\TowerPatch}{16\xspace}
\newcommand{\SeedImages}{20260928\xspace}
\newcommand{\SeedDraw}{20261011\xspace}
\newcommand{\SeedGauge}{30000\xspace}
\newcommand{\SeedShuffle}{40000\xspace}
\newcommand{\SeedTrainedFirst}{1\xspace}
\newcommand{\SeedTrainedLast}{5\xspace}
\newcommand{\SeedFrozenFirst}{11\xspace}
\newcommand{\SeedFrozenLast}{15\xspace}
\newcommand{\SeedSurrogate}{20260929\xspace}
\newcommand{\SeedRandomSets}{50000\xspace}
\newcommand{\ContrastPassCountA}{1,170\xspace}
\newcommand{\ContrastA}{0.0004\xspace}

\newcommand{\ContrastLowA}{-0.0034\xspace}
\newcommand{\ContrastHighA}{0.0042\xspace}
\newcommand{\LeaveSourceA}{4\xspace}
\newcommand{\LeaveDictA}{3\xspace}
\newcommand{\EnergyFloorTimesA}{182\xspace}

\newcommand{\NameShareLowA}{637\xspace}
\newcommand{\NameFixedA}{3\xspace}
\newcommand{\NameLostA}{0\xspace}
\newcommand{\NameMedianA}{0.10\xspace}

\newcommand{\SeedLeastA}{55\xspace}
\newcommand{\SeedMostA}{76\xspace}
\newcommand{\SecondsA}{22,154\xspace}
\newcommand{\VarAboveA}{4\xspace}
\newcommand{\SaAOne}{2\xspace}
\newcommand{\SaATwo}{0\xspace}
\newcommand{\SaAThree}{1\xspace}
\newcommand{\SaAFour}{1\xspace}
\newcommand{\SaAFive}{1\xspace}
\newcommand{\StrongPairsA}{512\xspace}
\newcommand{\ContrastPassCountB}{297\xspace}
\newcommand{\ContrastB}{-0.0028\xspace}

\newcommand{\ContrastLowB}{-0.0082\xspace}
\newcommand{\ContrastHighB}{0.0023\xspace}
\newcommand{\LeaveSourceB}{0\xspace}
\newcommand{\LeaveDictB}{0\xspace}
\newcommand{\EnergyFloorTimesB}{362\xspace}

\newcommand{\NameShareLowB}{622\xspace}
\newcommand{\NameFixedB}{18\xspace}
\newcommand{\NameLostB}{7\xspace}
\newcommand{\NameMedianB}{0.08\xspace}

\newcommand{\SeedLeastB}{188\xspace}
\newcommand{\SeedMostB}{234\xspace}
\newcommand{\SecondsB}{10,855\xspace}
\newcommand{\VarAboveB}{3\xspace}
\newcommand{\SaBOne}{1\xspace}
\newcommand{\SaBTwo}{0\xspace}
\newcommand{\SaBThree}{1\xspace}
\newcommand{\SaBFour}{1\xspace}
\newcommand{\SaBFive}{1\xspace}
\newcommand{\StrongPairsB}{1,338\xspace}
\newcommand{\ContrastPassCountC}{909\xspace}
\newcommand{\ContrastC}{-0.0004\xspace}

\newcommand{\ContrastLowC}{-0.0064\xspace}
\newcommand{\ContrastHighC}{0.0055\xspace}
\newcommand{\LeaveSourceC}{2\xspace}
\newcommand{\LeaveDictC}{2\xspace}
\newcommand{\EnergyFloorTimesC}{615\xspace}

\newcommand{\NameShareLowC}{575\xspace}
\newcommand{\NameFixedC}{65\xspace}
\newcommand{\NameLostC}{15\xspace}
\newcommand{\NameMedianC}{0.09\xspace}

\newcommand{\SeedLeastC}{346\xspace}
\newcommand{\SeedMostC}{385\xspace}
\newcommand{\SecondsC}{9,435\xspace}
\newcommand{\VarAboveC}{4\xspace}
\newcommand{\SaCOne}{2\xspace}
\newcommand{\SaCTwo}{3\xspace}
\newcommand{\SaCThree}{2\xspace}
\newcommand{\SaCFour}{1\xspace}
\newcommand{\SaCFive}{2\xspace}
\newcommand{\StrongPairsC}{2,016\xspace}

\newcommand{\ContrastAnyCount}{1,632\xspace}
\newcommand{\DinoCheckpoint}{facebook/dinov2-base\xspace}
\newcommand{\TowerCheckpoint}{google/siglip-base-patch16-224\xspace}
\newcommand{\SeedPairs}{512\xspace}
\newcommand{\SecondsDev}{14,358\xspace}
\newcommand{\AtomsApartMost}{3\xspace}
\newcommand{\ExHalfA}{28\xspace}
\newcommand{\ExCosA}{44\xspace}
\newcommand{\ExCosLowA}{0\xspace}
\newcommand{\ExCosHighA}{32\xspace}
\newcommand{\ExCosLeaveA}{5\xspace}
\newcommand{\BdTightA}{0.075\xspace}
\newcommand{\BdLooseA}{0.163\xspace}
\newcommand{\ClassGroupA}{304\xspace}
\newcommand{\ClassUpperA}{345\xspace}
\newcommand{\ExHalfB}{41\xspace}
\newcommand{\ExCosB}{47\xspace}
\newcommand{\ExCosLowB}{-16\xspace}
\newcommand{\ExCosHighB}{28\xspace}
\newcommand{\ExCosLeaveB}{5\xspace}
\newcommand{\BdTightB}{0.337\xspace}
\newcommand{\BdLooseB}{0.463\xspace}
\newcommand{\ClassGroupB}{1,058\xspace}
\newcommand{\ClassUpperB}{1,104\xspace}
\newcommand{\ExHalfC}{46\xspace}
\newcommand{\ExCosC}{67\xspace}
\newcommand{\ExCosLowC}{-3\xspace}
\newcommand{\ExCosHighC}{47\xspace}
\newcommand{\ExCosLeaveC}{5\xspace}
\newcommand{\BdTightC}{0.611\xspace}
\newcommand{\BdLooseC}{0.749\xspace}
\newcommand{\ClassGroupC}{1,795\xspace}
\newcommand{\ClassUpperC}{1,852\xspace}
\newcommand{\DevClassGroup}{1,042\xspace}
\newcommand{\DevClassUpper}{1,103\xspace}
\newcommand{\SecondsNone}{3\xspace}
\newcommand{\SeedSurrogateFirst}{1\xspace}
\newcommand{\SeedSurrogateLast}{5\xspace}
\newcommand{\SeedBoot}{20261012\xspace}
\newcommand{\SeedContrast}{20261030\xspace}
\newcommand{\SeedDepartures}{20261040\xspace}
\newcommand{\DinoMeanOne}{0.485\xspace}
\newcommand{\DinoStdOne}{0.229\xspace}
\newcommand{\DinoMeanTwo}{0.456\xspace}
\newcommand{\DinoStdTwo}{0.224\xspace}
\newcommand{\DinoMeanThree}{0.406\xspace}
\newcommand{\DinoStdThree}{0.225\xspace}
\newcommand{\TowerMean}{0.5\xspace}
\newcommand{\TowerStd}{0.5\xspace}
\newcommand{\Rescale}{255\xspace}
\newcommand{\SameSiteA}{14\xspace}
\newcommand{\SameBothA}{13\xspace}
\newcommand{\FrozenSiteA}{0\xspace}
\newcommand{\SameSiteLowA}{-26\xspace}
\newcommand{\SameSiteHighA}{-3\xspace}
\newcommand{\SameSiteLeaveA}{0\xspace}
\newcommand{\ImageContrastA}{37\xspace}
\newcommand{\SameRhoA}{0.869\xspace}
\newcommand{\SameDisagreeA}{28\xspace}
\newcommand{\SameHalvesA}{0.9881\xspace}
\newcommand{\MovesSelA}{6\xspace}
\newcommand{\MovesHalfA}{11\xspace}

\newcommand{\SelRhoA}{0.863\xspace}
\newcommand{\SameSiteB}{4\xspace}
\newcommand{\SameBothB}{2\xspace}
\newcommand{\FrozenSiteB}{0\xspace}
\newcommand{\SameSiteLowB}{-54\xspace}
\newcommand{\SameSiteHighB}{-22\xspace}
\newcommand{\SameSiteLeaveB}{0\xspace}
\newcommand{\ImageContrastB}{3\xspace}
\newcommand{\SameRhoB}{0.995\xspace}
\newcommand{\SameDisagreeB}{19\xspace}
\newcommand{\SameHalvesB}{0.9932\xspace}
\newcommand{\MovesSelB}{16\xspace}
\newcommand{\MovesHalfB}{38\xspace}

\newcommand{\SelRhoB}{0.983\xspace}
\newcommand{\SameSiteC}{7\xspace}
\newcommand{\SameBothC}{5\xspace}
\newcommand{\FrozenSiteC}{0\xspace}
\newcommand{\SameSiteLowC}{-56\xspace}
\newcommand{\SameSiteHighC}{-23\xspace}
\newcommand{\SameSiteLeaveC}{0\xspace}
\newcommand{\ImageContrastC}{11\xspace}
\newcommand{\SameRhoC}{0.991\xspace}
\newcommand{\SameDisagreeC}{16\xspace}
\newcommand{\SameHalvesC}{0.9908\xspace}
\newcommand{\MovesSelC}{28\xspace}
\newcommand{\MovesHalfC}{35\xspace}

\newcommand{\SelRhoC}{0.984\xspace}
\newcommand{\SeedSame}{20261050\xspace}
\newcommand{\SeedImageDraws}{20261060\xspace}
\newcommand{\SitePassCalls}{11\xspace}
\newcommand{\SitePassSeconds}{4,337\xspace}
\newcommand{\FreshImages}{500\xspace}
\newcommand{\FreshHalf}{250\xspace}
\newcommand{\FreshList}{118,287\xspace}
\newcommand{\SeedFreshSet}{20261101\xspace}
\newcommand{\SeedFreshDraws}{20261070\xspace}
\newcommand{\SeedAfterDraws}{20261080\xspace}
\newcommand{\FreshPassA}{316\xspace}
\newcommand{\FreshFourA}{2,500\xspace}
\newcommand{\FreshFourShareA}{0.977\xspace}

\newcommand{\FreshSelfA}{0.9867\xspace}
\newcommand{\FreshRuleA}{0.9742\xspace}
\newcommand{\FreshDiffA}{0.0125\xspace}

\newcommand{\FreshBoundA}{0.0072\xspace}
\newcommand{\FreshDiffLowA}{0.0064\xspace}
\newcommand{\FreshDiffHighA}{0.0193\xspace}

\newcommand{\FreshImageAboveA}{497\xspace}
\newcommand{\FreshSiteErrA}{42\xspace}
\newcommand{\FreshCosErrA}{65\xspace}
\newcommand{\FreshSplitA}{34\xspace}
\newcommand{\FreshSiteRepeatA}{17\xspace}
\newcommand{\FreshCosRepeatA}{49\xspace}

\newcommand{\FreshFrozenA}{0\xspace}
\newcommand{\FreshPassB}{1,045\xspace}
\newcommand{\FreshFourB}{2,493\xspace}
\newcommand{\FreshFourShareB}{0.974\xspace}

\newcommand{\FreshSelfB}{0.9871\xspace}
\newcommand{\FreshRuleB}{0.9771\xspace}
\newcommand{\FreshDiffB}{0.0100\xspace}

\newcommand{\FreshBoundB}{0.0049\xspace}
\newcommand{\FreshDiffLowB}{0.0041\xspace}
\newcommand{\FreshDiffHighB}{0.0164\xspace}
\newcommand{\FreshImageAboveB}{1\xspace}
\newcommand{\FreshSiteErrB}{27\xspace}
\newcommand{\FreshCosErrB}{56\xspace}
\newcommand{\FreshSplitB}{33\xspace}
\newcommand{\FreshSiteRepeatB}{4\xspace}
\newcommand{\FreshCosRepeatB}{42\xspace}

\newcommand{\FreshFrozenB}{0\xspace}
\newcommand{\SameHalvesLowA}{0.9830\xspace}
\newcommand{\SameHalvesHighA}{0.9924\xspace}
\newcommand{\BdTightLowA}{0.057\xspace}
\newcommand{\BdTightHighA}{0.094\xspace}
\newcommand{\BdLooseLowA}{0.137\xspace}
\newcommand{\BdLooseHighA}{0.189\xspace}
\newcommand{\ErrDiffA}{23\xspace}
\newcommand{\ErrDiffLowA}{10\xspace}
\newcommand{\ErrDiffHighA}{36\xspace}
\newcommand{\ImgSelfMeanA}{0.9829\xspace}
\newcommand{\ImgRuleMeanA}{0.9760\xspace}
\newcommand{\ImgSplitMeanA}{44\xspace}
\newcommand{\SameHalvesLowB}{0.9900\xspace}
\newcommand{\SameHalvesHighB}{0.9957\xspace}
\newcommand{\BdTightLowB}{0.301\xspace}
\newcommand{\BdTightHighB}{0.371\xspace}
\newcommand{\BdLooseLowB}{0.427\xspace}
\newcommand{\BdLooseHighB}{0.497\xspace}
\newcommand{\ErrDiffB}{43\xspace}
\newcommand{\ErrDiffLowB}{25\xspace}
\newcommand{\ErrDiffHighB}{63\xspace}

\newcommand{\ImgSplitMeanB}{52\xspace}
\newcommand{\SameHalvesLowC}{0.9871\xspace}
\newcommand{\SameHalvesHighC}{0.9941\xspace}
\newcommand{\BdTightLowC}{0.577\xspace}
\newcommand{\BdTightHighC}{0.644\xspace}
\newcommand{\BdLooseLowC}{0.719\xspace}
\newcommand{\BdLooseHighC}{0.778\xspace}
\newcommand{\ErrDiffC}{52\xspace}
\newcommand{\ErrDiffLowC}{33\xspace}
\newcommand{\ErrDiffHighC}{73\xspace}

\newcommand{\ImgSplitMeanC}{62\xspace}

\newcommand{\TimeSitePairA}{0.010\xspace}
\newcommand{\TimeSiteImageA}{0.008\xspace}
\newcommand{\TimeFullPairA}{1.081\xspace}

\newcommand{\TimeSitePairB}{0.008\xspace}
\newcommand{\TimeSiteImageB}{0.008\xspace}
\newcommand{\TimeFullPairB}{0.136\xspace}

\newcommand{\FreshCalls}{14\xspace}
\newcommand{\FreshSeconds}{6,720\xspace}
\newcommand{\PlantedImages}{100\xspace}
\newcommand{\PillowVersion}{11.0.0\xspace}
\newcommand{\NumpyVersion}{2.1.3\xspace}
\newcommand{\ScipyVersion}{1.18.0\xspace}
\newcommand{\SiteHalvesA}{0.9859\xspace}
\newcommand{\SiteHalvesLowA}{0.9805\xspace}
\newcommand{\SiteHalvesHighA}{0.9906\xspace}
\newcommand{\SiteDisagreeA}{34\xspace}
\newcommand{\ExSiteA}{22\xspace}
\newcommand{\ExBothA}{19\xspace}
\newcommand{\ExSiteLowA}{-19\xspace}
\newcommand{\ExSiteHighA}{6\xspace}
\newcommand{\ExSiteLeaveA}{1\xspace}
\newcommand{\ScreenSentA}{37\xspace}
\newcommand{\ScreenWrongA}{27\xspace}
\newcommand{\SiteHalvesB}{0.9881\xspace}
\newcommand{\SiteHalvesLowB}{0.9840\xspace}
\newcommand{\SiteHalvesHighB}{0.9918\xspace}
\newcommand{\SiteDisagreeB}{23\xspace}
\newcommand{\ExSiteB}{10\xspace}
\newcommand{\ExBothB}{5\xspace}
\newcommand{\ExSiteLowB}{-48\xspace}
\newcommand{\ExSiteHighB}{-15\xspace}
\newcommand{\ExSiteLeaveB}{0\xspace}
\newcommand{\ScreenSentB}{63\xspace}
\newcommand{\ScreenWrongB}{13\xspace}
\newcommand{\SiteHalvesC}{0.9840\xspace}
\newcommand{\SiteHalvesLowC}{0.9789\xspace}
\newcommand{\SiteHalvesHighC}{0.9885\xspace}
\newcommand{\SiteDisagreeC}{34\xspace}
\newcommand{\ExSiteC}{18\xspace}
\newcommand{\ExBothC}{11\xspace}
\newcommand{\ExSiteLowC}{-46\xspace}
\newcommand{\ExSiteHighC}{-11\xspace}
\newcommand{\ExSiteLeaveC}{0\xspace}
\newcommand{\ScreenSentC}{68\xspace}
\newcommand{\ScreenWrongC}{26\xspace}
\newcommand{\FreshFnMarginA}{0.0129\xspace}
\newcommand{\FreshFnMarginB}{0.0114\xspace}
\newcommand{\FreshFnMarginC}{0.0160\xspace}
\newcommand{\FreshPassC}{1,783\xspace}
\newcommand{\FreshFourC}{2,473\xspace}
\newcommand{\FreshFourShareC}{0.966\xspace}

\newcommand{\FreshSelfC}{0.9805\xspace}
\newcommand{\FreshRuleC}{0.9680\xspace}
\newcommand{\FreshDiffC}{0.0125\xspace}
\newcommand{\FreshBoundC}{0.0062\xspace}
\newcommand{\FreshDiffLowC}{0.0051\xspace}
\newcommand{\FreshDiffHighC}{0.0205\xspace}
\newcommand{\FreshImageAboveC}{16\xspace}
\newcommand{\FreshSiteErrC}{27\xspace}
\newcommand{\FreshCosErrC}{81\xspace}
\newcommand{\FreshSplitC}{50\xspace}
\newcommand{\FreshSiteRepeatC}{5\xspace}
\newcommand{\FreshCosRepeatC}{57\xspace}

\newcommand{\FreshFrozenC}{0\xspace}
\newcommand{\FifthCalls}{12\xspace}
\newcommand{\FifthSeconds}{1,526\xspace}

\newcommand{\SixthCalls}{2\xspace}
\newcommand{\SixthSeconds}{288.3\xspace}
\newcommand{\EightCount}{8\xspace}

\newcommand{\EightOneSource}{1\xspace}
\newcommand{\EightOneFeature}{1502\xspace}
\newcommand{\EightOneTarget}{5\xspace}

\newcommand{\EightOneSiteSelection}{0.264\xspace}

\newcommand{\EightOneSiteOffsetSelection}{0.014\xspace}

\newcommand{\EightTwoSource}{3\xspace}
\newcommand{\EightTwoFeature}{1443\xspace}
\newcommand{\EightTwoTarget}{1\xspace}

\newcommand{\EightTwoSiteSelection}{0.332\xspace}

\newcommand{\EightTwoSiteOffsetSelection}{0.082\xspace}

\newcommand{\EightThreeSource}{3\xspace}
\newcommand{\EightThreeFeature}{1443\xspace}
\newcommand{\EightThreeTarget}{4\xspace}

\newcommand{\EightThreeSiteSelection}{0.295\xspace}

\newcommand{\EightThreeSiteOffsetSelection}{0.045\xspace}

\newcommand{\EightFourSource}{4\xspace}
\newcommand{\EightFourFeature}{27\xspace}
\newcommand{\EightFourTarget}{5\xspace}

\newcommand{\EightFourSiteSelection}{0.311\xspace}

\newcommand{\EightFourSiteOffsetSelection}{0.061\xspace}

\newcommand{\EightFiveSource}{5\xspace}
\newcommand{\EightFiveFeature}{5381\xspace}
\newcommand{\EightFiveTarget}{1\xspace}

\newcommand{\EightFiveSiteSelection}{0.268\xspace}

\newcommand{\EightFiveSiteOffsetSelection}{0.018\xspace}

\newcommand{\EightSixSource}{3\xspace}
\newcommand{\EightSixFeature}{1124\xspace}
\newcommand{\EightSixTarget}{2\xspace}

\newcommand{\EightSixSiteSelection}{0.220\xspace}

\newcommand{\EightSixSiteOffsetSelection}{-0.030\xspace}

\newcommand{\EightSevenSource}{2\xspace}
\newcommand{\EightSevenFeature}{387\xspace}
\newcommand{\EightSevenTarget}{4\xspace}

\newcommand{\EightSevenSiteSelection}{0.173\xspace}

\newcommand{\EightSevenSiteOffsetSelection}{-0.077\xspace}

\newcommand{\EightEightSource}{5\xspace}
\newcommand{\EightEightFeature}{442\xspace}
\newcommand{\EightEightTarget}{3\xspace}

\newcommand{\EightEightSiteSelection}{0.280\xspace}

\newcommand{\EightEightSiteOffsetSelection}{0.030\xspace}

\newcommand{\FreshVarianceDifferenceA}{0.000227\xspace}
\newcommand{\FreshVarianceAOneSize}{12\xspace}
\newcommand{\FreshVarianceAOnePassing}{0.006052\xspace}
\newcommand{\FreshVarianceAOneCosine}{0.005820\xspace}
\newcommand{\FreshVarianceAOneDifference}{0.000232\xspace}
\newcommand{\FreshVarianceAOneBelow}{200\xspace}

\newcommand{\FreshVarianceATwoSize}{12\xspace}
\newcommand{\FreshVarianceATwoPassing}{0.006359\xspace}
\newcommand{\FreshVarianceATwoCosine}{0.006359\xspace}
\newcommand{\FreshVarianceATwoDifference}{0.000000\xspace}
\newcommand{\FreshVarianceATwoBelow}{200\xspace}

\newcommand{\FreshVarianceAThreeSize}{10\xspace}
\newcommand{\FreshVarianceAThreePassing}{0.004562\xspace}
\newcommand{\FreshVarianceAThreeCosine}{0.004330\xspace}
\newcommand{\FreshVarianceAThreeDifference}{0.000231\xspace}
\newcommand{\FreshVarianceAThreeBelow}{200\xspace}

\newcommand{\FreshVarianceAFourSize}{9\xspace}
\newcommand{\FreshVarianceAFourPassing}{0.003369\xspace}
\newcommand{\FreshVarianceAFourCosine}{0.003246\xspace}
\newcommand{\FreshVarianceAFourDifference}{0.000122\xspace}
\newcommand{\FreshVarianceAFourBelow}{200\xspace}

\newcommand{\FreshVarianceAFiveSize}{7\xspace}
\newcommand{\FreshVarianceAFivePassing}{0.004016\xspace}
\newcommand{\FreshVarianceAFiveCosine}{0.003466\xspace}
\newcommand{\FreshVarianceAFiveDifference}{0.000550\xspace}
\newcommand{\FreshVarianceAFiveBelow}{200\xspace}

\newcommand{\CopyGapA}{0.1178\xspace}
\newcommand{\ReplayMaximumA}{$1.8\times 10^{-12}$\xspace}
\newcommand{\ReplayWholeA}{$4.4\times 10^{-16}$\xspace}
\newcommand{\ReplayUndefinedA}{56\xspace}
\newcommand{\FreshVarianceDifferenceB}{0.000424\xspace}
\newcommand{\FreshVarianceBOneSize}{44\xspace}
\newcommand{\FreshVarianceBOnePassing}{0.007300\xspace}
\newcommand{\FreshVarianceBOneCosine}{0.007309\xspace}
\newcommand{\FreshVarianceBOneDifference}{-0.000009\xspace}
\newcommand{\FreshVarianceBOneBelow}{178\xspace}

\newcommand{\FreshVarianceBTwoSize}{45\xspace}
\newcommand{\FreshVarianceBTwoPassing}{0.006578\xspace}
\newcommand{\FreshVarianceBTwoCosine}{0.006578\xspace}
\newcommand{\FreshVarianceBTwoDifference}{0.000000\xspace}
\newcommand{\FreshVarianceBTwoBelow}{200\xspace}

\newcommand{\FreshVarianceBThreeSize}{41\xspace}
\newcommand{\FreshVarianceBThreePassing}{0.008981\xspace}
\newcommand{\FreshVarianceBThreeCosine}{0.008210\xspace}
\newcommand{\FreshVarianceBThreeDifference}{0.000770\xspace}
\newcommand{\FreshVarianceBThreeBelow}{200\xspace}

\newcommand{\FreshVarianceBFourSize}{45\xspace}
\newcommand{\FreshVarianceBFourPassing}{0.009603\xspace}
\newcommand{\FreshVarianceBFourCosine}{0.009564\xspace}
\newcommand{\FreshVarianceBFourDifference}{0.000039\xspace}
\newcommand{\FreshVarianceBFourBelow}{200\xspace}

\newcommand{\FreshVarianceBFiveSize}{36\xspace}
\newcommand{\FreshVarianceBFivePassing}{0.006336\xspace}
\newcommand{\FreshVarianceBFiveCosine}{0.005016\xspace}
\newcommand{\FreshVarianceBFiveDifference}{0.001320\xspace}
\newcommand{\FreshVarianceBFiveBelow}{191\xspace}

\newcommand{\CopyGapB}{0.0248\xspace}
\newcommand{\ReplayMaximumB}{$5.8\times 10^{-11}$\xspace}
\newcommand{\ReplayWholeB}{$4.4\times 10^{-16}$\xspace}
\newcommand{\ReplayUndefinedB}{1,372\xspace}
\newcommand{\FreshVarianceDifferenceC}{0.000070\xspace}
\newcommand{\FreshVarianceCOneSize}{70\xspace}
\newcommand{\FreshVarianceCOnePassing}{0.005784\xspace}
\newcommand{\FreshVarianceCOneCosine}{0.005794\xspace}
\newcommand{\FreshVarianceCOneDifference}{-0.000010\xspace}
\newcommand{\FreshVarianceCOneBelow}{192\xspace}

\newcommand{\FreshVarianceCTwoSize}{79\xspace}
\newcommand{\FreshVarianceCTwoPassing}{0.006550\xspace}
\newcommand{\FreshVarianceCTwoCosine}{0.006431\xspace}
\newcommand{\FreshVarianceCTwoDifference}{0.000119\xspace}
\newcommand{\FreshVarianceCTwoBelow}{42\xspace}

\newcommand{\FreshVarianceCThreeSize}{75\xspace}
\newcommand{\FreshVarianceCThreePassing}{0.005842\xspace}
\newcommand{\FreshVarianceCThreeCosine}{0.005889\xspace}
\newcommand{\FreshVarianceCThreeDifference}{-0.000047\xspace}
\newcommand{\FreshVarianceCThreeBelow}{171\xspace}

\newcommand{\FreshVarianceCFourSize}{73\xspace}
\newcommand{\FreshVarianceCFourPassing}{0.006105\xspace}
\newcommand{\FreshVarianceCFourCosine}{0.005946\xspace}
\newcommand{\FreshVarianceCFourDifference}{0.000159\xspace}
\newcommand{\FreshVarianceCFourBelow}{32\xspace}

\newcommand{\FreshVarianceCFiveSize}{71\xspace}
\newcommand{\FreshVarianceCFivePassing}{0.005933\xspace}
\newcommand{\FreshVarianceCFiveCosine}{0.005803\xspace}
\newcommand{\FreshVarianceCFiveDifference}{0.000129\xspace}
\newcommand{\FreshVarianceCFiveBelow}{200\xspace}

\newcommand{\CopyGapC}{0.0353\xspace}
\newcommand{\ReplayMaximumC}{$2.9\times 10^{-11}$\xspace}
\newcommand{\ReplayWholeC}{$1.1\times 10^{-16}$\xspace}
\newcommand{\ReplayUndefinedC}{3,012\xspace}
\newcommand{\CopyGapDev}{0.0586\xspace}
\newcommand{\CopyDuplicates}{0\xspace}
\newcommand{\SurrogateBMatch}{0\xspace}
\newcommand{\SurrogateBGreedy}{0\xspace}
\newcommand{\SurrogateBDense}{0\xspace}
\newcommand{\ReplayTolerance}{$1.0\times 10^{-5}$\xspace}
\newcommand{\ReplayFloor}{$1.0\times 10^{-12}$\xspace}
\newcommand{\TimingRepeats}{3\xspace}
\newcommand{\TimingThreads}{4\xspace}
\newcommand{\TimingBatch}{50\xspace}
\newcommand{\TimingDevice}{NVIDIA A100 80GB PCIe\xspace}

\newcommand{\TimeCosineA}{0.375\xspace}
\newcommand{\TimeCosineLowA}{0.369\xspace}
\newcommand{\TimeCosineHighA}{0.382\xspace}
\newcommand{\TimeSiteBatchA}{3.929\xspace}
\newcommand{\TimeSiteBatchLowA}{3.900\xspace}
\newcommand{\TimeSiteBatchHighA}{3.992\xspace}
\newcommand{\TimeSiteCodesA}{0.324\xspace}
\newcommand{\TimeSiteCodesLowA}{0.323\xspace}
\newcommand{\TimeSiteCodesHighA}{0.324\xspace}
\newcommand{\TimeFullBatchA}{0.836\xspace}
\newcommand{\TimeFullBatchLowA}{0.836\xspace}
\newcommand{\TimeFullBatchHighA}{0.838\xspace}
\newcommand{\TimeRemovalsA}{34.600\xspace}
\newcommand{\TimeRemovalsLowA}{34.145\xspace}
\newcommand{\TimeRemovalsHighA}{34.746\xspace}

\newcommand{\TimeCosineB}{0.349\xspace}
\newcommand{\TimeCosineLowB}{0.348\xspace}
\newcommand{\TimeCosineHighB}{0.351\xspace}
\newcommand{\TimeSiteBatchB}{3.764\xspace}
\newcommand{\TimeSiteBatchLowB}{3.754\xspace}
\newcommand{\TimeSiteBatchHighB}{3.777\xspace}
\newcommand{\TimeSiteCodesB}{0.255\xspace}
\newcommand{\TimeSiteCodesLowB}{0.254\xspace}
\newcommand{\TimeSiteCodesHighB}{0.257\xspace}
\newcommand{\TimeFullBatchB}{0.101\xspace}
\newcommand{\TimeFullBatchLowB}{0.100\xspace}
\newcommand{\TimeFullBatchHighB}{0.103\xspace}
\newcommand{\TimeRemovalsB}{4.338\xspace}
\newcommand{\TimeRemovalsLowB}{4.320\xspace}
\newcommand{\TimeRemovalsHighB}{4.340\xspace}

\newcommand{\SeventhFamilyShare}{0.99375\xspace}
\newcommand{\SeventhThirdImages}{250\xspace}
\newcommand{\SeventhOneScoringWholeOutput}{0.148\xspace}
\newcommand{\SeventhOneScoringWholeOutputLow}{0.092\xspace}
\newcommand{\SeventhOneScoringWholeOutputHigh}{0.224\xspace}
\newcommand{\SeventhOneScoringWholeSite}{0.262\xspace}
\newcommand{\SeventhOneScoringWholeSiteLow}{0.250\xspace}
\newcommand{\SeventhOneScoringWholeSiteHigh}{0.276\xspace}
\newcommand{\SeventhOneScoringWholeOutputDistance}{-0.102\xspace}
\newcommand{\SeventhOneScoringWholeSiteDistance}{0.012\xspace}
\newcommand{\SeventhOneScoringWholeOpposite}{0.9775\xspace}
\newcommand{\SeventhOneScoringWholeFixed}{0.9775\xspace}
\newcommand{\SeventhOneScoringEarlierOutput}{0.125\xspace}
\newcommand{\SeventhOneScoringEarlierOutputLow}{0.062\xspace}
\newcommand{\SeventhOneScoringEarlierOutputHigh}{0.270\xspace}
\newcommand{\SeventhOneScoringEarlierSite}{0.272\xspace}
\newcommand{\SeventhOneScoringEarlierSiteLow}{0.254\xspace}
\newcommand{\SeventhOneScoringEarlierSiteHigh}{0.295\xspace}
\newcommand{\SeventhOneScoringEarlierOutputDistance}{-0.125\xspace}
\newcommand{\SeventhOneScoringEarlierSiteDistance}{0.022\xspace}
\newcommand{\SeventhOneScoringEarlierOpposite}{0.8995\xspace}
\newcommand{\SeventhOneScoringEarlierFixed}{0.8995\xspace}
\newcommand{\SeventhOneScoringLaterOutput}{0.175\xspace}
\newcommand{\SeventhOneScoringLaterOutputLow}{0.141\xspace}
\newcommand{\SeventhOneScoringLaterOutputHigh}{0.220\xspace}
\newcommand{\SeventhOneScoringLaterSite}{0.253\xspace}
\newcommand{\SeventhOneScoringLaterSiteLow}{0.237\xspace}
\newcommand{\SeventhOneScoringLaterSiteHigh}{0.274\xspace}
\newcommand{\SeventhOneScoringLaterOutputDistance}{-0.075\xspace}
\newcommand{\SeventhOneScoringLaterSiteDistance}{0.003\xspace}
\newcommand{\SeventhOneScoringLaterOpposite}{0.6430\xspace}
\newcommand{\SeventhOneScoringLaterFixed}{0.6430\xspace}
\newcommand{\SeventhOneFreshWholeOutput}{0.211\xspace}
\newcommand{\SeventhOneFreshWholeOutputLow}{0.184\xspace}
\newcommand{\SeventhOneFreshWholeOutputHigh}{0.244\xspace}
\newcommand{\SeventhOneFreshWholeSite}{0.264\xspace}
\newcommand{\SeventhOneFreshWholeSiteLow}{0.249\xspace}
\newcommand{\SeventhOneFreshWholeSiteHigh}{0.283\xspace}
\newcommand{\SeventhOneFreshWholeOutputDistance}{-0.039\xspace}
\newcommand{\SeventhOneFreshWholeSiteDistance}{0.014\xspace}
\newcommand{\SeventhOneFreshWholeOpposite}{0.9505\xspace}
\newcommand{\SeventhOneFreshWholeFixed}{0.9505\xspace}
\newcommand{\SeventhOneFreshEarlierOutput}{0.211\xspace}
\newcommand{\SeventhOneFreshEarlierOutputLow}{0.173\xspace}
\newcommand{\SeventhOneFreshEarlierOutputHigh}{0.257\xspace}
\newcommand{\SeventhOneFreshEarlierSite}{0.270\xspace}
\newcommand{\SeventhOneFreshEarlierSiteLow}{0.248\xspace}
\newcommand{\SeventhOneFreshEarlierSiteHigh}{0.298\xspace}
\newcommand{\SeventhOneFreshEarlierOutputDistance}{-0.039\xspace}
\newcommand{\SeventhOneFreshEarlierSiteDistance}{0.020\xspace}
\newcommand{\SeventhOneFreshEarlierOpposite}{0.9170\xspace}
\newcommand{\SeventhOneFreshEarlierFixed}{0.9170\xspace}
\newcommand{\SeventhOneFreshLaterOutput}{0.211\xspace}
\newcommand{\SeventhOneFreshLaterOutputLow}{0.177\xspace}
\newcommand{\SeventhOneFreshLaterOutputHigh}{0.258\xspace}
\newcommand{\SeventhOneFreshLaterSite}{0.258\xspace}
\newcommand{\SeventhOneFreshLaterSiteLow}{0.237\xspace}
\newcommand{\SeventhOneFreshLaterSiteHigh}{0.283\xspace}
\newcommand{\SeventhOneFreshLaterOutputDistance}{-0.039\xspace}
\newcommand{\SeventhOneFreshLaterSiteDistance}{0.008\xspace}
\newcommand{\SeventhOneFreshLaterOpposite}{0.6975\xspace}
\newcommand{\SeventhOneFreshLaterFixed}{0.6975\xspace}
\newcommand{\SeventhOneThirdWholeOutput}{0.174\xspace}
\newcommand{\SeventhOneThirdWholeOutputLow}{0.111\xspace}
\newcommand{\SeventhOneThirdWholeOutputHigh}{0.264\xspace}
\newcommand{\SeventhOneThirdWholeSite}{0.267\xspace}
\newcommand{\SeventhOneThirdWholeSiteLow}{0.252\xspace}
\newcommand{\SeventhOneThirdWholeSiteHigh}{0.284\xspace}
\newcommand{\SeventhOneThirdWholeOutputDistance}{-0.076\xspace}
\newcommand{\SeventhOneThirdWholeSiteDistance}{0.017\xspace}
\newcommand{\SeventhOneThirdWholeOpposite}{0.9165\xspace}
\newcommand{\SeventhOneThirdWholeFixed}{0.9165\xspace}
\newcommand{\SeventhOneThirdEarlierOutput}{0.131\xspace}
\newcommand{\SeventhOneThirdEarlierOutputLow}{0.069\xspace}
\newcommand{\SeventhOneThirdEarlierOutputHigh}{0.261\xspace}
\newcommand{\SeventhOneThirdEarlierSite}{0.275\xspace}
\newcommand{\SeventhOneThirdEarlierSiteLow}{0.253\xspace}
\newcommand{\SeventhOneThirdEarlierSiteHigh}{0.304\xspace}
\newcommand{\SeventhOneThirdEarlierOutputDistance}{-0.119\xspace}
\newcommand{\SeventhOneThirdEarlierSiteDistance}{0.025\xspace}
\newcommand{\SeventhOneThirdEarlierOpposite}{0.9355\xspace}
\newcommand{\SeventhOneThirdEarlierFixed}{0.9350\xspace}
\newcommand{\SeventhOneThirdLaterOutput}{0.232\xspace}
\newcommand{\SeventhOneThirdLaterOutputLow}{0.170\xspace}
\newcommand{\SeventhOneThirdLaterOutputHigh}{0.315\xspace}
\newcommand{\SeventhOneThirdLaterSite}{0.260\xspace}
\newcommand{\SeventhOneThirdLaterSiteLow}{0.240\xspace}
\newcommand{\SeventhOneThirdLaterSiteHigh}{0.284\xspace}
\newcommand{\SeventhOneThirdLaterOutputDistance}{-0.018\xspace}
\newcommand{\SeventhOneThirdLaterSiteDistance}{0.010\xspace}
\newcommand{\SeventhOneThirdLaterOpposite}{0.5890\xspace}
\newcommand{\SeventhOneThirdLaterFixed}{0.5375\xspace}
\newcommand{\SeventhTwoScoringWholeOutput}{0.222\xspace}
\newcommand{\SeventhTwoScoringWholeOutputLow}{0.198\xspace}
\newcommand{\SeventhTwoScoringWholeOutputHigh}{0.249\xspace}
\newcommand{\SeventhTwoScoringWholeSite}{0.352\xspace}
\newcommand{\SeventhTwoScoringWholeSiteLow}{0.333\xspace}
\newcommand{\SeventhTwoScoringWholeSiteHigh}{0.373\xspace}
\newcommand{\SeventhTwoScoringWholeOutputDistance}{-0.028\xspace}
\newcommand{\SeventhTwoScoringWholeSiteDistance}{0.102\xspace}
\newcommand{\SeventhTwoScoringWholeOpposite}{0.9795\xspace}
\newcommand{\SeventhTwoScoringWholeFixed}{0.9795\xspace}
\newcommand{\SeventhTwoScoringEarlierOutput}{0.210\xspace}
\newcommand{\SeventhTwoScoringEarlierOutputLow}{0.178\xspace}
\newcommand{\SeventhTwoScoringEarlierOutputHigh}{0.253\xspace}
\newcommand{\SeventhTwoScoringEarlierSite}{0.347\xspace}
\newcommand{\SeventhTwoScoringEarlierSiteLow}{0.320\xspace}
\newcommand{\SeventhTwoScoringEarlierSiteHigh}{0.375\xspace}
\newcommand{\SeventhTwoScoringEarlierOutputDistance}{-0.040\xspace}
\newcommand{\SeventhTwoScoringEarlierSiteDistance}{0.097\xspace}
\newcommand{\SeventhTwoScoringEarlierOpposite}{0.9700\xspace}
\newcommand{\SeventhTwoScoringEarlierFixed}{0.9700\xspace}
\newcommand{\SeventhTwoScoringLaterOutput}{0.237\xspace}
\newcommand{\SeventhTwoScoringLaterOutputLow}{0.205\xspace}
\newcommand{\SeventhTwoScoringLaterOutputHigh}{0.276\xspace}
\newcommand{\SeventhTwoScoringLaterSite}{0.359\xspace}
\newcommand{\SeventhTwoScoringLaterSiteLow}{0.332\xspace}
\newcommand{\SeventhTwoScoringLaterSiteHigh}{0.390\xspace}
\newcommand{\SeventhTwoScoringLaterOutputDistance}{-0.013\xspace}
\newcommand{\SeventhTwoScoringLaterSiteDistance}{0.109\xspace}
\newcommand{\SeventhTwoScoringLaterOpposite}{0.7765\xspace}
\newcommand{\SeventhTwoScoringLaterFixed}{0.7765\xspace}
\newcommand{\SeventhTwoFreshWholeOutput}{0.201\xspace}
\newcommand{\SeventhTwoFreshWholeOutputLow}{0.178\xspace}
\newcommand{\SeventhTwoFreshWholeOutputHigh}{0.231\xspace}
\newcommand{\SeventhTwoFreshWholeSite}{0.317\xspace}
\newcommand{\SeventhTwoFreshWholeSiteLow}{0.299\xspace}
\newcommand{\SeventhTwoFreshWholeSiteHigh}{0.337\xspace}
\newcommand{\SeventhTwoFreshWholeOutputDistance}{-0.049\xspace}
\newcommand{\SeventhTwoFreshWholeSiteDistance}{0.067\xspace}
\newcommand{\SeventhTwoFreshWholeOpposite}{1.0000\xspace}
\newcommand{\SeventhTwoFreshWholeFixed}{1.0000\xspace}
\newcommand{\SeventhTwoFreshEarlierOutput}{0.188\xspace}
\newcommand{\SeventhTwoFreshEarlierOutputLow}{0.159\xspace}
\newcommand{\SeventhTwoFreshEarlierOutputHigh}{0.227\xspace}
\newcommand{\SeventhTwoFreshEarlierSite}{0.304\xspace}
\newcommand{\SeventhTwoFreshEarlierSiteLow}{0.281\xspace}
\newcommand{\SeventhTwoFreshEarlierSiteHigh}{0.333\xspace}
\newcommand{\SeventhTwoFreshEarlierOutputDistance}{-0.062\xspace}
\newcommand{\SeventhTwoFreshEarlierSiteDistance}{0.054\xspace}
\newcommand{\SeventhTwoFreshEarlierOpposite}{0.9995\xspace}
\newcommand{\SeventhTwoFreshEarlierFixed}{0.9995\xspace}
\newcommand{\SeventhTwoFreshLaterOutput}{0.216\xspace}
\newcommand{\SeventhTwoFreshLaterOutputLow}{0.181\xspace}
\newcommand{\SeventhTwoFreshLaterOutputHigh}{0.257\xspace}
\newcommand{\SeventhTwoFreshLaterSite}{0.331\xspace}
\newcommand{\SeventhTwoFreshLaterSiteLow}{0.306\xspace}
\newcommand{\SeventhTwoFreshLaterSiteHigh}{0.360\xspace}
\newcommand{\SeventhTwoFreshLaterOutputDistance}{-0.034\xspace}
\newcommand{\SeventhTwoFreshLaterSiteDistance}{0.081\xspace}
\newcommand{\SeventhTwoFreshLaterOpposite}{0.9480\xspace}
\newcommand{\SeventhTwoFreshLaterFixed}{0.9480\xspace}
\newcommand{\SeventhTwoThirdWholeOutput}{0.238\xspace}
\newcommand{\SeventhTwoThirdWholeOutputLow}{0.201\xspace}
\newcommand{\SeventhTwoThirdWholeOutputHigh}{0.284\xspace}
\newcommand{\SeventhTwoThirdWholeSite}{0.327\xspace}
\newcommand{\SeventhTwoThirdWholeSiteLow}{0.305\xspace}
\newcommand{\SeventhTwoThirdWholeSiteHigh}{0.350\xspace}
\newcommand{\SeventhTwoThirdWholeOutputDistance}{-0.012\xspace}
\newcommand{\SeventhTwoThirdWholeSiteDistance}{0.077\xspace}
\newcommand{\SeventhTwoThirdWholeOpposite}{0.7370\xspace}
\newcommand{\SeventhTwoThirdWholeFixed}{0.7370\xspace}
\newcommand{\SeventhTwoThirdEarlierOutput}{0.254\xspace}
\newcommand{\SeventhTwoThirdEarlierOutputLow}{0.205\xspace}
\newcommand{\SeventhTwoThirdEarlierOutputHigh}{0.321\xspace}
\newcommand{\SeventhTwoThirdEarlierSite}{0.331\xspace}
\newcommand{\SeventhTwoThirdEarlierSiteLow}{0.311\xspace}
\newcommand{\SeventhTwoThirdEarlierSiteHigh}{0.357\xspace}
\newcommand{\SeventhTwoThirdEarlierOutputDistance}{0.004\xspace}
\newcommand{\SeventhTwoThirdEarlierSiteDistance}{0.081\xspace}
\newcommand{\SeventhTwoThirdEarlierOpposite}{0.4690\xspace}
\newcommand{\SeventhTwoThirdEarlierFixed}{0.4690\xspace}
\newcommand{\SeventhTwoThirdLaterOutput}{0.226\xspace}
\newcommand{\SeventhTwoThirdLaterOutputLow}{0.179\xspace}
\newcommand{\SeventhTwoThirdLaterOutputHigh}{0.291\xspace}
\newcommand{\SeventhTwoThirdLaterSite}{0.323\xspace}
\newcommand{\SeventhTwoThirdLaterSiteLow}{0.291\xspace}
\newcommand{\SeventhTwoThirdLaterSiteHigh}{0.364\xspace}
\newcommand{\SeventhTwoThirdLaterOutputDistance}{-0.024\xspace}
\newcommand{\SeventhTwoThirdLaterSiteDistance}{0.073\xspace}
\newcommand{\SeventhTwoThirdLaterOpposite}{0.7810\xspace}
\newcommand{\SeventhTwoThirdLaterFixed}{0.7810\xspace}
\newcommand{\SeventhThreeScoringWholeOutput}{0.191\xspace}
\newcommand{\SeventhThreeScoringWholeOutputLow}{0.170\xspace}
\newcommand{\SeventhThreeScoringWholeOutputHigh}{0.215\xspace}
\newcommand{\SeventhThreeScoringWholeSite}{0.313\xspace}
\newcommand{\SeventhThreeScoringWholeSiteLow}{0.296\xspace}
\newcommand{\SeventhThreeScoringWholeSiteHigh}{0.331\xspace}
\newcommand{\SeventhThreeScoringWholeOutputDistance}{-0.059\xspace}
\newcommand{\SeventhThreeScoringWholeSiteDistance}{0.063\xspace}
\newcommand{\SeventhThreeScoringWholeOpposite}{1.0000\xspace}
\newcommand{\SeventhThreeScoringWholeFixed}{1.0000\xspace}
\newcommand{\SeventhThreeScoringEarlierOutput}{0.183\xspace}
\newcommand{\SeventhThreeScoringEarlierOutputLow}{0.156\xspace}
\newcommand{\SeventhThreeScoringEarlierOutputHigh}{0.218\xspace}
\newcommand{\SeventhThreeScoringEarlierSite}{0.308\xspace}
\newcommand{\SeventhThreeScoringEarlierSiteLow}{0.284\xspace}
\newcommand{\SeventhThreeScoringEarlierSiteHigh}{0.334\xspace}
\newcommand{\SeventhThreeScoringEarlierOutputDistance}{-0.067\xspace}
\newcommand{\SeventhThreeScoringEarlierSiteDistance}{0.058\xspace}
\newcommand{\SeventhThreeScoringEarlierOpposite}{0.9995\xspace}
\newcommand{\SeventhThreeScoringEarlierFixed}{0.9995\xspace}
\newcommand{\SeventhThreeScoringLaterOutput}{0.201\xspace}
\newcommand{\SeventhThreeScoringLaterOutputLow}{0.170\xspace}
\newcommand{\SeventhThreeScoringLaterOutputHigh}{0.241\xspace}
\newcommand{\SeventhThreeScoringLaterSite}{0.319\xspace}
\newcommand{\SeventhThreeScoringLaterSiteLow}{0.294\xspace}
\newcommand{\SeventhThreeScoringLaterSiteHigh}{0.344\xspace}
\newcommand{\SeventhThreeScoringLaterOutputDistance}{-0.049\xspace}
\newcommand{\SeventhThreeScoringLaterSiteDistance}{0.069\xspace}
\newcommand{\SeventhThreeScoringLaterOpposite}{0.9935\xspace}
\newcommand{\SeventhThreeScoringLaterFixed}{0.9935\xspace}
\newcommand{\SeventhThreeFreshWholeOutput}{0.181\xspace}
\newcommand{\SeventhThreeFreshWholeOutputLow}{0.160\xspace}
\newcommand{\SeventhThreeFreshWholeOutputHigh}{0.206\xspace}
\newcommand{\SeventhThreeFreshWholeSite}{0.287\xspace}
\newcommand{\SeventhThreeFreshWholeSiteLow}{0.270\xspace}
\newcommand{\SeventhThreeFreshWholeSiteHigh}{0.306\xspace}
\newcommand{\SeventhThreeFreshWholeOutputDistance}{-0.069\xspace}
\newcommand{\SeventhThreeFreshWholeSiteDistance}{0.037\xspace}
\newcommand{\SeventhThreeFreshWholeOpposite}{1.0000\xspace}
\newcommand{\SeventhThreeFreshWholeFixed}{1.0000\xspace}
\newcommand{\SeventhThreeFreshEarlierOutput}{0.173\xspace}
\newcommand{\SeventhThreeFreshEarlierOutputLow}{0.146\xspace}
\newcommand{\SeventhThreeFreshEarlierOutputHigh}{0.210\xspace}
\newcommand{\SeventhThreeFreshEarlierSite}{0.277\xspace}
\newcommand{\SeventhThreeFreshEarlierSiteLow}{0.255\xspace}
\newcommand{\SeventhThreeFreshEarlierSiteHigh}{0.304\xspace}
\newcommand{\SeventhThreeFreshEarlierOutputDistance}{-0.077\xspace}
\newcommand{\SeventhThreeFreshEarlierSiteDistance}{0.027\xspace}
\newcommand{\SeventhThreeFreshEarlierOpposite}{0.9925\xspace}
\newcommand{\SeventhThreeFreshEarlierFixed}{0.9925\xspace}
\newcommand{\SeventhThreeFreshLaterOutput}{0.190\xspace}
\newcommand{\SeventhThreeFreshLaterOutputLow}{0.162\xspace}
\newcommand{\SeventhThreeFreshLaterOutputHigh}{0.223\xspace}
\newcommand{\SeventhThreeFreshLaterSite}{0.299\xspace}
\newcommand{\SeventhThreeFreshLaterSiteLow}{0.277\xspace}
\newcommand{\SeventhThreeFreshLaterSiteHigh}{0.324\xspace}
\newcommand{\SeventhThreeFreshLaterOutputDistance}{-0.060\xspace}
\newcommand{\SeventhThreeFreshLaterSiteDistance}{0.049\xspace}
\newcommand{\SeventhThreeFreshLaterOpposite}{0.9995\xspace}
\newcommand{\SeventhThreeFreshLaterFixed}{0.9995\xspace}
\newcommand{\SeventhThreeThirdWholeOutput}{0.212\xspace}
\newcommand{\SeventhThreeThirdWholeOutputLow}{0.184\xspace}
\newcommand{\SeventhThreeThirdWholeOutputHigh}{0.246\xspace}
\newcommand{\SeventhThreeThirdWholeSite}{0.295\xspace}
\newcommand{\SeventhThreeThirdWholeSiteLow}{0.277\xspace}
\newcommand{\SeventhThreeThirdWholeSiteHigh}{0.314\xspace}
\newcommand{\SeventhThreeThirdWholeOutputDistance}{-0.038\xspace}
\newcommand{\SeventhThreeThirdWholeSiteDistance}{0.045\xspace}
\newcommand{\SeventhThreeThirdWholeOpposite}{0.9850\xspace}
\newcommand{\SeventhThreeThirdWholeFixed}{0.9850\xspace}
\newcommand{\SeventhThreeThirdEarlierOutput}{0.226\xspace}
\newcommand{\SeventhThreeThirdEarlierOutputLow}{0.187\xspace}
\newcommand{\SeventhThreeThirdEarlierOutputHigh}{0.271\xspace}
\newcommand{\SeventhThreeThirdEarlierSite}{0.302\xspace}
\newcommand{\SeventhThreeThirdEarlierSiteLow}{0.281\xspace}
\newcommand{\SeventhThreeThirdEarlierSiteHigh}{0.327\xspace}
\newcommand{\SeventhThreeThirdEarlierOutputDistance}{-0.024\xspace}
\newcommand{\SeventhThreeThirdEarlierSiteDistance}{0.052\xspace}
\newcommand{\SeventhThreeThirdEarlierOpposite}{0.8695\xspace}
\newcommand{\SeventhThreeThirdEarlierFixed}{0.8695\xspace}
\newcommand{\SeventhThreeThirdLaterOutput}{0.201\xspace}
\newcommand{\SeventhThreeThirdLaterOutputLow}{0.166\xspace}
\newcommand{\SeventhThreeThirdLaterOutputHigh}{0.247\xspace}
\newcommand{\SeventhThreeThirdLaterSite}{0.289\xspace}
\newcommand{\SeventhThreeThirdLaterSiteLow}{0.264\xspace}
\newcommand{\SeventhThreeThirdLaterSiteHigh}{0.318\xspace}
\newcommand{\SeventhThreeThirdLaterOutputDistance}{-0.049\xspace}
\newcommand{\SeventhThreeThirdLaterSiteDistance}{0.039\xspace}
\newcommand{\SeventhThreeThirdLaterOpposite}{0.9805\xspace}
\newcommand{\SeventhThreeThirdLaterFixed}{0.9805\xspace}
\newcommand{\SeventhFourScoringWholeOutput}{0.230\xspace}
\newcommand{\SeventhFourScoringWholeOutputLow}{0.203\xspace}
\newcommand{\SeventhFourScoringWholeOutputHigh}{0.262\xspace}
\newcommand{\SeventhFourScoringWholeSite}{0.309\xspace}
\newcommand{\SeventhFourScoringWholeSiteLow}{0.298\xspace}
\newcommand{\SeventhFourScoringWholeSiteHigh}{0.321\xspace}
\newcommand{\SeventhFourScoringWholeOutputDistance}{-0.020\xspace}
\newcommand{\SeventhFourScoringWholeSiteDistance}{0.059\xspace}
\newcommand{\SeventhFourScoringWholeOpposite}{0.8680\xspace}
\newcommand{\SeventhFourScoringWholeFixed}{0.8680\xspace}
\newcommand{\SeventhFourScoringEarlierOutput}{0.237\xspace}
\newcommand{\SeventhFourScoringEarlierOutputLow}{0.205\xspace}
\newcommand{\SeventhFourScoringEarlierOutputHigh}{0.270\xspace}
\newcommand{\SeventhFourScoringEarlierSite}{0.301\xspace}
\newcommand{\SeventhFourScoringEarlierSiteLow}{0.285\xspace}
\newcommand{\SeventhFourScoringEarlierSiteHigh}{0.319\xspace}
\newcommand{\SeventhFourScoringEarlierOutputDistance}{-0.013\xspace}
\newcommand{\SeventhFourScoringEarlierSiteDistance}{0.051\xspace}
\newcommand{\SeventhFourScoringEarlierOpposite}{0.7510\xspace}
\newcommand{\SeventhFourScoringEarlierFixed}{0.7510\xspace}
\newcommand{\SeventhFourScoringLaterOutput}{0.224\xspace}
\newcommand{\SeventhFourScoringLaterOutputLow}{0.190\xspace}
\newcommand{\SeventhFourScoringLaterOutputHigh}{0.275\xspace}
\newcommand{\SeventhFourScoringLaterSite}{0.317\xspace}
\newcommand{\SeventhFourScoringLaterSiteLow}{0.303\xspace}
\newcommand{\SeventhFourScoringLaterSiteHigh}{0.333\xspace}
\newcommand{\SeventhFourScoringLaterOutputDistance}{-0.026\xspace}
\newcommand{\SeventhFourScoringLaterSiteDistance}{0.067\xspace}
\newcommand{\SeventhFourScoringLaterOpposite}{0.7700\xspace}
\newcommand{\SeventhFourScoringLaterFixed}{0.7700\xspace}
\newcommand{\SeventhFourFreshWholeOutput}{0.206\xspace}
\newcommand{\SeventhFourFreshWholeOutputLow}{0.162\xspace}
\newcommand{\SeventhFourFreshWholeOutputHigh}{0.262\xspace}
\newcommand{\SeventhFourFreshWholeSite}{0.299\xspace}
\newcommand{\SeventhFourFreshWholeSiteLow}{0.288\xspace}
\newcommand{\SeventhFourFreshWholeSiteHigh}{0.311\xspace}
\newcommand{\SeventhFourFreshWholeOutputDistance}{-0.044\xspace}
\newcommand{\SeventhFourFreshWholeSiteDistance}{0.049\xspace}
\newcommand{\SeventhFourFreshWholeOpposite}{0.8860\xspace}
\newcommand{\SeventhFourFreshWholeFixed}{0.8860\xspace}
\newcommand{\SeventhFourFreshEarlierOutput}{0.246\xspace}
\newcommand{\SeventhFourFreshEarlierOutputLow}{0.218\xspace}
\newcommand{\SeventhFourFreshEarlierOutputHigh}{0.275\xspace}
\newcommand{\SeventhFourFreshEarlierSite}{0.303\xspace}
\newcommand{\SeventhFourFreshEarlierSiteLow}{0.288\xspace}
\newcommand{\SeventhFourFreshEarlierSiteHigh}{0.322\xspace}
\newcommand{\SeventhFourFreshEarlierOutputDistance}{-0.004\xspace}
\newcommand{\SeventhFourFreshEarlierSiteDistance}{0.053\xspace}
\newcommand{\SeventhFourFreshEarlierOpposite}{0.6085\xspace}
\newcommand{\SeventhFourFreshEarlierFixed}{0.6085\xspace}
\newcommand{\SeventhFourFreshLaterOutput}{0.179\xspace}
\newcommand{\SeventhFourFreshLaterOutputLow}{0.135\xspace}
\newcommand{\SeventhFourFreshLaterOutputHigh}{0.264\xspace}
\newcommand{\SeventhFourFreshLaterSite}{0.295\xspace}
\newcommand{\SeventhFourFreshLaterSiteLow}{0.282\xspace}
\newcommand{\SeventhFourFreshLaterSiteHigh}{0.309\xspace}
\newcommand{\SeventhFourFreshLaterOutputDistance}{-0.071\xspace}
\newcommand{\SeventhFourFreshLaterSiteDistance}{0.045\xspace}
\newcommand{\SeventhFourFreshLaterOpposite}{0.8445\xspace}
\newcommand{\SeventhFourFreshLaterFixed}{0.8445\xspace}
\newcommand{\SeventhFourThirdWholeOutput}{0.217\xspace}
\newcommand{\SeventhFourThirdWholeOutputLow}{0.194\xspace}
\newcommand{\SeventhFourThirdWholeOutputHigh}{0.250\xspace}
\newcommand{\SeventhFourThirdWholeSite}{0.306\xspace}
\newcommand{\SeventhFourThirdWholeSiteLow}{0.291\xspace}
\newcommand{\SeventhFourThirdWholeSiteHigh}{0.323\xspace}
\newcommand{\SeventhFourThirdWholeOutputDistance}{-0.033\xspace}
\newcommand{\SeventhFourThirdWholeSiteDistance}{0.056\xspace}
\newcommand{\SeventhFourThirdWholeOpposite}{0.9750\xspace}
\newcommand{\SeventhFourThirdWholeFixed}{0.9750\xspace}
\newcommand{\SeventhFourThirdEarlierOutput}{0.220\xspace}
\newcommand{\SeventhFourThirdEarlierOutputLow}{0.185\xspace}
\newcommand{\SeventhFourThirdEarlierOutputHigh}{0.274\xspace}
\newcommand{\SeventhFourThirdEarlierSite}{0.319\xspace}
\newcommand{\SeventhFourThirdEarlierSiteLow}{0.298\xspace}
\newcommand{\SeventhFourThirdEarlierSiteHigh}{0.346\xspace}
\newcommand{\SeventhFourThirdEarlierOutputDistance}{-0.030\xspace}
\newcommand{\SeventhFourThirdEarlierSiteDistance}{0.069\xspace}
\newcommand{\SeventhFourThirdEarlierOpposite}{0.8405\xspace}
\newcommand{\SeventhFourThirdEarlierFixed}{0.8405\xspace}
\newcommand{\SeventhFourThirdLaterOutput}{0.215\xspace}
\newcommand{\SeventhFourThirdLaterOutputLow}{0.185\xspace}
\newcommand{\SeventhFourThirdLaterOutputHigh}{0.255\xspace}
\newcommand{\SeventhFourThirdLaterSite}{0.296\xspace}
\newcommand{\SeventhFourThirdLaterSiteLow}{0.277\xspace}
\newcommand{\SeventhFourThirdLaterSiteHigh}{0.318\xspace}
\newcommand{\SeventhFourThirdLaterOutputDistance}{-0.035\xspace}
\newcommand{\SeventhFourThirdLaterSiteDistance}{0.046\xspace}
\newcommand{\SeventhFourThirdLaterOpposite}{0.9540\xspace}
\newcommand{\SeventhFourThirdLaterFixed}{0.9540\xspace}
\newcommand{\SeventhFiveScoringWholeOutput}{0.219\xspace}
\newcommand{\SeventhFiveScoringWholeOutputLow}{0.172\xspace}
\newcommand{\SeventhFiveScoringWholeOutputHigh}{0.272\xspace}
\newcommand{\SeventhFiveScoringWholeSite}{0.301\xspace}
\newcommand{\SeventhFiveScoringWholeSiteLow}{0.271\xspace}
\newcommand{\SeventhFiveScoringWholeSiteHigh}{0.338\xspace}
\newcommand{\SeventhFiveScoringWholeOutputDistance}{-0.031\xspace}
\newcommand{\SeventhFiveScoringWholeSiteDistance}{0.051\xspace}
\newcommand{\SeventhFiveScoringWholeOpposite}{0.8680\xspace}
\newcommand{\SeventhFiveScoringWholeFixed}{0.8680\xspace}
\newcommand{\SeventhFiveScoringEarlierOutput}{0.197\xspace}
\newcommand{\SeventhFiveScoringEarlierOutputLow}{0.160\xspace}
\newcommand{\SeventhFiveScoringEarlierOutputHigh}{0.245\xspace}
\newcommand{\SeventhFiveScoringEarlierSite}{0.262\xspace}
\newcommand{\SeventhFiveScoringEarlierSiteLow}{0.235\xspace}
\newcommand{\SeventhFiveScoringEarlierSiteHigh}{0.295\xspace}
\newcommand{\SeventhFiveScoringEarlierOutputDistance}{-0.053\xspace}
\newcommand{\SeventhFiveScoringEarlierSiteDistance}{0.012\xspace}
\newcommand{\SeventhFiveScoringEarlierOpposite}{0.7785\xspace}
\newcommand{\SeventhFiveScoringEarlierFixed}{0.7785\xspace}
\newcommand{\SeventhFiveScoringLaterOutput}{0.233\xspace}
\newcommand{\SeventhFiveScoringLaterOutputLow}{0.167\xspace}
\newcommand{\SeventhFiveScoringLaterOutputHigh}{0.320\xspace}
\newcommand{\SeventhFiveScoringLaterSite}{0.339\xspace}
\newcommand{\SeventhFiveScoringLaterSiteLow}{0.288\xspace}
\newcommand{\SeventhFiveScoringLaterSiteHigh}{0.398\xspace}
\newcommand{\SeventhFiveScoringLaterOutputDistance}{-0.017\xspace}
\newcommand{\SeventhFiveScoringLaterSiteDistance}{0.089\xspace}
\newcommand{\SeventhFiveScoringLaterOpposite}{0.6155\xspace}
\newcommand{\SeventhFiveScoringLaterFixed}{0.6155\xspace}
\newcommand{\SeventhFiveFreshWholeOutput}{0.221\xspace}
\newcommand{\SeventhFiveFreshWholeOutputLow}{0.186\xspace}
\newcommand{\SeventhFiveFreshWholeOutputHigh}{0.262\xspace}
\newcommand{\SeventhFiveFreshWholeSite}{0.280\xspace}
\newcommand{\SeventhFiveFreshWholeSiteLow}{0.263\xspace}
\newcommand{\SeventhFiveFreshWholeSiteHigh}{0.297\xspace}
\newcommand{\SeventhFiveFreshWholeOutputDistance}{-0.029\xspace}
\newcommand{\SeventhFiveFreshWholeSiteDistance}{0.030\xspace}
\newcommand{\SeventhFiveFreshWholeOpposite}{0.9245\xspace}
\newcommand{\SeventhFiveFreshWholeFixed}{0.9245\xspace}
\newcommand{\SeventhFiveFreshEarlierOutput}{0.225\xspace}
\newcommand{\SeventhFiveFreshEarlierOutputLow}{0.179\xspace}
\newcommand{\SeventhFiveFreshEarlierOutputHigh}{0.292\xspace}
\newcommand{\SeventhFiveFreshEarlierSite}{0.270\xspace}
\newcommand{\SeventhFiveFreshEarlierSiteLow}{0.251\xspace}
\newcommand{\SeventhFiveFreshEarlierSiteHigh}{0.293\xspace}
\newcommand{\SeventhFiveFreshEarlierOutputDistance}{-0.025\xspace}
\newcommand{\SeventhFiveFreshEarlierSiteDistance}{0.020\xspace}
\newcommand{\SeventhFiveFreshEarlierOpposite}{0.7735\xspace}
\newcommand{\SeventhFiveFreshEarlierFixed}{0.7735\xspace}
\newcommand{\SeventhFiveFreshLaterOutput}{0.218\xspace}
\newcommand{\SeventhFiveFreshLaterOutputLow}{0.173\xspace}
\newcommand{\SeventhFiveFreshLaterOutputHigh}{0.288\xspace}
\newcommand{\SeventhFiveFreshLaterSite}{0.287\xspace}
\newcommand{\SeventhFiveFreshLaterSiteLow}{0.265\xspace}
\newcommand{\SeventhFiveFreshLaterSiteHigh}{0.311\xspace}
\newcommand{\SeventhFiveFreshLaterOutputDistance}{-0.032\xspace}
\newcommand{\SeventhFiveFreshLaterSiteDistance}{0.037\xspace}
\newcommand{\SeventhFiveFreshLaterOpposite}{0.8420\xspace}
\newcommand{\SeventhFiveFreshLaterFixed}{0.8420\xspace}
\newcommand{\SeventhFiveThirdWholeOutput}{0.215\xspace}
\newcommand{\SeventhFiveThirdWholeOutputLow}{0.156\xspace}
\newcommand{\SeventhFiveThirdWholeOutputHigh}{0.309\xspace}
\newcommand{\SeventhFiveThirdWholeSite}{0.260\xspace}
\newcommand{\SeventhFiveThirdWholeSiteLow}{0.242\xspace}
\newcommand{\SeventhFiveThirdWholeSiteHigh}{0.280\xspace}
\newcommand{\SeventhFiveThirdWholeOutputDistance}{-0.035\xspace}
\newcommand{\SeventhFiveThirdWholeSiteDistance}{0.010\xspace}
\newcommand{\SeventhFiveThirdWholeOpposite}{0.6940\xspace}
\newcommand{\SeventhFiveThirdWholeFixed}{0.6710\xspace}
\newcommand{\SeventhFiveThirdEarlierOutput}{0.276\xspace}
\newcommand{\SeventhFiveThirdEarlierOutputLow}{0.174\xspace}
\newcommand{\SeventhFiveThirdEarlierOutputHigh}{0.441\xspace}
\newcommand{\SeventhFiveThirdEarlierSite}{0.259\xspace}
\newcommand{\SeventhFiveThirdEarlierSiteLow}{0.233\xspace}
\newcommand{\SeventhFiveThirdEarlierSiteHigh}{0.291\xspace}
\newcommand{\SeventhFiveThirdEarlierOutputDistance}{0.026\xspace}
\newcommand{\SeventhFiveThirdEarlierSiteDistance}{0.009\xspace}
\newcommand{\SeventhFiveThirdEarlierOpposite}{0.4430\xspace}
\newcommand{\SeventhFiveThirdEarlierFixed}{0.3060\xspace}
\newcommand{\SeventhFiveThirdLaterOutput}{0.153\xspace}
\newcommand{\SeventhFiveThirdLaterOutputLow}{0.117\xspace}
\newcommand{\SeventhFiveThirdLaterOutputHigh}{0.201\xspace}
\newcommand{\SeventhFiveThirdLaterSite}{0.261\xspace}
\newcommand{\SeventhFiveThirdLaterSiteLow}{0.238\xspace}
\newcommand{\SeventhFiveThirdLaterSiteHigh}{0.290\xspace}
\newcommand{\SeventhFiveThirdLaterOutputDistance}{-0.097\xspace}
\newcommand{\SeventhFiveThirdLaterSiteDistance}{0.011\xspace}
\newcommand{\SeventhFiveThirdLaterOpposite}{0.8105\xspace}
\newcommand{\SeventhFiveThirdLaterFixed}{0.8105\xspace}
\newcommand{\SeventhSixScoringWholeOutput}{0.267\xspace}
\newcommand{\SeventhSixScoringWholeOutputLow}{0.256\xspace}
\newcommand{\SeventhSixScoringWholeOutputHigh}{0.279\xspace}
\newcommand{\SeventhSixScoringWholeSite}{0.224\xspace}
\newcommand{\SeventhSixScoringWholeSiteLow}{0.214\xspace}
\newcommand{\SeventhSixScoringWholeSiteHigh}{0.236\xspace}
\newcommand{\SeventhSixScoringWholeOutputDistance}{0.017\xspace}
\newcommand{\SeventhSixScoringWholeSiteDistance}{-0.026\xspace}
\newcommand{\SeventhSixScoringWholeOpposite}{0.9995\xspace}
\newcommand{\SeventhSixScoringWholeFixed}{0.9995\xspace}
\newcommand{\SeventhSixScoringEarlierOutput}{0.271\xspace}
\newcommand{\SeventhSixScoringEarlierOutputLow}{0.256\xspace}
\newcommand{\SeventhSixScoringEarlierOutputHigh}{0.287\xspace}
\newcommand{\SeventhSixScoringEarlierSite}{0.225\xspace}
\newcommand{\SeventhSixScoringEarlierSiteLow}{0.212\xspace}
\newcommand{\SeventhSixScoringEarlierSiteHigh}{0.241\xspace}
\newcommand{\SeventhSixScoringEarlierOutputDistance}{0.021\xspace}
\newcommand{\SeventhSixScoringEarlierSiteDistance}{-0.025\xspace}
\newcommand{\SeventhSixScoringEarlierOpposite}{0.9975\xspace}
\newcommand{\SeventhSixScoringEarlierFixed}{0.9975\xspace}
\newcommand{\SeventhSixScoringLaterOutput}{0.262\xspace}
\newcommand{\SeventhSixScoringLaterOutputLow}{0.247\xspace}
\newcommand{\SeventhSixScoringLaterOutputHigh}{0.279\xspace}
\newcommand{\SeventhSixScoringLaterSite}{0.224\xspace}
\newcommand{\SeventhSixScoringLaterSiteLow}{0.209\xspace}
\newcommand{\SeventhSixScoringLaterSiteHigh}{0.240\xspace}
\newcommand{\SeventhSixScoringLaterOutputDistance}{0.012\xspace}
\newcommand{\SeventhSixScoringLaterSiteDistance}{-0.026\xspace}
\newcommand{\SeventhSixScoringLaterOpposite}{0.9430\xspace}
\newcommand{\SeventhSixScoringLaterFixed}{0.9430\xspace}
\newcommand{\SeventhSixFreshWholeOutput}{0.273\xspace}
\newcommand{\SeventhSixFreshWholeOutputLow}{0.263\xspace}
\newcommand{\SeventhSixFreshWholeOutputHigh}{0.284\xspace}
\newcommand{\SeventhSixFreshWholeSite}{0.227\xspace}
\newcommand{\SeventhSixFreshWholeSiteLow}{0.216\xspace}
\newcommand{\SeventhSixFreshWholeSiteHigh}{0.239\xspace}
\newcommand{\SeventhSixFreshWholeOutputDistance}{0.023\xspace}
\newcommand{\SeventhSixFreshWholeSiteDistance}{-0.023\xspace}
\newcommand{\SeventhSixFreshWholeOpposite}{0.9995\xspace}
\newcommand{\SeventhSixFreshWholeFixed}{0.9995\xspace}
\newcommand{\SeventhSixFreshEarlierOutput}{0.270\xspace}
\newcommand{\SeventhSixFreshEarlierOutputLow}{0.257\xspace}
\newcommand{\SeventhSixFreshEarlierOutputHigh}{0.284\xspace}
\newcommand{\SeventhSixFreshEarlierSite}{0.223\xspace}
\newcommand{\SeventhSixFreshEarlierSiteLow}{0.210\xspace}
\newcommand{\SeventhSixFreshEarlierSiteHigh}{0.238\xspace}
\newcommand{\SeventhSixFreshEarlierOutputDistance}{0.020\xspace}
\newcommand{\SeventhSixFreshEarlierSiteDistance}{-0.027\xspace}
\newcommand{\SeventhSixFreshEarlierOpposite}{0.9995\xspace}
\newcommand{\SeventhSixFreshEarlierFixed}{0.9995\xspace}
\newcommand{\SeventhSixFreshLaterOutput}{0.277\xspace}
\newcommand{\SeventhSixFreshLaterOutputLow}{0.261\xspace}
\newcommand{\SeventhSixFreshLaterOutputHigh}{0.295\xspace}
\newcommand{\SeventhSixFreshLaterSite}{0.231\xspace}
\newcommand{\SeventhSixFreshLaterSiteLow}{0.214\xspace}
\newcommand{\SeventhSixFreshLaterSiteHigh}{0.251\xspace}
\newcommand{\SeventhSixFreshLaterOutputDistance}{0.027\xspace}
\newcommand{\SeventhSixFreshLaterSiteDistance}{-0.019\xspace}
\newcommand{\SeventhSixFreshLaterOpposite}{0.9680\xspace}
\newcommand{\SeventhSixFreshLaterFixed}{0.9680\xspace}
\newcommand{\SeventhSixThirdWholeOutput}{0.273\xspace}
\newcommand{\SeventhSixThirdWholeOutputLow}{0.256\xspace}
\newcommand{\SeventhSixThirdWholeOutputHigh}{0.291\xspace}
\newcommand{\SeventhSixThirdWholeSite}{0.225\xspace}
\newcommand{\SeventhSixThirdWholeSiteLow}{0.210\xspace}
\newcommand{\SeventhSixThirdWholeSiteHigh}{0.244\xspace}
\newcommand{\SeventhSixThirdWholeOutputDistance}{0.023\xspace}
\newcommand{\SeventhSixThirdWholeSiteDistance}{-0.025\xspace}
\newcommand{\SeventhSixThirdWholeOpposite}{0.9960\xspace}
\newcommand{\SeventhSixThirdWholeFixed}{0.9960\xspace}
\newcommand{\SeventhSixThirdEarlierOutput}{0.271\xspace}
\newcommand{\SeventhSixThirdEarlierOutputLow}{0.252\xspace}
\newcommand{\SeventhSixThirdEarlierOutputHigh}{0.292\xspace}
\newcommand{\SeventhSixThirdEarlierSite}{0.218\xspace}
\newcommand{\SeventhSixThirdEarlierSiteLow}{0.199\xspace}
\newcommand{\SeventhSixThirdEarlierSiteHigh}{0.241\xspace}
\newcommand{\SeventhSixThirdEarlierOutputDistance}{0.021\xspace}
\newcommand{\SeventhSixThirdEarlierSiteDistance}{-0.032\xspace}
\newcommand{\SeventhSixThirdEarlierOpposite}{0.9810\xspace}
\newcommand{\SeventhSixThirdEarlierFixed}{0.9810\xspace}
\newcommand{\SeventhSixThirdLaterOutput}{0.274\xspace}
\newcommand{\SeventhSixThirdLaterOutputLow}{0.250\xspace}
\newcommand{\SeventhSixThirdLaterOutputHigh}{0.302\xspace}
\newcommand{\SeventhSixThirdLaterSite}{0.232\xspace}
\newcommand{\SeventhSixThirdLaterSiteLow}{0.209\xspace}
\newcommand{\SeventhSixThirdLaterSiteHigh}{0.260\xspace}
\newcommand{\SeventhSixThirdLaterOutputDistance}{0.024\xspace}
\newcommand{\SeventhSixThirdLaterSiteDistance}{-0.018\xspace}
\newcommand{\SeventhSixThirdLaterOpposite}{0.8635\xspace}
\newcommand{\SeventhSixThirdLaterFixed}{0.8635\xspace}
\newcommand{\SeventhSevenScoringWholeOutput}{0.442\xspace}
\newcommand{\SeventhSevenScoringWholeOutputLow}{0.415\xspace}
\newcommand{\SeventhSevenScoringWholeOutputHigh}{0.475\xspace}
\newcommand{\SeventhSevenScoringWholeSite}{0.179\xspace}
\newcommand{\SeventhSevenScoringWholeSiteLow}{0.163\xspace}
\newcommand{\SeventhSevenScoringWholeSiteHigh}{0.200\xspace}
\newcommand{\SeventhSevenScoringWholeOutputDistance}{0.192\xspace}
\newcommand{\SeventhSevenScoringWholeSiteDistance}{-0.071\xspace}
\newcommand{\SeventhSevenScoringWholeOpposite}{1.0000\xspace}
\newcommand{\SeventhSevenScoringWholeFixed}{1.0000\xspace}
\newcommand{\SeventhSevenScoringEarlierOutput}{0.448\xspace}
\newcommand{\SeventhSevenScoringEarlierOutputLow}{0.412\xspace}
\newcommand{\SeventhSevenScoringEarlierOutputHigh}{0.498\xspace}
\newcommand{\SeventhSevenScoringEarlierSite}{0.177\xspace}
\newcommand{\SeventhSevenScoringEarlierSiteLow}{0.154\xspace}
\newcommand{\SeventhSevenScoringEarlierSiteHigh}{0.205\xspace}
\newcommand{\SeventhSevenScoringEarlierOutputDistance}{0.198\xspace}
\newcommand{\SeventhSevenScoringEarlierSiteDistance}{-0.073\xspace}
\newcommand{\SeventhSevenScoringEarlierOpposite}{1.0000\xspace}
\newcommand{\SeventhSevenScoringEarlierFixed}{1.0000\xspace}
\newcommand{\SeventhSevenScoringLaterOutput}{0.435\xspace}
\newcommand{\SeventhSevenScoringLaterOutputLow}{0.398\xspace}
\newcommand{\SeventhSevenScoringLaterOutputHigh}{0.481\xspace}
\newcommand{\SeventhSevenScoringLaterSite}{0.182\xspace}
\newcommand{\SeventhSevenScoringLaterSiteLow}{0.160\xspace}
\newcommand{\SeventhSevenScoringLaterSiteHigh}{0.211\xspace}
\newcommand{\SeventhSevenScoringLaterOutputDistance}{0.185\xspace}
\newcommand{\SeventhSevenScoringLaterSiteDistance}{-0.068\xspace}
\newcommand{\SeventhSevenScoringLaterOpposite}{1.0000\xspace}
\newcommand{\SeventhSevenScoringLaterFixed}{1.0000\xspace}
\newcommand{\SeventhSevenFreshWholeOutput}{0.417\xspace}
\newcommand{\SeventhSevenFreshWholeOutputLow}{0.391\xspace}
\newcommand{\SeventhSevenFreshWholeOutputHigh}{0.453\xspace}
\newcommand{\SeventhSevenFreshWholeSite}{0.174\xspace}
\newcommand{\SeventhSevenFreshWholeSiteLow}{0.156\xspace}
\newcommand{\SeventhSevenFreshWholeSiteHigh}{0.196\xspace}
\newcommand{\SeventhSevenFreshWholeOutputDistance}{0.167\xspace}
\newcommand{\SeventhSevenFreshWholeSiteDistance}{-0.076\xspace}
\newcommand{\SeventhSevenFreshWholeOpposite}{1.0000\xspace}
\newcommand{\SeventhSevenFreshWholeFixed}{1.0000\xspace}
\newcommand{\SeventhSevenFreshEarlierOutput}{0.406\xspace}
\newcommand{\SeventhSevenFreshEarlierOutputLow}{0.368\xspace}
\newcommand{\SeventhSevenFreshEarlierOutputHigh}{0.464\xspace}
\newcommand{\SeventhSevenFreshEarlierSite}{0.166\xspace}
\newcommand{\SeventhSevenFreshEarlierSiteLow}{0.143\xspace}
\newcommand{\SeventhSevenFreshEarlierSiteHigh}{0.198\xspace}
\newcommand{\SeventhSevenFreshEarlierOutputDistance}{0.156\xspace}
\newcommand{\SeventhSevenFreshEarlierSiteDistance}{-0.084\xspace}
\newcommand{\SeventhSevenFreshEarlierOpposite}{1.0000\xspace}
\newcommand{\SeventhSevenFreshEarlierFixed}{1.0000\xspace}
\newcommand{\SeventhSevenFreshLaterOutput}{0.428\xspace}
\newcommand{\SeventhSevenFreshLaterOutputLow}{0.395\xspace}
\newcommand{\SeventhSevenFreshLaterOutputHigh}{0.473\xspace}
\newcommand{\SeventhSevenFreshLaterSite}{0.181\xspace}
\newcommand{\SeventhSevenFreshLaterSiteLow}{0.158\xspace}
\newcommand{\SeventhSevenFreshLaterSiteHigh}{0.211\xspace}
\newcommand{\SeventhSevenFreshLaterOutputDistance}{0.178\xspace}
\newcommand{\SeventhSevenFreshLaterSiteDistance}{-0.069\xspace}
\newcommand{\SeventhSevenFreshLaterOpposite}{1.0000\xspace}
\newcommand{\SeventhSevenFreshLaterFixed}{1.0000\xspace}
\newcommand{\SeventhSevenThirdWholeOutput}{0.451\xspace}
\newcommand{\SeventhSevenThirdWholeOutputLow}{0.399\xspace}
\newcommand{\SeventhSevenThirdWholeOutputHigh}{0.510\xspace}
\newcommand{\SeventhSevenThirdWholeSite}{0.178\xspace}
\newcommand{\SeventhSevenThirdWholeSiteLow}{0.157\xspace}
\newcommand{\SeventhSevenThirdWholeSiteHigh}{0.207\xspace}
\newcommand{\SeventhSevenThirdWholeOutputDistance}{0.201\xspace}
\newcommand{\SeventhSevenThirdWholeSiteDistance}{-0.072\xspace}
\newcommand{\SeventhSevenThirdWholeOpposite}{1.0000\xspace}
\newcommand{\SeventhSevenThirdWholeFixed}{1.0000\xspace}
\newcommand{\SeventhSevenThirdEarlierOutput}{0.474\xspace}
\newcommand{\SeventhSevenThirdEarlierOutputLow}{0.409\xspace}
\newcommand{\SeventhSevenThirdEarlierOutputHigh}{0.562\xspace}
\newcommand{\SeventhSevenThirdEarlierSite}{0.176\xspace}
\newcommand{\SeventhSevenThirdEarlierSiteLow}{0.149\xspace}
\newcommand{\SeventhSevenThirdEarlierSiteHigh}{0.220\xspace}
\newcommand{\SeventhSevenThirdEarlierOutputDistance}{0.224\xspace}
\newcommand{\SeventhSevenThirdEarlierSiteDistance}{-0.074\xspace}
\newcommand{\SeventhSevenThirdEarlierOpposite}{0.9975\xspace}
\newcommand{\SeventhSevenThirdEarlierFixed}{0.9975\xspace}
\newcommand{\SeventhSevenThirdLaterOutput}{0.413\xspace}
\newcommand{\SeventhSevenThirdLaterOutputLow}{0.350\xspace}
\newcommand{\SeventhSevenThirdLaterOutputHigh}{0.507\xspace}
\newcommand{\SeventhSevenThirdLaterSite}{0.181\xspace}
\newcommand{\SeventhSevenThirdLaterSiteLow}{0.150\xspace}
\newcommand{\SeventhSevenThirdLaterSiteHigh}{0.226\xspace}
\newcommand{\SeventhSevenThirdLaterOutputDistance}{0.163\xspace}
\newcommand{\SeventhSevenThirdLaterSiteDistance}{-0.069\xspace}
\newcommand{\SeventhSevenThirdLaterOpposite}{0.9970\xspace}
\newcommand{\SeventhSevenThirdLaterFixed}{0.9970\xspace}
\newcommand{\SeventhEightScoringWholeOutput}{0.180\xspace}
\newcommand{\SeventhEightScoringWholeOutputLow}{0.152\xspace}
\newcommand{\SeventhEightScoringWholeOutputHigh}{0.227\xspace}
\newcommand{\SeventhEightScoringWholeSite}{0.268\xspace}
\newcommand{\SeventhEightScoringWholeSiteLow}{0.236\xspace}
\newcommand{\SeventhEightScoringWholeSiteHigh}{0.311\xspace}
\newcommand{\SeventhEightScoringWholeOutputDistance}{-0.070\xspace}
\newcommand{\SeventhEightScoringWholeSiteDistance}{0.018\xspace}
\newcommand{\SeventhEightScoringWholeOpposite}{0.8540\xspace}
\newcommand{\SeventhEightScoringWholeFixed}{0.8540\xspace}
\newcommand{\SeventhEightScoringEarlierOutput}{0.161\xspace}
\newcommand{\SeventhEightScoringEarlierOutputLow}{0.130\xspace}
\newcommand{\SeventhEightScoringEarlierOutputHigh}{0.230\xspace}
\newcommand{\SeventhEightScoringEarlierSite}{0.251\xspace}
\newcommand{\SeventhEightScoringEarlierSiteLow}{0.210\xspace}
\newcommand{\SeventhEightScoringEarlierSiteHigh}{0.316\xspace}
\newcommand{\SeventhEightScoringEarlierOutputDistance}{-0.089\xspace}
\newcommand{\SeventhEightScoringEarlierSiteDistance}{0.001\xspace}
\newcommand{\SeventhEightScoringEarlierOpposite}{0.5475\xspace}
\newcommand{\SeventhEightScoringEarlierFixed}{0.5475\xspace}
\newcommand{\SeventhEightScoringLaterOutput}{0.212\xspace}
\newcommand{\SeventhEightScoringLaterOutputLow}{0.178\xspace}
\newcommand{\SeventhEightScoringLaterOutputHigh}{0.262\xspace}
\newcommand{\SeventhEightScoringLaterSite}{0.294\xspace}
\newcommand{\SeventhEightScoringLaterSiteLow}{0.258\xspace}
\newcommand{\SeventhEightScoringLaterSiteHigh}{0.342\xspace}
\newcommand{\SeventhEightScoringLaterOutputDistance}{-0.038\xspace}
\newcommand{\SeventhEightScoringLaterSiteDistance}{0.044\xspace}
\newcommand{\SeventhEightScoringLaterOpposite}{0.9390\xspace}
\newcommand{\SeventhEightScoringLaterFixed}{0.9390\xspace}
\newcommand{\SeventhEightFreshWholeOutput}{0.225\xspace}
\newcommand{\SeventhEightFreshWholeOutputLow}{0.196\xspace}
\newcommand{\SeventhEightFreshWholeOutputHigh}{0.262\xspace}
\newcommand{\SeventhEightFreshWholeSite}{0.291\xspace}
\newcommand{\SeventhEightFreshWholeSiteLow}{0.264\xspace}
\newcommand{\SeventhEightFreshWholeSiteHigh}{0.325\xspace}
\newcommand{\SeventhEightFreshWholeOutputDistance}{-0.025\xspace}
\newcommand{\SeventhEightFreshWholeSiteDistance}{0.041\xspace}
\newcommand{\SeventhEightFreshWholeOpposite}{0.9020\xspace}
\newcommand{\SeventhEightFreshWholeFixed}{0.9020\xspace}
\newcommand{\SeventhEightFreshEarlierOutput}{0.211\xspace}
\newcommand{\SeventhEightFreshEarlierOutputLow}{0.178\xspace}
\newcommand{\SeventhEightFreshEarlierOutputHigh}{0.253\xspace}
\newcommand{\SeventhEightFreshEarlierSite}{0.276\xspace}
\newcommand{\SeventhEightFreshEarlierSiteLow}{0.246\xspace}
\newcommand{\SeventhEightFreshEarlierSiteHigh}{0.315\xspace}
\newcommand{\SeventhEightFreshEarlierOutputDistance}{-0.039\xspace}
\newcommand{\SeventhEightFreshEarlierSiteDistance}{0.026\xspace}
\newcommand{\SeventhEightFreshEarlierOpposite}{0.9120\xspace}
\newcommand{\SeventhEightFreshEarlierFixed}{0.9120\xspace}
\newcommand{\SeventhEightFreshLaterOutput}{0.247\xspace}
\newcommand{\SeventhEightFreshLaterOutputLow}{0.195\xspace}
\newcommand{\SeventhEightFreshLaterOutputHigh}{0.319\xspace}
\newcommand{\SeventhEightFreshLaterSite}{0.313\xspace}
\newcommand{\SeventhEightFreshLaterSiteLow}{0.261\xspace}
\newcommand{\SeventhEightFreshLaterSiteHigh}{0.381\xspace}
\newcommand{\SeventhEightFreshLaterOutputDistance}{-0.003\xspace}
\newcommand{\SeventhEightFreshLaterSiteDistance}{0.063\xspace}
\newcommand{\SeventhEightFreshLaterOpposite}{0.4950\xspace}
\newcommand{\SeventhEightFreshLaterFixed}{0.4950\xspace}
\newcommand{\SeventhEightThirdWholeOutput}{0.169\xspace}
\newcommand{\SeventhEightThirdWholeOutputLow}{0.133\xspace}
\newcommand{\SeventhEightThirdWholeOutputHigh}{0.255\xspace}
\newcommand{\SeventhEightThirdWholeSite}{0.258\xspace}
\newcommand{\SeventhEightThirdWholeSiteLow}{0.220\xspace}
\newcommand{\SeventhEightThirdWholeSiteHigh}{0.325\xspace}
\newcommand{\SeventhEightThirdWholeOutputDistance}{-0.081\xspace}
\newcommand{\SeventhEightThirdWholeSiteDistance}{0.008\xspace}
\newcommand{\SeventhEightThirdWholeOpposite}{0.6265\xspace}
\newcommand{\SeventhEightThirdWholeFixed}{0.6265\xspace}
\newcommand{\SeventhEightThirdEarlierOutput}{0.245\xspace}
\newcommand{\SeventhEightThirdEarlierOutputLow}{0.184\xspace}
\newcommand{\SeventhEightThirdEarlierOutputHigh}{0.332\xspace}
\newcommand{\SeventhEightThirdEarlierSite}{0.306\xspace}
\newcommand{\SeventhEightThirdEarlierSiteLow}{0.253\xspace}
\newcommand{\SeventhEightThirdEarlierSiteHigh}{0.378\xspace}
\newcommand{\SeventhEightThirdEarlierOutputDistance}{-0.005\xspace}
\newcommand{\SeventhEightThirdEarlierSiteDistance}{0.056\xspace}
\newcommand{\SeventhEightThirdEarlierOpposite}{0.5105\xspace}
\newcommand{\SeventhEightThirdEarlierFixed}{0.5105\xspace}
\newcommand{\SeventhEightThirdLaterOutput}{0.144\xspace}
\newcommand{\SeventhEightThirdLaterOutputLow}{0.114\xspace}
\newcommand{\SeventhEightThirdLaterOutputHigh}{0.244\xspace}
\newcommand{\SeventhEightThirdLaterSite}{0.236\xspace}
\newcommand{\SeventhEightThirdLaterSiteLow}{0.200\xspace}
\newcommand{\SeventhEightThirdLaterSiteHigh}{0.329\xspace}
\newcommand{\SeventhEightThirdLaterOutputDistance}{-0.106\xspace}
\newcommand{\SeventhEightThirdLaterSiteDistance}{-0.014\xspace}
\newcommand{\SeventhEightThirdLaterOpposite}{0.3395\xspace}
\newcommand{\SeventhEightThirdLaterFixed}{0.3395\xspace}
\newcommand{\SeventhFreshUpper}{82\xspace}

\newcommand{\SeventhBoundOneAOneSelectionScoring}{17\xspace}
\newcommand{\SeventhBoundOneAOneSelectionScoringLow}{13.0\xspace}
\newcommand{\SeventhBoundOneAOneSelectionScoringHigh}{22.0\xspace}
\newcommand{\SeventhBoundOneAOneSelectionFresh}{12\xspace}
\newcommand{\SeventhBoundOneAOneSelectionFreshLow}{10.0\xspace}
\newcommand{\SeventhBoundOneAOneSelectionFreshHigh}{17.0\xspace}
\newcommand{\SeventhBoundOneAOneSelectionStable}{4\xspace}
\newcommand{\SeventhBoundOneAOneSelectionStableLow}{2.0\xspace}
\newcommand{\SeventhBoundOneAOneSelectionStableHigh}{6.0\xspace}
\newcommand{\SeventhBoundOneAOneSameScoring}{14\xspace}
\newcommand{\SeventhBoundOneAOneSameScoringLow}{11.0\xspace}
\newcommand{\SeventhBoundOneAOneSameScoringHigh}{20.0\xspace}
\newcommand{\SeventhBoundOneAOneSameFresh}{15\xspace}
\newcommand{\SeventhBoundOneAOneSameFreshLow}{11.0\xspace}
\newcommand{\SeventhBoundOneAOneSameFreshHigh}{19.0\xspace}
\newcommand{\SeventhBoundOneAOneSameStable}{5\xspace}
\newcommand{\SeventhBoundOneAOneSameStableLow}{0.0\xspace}
\newcommand{\SeventhBoundOneAOneSameStableHigh}{5.0\xspace}
\newcommand{\SeventhBoundOneATwoSelectionScoring}{6\xspace}
\newcommand{\SeventhBoundOneATwoSelectionScoringLow}{3.0\xspace}
\newcommand{\SeventhBoundOneATwoSelectionScoringHigh}{8.0\xspace}
\newcommand{\SeventhBoundOneATwoSelectionFresh}{5\xspace}
\newcommand{\SeventhBoundOneATwoSelectionFreshLow}{2.0\xspace}
\newcommand{\SeventhBoundOneATwoSelectionFreshHigh}{8.0\xspace}
\newcommand{\SeventhBoundOneATwoSelectionStable}{3\xspace}
\newcommand{\SeventhBoundOneATwoSelectionStableLow}{0.0\xspace}
\newcommand{\SeventhBoundOneATwoSelectionStableHigh}{3.0\xspace}
\newcommand{\SeventhBoundOneATwoSameScoring}{4\xspace}
\newcommand{\SeventhBoundOneATwoSameScoringLow}{2.0\xspace}
\newcommand{\SeventhBoundOneATwoSameScoringHigh}{7.0\xspace}
\newcommand{\SeventhBoundOneATwoSameFresh}{5\xspace}
\newcommand{\SeventhBoundOneATwoSameFreshLow}{3.0\xspace}
\newcommand{\SeventhBoundOneATwoSameFreshHigh}{9.0\xspace}
\newcommand{\SeventhBoundOneATwoSameStable}{1\xspace}
\newcommand{\SeventhBoundOneATwoSameStableLow}{0.0\xspace}
\newcommand{\SeventhBoundOneATwoSameStableHigh}{2.0\xspace}
\newcommand{\SeventhBoundOneAThreeSelectionScoring}{7\xspace}
\newcommand{\SeventhBoundOneAThreeSelectionScoringLow}{3.0\xspace}
\newcommand{\SeventhBoundOneAThreeSelectionScoringHigh}{9.0\xspace}
\newcommand{\SeventhBoundOneAThreeSelectionFresh}{4\xspace}
\newcommand{\SeventhBoundOneAThreeSelectionFreshLow}{4.0\xspace}
\newcommand{\SeventhBoundOneAThreeSelectionFreshHigh}{7.0\xspace}
\newcommand{\SeventhBoundOneAThreeSelectionStable}{3\xspace}
\newcommand{\SeventhBoundOneAThreeSelectionStableLow}{0.0\xspace}
\newcommand{\SeventhBoundOneAThreeSelectionStableHigh}{4.0\xspace}
\newcommand{\SeventhBoundOneAThreeSameScoring}{7\xspace}
\newcommand{\SeventhBoundOneAThreeSameScoringLow}{3.0\xspace}
\newcommand{\SeventhBoundOneAThreeSameScoringHigh}{9.0\xspace}
\newcommand{\SeventhBoundOneAThreeSameFresh}{4\xspace}
\newcommand{\SeventhBoundOneAThreeSameFreshLow}{3.0\xspace}
\newcommand{\SeventhBoundOneAThreeSameFreshHigh}{7.0\xspace}
\newcommand{\SeventhBoundOneAThreeSameStable}{3\xspace}
\newcommand{\SeventhBoundOneAThreeSameStableLow}{0.0\xspace}
\newcommand{\SeventhBoundOneAThreeSameStableHigh}{4.0\xspace}
\newcommand{\SeventhBoundOneAFourSelectionScoring}{12\xspace}
\newcommand{\SeventhBoundOneAFourSelectionScoringLow}{8.0\xspace}
\newcommand{\SeventhBoundOneAFourSelectionScoringHigh}{16.0\xspace}
\newcommand{\SeventhBoundOneAFourSelectionFresh}{8\xspace}
\newcommand{\SeventhBoundOneAFourSelectionFreshLow}{6.0\xspace}
\newcommand{\SeventhBoundOneAFourSelectionFreshHigh}{12.0\xspace}
\newcommand{\SeventhBoundOneAFourSelectionStable}{5\xspace}
\newcommand{\SeventhBoundOneAFourSelectionStableLow}{2.0\xspace}
\newcommand{\SeventhBoundOneAFourSelectionStableHigh}{6.0\xspace}
\newcommand{\SeventhBoundOneAFourSameScoring}{11\xspace}
\newcommand{\SeventhBoundOneAFourSameScoringLow}{7.0\xspace}
\newcommand{\SeventhBoundOneAFourSameScoringHigh}{15.0\xspace}
\newcommand{\SeventhBoundOneAFourSameFresh}{6\xspace}
\newcommand{\SeventhBoundOneAFourSameFreshLow}{4.0\xspace}
\newcommand{\SeventhBoundOneAFourSameFreshHigh}{11.0\xspace}
\newcommand{\SeventhBoundOneAFourSameStable}{3\xspace}
\newcommand{\SeventhBoundOneAFourSameStableLow}{0.0\xspace}
\newcommand{\SeventhBoundOneAFourSameStableHigh}{4.0\xspace}
\newcommand{\SeventhBoundOneAFiveSelectionScoring}{10\xspace}
\newcommand{\SeventhBoundOneAFiveSelectionScoringLow}{5.0\xspace}
\newcommand{\SeventhBoundOneAFiveSelectionScoringHigh}{13.0\xspace}
\newcommand{\SeventhBoundOneAFiveSelectionFresh}{7\xspace}
\newcommand{\SeventhBoundOneAFiveSelectionFreshLow}{6.0\xspace}
\newcommand{\SeventhBoundOneAFiveSelectionFreshHigh}{10.0\xspace}
\newcommand{\SeventhBoundOneAFiveSelectionStable}{4\xspace}
\newcommand{\SeventhBoundOneAFiveSelectionStableLow}{2.0\xspace}
\newcommand{\SeventhBoundOneAFiveSelectionStableHigh}{6.0\xspace}
\newcommand{\SeventhBoundOneAFiveSameScoring}{10\xspace}
\newcommand{\SeventhBoundOneAFiveSameScoringLow}{5.0\xspace}
\newcommand{\SeventhBoundOneAFiveSameScoringHigh}{13.0\xspace}
\newcommand{\SeventhBoundOneAFiveSameFresh}{7\xspace}
\newcommand{\SeventhBoundOneAFiveSameFreshLow}{6.0\xspace}
\newcommand{\SeventhBoundOneAFiveSameFreshHigh}{10.0\xspace}
\newcommand{\SeventhBoundOneAFiveSameStable}{4\xspace}
\newcommand{\SeventhBoundOneAFiveSameStableLow}{1.0\xspace}
\newcommand{\SeventhBoundOneAFiveSameStableHigh}{5.0\xspace}
\newcommand{\SeventhBoundOneAAllSelectionScoring}{52\xspace}
\newcommand{\SeventhBoundOneAAllSelectionScoringLow}{41.0\xspace}
\newcommand{\SeventhBoundOneAAllSelectionScoringHigh}{59.0\xspace}
\newcommand{\SeventhBoundOneAAllSelectionFresh}{36\xspace}
\newcommand{\SeventhBoundOneAAllSelectionFreshLow}{34.0\xspace}
\newcommand{\SeventhBoundOneAAllSelectionFreshHigh}{47.0\xspace}
\newcommand{\SeventhBoundOneAAllSelectionStable}{19\xspace}
\newcommand{\SeventhBoundOneAAllSelectionStableLow}{10.0\xspace}
\newcommand{\SeventhBoundOneAAllSelectionStableHigh}{19.0\xspace}
\newcommand{\SeventhBoundOneAAllSameScoring}{46\xspace}
\newcommand{\SeventhBoundOneAAllSameScoringLow}{36.0\xspace}
\newcommand{\SeventhBoundOneAAllSameScoringHigh}{55.0\xspace}
\newcommand{\SeventhBoundOneAAllSameFresh}{37\xspace}
\newcommand{\SeventhBoundOneAAllSameFreshLow}{34.0\xspace}
\newcommand{\SeventhBoundOneAAllSameFreshHigh}{48.0\xspace}
\newcommand{\SeventhBoundOneAAllSameStable}{16\xspace}
\newcommand{\SeventhBoundOneAAllSameStableLow}{6.0\xspace}
\newcommand{\SeventhBoundOneAAllSameStableHigh}{15.0\xspace}
\newcommand{\SeventhBoundOneBOneSelectionScoring}{19\xspace}
\newcommand{\SeventhBoundOneBOneSelectionScoringLow}{13.0\xspace}
\newcommand{\SeventhBoundOneBOneSelectionScoringHigh}{27.0\xspace}
\newcommand{\SeventhBoundOneBOneSelectionFresh}{15\xspace}
\newcommand{\SeventhBoundOneBOneSelectionFreshLow}{12.0\xspace}
\newcommand{\SeventhBoundOneBOneSelectionFreshHigh}{22.0\xspace}
\newcommand{\SeventhBoundOneBOneSelectionStable}{3\xspace}
\newcommand{\SeventhBoundOneBOneSelectionStableLow}{2.0\xspace}
\newcommand{\SeventhBoundOneBOneSelectionStableHigh}{6.0\xspace}
\newcommand{\SeventhBoundOneBOneSameScoring}{11\xspace}
\newcommand{\SeventhBoundOneBOneSameScoringLow}{5.0\xspace}
\newcommand{\SeventhBoundOneBOneSameScoringHigh}{16.0\xspace}
\newcommand{\SeventhBoundOneBOneSameFresh}{10\xspace}
\newcommand{\SeventhBoundOneBOneSameFreshLow}{5.0\xspace}
\newcommand{\SeventhBoundOneBOneSameFreshHigh}{15.0\xspace}
\newcommand{\SeventhBoundOneBOneSameStable}{0\xspace}
\newcommand{\SeventhBoundOneBOneSameStableLow}{0.0\xspace}
\newcommand{\SeventhBoundOneBOneSameStableHigh}{1.0\xspace}
\newcommand{\SeventhBoundOneBTwoSelectionScoring}{8\xspace}
\newcommand{\SeventhBoundOneBTwoSelectionScoringLow}{5.0\xspace}
\newcommand{\SeventhBoundOneBTwoSelectionScoringHigh}{15.0\xspace}
\newcommand{\SeventhBoundOneBTwoSelectionFresh}{17\xspace}
\newcommand{\SeventhBoundOneBTwoSelectionFreshLow}{10.0\xspace}
\newcommand{\SeventhBoundOneBTwoSelectionFreshHigh}{23.0\xspace}
\newcommand{\SeventhBoundOneBTwoSelectionStable}{4\xspace}
\newcommand{\SeventhBoundOneBTwoSelectionStableLow}{2.0\xspace}
\newcommand{\SeventhBoundOneBTwoSelectionStableHigh}{6.0\xspace}
\newcommand{\SeventhBoundOneBTwoSameScoring}{5\xspace}
\newcommand{\SeventhBoundOneBTwoSameScoringLow}{2.0\xspace}
\newcommand{\SeventhBoundOneBTwoSameScoringHigh}{12.0\xspace}
\newcommand{\SeventhBoundOneBTwoSameFresh}{7\xspace}
\newcommand{\SeventhBoundOneBTwoSameFreshLow}{2.0\xspace}
\newcommand{\SeventhBoundOneBTwoSameFreshHigh}{13.0\xspace}
\newcommand{\SeventhBoundOneBTwoSameStable}{0\xspace}
\newcommand{\SeventhBoundOneBTwoSameStableLow}{0.0\xspace}
\newcommand{\SeventhBoundOneBTwoSameStableHigh}{1.0\xspace}
\newcommand{\SeventhBoundOneBThreeSelectionScoring}{11\xspace}
\newcommand{\SeventhBoundOneBThreeSelectionScoringLow}{6.0\xspace}
\newcommand{\SeventhBoundOneBThreeSelectionScoringHigh}{13.0\xspace}
\newcommand{\SeventhBoundOneBThreeSelectionFresh}{12\xspace}
\newcommand{\SeventhBoundOneBThreeSelectionFreshLow}{9.0\xspace}
\newcommand{\SeventhBoundOneBThreeSelectionFreshHigh}{19.0\xspace}
\newcommand{\SeventhBoundOneBThreeSelectionStable}{3\xspace}
\newcommand{\SeventhBoundOneBThreeSelectionStableLow}{1.0\xspace}
\newcommand{\SeventhBoundOneBThreeSelectionStableHigh}{5.0\xspace}
\newcommand{\SeventhBoundOneBThreeSameScoring}{8\xspace}
\newcommand{\SeventhBoundOneBThreeSameScoringLow}{3.0\xspace}
\newcommand{\SeventhBoundOneBThreeSameScoringHigh}{11.0\xspace}
\newcommand{\SeventhBoundOneBThreeSameFresh}{4\xspace}
\newcommand{\SeventhBoundOneBThreeSameFreshLow}{1.0\xspace}
\newcommand{\SeventhBoundOneBThreeSameFreshHigh}{9.0\xspace}
\newcommand{\SeventhBoundOneBThreeSameStable}{0\xspace}
\newcommand{\SeventhBoundOneBThreeSameStableLow}{0.0\xspace}
\newcommand{\SeventhBoundOneBThreeSameStableHigh}{1.0\xspace}
\newcommand{\SeventhBoundOneBFourSelectionScoring}{6\xspace}
\newcommand{\SeventhBoundOneBFourSelectionScoringLow}{3.0\xspace}
\newcommand{\SeventhBoundOneBFourSelectionScoringHigh}{9.0\xspace}
\newcommand{\SeventhBoundOneBFourSelectionFresh}{8\xspace}
\newcommand{\SeventhBoundOneBFourSelectionFreshLow}{6.0\xspace}
\newcommand{\SeventhBoundOneBFourSelectionFreshHigh}{18.0\xspace}
\newcommand{\SeventhBoundOneBFourSelectionStable}{2\xspace}
\newcommand{\SeventhBoundOneBFourSelectionStableLow}{0.0\xspace}
\newcommand{\SeventhBoundOneBFourSelectionStableHigh}{3.0\xspace}
\newcommand{\SeventhBoundOneBFourSameScoring}{2\xspace}
\newcommand{\SeventhBoundOneBFourSameScoringLow}{1.0\xspace}
\newcommand{\SeventhBoundOneBFourSameScoringHigh}{7.0\xspace}
\newcommand{\SeventhBoundOneBFourSameFresh}{5\xspace}
\newcommand{\SeventhBoundOneBFourSameFreshLow}{1.0\xspace}
\newcommand{\SeventhBoundOneBFourSameFreshHigh}{9.0\xspace}
\newcommand{\SeventhBoundOneBFourSameStable}{0\xspace}
\newcommand{\SeventhBoundOneBFourSameStableLow}{0.0\xspace}
\newcommand{\SeventhBoundOneBFourSameStableHigh}{0.0\xspace}
\newcommand{\SeventhBoundOneBFiveSelectionScoring}{8\xspace}
\newcommand{\SeventhBoundOneBFiveSelectionScoringLow}{7.0\xspace}
\newcommand{\SeventhBoundOneBFiveSelectionScoringHigh}{17.0\xspace}
\newcommand{\SeventhBoundOneBFiveSelectionFresh}{11\xspace}
\newcommand{\SeventhBoundOneBFiveSelectionFreshLow}{7.0\xspace}
\newcommand{\SeventhBoundOneBFiveSelectionFreshHigh}{15.0\xspace}
\newcommand{\SeventhBoundOneBFiveSelectionStable}{4\xspace}
\newcommand{\SeventhBoundOneBFiveSelectionStableLow}{0.0\xspace}
\newcommand{\SeventhBoundOneBFiveSelectionStableHigh}{5.0\xspace}
\newcommand{\SeventhBoundOneBFiveSameScoring}{10\xspace}
\newcommand{\SeventhBoundOneBFiveSameScoringLow}{5.0\xspace}
\newcommand{\SeventhBoundOneBFiveSameScoringHigh}{14.0\xspace}
\newcommand{\SeventhBoundOneBFiveSameFresh}{7\xspace}
\newcommand{\SeventhBoundOneBFiveSameFreshLow}{5.0\xspace}
\newcommand{\SeventhBoundOneBFiveSameFreshHigh}{12.0\xspace}
\newcommand{\SeventhBoundOneBFiveSameStable}{2\xspace}
\newcommand{\SeventhBoundOneBFiveSameStableLow}{0.0\xspace}
\newcommand{\SeventhBoundOneBFiveSameStableHigh}{2.0\xspace}
\newcommand{\SeventhBoundOneBAllSelectionScoring}{52\xspace}
\newcommand{\SeventhBoundOneBAllSelectionScoringLow}{45.0\xspace}
\newcommand{\SeventhBoundOneBAllSelectionScoringHigh}{68.0\xspace}
\newcommand{\SeventhBoundOneBAllSelectionFresh}{63\xspace}
\newcommand{\SeventhBoundOneBAllSelectionFreshLow}{57.0\xspace}
\newcommand{\SeventhBoundOneBAllSelectionFreshHigh}{82.0\xspace}
\newcommand{\SeventhBoundOneBAllSelectionStable}{16\xspace}
\newcommand{\SeventhBoundOneBAllSelectionStableLow}{8.0\xspace}
\newcommand{\SeventhBoundOneBAllSelectionStableHigh}{18.0\xspace}
\newcommand{\SeventhBoundOneBAllSameScoring}{36\xspace}
\newcommand{\SeventhBoundOneBAllSameScoringLow}{27.0\xspace}
\newcommand{\SeventhBoundOneBAllSameScoringHigh}{47.0\xspace}
\newcommand{\SeventhBoundOneBAllSameFresh}{33\xspace}
\newcommand{\SeventhBoundOneBAllSameFreshLow}{25.0\xspace}
\newcommand{\SeventhBoundOneBAllSameFreshHigh}{46.0\xspace}
\newcommand{\SeventhBoundOneBAllSameStable}{2\xspace}
\newcommand{\SeventhBoundOneBAllSameStableLow}{0.0\xspace}
\newcommand{\SeventhBoundOneBAllSameStableHigh}{2.0\xspace}
\newcommand{\SeventhBoundOneCOneSelectionScoring}{8\xspace}
\newcommand{\SeventhBoundOneCOneSelectionScoringLow}{7.0\xspace}
\newcommand{\SeventhBoundOneCOneSelectionScoringHigh}{17.0\xspace}
\newcommand{\SeventhBoundOneCOneSelectionFresh}{6\xspace}
\newcommand{\SeventhBoundOneCOneSelectionFreshLow}{4.0\xspace}
\newcommand{\SeventhBoundOneCOneSelectionFreshHigh}{12.0\xspace}
\newcommand{\SeventhBoundOneCOneSelectionStable}{2\xspace}
\newcommand{\SeventhBoundOneCOneSelectionStableLow}{0.0\xspace}
\newcommand{\SeventhBoundOneCOneSelectionStableHigh}{3.0\xspace}
\newcommand{\SeventhBoundOneCOneSameScoring}{5\xspace}
\newcommand{\SeventhBoundOneCOneSameScoringLow}{3.0\xspace}
\newcommand{\SeventhBoundOneCOneSameScoringHigh}{11.0\xspace}
\newcommand{\SeventhBoundOneCOneSameFresh}{6\xspace}
\newcommand{\SeventhBoundOneCOneSameFreshLow}{3.0\xspace}
\newcommand{\SeventhBoundOneCOneSameFreshHigh}{10.0\xspace}
\newcommand{\SeventhBoundOneCOneSameStable}{0\xspace}
\newcommand{\SeventhBoundOneCOneSameStableLow}{0.0\xspace}
\newcommand{\SeventhBoundOneCOneSameStableHigh}{1.0\xspace}
\newcommand{\SeventhBoundOneCTwoSelectionScoring}{10\xspace}
\newcommand{\SeventhBoundOneCTwoSelectionScoringLow}{8.0\xspace}
\newcommand{\SeventhBoundOneCTwoSelectionScoringHigh}{15.0\xspace}
\newcommand{\SeventhBoundOneCTwoSelectionFresh}{11\xspace}
\newcommand{\SeventhBoundOneCTwoSelectionFreshLow}{7.0\xspace}
\newcommand{\SeventhBoundOneCTwoSelectionFreshHigh}{16.0\xspace}
\newcommand{\SeventhBoundOneCTwoSelectionStable}{1\xspace}
\newcommand{\SeventhBoundOneCTwoSelectionStableLow}{0.0\xspace}
\newcommand{\SeventhBoundOneCTwoSelectionStableHigh}{3.0\xspace}
\newcommand{\SeventhBoundOneCTwoSameScoring}{5\xspace}
\newcommand{\SeventhBoundOneCTwoSameScoringLow}{3.0\xspace}
\newcommand{\SeventhBoundOneCTwoSameScoringHigh}{12.0\xspace}
\newcommand{\SeventhBoundOneCTwoSameFresh}{10\xspace}
\newcommand{\SeventhBoundOneCTwoSameFreshLow}{4.0\xspace}
\newcommand{\SeventhBoundOneCTwoSameFreshHigh}{14.0\xspace}
\newcommand{\SeventhBoundOneCTwoSameStable}{0\xspace}
\newcommand{\SeventhBoundOneCTwoSameStableLow}{0.0\xspace}
\newcommand{\SeventhBoundOneCTwoSameStableHigh}{1.0\xspace}
\newcommand{\SeventhBoundOneCThreeSelectionScoring}{11\xspace}
\newcommand{\SeventhBoundOneCThreeSelectionScoringLow}{7.0\xspace}
\newcommand{\SeventhBoundOneCThreeSelectionScoringHigh}{16.0\xspace}
\newcommand{\SeventhBoundOneCThreeSelectionFresh}{8\xspace}
\newcommand{\SeventhBoundOneCThreeSelectionFreshLow}{4.0\xspace}
\newcommand{\SeventhBoundOneCThreeSelectionFreshHigh}{13.0\xspace}
\newcommand{\SeventhBoundOneCThreeSelectionStable}{1\xspace}
\newcommand{\SeventhBoundOneCThreeSelectionStableLow}{0.0\xspace}
\newcommand{\SeventhBoundOneCThreeSelectionStableHigh}{3.0\xspace}
\newcommand{\SeventhBoundOneCThreeSameScoring}{7\xspace}
\newcommand{\SeventhBoundOneCThreeSameScoringLow}{4.0\xspace}
\newcommand{\SeventhBoundOneCThreeSameScoringHigh}{11.0\xspace}
\newcommand{\SeventhBoundOneCThreeSameFresh}{9\xspace}
\newcommand{\SeventhBoundOneCThreeSameFreshLow}{4.0\xspace}
\newcommand{\SeventhBoundOneCThreeSameFreshHigh}{12.0\xspace}
\newcommand{\SeventhBoundOneCThreeSameStable}{1\xspace}
\newcommand{\SeventhBoundOneCThreeSameStableLow}{0.0\xspace}
\newcommand{\SeventhBoundOneCThreeSameStableHigh}{2.0\xspace}
\newcommand{\SeventhBoundOneCFourSelectionScoring}{11\xspace}
\newcommand{\SeventhBoundOneCFourSelectionScoringLow}{8.0\xspace}
\newcommand{\SeventhBoundOneCFourSelectionScoringHigh}{18.0\xspace}
\newcommand{\SeventhBoundOneCFourSelectionFresh}{11\xspace}
\newcommand{\SeventhBoundOneCFourSelectionFreshLow}{8.0\xspace}
\newcommand{\SeventhBoundOneCFourSelectionFreshHigh}{16.0\xspace}
\newcommand{\SeventhBoundOneCFourSelectionStable}{2\xspace}
\newcommand{\SeventhBoundOneCFourSelectionStableLow}{0.0\xspace}
\newcommand{\SeventhBoundOneCFourSelectionStableHigh}{4.0\xspace}
\newcommand{\SeventhBoundOneCFourSameScoring}{3\xspace}
\newcommand{\SeventhBoundOneCFourSameScoringLow}{3.0\xspace}
\newcommand{\SeventhBoundOneCFourSameScoringHigh}{11.0\xspace}
\newcommand{\SeventhBoundOneCFourSameFresh}{6\xspace}
\newcommand{\SeventhBoundOneCFourSameFreshLow}{3.0\xspace}
\newcommand{\SeventhBoundOneCFourSameFreshHigh}{11.0\xspace}
\newcommand{\SeventhBoundOneCFourSameStable}{1\xspace}
\newcommand{\SeventhBoundOneCFourSameStableLow}{0.0\xspace}
\newcommand{\SeventhBoundOneCFourSameStableHigh}{1.0\xspace}
\newcommand{\SeventhBoundOneCFiveSelectionScoring}{10\xspace}
\newcommand{\SeventhBoundOneCFiveSelectionScoringLow}{7.0\xspace}
\newcommand{\SeventhBoundOneCFiveSelectionScoringHigh}{17.0\xspace}
\newcommand{\SeventhBoundOneCFiveSelectionFresh}{6\xspace}
\newcommand{\SeventhBoundOneCFiveSelectionFreshLow}{5.0\xspace}
\newcommand{\SeventhBoundOneCFiveSelectionFreshHigh}{14.0\xspace}
\newcommand{\SeventhBoundOneCFiveSelectionStable}{1\xspace}
\newcommand{\SeventhBoundOneCFiveSelectionStableLow}{0.0\xspace}
\newcommand{\SeventhBoundOneCFiveSelectionStableHigh}{3.0\xspace}
\newcommand{\SeventhBoundOneCFiveSameScoring}{7\xspace}
\newcommand{\SeventhBoundOneCFiveSameScoringLow}{5.0\xspace}
\newcommand{\SeventhBoundOneCFiveSameScoringHigh}{14.0\xspace}
\newcommand{\SeventhBoundOneCFiveSameFresh}{4\xspace}
\newcommand{\SeventhBoundOneCFiveSameFreshLow}{4.0\xspace}
\newcommand{\SeventhBoundOneCFiveSameFreshHigh}{11.0\xspace}
\newcommand{\SeventhBoundOneCFiveSameStable}{1\xspace}
\newcommand{\SeventhBoundOneCFiveSameStableLow}{0.0\xspace}
\newcommand{\SeventhBoundOneCFiveSameStableHigh}{2.0\xspace}
\newcommand{\SeventhBoundOneCAllSelectionScoring}{50\xspace}
\newcommand{\SeventhBoundOneCAllSelectionScoringLow}{49.0\xspace}
\newcommand{\SeventhBoundOneCAllSelectionScoringHigh}{70.0\xspace}
\newcommand{\SeventhBoundOneCAllSelectionFresh}{42\xspace}
\newcommand{\SeventhBoundOneCAllSelectionFreshLow}{39.0\xspace}
\newcommand{\SeventhBoundOneCAllSelectionFreshHigh}{60.0\xspace}
\newcommand{\SeventhBoundOneCAllSelectionStable}{7\xspace}
\newcommand{\SeventhBoundOneCAllSelectionStableLow}{4.0\xspace}
\newcommand{\SeventhBoundOneCAllSelectionStableHigh}{11.0\xspace}
\newcommand{\SeventhBoundOneCAllSameScoring}{27\xspace}
\newcommand{\SeventhBoundOneCAllSameScoringLow}{27.0\xspace}
\newcommand{\SeventhBoundOneCAllSameScoringHigh}{47.0\xspace}
\newcommand{\SeventhBoundOneCAllSameFresh}{35\xspace}
\newcommand{\SeventhBoundOneCAllSameFreshLow}{27.0\xspace}
\newcommand{\SeventhBoundOneCAllSameFreshHigh}{46.0\xspace}
\newcommand{\SeventhBoundOneCAllSameStable}{3\xspace}
\newcommand{\SeventhBoundOneCAllSameStableLow}{0.0\xspace}
\newcommand{\SeventhBoundOneCAllSameStableHigh}{4.0\xspace}

\newcommand{\SeventhBoundTwoAOneSelectionScoring}{10\xspace}
\newcommand{\SeventhBoundTwoAOneSelectionScoringLow}{7.0\xspace}
\newcommand{\SeventhBoundTwoAOneSelectionScoringHigh}{15.0\xspace}
\newcommand{\SeventhBoundTwoAOneSelectionFresh}{15\xspace}
\newcommand{\SeventhBoundTwoAOneSelectionFreshLow}{11.0\xspace}
\newcommand{\SeventhBoundTwoAOneSelectionFreshHigh}{17.0\xspace}
\newcommand{\SeventhBoundTwoAOneSelectionStable}{6\xspace}
\newcommand{\SeventhBoundTwoAOneSelectionStableLow}{2.0\xspace}
\newcommand{\SeventhBoundTwoAOneSelectionStableHigh}{7.0\xspace}
\newcommand{\SeventhBoundTwoAOneSameScoring}{7\xspace}
\newcommand{\SeventhBoundTwoAOneSameScoringLow}{5.0\xspace}
\newcommand{\SeventhBoundTwoAOneSameScoringHigh}{13.0\xspace}
\newcommand{\SeventhBoundTwoAOneSameFresh}{13\xspace}
\newcommand{\SeventhBoundTwoAOneSameFreshLow}{9.0\xspace}
\newcommand{\SeventhBoundTwoAOneSameFreshHigh}{16.0\xspace}
\newcommand{\SeventhBoundTwoAOneSameStable}{1\xspace}
\newcommand{\SeventhBoundTwoAOneSameStableLow}{0.0\xspace}
\newcommand{\SeventhBoundTwoAOneSameStableHigh}{3.0\xspace}
\newcommand{\SeventhBoundTwoATwoSelectionScoring}{5\xspace}
\newcommand{\SeventhBoundTwoATwoSelectionScoringLow}{1.0\xspace}
\newcommand{\SeventhBoundTwoATwoSelectionScoringHigh}{6.0\xspace}
\newcommand{\SeventhBoundTwoATwoSelectionFresh}{3\xspace}
\newcommand{\SeventhBoundTwoATwoSelectionFreshLow}{0.0\xspace}
\newcommand{\SeventhBoundTwoATwoSelectionFreshHigh}{4.0\xspace}
\newcommand{\SeventhBoundTwoATwoSelectionStable}{0\xspace}
\newcommand{\SeventhBoundTwoATwoSelectionStableLow}{0.0\xspace}
\newcommand{\SeventhBoundTwoATwoSelectionStableHigh}{1.0\xspace}
\newcommand{\SeventhBoundTwoATwoSameScoring}{5\xspace}
\newcommand{\SeventhBoundTwoATwoSameScoringLow}{1.0\xspace}
\newcommand{\SeventhBoundTwoATwoSameScoringHigh}{6.0\xspace}
\newcommand{\SeventhBoundTwoATwoSameFresh}{3\xspace}
\newcommand{\SeventhBoundTwoATwoSameFreshLow}{0.0\xspace}
\newcommand{\SeventhBoundTwoATwoSameFreshHigh}{4.0\xspace}
\newcommand{\SeventhBoundTwoATwoSameStable}{0\xspace}
\newcommand{\SeventhBoundTwoATwoSameStableLow}{0.0\xspace}
\newcommand{\SeventhBoundTwoATwoSameStableHigh}{1.0\xspace}
\newcommand{\SeventhBoundTwoAThreeSelectionScoring}{6\xspace}
\newcommand{\SeventhBoundTwoAThreeSelectionScoringLow}{4.0\xspace}
\newcommand{\SeventhBoundTwoAThreeSelectionScoringHigh}{11.0\xspace}
\newcommand{\SeventhBoundTwoAThreeSelectionFresh}{8\xspace}
\newcommand{\SeventhBoundTwoAThreeSelectionFreshLow}{5.0\xspace}
\newcommand{\SeventhBoundTwoAThreeSelectionFreshHigh}{10.0\xspace}
\newcommand{\SeventhBoundTwoAThreeSelectionStable}{2\xspace}
\newcommand{\SeventhBoundTwoAThreeSelectionStableLow}{1.0\xspace}
\newcommand{\SeventhBoundTwoAThreeSelectionStableHigh}{4.0\xspace}
\newcommand{\SeventhBoundTwoAThreeSameScoring}{6\xspace}
\newcommand{\SeventhBoundTwoAThreeSameScoringLow}{4.0\xspace}
\newcommand{\SeventhBoundTwoAThreeSameScoringHigh}{11.0\xspace}
\newcommand{\SeventhBoundTwoAThreeSameFresh}{8\xspace}
\newcommand{\SeventhBoundTwoAThreeSameFreshLow}{5.0\xspace}
\newcommand{\SeventhBoundTwoAThreeSameFreshHigh}{10.0\xspace}
\newcommand{\SeventhBoundTwoAThreeSameStable}{2\xspace}
\newcommand{\SeventhBoundTwoAThreeSameStableLow}{1.0\xspace}
\newcommand{\SeventhBoundTwoAThreeSameStableHigh}{4.0\xspace}
\newcommand{\SeventhBoundTwoAFourSelectionScoring}{8\xspace}
\newcommand{\SeventhBoundTwoAFourSelectionScoringLow}{5.0\xspace}
\newcommand{\SeventhBoundTwoAFourSelectionScoringHigh}{11.0\xspace}
\newcommand{\SeventhBoundTwoAFourSelectionFresh}{8\xspace}
\newcommand{\SeventhBoundTwoAFourSelectionFreshLow}{6.0\xspace}
\newcommand{\SeventhBoundTwoAFourSelectionFreshHigh}{12.0\xspace}
\newcommand{\SeventhBoundTwoAFourSelectionStable}{3\xspace}
\newcommand{\SeventhBoundTwoAFourSelectionStableLow}{1.0\xspace}
\newcommand{\SeventhBoundTwoAFourSelectionStableHigh}{5.0\xspace}
\newcommand{\SeventhBoundTwoAFourSameScoring}{6\xspace}
\newcommand{\SeventhBoundTwoAFourSameScoringLow}{3.0\xspace}
\newcommand{\SeventhBoundTwoAFourSameScoringHigh}{9.0\xspace}
\newcommand{\SeventhBoundTwoAFourSameFresh}{6\xspace}
\newcommand{\SeventhBoundTwoAFourSameFreshLow}{4.0\xspace}
\newcommand{\SeventhBoundTwoAFourSameFreshHigh}{11.0\xspace}
\newcommand{\SeventhBoundTwoAFourSameStable}{1\xspace}
\newcommand{\SeventhBoundTwoAFourSameStableLow}{0.0\xspace}
\newcommand{\SeventhBoundTwoAFourSameStableHigh}{2.0\xspace}
\newcommand{\SeventhBoundTwoAFiveSelectionScoring}{5\xspace}
\newcommand{\SeventhBoundTwoAFiveSelectionScoringLow}{2.0\xspace}
\newcommand{\SeventhBoundTwoAFiveSelectionScoringHigh}{11.0\xspace}
\newcommand{\SeventhBoundTwoAFiveSelectionFresh}{8\xspace}
\newcommand{\SeventhBoundTwoAFiveSelectionFreshLow}{7.0\xspace}
\newcommand{\SeventhBoundTwoAFiveSelectionFreshHigh}{13.0\xspace}
\newcommand{\SeventhBoundTwoAFiveSelectionStable}{3\xspace}
\newcommand{\SeventhBoundTwoAFiveSelectionStableLow}{0.0\xspace}
\newcommand{\SeventhBoundTwoAFiveSelectionStableHigh}{4.0\xspace}
\newcommand{\SeventhBoundTwoAFiveSameScoring}{4\xspace}
\newcommand{\SeventhBoundTwoAFiveSameScoringLow}{1.0\xspace}
\newcommand{\SeventhBoundTwoAFiveSameScoringHigh}{10.0\xspace}
\newcommand{\SeventhBoundTwoAFiveSameFresh}{7\xspace}
\newcommand{\SeventhBoundTwoAFiveSameFreshLow}{5.0\xspace}
\newcommand{\SeventhBoundTwoAFiveSameFreshHigh}{12.0\xspace}
\newcommand{\SeventhBoundTwoAFiveSameStable}{1\xspace}
\newcommand{\SeventhBoundTwoAFiveSameStableLow}{0.0\xspace}
\newcommand{\SeventhBoundTwoAFiveSameStableHigh}{2.0\xspace}
\newcommand{\SeventhBoundTwoAAllSelectionScoring}{34\xspace}
\newcommand{\SeventhBoundTwoAAllSelectionScoringLow}{28.0\xspace}
\newcommand{\SeventhBoundTwoAAllSelectionScoringHigh}{44.0\xspace}
\newcommand{\SeventhBoundTwoAAllSelectionFresh}{42\xspace}
\newcommand{\SeventhBoundTwoAAllSelectionFreshLow}{37.0\xspace}
\newcommand{\SeventhBoundTwoAAllSelectionFreshHigh}{49.0\xspace}
\newcommand{\SeventhBoundTwoAAllSelectionStable}{14\xspace}
\newcommand{\SeventhBoundTwoAAllSelectionStableLow}{7.0\xspace}
\newcommand{\SeventhBoundTwoAAllSelectionStableHigh}{16.0\xspace}
\newcommand{\SeventhBoundTwoAAllSameScoring}{28\xspace}
\newcommand{\SeventhBoundTwoAAllSameScoringLow}{22.0\xspace}
\newcommand{\SeventhBoundTwoAAllSameScoringHigh}{39.0\xspace}
\newcommand{\SeventhBoundTwoAAllSameFresh}{37\xspace}
\newcommand{\SeventhBoundTwoAAllSameFreshLow}{30.0\xspace}
\newcommand{\SeventhBoundTwoAAllSameFreshHigh}{44.0\xspace}
\newcommand{\SeventhBoundTwoAAllSameStable}{5\xspace}
\newcommand{\SeventhBoundTwoAAllSameStableLow}{2.0\xspace}
\newcommand{\SeventhBoundTwoAAllSameStableHigh}{8.0\xspace}
\newcommand{\SeventhBoundTwoBOneSelectionScoring}{5\xspace}
\newcommand{\SeventhBoundTwoBOneSelectionScoringLow}{2.0\xspace}
\newcommand{\SeventhBoundTwoBOneSelectionScoringHigh}{9.0\xspace}
\newcommand{\SeventhBoundTwoBOneSelectionFresh}{3\xspace}
\newcommand{\SeventhBoundTwoBOneSelectionFreshLow}{1.0\xspace}
\newcommand{\SeventhBoundTwoBOneSelectionFreshHigh}{6.0\xspace}
\newcommand{\SeventhBoundTwoBOneSelectionStable}{0\xspace}
\newcommand{\SeventhBoundTwoBOneSelectionStableLow}{0.0\xspace}
\newcommand{\SeventhBoundTwoBOneSelectionStableHigh}{1.0\xspace}
\newcommand{\SeventhBoundTwoBOneSameScoring}{3\xspace}
\newcommand{\SeventhBoundTwoBOneSameScoringLow}{1.0\xspace}
\newcommand{\SeventhBoundTwoBOneSameScoringHigh}{7.0\xspace}
\newcommand{\SeventhBoundTwoBOneSameFresh}{3\xspace}
\newcommand{\SeventhBoundTwoBOneSameFreshLow}{0.0\xspace}
\newcommand{\SeventhBoundTwoBOneSameFreshHigh}{7.0\xspace}
\newcommand{\SeventhBoundTwoBOneSameStable}{0\xspace}
\newcommand{\SeventhBoundTwoBOneSameStableLow}{0.0\xspace}
\newcommand{\SeventhBoundTwoBOneSameStableHigh}{0.0\xspace}
\newcommand{\SeventhBoundTwoBTwoSelectionScoring}{3\xspace}
\newcommand{\SeventhBoundTwoBTwoSelectionScoringLow}{2.0\xspace}
\newcommand{\SeventhBoundTwoBTwoSelectionScoringHigh}{8.0\xspace}
\newcommand{\SeventhBoundTwoBTwoSelectionFresh}{7\xspace}
\newcommand{\SeventhBoundTwoBTwoSelectionFreshLow}{3.0\xspace}
\newcommand{\SeventhBoundTwoBTwoSelectionFreshHigh}{10.0\xspace}
\newcommand{\SeventhBoundTwoBTwoSelectionStable}{1\xspace}
\newcommand{\SeventhBoundTwoBTwoSelectionStableLow}{0.0\xspace}
\newcommand{\SeventhBoundTwoBTwoSelectionStableHigh}{2.0\xspace}
\newcommand{\SeventhBoundTwoBTwoSameScoring}{2\xspace}
\newcommand{\SeventhBoundTwoBTwoSameScoringLow}{0.0\xspace}
\newcommand{\SeventhBoundTwoBTwoSameScoringHigh}{6.0\xspace}
\newcommand{\SeventhBoundTwoBTwoSameFresh}{3\xspace}
\newcommand{\SeventhBoundTwoBTwoSameFreshLow}{0.0\xspace}
\newcommand{\SeventhBoundTwoBTwoSameFreshHigh}{6.0\xspace}
\newcommand{\SeventhBoundTwoBTwoSameStable}{0\xspace}
\newcommand{\SeventhBoundTwoBTwoSameStableLow}{0.0\xspace}
\newcommand{\SeventhBoundTwoBTwoSameStableHigh}{0.0\xspace}
\newcommand{\SeventhBoundTwoBThreeSelectionScoring}{4\xspace}
\newcommand{\SeventhBoundTwoBThreeSelectionScoringLow}{3.0\xspace}
\newcommand{\SeventhBoundTwoBThreeSelectionScoringHigh}{10.0\xspace}
\newcommand{\SeventhBoundTwoBThreeSelectionFresh}{5\xspace}
\newcommand{\SeventhBoundTwoBThreeSelectionFreshLow}{3.0\xspace}
\newcommand{\SeventhBoundTwoBThreeSelectionFreshHigh}{11.0\xspace}
\newcommand{\SeventhBoundTwoBThreeSelectionStable}{1\xspace}
\newcommand{\SeventhBoundTwoBThreeSelectionStableLow}{0.0\xspace}
\newcommand{\SeventhBoundTwoBThreeSelectionStableHigh}{3.0\xspace}
\newcommand{\SeventhBoundTwoBThreeSameScoring}{5\xspace}
\newcommand{\SeventhBoundTwoBThreeSameScoringLow}{3.0\xspace}
\newcommand{\SeventhBoundTwoBThreeSameScoringHigh}{10.0\xspace}
\newcommand{\SeventhBoundTwoBThreeSameFresh}{8\xspace}
\newcommand{\SeventhBoundTwoBThreeSameFreshLow}{3.0\xspace}
\newcommand{\SeventhBoundTwoBThreeSameFreshHigh}{9.0\xspace}
\newcommand{\SeventhBoundTwoBThreeSameStable}{1\xspace}
\newcommand{\SeventhBoundTwoBThreeSameStableLow}{0.0\xspace}
\newcommand{\SeventhBoundTwoBThreeSameStableHigh}{2.0\xspace}
\newcommand{\SeventhBoundTwoBFourSelectionScoring}{9\xspace}
\newcommand{\SeventhBoundTwoBFourSelectionScoringLow}{6.0\xspace}
\newcommand{\SeventhBoundTwoBFourSelectionScoringHigh}{12.0\xspace}
\newcommand{\SeventhBoundTwoBFourSelectionFresh}{9\xspace}
\newcommand{\SeventhBoundTwoBFourSelectionFreshLow}{6.0\xspace}
\newcommand{\SeventhBoundTwoBFourSelectionFreshHigh}{14.0\xspace}
\newcommand{\SeventhBoundTwoBFourSelectionStable}{3\xspace}
\newcommand{\SeventhBoundTwoBFourSelectionStableLow}{0.0\xspace}
\newcommand{\SeventhBoundTwoBFourSelectionStableHigh}{4.0\xspace}
\newcommand{\SeventhBoundTwoBFourSameScoring}{7\xspace}
\newcommand{\SeventhBoundTwoBFourSameScoringLow}{2.0\xspace}
\newcommand{\SeventhBoundTwoBFourSameScoringHigh}{10.0\xspace}
\newcommand{\SeventhBoundTwoBFourSameFresh}{2\xspace}
\newcommand{\SeventhBoundTwoBFourSameFreshLow}{1.0\xspace}
\newcommand{\SeventhBoundTwoBFourSameFreshHigh}{7.0\xspace}
\newcommand{\SeventhBoundTwoBFourSameStable}{0\xspace}
\newcommand{\SeventhBoundTwoBFourSameStableLow}{0.0\xspace}
\newcommand{\SeventhBoundTwoBFourSameStableHigh}{0.0\xspace}
\newcommand{\SeventhBoundTwoBFiveSelectionScoring}{2\xspace}
\newcommand{\SeventhBoundTwoBFiveSelectionScoringLow}{1.0\xspace}
\newcommand{\SeventhBoundTwoBFiveSelectionScoringHigh}{7.0\xspace}
\newcommand{\SeventhBoundTwoBFiveSelectionFresh}{3\xspace}
\newcommand{\SeventhBoundTwoBFiveSelectionFreshLow}{1.0\xspace}
\newcommand{\SeventhBoundTwoBFiveSelectionFreshHigh}{6.0\xspace}
\newcommand{\SeventhBoundTwoBFiveSelectionStable}{1\xspace}
\newcommand{\SeventhBoundTwoBFiveSelectionStableLow}{0.0\xspace}
\newcommand{\SeventhBoundTwoBFiveSelectionStableHigh}{2.0\xspace}
\newcommand{\SeventhBoundTwoBFiveSameScoring}{2\xspace}
\newcommand{\SeventhBoundTwoBFiveSameScoringLow}{0.0\xspace}
\newcommand{\SeventhBoundTwoBFiveSameScoringHigh}{4.0\xspace}
\newcommand{\SeventhBoundTwoBFiveSameFresh}{4\xspace}
\newcommand{\SeventhBoundTwoBFiveSameFreshLow}{1.0\xspace}
\newcommand{\SeventhBoundTwoBFiveSameFreshHigh}{7.0\xspace}
\newcommand{\SeventhBoundTwoBFiveSameStable}{0\xspace}
\newcommand{\SeventhBoundTwoBFiveSameStableLow}{0.0\xspace}
\newcommand{\SeventhBoundTwoBFiveSameStableHigh}{0.0\xspace}
\newcommand{\SeventhBoundTwoBAllSelectionScoring}{23\xspace}
\newcommand{\SeventhBoundTwoBAllSelectionScoringLow}{21.0\xspace}
\newcommand{\SeventhBoundTwoBAllSelectionScoringHigh}{36.0\xspace}
\newcommand{\SeventhBoundTwoBAllSelectionFresh}{27\xspace}
\newcommand{\SeventhBoundTwoBAllSelectionFreshLow}{23.0\xspace}
\newcommand{\SeventhBoundTwoBAllSelectionFreshHigh}{38.0\xspace}
\newcommand{\SeventhBoundTwoBAllSelectionStable}{6\xspace}
\newcommand{\SeventhBoundTwoBAllSelectionStableLow}{2.0\xspace}
\newcommand{\SeventhBoundTwoBAllSelectionStableHigh}{8.0\xspace}
\newcommand{\SeventhBoundTwoBAllSameScoring}{19\xspace}
\newcommand{\SeventhBoundTwoBAllSameScoringLow}{13.0\xspace}
\newcommand{\SeventhBoundTwoBAllSameScoringHigh}{28.0\xspace}
\newcommand{\SeventhBoundTwoBAllSameFresh}{20\xspace}
\newcommand{\SeventhBoundTwoBAllSameFreshLow}{11.0\xspace}
\newcommand{\SeventhBoundTwoBAllSameFreshHigh}{26.0\xspace}
\newcommand{\SeventhBoundTwoBAllSameStable}{1\xspace}
\newcommand{\SeventhBoundTwoBAllSameStableLow}{0.0\xspace}
\newcommand{\SeventhBoundTwoBAllSameStableHigh}{2.0\xspace}
\newcommand{\SeventhBoundTwoCOneSelectionScoring}{10\xspace}
\newcommand{\SeventhBoundTwoCOneSelectionScoringLow}{6.0\xspace}
\newcommand{\SeventhBoundTwoCOneSelectionScoringHigh}{13.0\xspace}
\newcommand{\SeventhBoundTwoCOneSelectionFresh}{3\xspace}
\newcommand{\SeventhBoundTwoCOneSelectionFreshLow}{2.0\xspace}
\newcommand{\SeventhBoundTwoCOneSelectionFreshHigh}{8.0\xspace}
\newcommand{\SeventhBoundTwoCOneSelectionStable}{0\xspace}
\newcommand{\SeventhBoundTwoCOneSelectionStableLow}{0.0\xspace}
\newcommand{\SeventhBoundTwoCOneSelectionStableHigh}{2.0\xspace}
\newcommand{\SeventhBoundTwoCOneSameScoring}{3\xspace}
\newcommand{\SeventhBoundTwoCOneSameScoringLow}{2.0\xspace}
\newcommand{\SeventhBoundTwoCOneSameScoringHigh}{8.0\xspace}
\newcommand{\SeventhBoundTwoCOneSameFresh}{1\xspace}
\newcommand{\SeventhBoundTwoCOneSameFreshLow}{1.0\xspace}
\newcommand{\SeventhBoundTwoCOneSameFreshHigh}{6.0\xspace}
\newcommand{\SeventhBoundTwoCOneSameStable}{0\xspace}
\newcommand{\SeventhBoundTwoCOneSameStableLow}{0.0\xspace}
\newcommand{\SeventhBoundTwoCOneSameStableHigh}{1.0\xspace}
\newcommand{\SeventhBoundTwoCTwoSelectionScoring}{6\xspace}
\newcommand{\SeventhBoundTwoCTwoSelectionScoringLow}{4.0\xspace}
\newcommand{\SeventhBoundTwoCTwoSelectionScoringHigh}{13.0\xspace}
\newcommand{\SeventhBoundTwoCTwoSelectionFresh}{5\xspace}
\newcommand{\SeventhBoundTwoCTwoSelectionFreshLow}{4.0\xspace}
\newcommand{\SeventhBoundTwoCTwoSelectionFreshHigh}{11.0\xspace}
\newcommand{\SeventhBoundTwoCTwoSelectionStable}{1\xspace}
\newcommand{\SeventhBoundTwoCTwoSelectionStableLow}{1.0\xspace}
\newcommand{\SeventhBoundTwoCTwoSelectionStableHigh}{3.0\xspace}
\newcommand{\SeventhBoundTwoCTwoSameScoring}{8\xspace}
\newcommand{\SeventhBoundTwoCTwoSameScoringLow}{2.0\xspace}
\newcommand{\SeventhBoundTwoCTwoSameScoringHigh}{10.0\xspace}
\newcommand{\SeventhBoundTwoCTwoSameFresh}{6\xspace}
\newcommand{\SeventhBoundTwoCTwoSameFreshLow}{2.0\xspace}
\newcommand{\SeventhBoundTwoCTwoSameFreshHigh}{8.0\xspace}
\newcommand{\SeventhBoundTwoCTwoSameStable}{1\xspace}
\newcommand{\SeventhBoundTwoCTwoSameStableLow}{1.0\xspace}
\newcommand{\SeventhBoundTwoCTwoSameStableHigh}{1.0\xspace}
\newcommand{\SeventhBoundTwoCThreeSelectionScoring}{2\xspace}
\newcommand{\SeventhBoundTwoCThreeSelectionScoringLow}{1.0\xspace}
\newcommand{\SeventhBoundTwoCThreeSelectionScoringHigh}{6.0\xspace}
\newcommand{\SeventhBoundTwoCThreeSelectionFresh}{1\xspace}
\newcommand{\SeventhBoundTwoCThreeSelectionFreshLow}{1.0\xspace}
\newcommand{\SeventhBoundTwoCThreeSelectionFreshHigh}{6.0\xspace}
\newcommand{\SeventhBoundTwoCThreeSelectionStable}{0\xspace}
\newcommand{\SeventhBoundTwoCThreeSelectionStableLow}{0.0\xspace}
\newcommand{\SeventhBoundTwoCThreeSelectionStableHigh}{1.0\xspace}
\newcommand{\SeventhBoundTwoCThreeSameScoring}{0\xspace}
\newcommand{\SeventhBoundTwoCThreeSameScoringLow}{0.0\xspace}
\newcommand{\SeventhBoundTwoCThreeSameScoringHigh}{4.0\xspace}
\newcommand{\SeventhBoundTwoCThreeSameFresh}{1\xspace}
\newcommand{\SeventhBoundTwoCThreeSameFreshLow}{0.0\xspace}
\newcommand{\SeventhBoundTwoCThreeSameFreshHigh}{4.0\xspace}
\newcommand{\SeventhBoundTwoCThreeSameStable}{0\xspace}
\newcommand{\SeventhBoundTwoCThreeSameStableLow}{0.0\xspace}
\newcommand{\SeventhBoundTwoCThreeSameStableHigh}{0.0\xspace}
\newcommand{\SeventhBoundTwoCFourSelectionScoring}{7\xspace}
\newcommand{\SeventhBoundTwoCFourSelectionScoringLow}{4.0\xspace}
\newcommand{\SeventhBoundTwoCFourSelectionScoringHigh}{10.0\xspace}
\newcommand{\SeventhBoundTwoCFourSelectionFresh}{12\xspace}
\newcommand{\SeventhBoundTwoCFourSelectionFreshLow}{6.0\xspace}
\newcommand{\SeventhBoundTwoCFourSelectionFreshHigh}{14.0\xspace}
\newcommand{\SeventhBoundTwoCFourSelectionStable}{1\xspace}
\newcommand{\SeventhBoundTwoCFourSelectionStableLow}{0.0\xspace}
\newcommand{\SeventhBoundTwoCFourSelectionStableHigh}{3.0\xspace}
\newcommand{\SeventhBoundTwoCFourSameScoring}{3\xspace}
\newcommand{\SeventhBoundTwoCFourSameScoringLow}{1.0\xspace}
\newcommand{\SeventhBoundTwoCFourSameScoringHigh}{7.0\xspace}
\newcommand{\SeventhBoundTwoCFourSameFresh}{9\xspace}
\newcommand{\SeventhBoundTwoCFourSameFreshLow}{3.0\xspace}
\newcommand{\SeventhBoundTwoCFourSameFreshHigh}{10.0\xspace}
\newcommand{\SeventhBoundTwoCFourSameStable}{0\xspace}
\newcommand{\SeventhBoundTwoCFourSameStableLow}{0.0\xspace}
\newcommand{\SeventhBoundTwoCFourSameStableHigh}{1.0\xspace}
\newcommand{\SeventhBoundTwoCFiveSelectionScoring}{9\xspace}
\newcommand{\SeventhBoundTwoCFiveSelectionScoringLow}{7.0\xspace}
\newcommand{\SeventhBoundTwoCFiveSelectionScoringHigh}{15.0\xspace}
\newcommand{\SeventhBoundTwoCFiveSelectionFresh}{6\xspace}
\newcommand{\SeventhBoundTwoCFiveSelectionFreshLow}{4.0\xspace}
\newcommand{\SeventhBoundTwoCFiveSelectionFreshHigh}{13.0\xspace}
\newcommand{\SeventhBoundTwoCFiveSelectionStable}{2\xspace}
\newcommand{\SeventhBoundTwoCFiveSelectionStableLow}{0.0\xspace}
\newcommand{\SeventhBoundTwoCFiveSelectionStableHigh}{3.0\xspace}
\newcommand{\SeventhBoundTwoCFiveSameScoring}{2\xspace}
\newcommand{\SeventhBoundTwoCFiveSameScoringLow}{1.0\xspace}
\newcommand{\SeventhBoundTwoCFiveSameScoringHigh}{8.0\xspace}
\newcommand{\SeventhBoundTwoCFiveSameFresh}{6\xspace}
\newcommand{\SeventhBoundTwoCFiveSameFreshLow}{2.0\xspace}
\newcommand{\SeventhBoundTwoCFiveSameFreshHigh}{10.0\xspace}
\newcommand{\SeventhBoundTwoCFiveSameStable}{1\xspace}
\newcommand{\SeventhBoundTwoCFiveSameStableLow}{0.0\xspace}
\newcommand{\SeventhBoundTwoCFiveSameStableHigh}{1.0\xspace}
\newcommand{\SeventhBoundTwoCAllSelectionScoring}{34\xspace}
\newcommand{\SeventhBoundTwoCAllSelectionScoringLow}{31.0\xspace}
\newcommand{\SeventhBoundTwoCAllSelectionScoringHigh}{47.0\xspace}
\newcommand{\SeventhBoundTwoCAllSelectionFresh}{27\xspace}
\newcommand{\SeventhBoundTwoCAllSelectionFreshLow}{25.0\xspace}
\newcommand{\SeventhBoundTwoCAllSelectionFreshHigh}{41.0\xspace}
\newcommand{\SeventhBoundTwoCAllSelectionStable}{4\xspace}
\newcommand{\SeventhBoundTwoCAllSelectionStableLow}{1.0\xspace}
\newcommand{\SeventhBoundTwoCAllSelectionStableHigh}{7.0\xspace}
\newcommand{\SeventhBoundTwoCAllSameScoring}{16\xspace}
\newcommand{\SeventhBoundTwoCAllSameScoringLow}{13.0\xspace}
\newcommand{\SeventhBoundTwoCAllSameScoringHigh}{27.0\xspace}
\newcommand{\SeventhBoundTwoCAllSameFresh}{23\xspace}
\newcommand{\SeventhBoundTwoCAllSameFreshLow}{15.0\xspace}
\newcommand{\SeventhBoundTwoCAllSameFreshHigh}{29.0\xspace}
\newcommand{\SeventhBoundTwoCAllSameStable}{2\xspace}
\newcommand{\SeventhBoundTwoCAllSameStableLow}{1.0\xspace}
\newcommand{\SeventhBoundTwoCAllSameStableHigh}{3.0\xspace}

\newcommand{\SeventhBoundThreeAOneSelectionScoring}{9\xspace}
\newcommand{\SeventhBoundThreeAOneSelectionScoringLow}{6.0\xspace}
\newcommand{\SeventhBoundThreeAOneSelectionScoringHigh}{15.0\xspace}
\newcommand{\SeventhBoundThreeAOneSelectionFresh}{7\xspace}
\newcommand{\SeventhBoundThreeAOneSelectionFreshLow}{6.0\xspace}
\newcommand{\SeventhBoundThreeAOneSelectionFreshHigh}{13.0\xspace}
\newcommand{\SeventhBoundThreeAOneSelectionStable}{1\xspace}
\newcommand{\SeventhBoundThreeAOneSelectionStableLow}{0.0\xspace}
\newcommand{\SeventhBoundThreeAOneSelectionStableHigh}{3.0\xspace}
\newcommand{\SeventhBoundThreeAOneSameScoring}{8\xspace}
\newcommand{\SeventhBoundThreeAOneSameScoringLow}{4.0\xspace}
\newcommand{\SeventhBoundThreeAOneSameScoringHigh}{14.0\xspace}
\newcommand{\SeventhBoundThreeAOneSameFresh}{7\xspace}
\newcommand{\SeventhBoundThreeAOneSameFreshLow}{5.0\xspace}
\newcommand{\SeventhBoundThreeAOneSameFreshHigh}{13.0\xspace}
\newcommand{\SeventhBoundThreeAOneSameStable}{1\xspace}
\newcommand{\SeventhBoundThreeAOneSameStableLow}{0.0\xspace}
\newcommand{\SeventhBoundThreeAOneSameStableHigh}{3.0\xspace}
\newcommand{\SeventhBoundThreeATwoSelectionScoring}{3\xspace}
\newcommand{\SeventhBoundThreeATwoSelectionScoringLow}{2.0\xspace}
\newcommand{\SeventhBoundThreeATwoSelectionScoringHigh}{9.0\xspace}
\newcommand{\SeventhBoundThreeATwoSelectionFresh}{9\xspace}
\newcommand{\SeventhBoundThreeATwoSelectionFreshLow}{6.0\xspace}
\newcommand{\SeventhBoundThreeATwoSelectionFreshHigh}{12.0\xspace}
\newcommand{\SeventhBoundThreeATwoSelectionStable}{1\xspace}
\newcommand{\SeventhBoundThreeATwoSelectionStableLow}{0.0\xspace}
\newcommand{\SeventhBoundThreeATwoSelectionStableHigh}{3.0\xspace}
\newcommand{\SeventhBoundThreeATwoSameScoring}{6\xspace}
\newcommand{\SeventhBoundThreeATwoSameScoringLow}{3.0\xspace}
\newcommand{\SeventhBoundThreeATwoSameScoringHigh}{10.0\xspace}
\newcommand{\SeventhBoundThreeATwoSameFresh}{8\xspace}
\newcommand{\SeventhBoundThreeATwoSameFreshLow}{5.0\xspace}
\newcommand{\SeventhBoundThreeATwoSameFreshHigh}{11.0\xspace}
\newcommand{\SeventhBoundThreeATwoSameStable}{1\xspace}
\newcommand{\SeventhBoundThreeATwoSameStableLow}{0.0\xspace}
\newcommand{\SeventhBoundThreeATwoSameStableHigh}{2.0\xspace}
\newcommand{\SeventhBoundThreeAThreeSelectionScoring}{9\xspace}
\newcommand{\SeventhBoundThreeAThreeSelectionScoringLow}{5.0\xspace}
\newcommand{\SeventhBoundThreeAThreeSelectionScoringHigh}{13.0\xspace}
\newcommand{\SeventhBoundThreeAThreeSelectionFresh}{7\xspace}
\newcommand{\SeventhBoundThreeAThreeSelectionFreshLow}{6.0\xspace}
\newcommand{\SeventhBoundThreeAThreeSelectionFreshHigh}{11.0\xspace}
\newcommand{\SeventhBoundThreeAThreeSelectionStable}{2\xspace}
\newcommand{\SeventhBoundThreeAThreeSelectionStableLow}{1.0\xspace}
\newcommand{\SeventhBoundThreeAThreeSelectionStableHigh}{4.0\xspace}
\newcommand{\SeventhBoundThreeAThreeSameScoring}{11\xspace}
\newcommand{\SeventhBoundThreeAThreeSameScoringLow}{6.0\xspace}
\newcommand{\SeventhBoundThreeAThreeSameScoringHigh}{15.0\xspace}
\newcommand{\SeventhBoundThreeAThreeSameFresh}{8\xspace}
\newcommand{\SeventhBoundThreeAThreeSameFreshLow}{6.0\xspace}
\newcommand{\SeventhBoundThreeAThreeSameFreshHigh}{12.0\xspace}
\newcommand{\SeventhBoundThreeAThreeSameStable}{2\xspace}
\newcommand{\SeventhBoundThreeAThreeSameStableLow}{0.0\xspace}
\newcommand{\SeventhBoundThreeAThreeSameStableHigh}{4.0\xspace}
\newcommand{\SeventhBoundThreeAFourSelectionScoring}{4\xspace}
\newcommand{\SeventhBoundThreeAFourSelectionScoringLow}{3.0\xspace}
\newcommand{\SeventhBoundThreeAFourSelectionScoringHigh}{8.0\xspace}
\newcommand{\SeventhBoundThreeAFourSelectionFresh}{5\xspace}
\newcommand{\SeventhBoundThreeAFourSelectionFreshLow}{4.0\xspace}
\newcommand{\SeventhBoundThreeAFourSelectionFreshHigh}{7.0\xspace}
\newcommand{\SeventhBoundThreeAFourSelectionStable}{2\xspace}
\newcommand{\SeventhBoundThreeAFourSelectionStableLow}{0.0\xspace}
\newcommand{\SeventhBoundThreeAFourSelectionStableHigh}{3.0\xspace}
\newcommand{\SeventhBoundThreeAFourSameScoring}{5\xspace}
\newcommand{\SeventhBoundThreeAFourSameScoringLow}{3.0\xspace}
\newcommand{\SeventhBoundThreeAFourSameScoringHigh}{9.0\xspace}
\newcommand{\SeventhBoundThreeAFourSameFresh}{4\xspace}
\newcommand{\SeventhBoundThreeAFourSameFreshLow}{3.0\xspace}
\newcommand{\SeventhBoundThreeAFourSameFreshHigh}{7.0\xspace}
\newcommand{\SeventhBoundThreeAFourSameStable}{3\xspace}
\newcommand{\SeventhBoundThreeAFourSameStableLow}{0.0\xspace}
\newcommand{\SeventhBoundThreeAFourSameStableHigh}{3.0\xspace}
\newcommand{\SeventhBoundThreeAFiveSelectionScoring}{2\xspace}
\newcommand{\SeventhBoundThreeAFiveSelectionScoringLow}{2.0\xspace}
\newcommand{\SeventhBoundThreeAFiveSelectionScoringHigh}{6.0\xspace}
\newcommand{\SeventhBoundThreeAFiveSelectionFresh}{2\xspace}
\newcommand{\SeventhBoundThreeAFiveSelectionFreshLow}{2.0\xspace}
\newcommand{\SeventhBoundThreeAFiveSelectionFreshHigh}{7.0\xspace}
\newcommand{\SeventhBoundThreeAFiveSelectionStable}{1\xspace}
\newcommand{\SeventhBoundThreeAFiveSelectionStableLow}{0.0\xspace}
\newcommand{\SeventhBoundThreeAFiveSelectionStableHigh}{2.0\xspace}
\newcommand{\SeventhBoundThreeAFiveSameScoring}{1\xspace}
\newcommand{\SeventhBoundThreeAFiveSameScoringLow}{1.0\xspace}
\newcommand{\SeventhBoundThreeAFiveSameScoringHigh}{5.0\xspace}
\newcommand{\SeventhBoundThreeAFiveSameFresh}{1\xspace}
\newcommand{\SeventhBoundThreeAFiveSameFreshLow}{1.0\xspace}
\newcommand{\SeventhBoundThreeAFiveSameFreshHigh}{6.0\xspace}
\newcommand{\SeventhBoundThreeAFiveSameStable}{0\xspace}
\newcommand{\SeventhBoundThreeAFiveSameStableLow}{0.0\xspace}
\newcommand{\SeventhBoundThreeAFiveSameStableHigh}{1.0\xspace}
\newcommand{\SeventhBoundThreeAAllSelectionScoring}{27\xspace}
\newcommand{\SeventhBoundThreeAAllSelectionScoringLow}{25.0\xspace}
\newcommand{\SeventhBoundThreeAAllSelectionScoringHigh}{41.0\xspace}
\newcommand{\SeventhBoundThreeAAllSelectionFresh}{30\xspace}
\newcommand{\SeventhBoundThreeAAllSelectionFreshLow}{29.0\xspace}
\newcommand{\SeventhBoundThreeAAllSelectionFreshHigh}{43.0\xspace}
\newcommand{\SeventhBoundThreeAAllSelectionStable}{7\xspace}
\newcommand{\SeventhBoundThreeAAllSelectionStableLow}{4.0\xspace}
\newcommand{\SeventhBoundThreeAAllSelectionStableHigh}{11.0\xspace}
\newcommand{\SeventhBoundThreeAAllSameScoring}{31\xspace}
\newcommand{\SeventhBoundThreeAAllSameScoringLow}{26.0\xspace}
\newcommand{\SeventhBoundThreeAAllSameScoringHigh}{43.0\xspace}
\newcommand{\SeventhBoundThreeAAllSameFresh}{28\xspace}
\newcommand{\SeventhBoundThreeAAllSameFreshLow}{27.0\xspace}
\newcommand{\SeventhBoundThreeAAllSameFreshHigh}{41.0\xspace}
\newcommand{\SeventhBoundThreeAAllSameStable}{7\xspace}
\newcommand{\SeventhBoundThreeAAllSameStableLow}{2.0\xspace}
\newcommand{\SeventhBoundThreeAAllSameStableHigh}{8.0\xspace}
\newcommand{\SeventhBoundThreeBOneSelectionScoring}{11\xspace}
\newcommand{\SeventhBoundThreeBOneSelectionScoringLow}{7.0\xspace}
\newcommand{\SeventhBoundThreeBOneSelectionScoringHigh}{15.0\xspace}
\newcommand{\SeventhBoundThreeBOneSelectionFresh}{12\xspace}
\newcommand{\SeventhBoundThreeBOneSelectionFreshLow}{9.0\xspace}
\newcommand{\SeventhBoundThreeBOneSelectionFreshHigh}{16.0\xspace}
\newcommand{\SeventhBoundThreeBOneSelectionStable}{2\xspace}
\newcommand{\SeventhBoundThreeBOneSelectionStableLow}{1.0\xspace}
\newcommand{\SeventhBoundThreeBOneSelectionStableHigh}{5.0\xspace}
\newcommand{\SeventhBoundThreeBOneSameScoring}{6\xspace}
\newcommand{\SeventhBoundThreeBOneSameScoringLow}{2.0\xspace}
\newcommand{\SeventhBoundThreeBOneSameScoringHigh}{10.0\xspace}
\newcommand{\SeventhBoundThreeBOneSameFresh}{5\xspace}
\newcommand{\SeventhBoundThreeBOneSameFreshLow}{2.0\xspace}
\newcommand{\SeventhBoundThreeBOneSameFreshHigh}{8.0\xspace}
\newcommand{\SeventhBoundThreeBOneSameStable}{0\xspace}
\newcommand{\SeventhBoundThreeBOneSameStableLow}{0.0\xspace}
\newcommand{\SeventhBoundThreeBOneSameStableHigh}{1.0\xspace}
\newcommand{\SeventhBoundThreeBTwoSelectionScoring}{0\xspace}
\newcommand{\SeventhBoundThreeBTwoSelectionScoringLow}{0.0\xspace}
\newcommand{\SeventhBoundThreeBTwoSelectionScoringHigh}{5.0\xspace}
\newcommand{\SeventhBoundThreeBTwoSelectionFresh}{4\xspace}
\newcommand{\SeventhBoundThreeBTwoSelectionFreshLow}{2.0\xspace}
\newcommand{\SeventhBoundThreeBTwoSelectionFreshHigh}{8.0\xspace}
\newcommand{\SeventhBoundThreeBTwoSelectionStable}{0\xspace}
\newcommand{\SeventhBoundThreeBTwoSelectionStableLow}{0.0\xspace}
\newcommand{\SeventhBoundThreeBTwoSelectionStableHigh}{1.0\xspace}
\newcommand{\SeventhBoundThreeBTwoSameScoring}{0\xspace}
\newcommand{\SeventhBoundThreeBTwoSameScoringLow}{0.0\xspace}
\newcommand{\SeventhBoundThreeBTwoSameScoringHigh}{5.0\xspace}
\newcommand{\SeventhBoundThreeBTwoSameFresh}{2\xspace}
\newcommand{\SeventhBoundThreeBTwoSameFreshLow}{0.0\xspace}
\newcommand{\SeventhBoundThreeBTwoSameFreshHigh}{5.0\xspace}
\newcommand{\SeventhBoundThreeBTwoSameStable}{0\xspace}
\newcommand{\SeventhBoundThreeBTwoSameStableLow}{0.0\xspace}
\newcommand{\SeventhBoundThreeBTwoSameStableHigh}{0.0\xspace}
\newcommand{\SeventhBoundThreeBThreeSelectionScoring}{3\xspace}
\newcommand{\SeventhBoundThreeBThreeSelectionScoringLow}{2.0\xspace}
\newcommand{\SeventhBoundThreeBThreeSelectionScoringHigh}{11.0\xspace}
\newcommand{\SeventhBoundThreeBThreeSelectionFresh}{4\xspace}
\newcommand{\SeventhBoundThreeBThreeSelectionFreshLow}{3.0\xspace}
\newcommand{\SeventhBoundThreeBThreeSelectionFreshHigh}{13.0\xspace}
\newcommand{\SeventhBoundThreeBThreeSelectionStable}{0\xspace}
\newcommand{\SeventhBoundThreeBThreeSelectionStableLow}{0.0\xspace}
\newcommand{\SeventhBoundThreeBThreeSelectionStableHigh}{3.0\xspace}
\newcommand{\SeventhBoundThreeBThreeSameScoring}{5\xspace}
\newcommand{\SeventhBoundThreeBThreeSameScoringLow}{2.0\xspace}
\newcommand{\SeventhBoundThreeBThreeSameScoringHigh}{10.0\xspace}
\newcommand{\SeventhBoundThreeBThreeSameFresh}{5\xspace}
\newcommand{\SeventhBoundThreeBThreeSameFreshLow}{1.0\xspace}
\newcommand{\SeventhBoundThreeBThreeSameFreshHigh}{9.0\xspace}
\newcommand{\SeventhBoundThreeBThreeSameStable}{0\xspace}
\newcommand{\SeventhBoundThreeBThreeSameStableLow}{0.0\xspace}
\newcommand{\SeventhBoundThreeBThreeSameStableHigh}{1.0\xspace}
\newcommand{\SeventhBoundThreeBFourSelectionScoring}{17\xspace}
\newcommand{\SeventhBoundThreeBFourSelectionScoringLow}{11.0\xspace}
\newcommand{\SeventhBoundThreeBFourSelectionScoringHigh}{19.0\xspace}
\newcommand{\SeventhBoundThreeBFourSelectionFresh}{8\xspace}
\newcommand{\SeventhBoundThreeBFourSelectionFreshLow}{8.0\xspace}
\newcommand{\SeventhBoundThreeBFourSelectionFreshHigh}{16.0\xspace}
\newcommand{\SeventhBoundThreeBFourSelectionStable}{4\xspace}
\newcommand{\SeventhBoundThreeBFourSelectionStableLow}{1.0\xspace}
\newcommand{\SeventhBoundThreeBFourSelectionStableHigh}{6.0\xspace}
\newcommand{\SeventhBoundThreeBFourSameScoring}{7\xspace}
\newcommand{\SeventhBoundThreeBFourSameScoringLow}{3.0\xspace}
\newcommand{\SeventhBoundThreeBFourSameScoringHigh}{10.0\xspace}
\newcommand{\SeventhBoundThreeBFourSameFresh}{5\xspace}
\newcommand{\SeventhBoundThreeBFourSameFreshLow}{2.0\xspace}
\newcommand{\SeventhBoundThreeBFourSameFreshHigh}{10.0\xspace}
\newcommand{\SeventhBoundThreeBFourSameStable}{1\xspace}
\newcommand{\SeventhBoundThreeBFourSameStableLow}{0.0\xspace}
\newcommand{\SeventhBoundThreeBFourSameStableHigh}{1.0\xspace}
\newcommand{\SeventhBoundThreeBFiveSelectionScoring}{9\xspace}
\newcommand{\SeventhBoundThreeBFiveSelectionScoringLow}{6.0\xspace}
\newcommand{\SeventhBoundThreeBFiveSelectionScoringHigh}{14.0\xspace}
\newcommand{\SeventhBoundThreeBFiveSelectionFresh}{8\xspace}
\newcommand{\SeventhBoundThreeBFiveSelectionFreshLow}{5.0\xspace}
\newcommand{\SeventhBoundThreeBFiveSelectionFreshHigh}{13.0\xspace}
\newcommand{\SeventhBoundThreeBFiveSelectionStable}{2\xspace}
\newcommand{\SeventhBoundThreeBFiveSelectionStableLow}{0.0\xspace}
\newcommand{\SeventhBoundThreeBFiveSelectionStableHigh}{3.0\xspace}
\newcommand{\SeventhBoundThreeBFiveSameScoring}{4\xspace}
\newcommand{\SeventhBoundThreeBFiveSameScoringLow}{2.0\xspace}
\newcommand{\SeventhBoundThreeBFiveSameScoringHigh}{10.0\xspace}
\newcommand{\SeventhBoundThreeBFiveSameFresh}{7\xspace}
\newcommand{\SeventhBoundThreeBFiveSameFreshLow}{4.0\xspace}
\newcommand{\SeventhBoundThreeBFiveSameFreshHigh}{11.0\xspace}
\newcommand{\SeventhBoundThreeBFiveSameStable}{1\xspace}
\newcommand{\SeventhBoundThreeBFiveSameStableLow}{0.0\xspace}
\newcommand{\SeventhBoundThreeBFiveSameStableHigh}{1.0\xspace}
\newcommand{\SeventhBoundThreeBAllSelectionScoring}{40\xspace}
\newcommand{\SeventhBoundThreeBAllSelectionScoringLow}{36.0\xspace}
\newcommand{\SeventhBoundThreeBAllSelectionScoringHigh}{54.0\xspace}
\newcommand{\SeventhBoundThreeBAllSelectionFresh}{36\xspace}
\newcommand{\SeventhBoundThreeBAllSelectionFreshLow}{36.0\xspace}
\newcommand{\SeventhBoundThreeBAllSelectionFreshHigh}{55.0\xspace}
\newcommand{\SeventhBoundThreeBAllSelectionStable}{8\xspace}
\newcommand{\SeventhBoundThreeBAllSelectionStableLow}{4.0\xspace}
\newcommand{\SeventhBoundThreeBAllSelectionStableHigh}{12.0\xspace}
\newcommand{\SeventhBoundThreeBAllSameScoring}{22\xspace}
\newcommand{\SeventhBoundThreeBAllSameScoringLow}{18.0\xspace}
\newcommand{\SeventhBoundThreeBAllSameScoringHigh}{34.0\xspace}
\newcommand{\SeventhBoundThreeBAllSameFresh}{24\xspace}
\newcommand{\SeventhBoundThreeBAllSameFreshLow}{17.0\xspace}
\newcommand{\SeventhBoundThreeBAllSameFreshHigh}{33.0\xspace}
\newcommand{\SeventhBoundThreeBAllSameStable}{2\xspace}
\newcommand{\SeventhBoundThreeBAllSameStableLow}{0.0\xspace}
\newcommand{\SeventhBoundThreeBAllSameStableHigh}{2.0\xspace}
\newcommand{\SeventhBoundThreeCOneSelectionScoring}{9\xspace}
\newcommand{\SeventhBoundThreeCOneSelectionScoringLow}{6.0\xspace}
\newcommand{\SeventhBoundThreeCOneSelectionScoringHigh}{15.0\xspace}
\newcommand{\SeventhBoundThreeCOneSelectionFresh}{12\xspace}
\newcommand{\SeventhBoundThreeCOneSelectionFreshLow}{7.0\xspace}
\newcommand{\SeventhBoundThreeCOneSelectionFreshHigh}{16.0\xspace}
\newcommand{\SeventhBoundThreeCOneSelectionStable}{1\xspace}
\newcommand{\SeventhBoundThreeCOneSelectionStableLow}{0.0\xspace}
\newcommand{\SeventhBoundThreeCOneSelectionStableHigh}{5.0\xspace}
\newcommand{\SeventhBoundThreeCOneSameScoring}{4\xspace}
\newcommand{\SeventhBoundThreeCOneSameScoringLow}{2.0\xspace}
\newcommand{\SeventhBoundThreeCOneSameScoringHigh}{9.0\xspace}
\newcommand{\SeventhBoundThreeCOneSameFresh}{4\xspace}
\newcommand{\SeventhBoundThreeCOneSameFreshLow}{2.0\xspace}
\newcommand{\SeventhBoundThreeCOneSameFreshHigh}{8.0\xspace}
\newcommand{\SeventhBoundThreeCOneSameStable}{0\xspace}
\newcommand{\SeventhBoundThreeCOneSameStableLow}{0.0\xspace}
\newcommand{\SeventhBoundThreeCOneSameStableHigh}{0.0\xspace}
\newcommand{\SeventhBoundThreeCTwoSelectionScoring}{4\xspace}
\newcommand{\SeventhBoundThreeCTwoSelectionScoringLow}{2.0\xspace}
\newcommand{\SeventhBoundThreeCTwoSelectionScoringHigh}{7.0\xspace}
\newcommand{\SeventhBoundThreeCTwoSelectionFresh}{5\xspace}
\newcommand{\SeventhBoundThreeCTwoSelectionFreshLow}{2.0\xspace}
\newcommand{\SeventhBoundThreeCTwoSelectionFreshHigh}{7.0\xspace}
\newcommand{\SeventhBoundThreeCTwoSelectionStable}{1\xspace}
\newcommand{\SeventhBoundThreeCTwoSelectionStableLow}{1.0\xspace}
\newcommand{\SeventhBoundThreeCTwoSelectionStableHigh}{2.0\xspace}
\newcommand{\SeventhBoundThreeCTwoSameScoring}{3\xspace}
\newcommand{\SeventhBoundThreeCTwoSameScoringLow}{1.0\xspace}
\newcommand{\SeventhBoundThreeCTwoSameScoringHigh}{6.0\xspace}
\newcommand{\SeventhBoundThreeCTwoSameFresh}{4\xspace}
\newcommand{\SeventhBoundThreeCTwoSameFreshLow}{1.0\xspace}
\newcommand{\SeventhBoundThreeCTwoSameFreshHigh}{6.0\xspace}
\newcommand{\SeventhBoundThreeCTwoSameStable}{1\xspace}
\newcommand{\SeventhBoundThreeCTwoSameStableLow}{0.0\xspace}
\newcommand{\SeventhBoundThreeCTwoSameStableHigh}{2.0\xspace}
\newcommand{\SeventhBoundThreeCThreeSelectionScoring}{11\xspace}
\newcommand{\SeventhBoundThreeCThreeSelectionScoringLow}{8.0\xspace}
\newcommand{\SeventhBoundThreeCThreeSelectionScoringHigh}{13.0\xspace}
\newcommand{\SeventhBoundThreeCThreeSelectionFresh}{7\xspace}
\newcommand{\SeventhBoundThreeCThreeSelectionFreshLow}{6.0\xspace}
\newcommand{\SeventhBoundThreeCThreeSelectionFreshHigh}{11.0\xspace}
\newcommand{\SeventhBoundThreeCThreeSelectionStable}{6\xspace}
\newcommand{\SeventhBoundThreeCThreeSelectionStableLow}{2.0\xspace}
\newcommand{\SeventhBoundThreeCThreeSelectionStableHigh}{6.0\xspace}
\newcommand{\SeventhBoundThreeCThreeSameScoring}{4\xspace}
\newcommand{\SeventhBoundThreeCThreeSameScoringLow}{2.0\xspace}
\newcommand{\SeventhBoundThreeCThreeSameScoringHigh}{7.0\xspace}
\newcommand{\SeventhBoundThreeCThreeSameFresh}{5\xspace}
\newcommand{\SeventhBoundThreeCThreeSameFreshLow}{2.0\xspace}
\newcommand{\SeventhBoundThreeCThreeSameFreshHigh}{8.0\xspace}
\newcommand{\SeventhBoundThreeCThreeSameStable}{2\xspace}
\newcommand{\SeventhBoundThreeCThreeSameStableLow}{0.0\xspace}
\newcommand{\SeventhBoundThreeCThreeSameStableHigh}{2.0\xspace}
\newcommand{\SeventhBoundThreeCFourSelectionScoring}{10\xspace}
\newcommand{\SeventhBoundThreeCFourSelectionScoringLow}{7.0\xspace}
\newcommand{\SeventhBoundThreeCFourSelectionScoringHigh}{12.0\xspace}
\newcommand{\SeventhBoundThreeCFourSelectionFresh}{10\xspace}
\newcommand{\SeventhBoundThreeCFourSelectionFreshLow}{7.0\xspace}
\newcommand{\SeventhBoundThreeCFourSelectionFreshHigh}{12.0\xspace}
\newcommand{\SeventhBoundThreeCFourSelectionStable}{4\xspace}
\newcommand{\SeventhBoundThreeCFourSelectionStableLow}{3.0\xspace}
\newcommand{\SeventhBoundThreeCFourSelectionStableHigh}{5.0\xspace}
\newcommand{\SeventhBoundThreeCFourSameScoring}{7\xspace}
\newcommand{\SeventhBoundThreeCFourSameScoringLow}{4.0\xspace}
\newcommand{\SeventhBoundThreeCFourSameScoringHigh}{10.0\xspace}
\newcommand{\SeventhBoundThreeCFourSameFresh}{7\xspace}
\newcommand{\SeventhBoundThreeCFourSameFreshLow}{4.0\xspace}
\newcommand{\SeventhBoundThreeCFourSameFreshHigh}{10.0\xspace}
\newcommand{\SeventhBoundThreeCFourSameStable}{4\xspace}
\newcommand{\SeventhBoundThreeCFourSameStableLow}{2.0\xspace}
\newcommand{\SeventhBoundThreeCFourSameStableHigh}{4.0\xspace}
\newcommand{\SeventhBoundThreeCFiveSelectionScoring}{5\xspace}
\newcommand{\SeventhBoundThreeCFiveSelectionScoringLow}{3.0\xspace}
\newcommand{\SeventhBoundThreeCFiveSelectionScoringHigh}{10.0\xspace}
\newcommand{\SeventhBoundThreeCFiveSelectionFresh}{3\xspace}
\newcommand{\SeventhBoundThreeCFiveSelectionFreshLow}{1.0\xspace}
\newcommand{\SeventhBoundThreeCFiveSelectionFreshHigh}{7.0\xspace}
\newcommand{\SeventhBoundThreeCFiveSelectionStable}{2\xspace}
\newcommand{\SeventhBoundThreeCFiveSelectionStableLow}{0.0\xspace}
\newcommand{\SeventhBoundThreeCFiveSelectionStableHigh}{3.0\xspace}
\newcommand{\SeventhBoundThreeCFiveSameScoring}{4\xspace}
\newcommand{\SeventhBoundThreeCFiveSameScoringLow}{2.0\xspace}
\newcommand{\SeventhBoundThreeCFiveSameScoringHigh}{8.0\xspace}
\newcommand{\SeventhBoundThreeCFiveSameFresh}{3\xspace}
\newcommand{\SeventhBoundThreeCFiveSameFreshLow}{1.0\xspace}
\newcommand{\SeventhBoundThreeCFiveSameFreshHigh}{6.0\xspace}
\newcommand{\SeventhBoundThreeCFiveSameStable}{0\xspace}
\newcommand{\SeventhBoundThreeCFiveSameStableLow}{0.0\xspace}
\newcommand{\SeventhBoundThreeCFiveSameStableHigh}{1.0\xspace}
\newcommand{\SeventhBoundThreeCAllSelectionScoring}{39\xspace}
\newcommand{\SeventhBoundThreeCAllSelectionScoringLow}{34.0\xspace}
\newcommand{\SeventhBoundThreeCAllSelectionScoringHigh}{48.0\xspace}
\newcommand{\SeventhBoundThreeCAllSelectionFresh}{37\xspace}
\newcommand{\SeventhBoundThreeCAllSelectionFreshLow}{31.0\xspace}
\newcommand{\SeventhBoundThreeCAllSelectionFreshHigh}{45.0\xspace}
\newcommand{\SeventhBoundThreeCAllSelectionStable}{14\xspace}
\newcommand{\SeventhBoundThreeCAllSelectionStableLow}{9.0\xspace}
\newcommand{\SeventhBoundThreeCAllSelectionStableHigh}{16.0\xspace}
\newcommand{\SeventhBoundThreeCAllSameScoring}{22\xspace}
\newcommand{\SeventhBoundThreeCAllSameScoringLow}{18.0\xspace}
\newcommand{\SeventhBoundThreeCAllSameScoringHigh}{31.0\xspace}
\newcommand{\SeventhBoundThreeCAllSameFresh}{23\xspace}
\newcommand{\SeventhBoundThreeCAllSameFreshLow}{17.0\xspace}
\newcommand{\SeventhBoundThreeCAllSameFreshHigh}{29.0\xspace}
\newcommand{\SeventhBoundThreeCAllSameStable}{7\xspace}
\newcommand{\SeventhBoundThreeCAllSameStableLow}{4.0\xspace}
\newcommand{\SeventhBoundThreeCAllSameStableHigh}{7.0\xspace}
\newcommand{\SeventhProcessor}{Apple M4\xspace}
\newcommand{\SeventhScreenAssignmentA}{3.079\xspace}
\newcommand{\SeventhScreenAssignmentALow}{3.065\xspace}
\newcommand{\SeventhScreenAssignmentAHigh}{3.542\xspace}
\newcommand{\SeventhScreenAssignmentB}{1.959\xspace}
\newcommand{\SeventhScreenAssignmentBLow}{1.850\xspace}
\newcommand{\SeventhScreenAssignmentBHigh}{2.154\xspace}
\newcommand{\SeventhScreenCallsA}{0.224\xspace}
\newcommand{\SeventhScreenCallsALow}{0.223\xspace}
\newcommand{\SeventhScreenCallsAHigh}{0.233\xspace}
\newcommand{\SeventhScreenCallsB}{0.233\xspace}
\newcommand{\SeventhScreenCallsBLow}{0.232\xspace}
\newcommand{\SeventhScreenCallsBHigh}{0.242\xspace}
\newcommand{\SeventhScreenTotalA}{3.302\xspace}
\newcommand{\SeventhMacA}{1.708\xspace}
\newcommand{\SeventhMacALow}{1.704\xspace}
\newcommand{\SeventhMacAHigh}{1.862\xspace}
\newcommand{\SeventhScreenTotalB}{2.193\xspace}
\newcommand{\SeventhMacB}{1.197\xspace}
\newcommand{\SeventhMacBLow}{1.167\xspace}
\newcommand{\SeventhMacBHigh}{1.202\xspace}

\newcommand{\FittedSiteExact}{0.2463317291694092\xspace}
\newcommand{\FittedAgree}{2,527\xspace}
\newcommand{\FittedAgreement}{0.987109\xspace}
\newcommand{\AdditionalSeconds}{193.083\xspace}
\newcommand{\ThirdSeed}{2026110207\xspace}
\newcommand{\DinoCorpus}{LVD-142M\xspace}
\newcommand{\PairImageSeed}{2026110215\xspace}
\newcommand{\CasePartner}{744\xspace}
\newcommand{\FittedSeedA}{2026110340\xspace}
\newcommand{\FittedCountA}{42\xspace}
\newcommand{\FittedSeedB}{2026110341\xspace}
\newcommand{\FittedCountB}{27\xspace}
\newcommand{\FittedSeedC}{2026110342\xspace}
\newcommand{\FittedCountC}{28\xspace}
\newcommand{\CaseScoreError}{0.441748\xspace}
\newcommand{\CaseScoreTopShare}{0.201588\xspace}
\newcommand{\CaseScoreTopEnergy}{0.231254\xspace}
\newcommand{\CaseScoreRemoved}{0.446113\xspace}

\newcommand{\CaseFreshError}{0.417312\xspace}
\newcommand{\CaseFreshTopShare}{0.227587\xspace}
\newcommand{\CaseFreshTopEnergy}{0.283376\xspace}
\newcommand{\CaseFreshRemoved}{0.425597\xspace}
\newcommand{\CaseFreshPatch}{0.9978\xspace}
\newcommand{\CaseThirdError}{0.450594\xspace}
\newcommand{\CaseThirdTopShare}{0.451760\xspace}
\newcommand{\CaseThirdTopEnergy}{0.460655\xspace}
\newcommand{\CaseThirdRemoved}{0.446269\xspace}

\newcommand{\EighthCellAE}{4\xspace}
\newcommand{\EighthCellAF}{2\xspace}
\newcommand{\EighthCellAG}{6\xspace}
\newcommand{\EighthCellAH}{4\xspace}
\newcommand{\EighthCellAI}{2\xspace}
\newcommand{\EighthCellAJ}{6\xspace}
\newcommand{\EighthCellAK}{11\xspace}
\newcommand{\EighthCellAL}{8\xspace}
\newcommand{\EighthCellAM}{13\xspace}
\newcommand{\EighthCellAN}{11\xspace}
\newcommand{\EighthCellAO}{8\xspace}
\newcommand{\EighthCellAP}{13\xspace}
\newcommand{\EighthCellAQ}{1\xspace}
\newcommand{\EighthCellAR}{0\xspace}
\newcommand{\EighthCellAS}{1\xspace}
\newcommand{\EighthCellAT}{1\xspace}
\newcommand{\EighthCellAU}{0\xspace}
\newcommand{\EighthCellAV}{1\xspace}
\newcommand{\EighthCellAW}{2\xspace}
\newcommand{\EighthCellAX}{0\xspace}
\newcommand{\EighthCellAY}{4\xspace}
\newcommand{\EighthCellAZ}{2\xspace}
\newcommand{\EighthCellBA}{0\xspace}
\newcommand{\EighthCellBB}{4\xspace}
\newcommand{\EighthCellBC}{0\xspace}
\newcommand{\EighthCellBD}{0\xspace}
\newcommand{\EighthCellBE}{1\xspace}
\newcommand{\EighthCellBF}{0\xspace}
\newcommand{\EighthCellBG}{0\xspace}
\newcommand{\EighthCellBH}{1\xspace}
\newcommand{\EighthCellBI}{8\xspace}
\newcommand{\EighthCellBJ}{5\xspace}
\newcommand{\EighthCellBK}{10\xspace}
\newcommand{\EighthCellBL}{8\xspace}
\newcommand{\EighthCellBM}{5\xspace}
\newcommand{\EighthCellBN}{10\xspace}
\newcommand{\EighthCellBO}{2\xspace}
\newcommand{\EighthCellBP}{1\xspace}
\newcommand{\EighthCellBQ}{5\xspace}
\newcommand{\EighthCellBR}{2\xspace}
\newcommand{\EighthCellBS}{1\xspace}
\newcommand{\EighthCellBT}{5\xspace}
\newcommand{\EighthCellBU}{6\xspace}
\newcommand{\EighthCellBV}{4\xspace}
\newcommand{\EighthCellBW}{8\xspace}
\newcommand{\EighthCellBX}{6\xspace}
\newcommand{\EighthCellBY}{4\xspace}
\newcommand{\EighthCellBZ}{8\xspace}
\newcommand{\EighthCellCA}{1\xspace}
\newcommand{\EighthCellCB}{0\xspace}
\newcommand{\EighthCellCC}{4\xspace}
\newcommand{\EighthCellCD}{1\xspace}
\newcommand{\EighthCellCE}{0\xspace}
\newcommand{\EighthCellCF}{4\xspace}
\newcommand{\EighthCellCG}{7\xspace}
\newcommand{\EighthCellCH}{6\xspace}
\newcommand{\EighthCellCI}{11\xspace}
\newcommand{\EighthCellCJ}{7\xspace}
\newcommand{\EighthCellCK}{6\xspace}
\newcommand{\EighthCellCL}{11\xspace}
\newcommand{\EighthCellCM}{8\xspace}
\newcommand{\EighthCellCN}{5\xspace}
\newcommand{\EighthCellCO}{14\xspace}
\newcommand{\EighthCellCP}{8\xspace}
\newcommand{\EighthCellCQ}{5\xspace}
\newcommand{\EighthCellCR}{14\xspace}
\newcommand{\EighthCellCS}{34\xspace}
\newcommand{\EighthCellCT}{29\xspace}
\newcommand{\EighthCellCU}{39\xspace}
\newcommand{\EighthCellCV}{34\xspace}
\newcommand{\EighthCellCW}{29\xspace}
\newcommand{\EighthCellCX}{39\xspace}
\newcommand{\EighthCellCY}{1\xspace}
\newcommand{\EighthCellCZ}{0\xspace}
\newcommand{\EighthCellDA}{3\xspace}
\newcommand{\EighthCellDB}{1\xspace}
\newcommand{\EighthCellDC}{0\xspace}
\newcommand{\EighthCellDD}{3\xspace}
\newcommand{\EighthCellDE}{2\xspace}
\newcommand{\EighthCellDF}{1\xspace}
\newcommand{\EighthCellDG}{4\xspace}
\newcommand{\EighthCellDH}{2\xspace}
\newcommand{\EighthCellDI}{1\xspace}
\newcommand{\EighthCellDJ}{4\xspace}
\newcommand{\EighthCellDK}{3\xspace}
\newcommand{\EighthCellDL}{0\xspace}
\newcommand{\EighthCellDM}{7\xspace}
\newcommand{\EighthCellDN}{3\xspace}
\newcommand{\EighthCellDO}{0\xspace}
\newcommand{\EighthCellDP}{7\xspace}
\newcommand{\EighthCellDQ}{4\xspace}
\newcommand{\EighthCellDR}{1\xspace}
\newcommand{\EighthCellDS}{5\xspace}
\newcommand{\EighthCellDT}{4\xspace}
\newcommand{\EighthCellDU}{1\xspace}
\newcommand{\EighthCellDV}{5\xspace}
\newcommand{\EighthCellDW}{4\xspace}
\newcommand{\EighthCellDX}{2\xspace}
\newcommand{\EighthCellDY}{9\xspace}
\newcommand{\EighthCellDZ}{4\xspace}
\newcommand{\EighthCellEA}{2\xspace}
\newcommand{\EighthCellEB}{9\xspace}
\newcommand{\EighthCellEC}{1\xspace}
\newcommand{\EighthCellED}{0\xspace}
\newcommand{\EighthCellEE}{4\xspace}
\newcommand{\EighthCellEF}{1\xspace}
\newcommand{\EighthCellEG}{0\xspace}
\newcommand{\EighthCellEH}{4\xspace}
\newcommand{\EighthCellEI}{7\xspace}
\newcommand{\EighthCellEJ}{4\xspace}
\newcommand{\EighthCellEK}{10\xspace}
\newcommand{\EighthCellEL}{7\xspace}
\newcommand{\EighthCellEM}{4\xspace}
\newcommand{\EighthCellEN}{10\xspace}
\newcommand{\EighthCellEO}{2\xspace}
\newcommand{\EighthCellEP}{2\xspace}
\newcommand{\EighthCellEQ}{5\xspace}
\newcommand{\EighthCellER}{2\xspace}
\newcommand{\EighthCellES}{2\xspace}
\newcommand{\EighthCellET}{5\xspace}
\newcommand{\EighthCellEU}{3\xspace}
\newcommand{\EighthCellEV}{1\xspace}
\newcommand{\EighthCellEW}{5\xspace}
\newcommand{\EighthCellEX}{3\xspace}
\newcommand{\EighthCellEY}{1\xspace}
\newcommand{\EighthCellEZ}{5\xspace}
\newcommand{\EighthCellFA}{0\xspace}
\newcommand{\EighthCellFB}{0\xspace}
\newcommand{\EighthCellFC}{3\xspace}
\newcommand{\EighthCellFD}{0\xspace}
\newcommand{\EighthCellFE}{0\xspace}
\newcommand{\EighthCellFF}{3\xspace}
\newcommand{\EighthCellFG}{18\xspace}
\newcommand{\EighthCellFH}{13\xspace}
\newcommand{\EighthCellFI}{26\xspace}
\newcommand{\EighthCellFJ}{18\xspace}
\newcommand{\EighthCellFK}{13\xspace}
\newcommand{\EighthCellFL}{26\xspace}
\newcommand{\EighthCellFM}{9\xspace}
\newcommand{\EighthCellFN}{7\xspace}
\newcommand{\EighthCellFO}{15\xspace}
\newcommand{\EighthCellFP}{9\xspace}
\newcommand{\EighthCellFQ}{7\xspace}
\newcommand{\EighthCellFR}{15\xspace}
\newcommand{\EighthCellFS}{1\xspace}
\newcommand{\EighthCellFT}{0\xspace}
\newcommand{\EighthCellFU}{3\xspace}
\newcommand{\EighthCellFV}{1\xspace}
\newcommand{\EighthCellFW}{0\xspace}
\newcommand{\EighthCellFX}{3\xspace}
\newcommand{\EighthCellFY}{2\xspace}
\newcommand{\EighthCellFZ}{1\xspace}
\newcommand{\EighthCellGA}{6\xspace}
\newcommand{\EighthCellGB}{2\xspace}
\newcommand{\EighthCellGC}{1\xspace}
\newcommand{\EighthCellGD}{6\xspace}
\newcommand{\EighthCellGE}{4\xspace}
\newcommand{\EighthCellGF}{3\xspace}
\newcommand{\EighthCellGG}{8\xspace}
\newcommand{\EighthCellGH}{4\xspace}
\newcommand{\EighthCellGI}{3\xspace}
\newcommand{\EighthCellGJ}{8\xspace}
\newcommand{\EighthCellGK}{1\xspace}
\newcommand{\EighthCellGL}{0\xspace}
\newcommand{\EighthCellGM}{4\xspace}
\newcommand{\EighthCellGN}{1\xspace}
\newcommand{\EighthCellGO}{0\xspace}
\newcommand{\EighthCellGP}{4\xspace}
\newcommand{\EighthCellGQ}{1\xspace}
\newcommand{\EighthCellGR}{0\xspace}
\newcommand{\EighthCellGS}{5\xspace}
\newcommand{\EighthCellGT}{1\xspace}
\newcommand{\EighthCellGU}{0\xspace}
\newcommand{\EighthCellGV}{5\xspace}
\newcommand{\EighthCellGW}{0\xspace}
\newcommand{\EighthCellGX}{0\xspace}
\newcommand{\EighthCellGY}{3\xspace}
\newcommand{\EighthCellGZ}{0\xspace}
\newcommand{\EighthCellHA}{0\xspace}
\newcommand{\EighthCellHB}{3\xspace}
\newcommand{\EighthCellHC}{4\xspace}
\newcommand{\EighthCellHD}{1\xspace}
\newcommand{\EighthCellHE}{5.025\xspace}
\newcommand{\EighthCellHF}{4\xspace}
\newcommand{\EighthCellHG}{1\xspace}
\newcommand{\EighthCellHH}{5.025\xspace}
\newcommand{\EighthCellHI}{8\xspace}
\newcommand{\EighthCellHJ}{4\xspace}
\newcommand{\EighthCellHK}{10\xspace}
\newcommand{\EighthCellHL}{8\xspace}
\newcommand{\EighthCellHM}{4\xspace}
\newcommand{\EighthCellHN}{10\xspace}
\newcommand{\EighthCellHO}{1\xspace}
\newcommand{\EighthCellHP}{1\xspace}
\newcommand{\EighthCellHQ}{7\xspace}
\newcommand{\EighthCellHR}{1\xspace}
\newcommand{\EighthCellHS}{1\xspace}
\newcommand{\EighthCellHT}{6\xspace}
\newcommand{\EighthCellHU}{5\xspace}
\newcommand{\EighthCellHV}{2\xspace}
\newcommand{\EighthCellHW}{8\xspace}
\newcommand{\EighthCellHX}{6\xspace}
\newcommand{\EighthCellHY}{2\xspace}
\newcommand{\EighthCellHZ}{9\xspace}
\newcommand{\EighthCellIA}{11\xspace}
\newcommand{\EighthCellIB}{10\xspace}
\newcommand{\EighthCellIC}{22\xspace}
\newcommand{\EighthCellID}{11\xspace}
\newcommand{\EighthCellIE}{10\xspace}
\newcommand{\EighthCellIF}{21\xspace}
\newcommand{\EighthCellIG}{16\xspace}
\newcommand{\EighthCellIH}{12\xspace}
\newcommand{\EighthCellII}{24\xspace}
\newcommand{\EighthCellIJ}{17\xspace}
\newcommand{\EighthCellIK}{13\xspace}
\newcommand{\EighthCellIL}{24\xspace}
\newcommand{\EighthCellIM}{1\xspace}
\newcommand{\EighthCellIN}{0\xspace}
\newcommand{\EighthCellIO}{3\xspace}
\newcommand{\EighthCellIP}{1\xspace}
\newcommand{\EighthCellIQ}{0\xspace}
\newcommand{\EighthCellIR}{2\xspace}
\newcommand{\EighthCellIS}{12\xspace}
\newcommand{\EighthCellIT}{8\xspace}
\newcommand{\EighthCellIU}{14\xspace}
\newcommand{\EighthCellIV}{12\xspace}
\newcommand{\EighthCellIW}{8\xspace}
\newcommand{\EighthCellIX}{15\xspace}
\newcommand{\EighthCellIY}{1\xspace}
\newcommand{\EighthCellIZ}{0\xspace}
\newcommand{\EighthCellJA}{1\xspace}
\newcommand{\EighthCellJB}{1\xspace}
\newcommand{\EighthCellJC}{0\xspace}
\newcommand{\EighthCellJD}{1\xspace}
\newcommand{\EighthCellJE}{2\xspace}
\newcommand{\EighthCellJF}{0\xspace}
\newcommand{\EighthCellJG}{4\xspace}
\newcommand{\EighthCellJH}{2\xspace}
\newcommand{\EighthCellJI}{0\xspace}
\newcommand{\EighthCellJJ}{4\xspace}
\newcommand{\EighthCellJK}{0\xspace}
\newcommand{\EighthCellJL}{0\xspace}
\newcommand{\EighthCellJM}{1\xspace}
\newcommand{\EighthCellJN}{0\xspace}
\newcommand{\EighthCellJO}{0\xspace}
\newcommand{\EighthCellJP}{1\xspace}
\newcommand{\EighthCellJQ}{8\xspace}
\newcommand{\EighthCellJR}{5\xspace}
\newcommand{\EighthCellJS}{10\xspace}
\newcommand{\EighthCellJT}{8\xspace}
\newcommand{\EighthCellJU}{5\xspace}
\newcommand{\EighthCellJV}{10\xspace}
\newcommand{\EighthCellJW}{2\xspace}
\newcommand{\EighthCellJX}{1\xspace}
\newcommand{\EighthCellJY}{5\xspace}
\newcommand{\EighthCellJZ}{2\xspace}
\newcommand{\EighthCellKA}{0\xspace}
\newcommand{\EighthCellKB}{5\xspace}
\newcommand{\EighthCellKC}{4\xspace}
\newcommand{\EighthCellKD}{1\xspace}
\newcommand{\EighthCellKE}{7\xspace}
\newcommand{\EighthCellKF}{4\xspace}
\newcommand{\EighthCellKG}{2\xspace}
\newcommand{\EighthCellKH}{7\xspace}
\newcommand{\EighthCellKI}{1\xspace}
\newcommand{\EighthCellKJ}{0\xspace}
\newcommand{\EighthCellKK}{4\xspace}
\newcommand{\EighthCellKL}{1\xspace}
\newcommand{\EighthCellKM}{0\xspace}
\newcommand{\EighthCellKN}{4\xspace}
\newcommand{\EighthCellKO}{6\xspace}
\newcommand{\EighthCellKP}{4\xspace}
\newcommand{\EighthCellKQ}{10\xspace}
\newcommand{\EighthCellKR}{6\xspace}
\newcommand{\EighthCellKS}{5\xspace}
\newcommand{\EighthCellKT}{11\xspace}
\newcommand{\EighthCellKU}{5\xspace}
\newcommand{\EighthCellKV}{3\xspace}
\newcommand{\EighthCellKW}{11\xspace}
\newcommand{\EighthCellKX}{5\xspace}
\newcommand{\EighthCellKY}{2\xspace}
\newcommand{\EighthCellKZ}{11\xspace}
\newcommand{\EighthCellLA}{32\xspace}
\newcommand{\EighthCellLB}{25\xspace}
\newcommand{\EighthCellLC}{37\xspace}
\newcommand{\EighthCellLD}{32\xspace}
\newcommand{\EighthCellLE}{27\xspace}
\newcommand{\EighthCellLF}{39\xspace}
\newcommand{\EighthCellLG}{2\xspace}
\newcommand{\EighthCellLH}{0\xspace}
\newcommand{\EighthCellLI}{5\xspace}
\newcommand{\EighthCellLJ}{1\xspace}
\newcommand{\EighthCellLK}{0\xspace}
\newcommand{\EighthCellLL}{5\xspace}
\newcommand{\EighthCellLM}{1\xspace}
\newcommand{\EighthCellLN}{0\xspace}
\newcommand{\EighthCellLO}{3\xspace}
\newcommand{\EighthCellLP}{1\xspace}
\newcommand{\EighthCellLQ}{0\xspace}
\newcommand{\EighthCellLR}{4\xspace}
\newcommand{\EighthCellLS}{2\xspace}
\newcommand{\EighthCellLT}{0\xspace}
\newcommand{\EighthCellLU}{4\xspace}
\newcommand{\EighthCellLV}{2\xspace}
\newcommand{\EighthCellLW}{0\xspace}
\newcommand{\EighthCellLX}{4\xspace}
\newcommand{\EighthCellLY}{1\xspace}
\newcommand{\EighthCellLZ}{0\xspace}
\newcommand{\EighthCellMA}{4\xspace}
\newcommand{\EighthCellMB}{1\xspace}
\newcommand{\EighthCellMC}{0\xspace}
\newcommand{\EighthCellMD}{4\xspace}
\newcommand{\EighthCellME}{4\xspace}
\newcommand{\EighthCellMF}{1\xspace}
\newcommand{\EighthCellMG}{7\xspace}
\newcommand{\EighthCellMH}{4\xspace}
\newcommand{\EighthCellMI}{1\xspace}
\newcommand{\EighthCellMJ}{6\xspace}
\newcommand{\EighthCellMK}{4\xspace}
\newcommand{\EighthCellML}{0\xspace}
\newcommand{\EighthCellMM}{4\xspace}
\newcommand{\EighthCellMN}{4\xspace}
\newcommand{\EighthCellMO}{0\xspace}
\newcommand{\EighthCellMP}{5\xspace}
\newcommand{\EighthCellMQ}{2\xspace}
\newcommand{\EighthCellMR}{0\xspace}
\newcommand{\EighthCellMS}{4\xspace}
\newcommand{\EighthCellMT}{1\xspace}
\newcommand{\EighthCellMU}{0\xspace}
\newcommand{\EighthCellMV}{4\xspace}
\newcommand{\EighthCellMW}{0\xspace}
\newcommand{\EighthCellMX}{0\xspace}
\newcommand{\EighthCellMY}{5\xspace}
\newcommand{\EighthCellMZ}{0\xspace}
\newcommand{\EighthCellNA}{0\xspace}
\newcommand{\EighthCellNB}{5\xspace}
\newcommand{\EighthCellNC}{3\xspace}
\newcommand{\EighthCellND}{0\xspace}
\newcommand{\EighthCellNE}{4\xspace}
\newcommand{\EighthCellNF}{2\xspace}
\newcommand{\EighthCellNG}{0\xspace}
\newcommand{\EighthCellNH}{3\xspace}
\newcommand{\EighthCellNI}{1\xspace}
\newcommand{\EighthCellNJ}{0\xspace}
\newcommand{\EighthCellNK}{4\xspace}
\newcommand{\EighthCellNL}{1\xspace}
\newcommand{\EighthCellNM}{0\xspace}
\newcommand{\EighthCellNN}{5\xspace}
\newcommand{\EighthCellNO}{13\xspace}
\newcommand{\EighthCellNP}{7\xspace}
\newcommand{\EighthCellNQ}{17\xspace}
\newcommand{\EighthCellNR}{10\xspace}
\newcommand{\EighthCellNS}{5\xspace}
\newcommand{\EighthCellNT}{15\xspace}
\newcommand{\EighthCellNU}{7\xspace}
\newcommand{\EighthCellNV}{3\xspace}
\newcommand{\EighthCellNW}{13\xspace}
\newcommand{\EighthCellNX}{7\xspace}
\newcommand{\EighthCellNY}{4\xspace}
\newcommand{\EighthCellNZ}{14.025\xspace}
\newcommand{\EighthCellOA}{0\xspace}
\newcommand{\EighthCellOB}{0\xspace}
\newcommand{\EighthCellOC}{3\xspace}
\newcommand{\EighthCellOD}{0\xspace}
\newcommand{\EighthCellOE}{0\xspace}
\newcommand{\EighthCellOF}{2\xspace}
\newcommand{\EighthCellOG}{1\xspace}
\newcommand{\EighthCellOH}{0\xspace}
\newcommand{\EighthCellOI}{5\xspace}
\newcommand{\EighthCellOJ}{1\xspace}
\newcommand{\EighthCellOK}{0\xspace}
\newcommand{\EighthCellOL}{5\xspace}
\newcommand{\EighthCellOM}{5\xspace}
\newcommand{\EighthCellON}{1\xspace}
\newcommand{\EighthCellOO}{6\xspace}
\newcommand{\EighthCellOP}{3\xspace}
\newcommand{\EighthCellOQ}{1\xspace}
\newcommand{\EighthCellOR}{6\xspace}
\newcommand{\EighthCellOS}{1\xspace}
\newcommand{\EighthCellOT}{0\xspace}
\newcommand{\EighthCellOU}{3\xspace}
\newcommand{\EighthCellOV}{1\xspace}
\newcommand{\EighthCellOW}{0\xspace}
\newcommand{\EighthCellOX}{4\xspace}
\newcommand{\EighthCellOY}{1\xspace}
\newcommand{\EighthCellOZ}{0\xspace}
\newcommand{\EighthCellPA}{3\xspace}
\newcommand{\EighthCellPB}{1\xspace}
\newcommand{\EighthCellPC}{0\xspace}
\newcommand{\EighthCellPD}{3\xspace}
\newcommand{\EighthCellPE}{0\xspace}
\newcommand{\EighthCellPF}{0\xspace}
\newcommand{\EighthCellPG}{3\xspace}
\newcommand{\EighthCellPH}{0\xspace}
\newcommand{\EighthCellPI}{0\xspace}
\newcommand{\EighthCellPJ}{3\xspace}
\newcommand{\EighthCellPK}{4\xspace}
\newcommand{\EighthCellPL}{0\xspace}
\newcommand{\EighthCellPM}{5\xspace}
\newcommand{\EighthCellPN}{3\xspace}
\newcommand{\EighthCellPO}{0\xspace}
\newcommand{\EighthCellPP}{5\xspace}
\newcommand{\EighthCellPQ}{5\xspace}
\newcommand{\EighthCellPR}{1\xspace}
\newcommand{\EighthCellPS}{6\xspace}
\newcommand{\EighthCellPT}{6\xspace}
\newcommand{\EighthCellPU}{1\xspace}
\newcommand{\EighthCellPV}{7\xspace}
\newcommand{\EighthCellPW}{1\xspace}
\newcommand{\EighthCellPX}{0\xspace}
\newcommand{\EighthCellPY}{5\xspace}
\newcommand{\EighthCellPZ}{1\xspace}
\newcommand{\EighthCellQA}{0\xspace}
\newcommand{\EighthCellQB}{4\xspace}
\newcommand{\EighthCellQC}{5\xspace}
\newcommand{\EighthCellQD}{1\xspace}
\newcommand{\EighthCellQE}{7\xspace}
\newcommand{\EighthCellQF}{5\xspace}
\newcommand{\EighthCellQG}{1\xspace}
\newcommand{\EighthCellQH}{8\xspace}
\newcommand{\EighthCellQI}{11\xspace}
\newcommand{\EighthCellQJ}{6\xspace}
\newcommand{\EighthCellQK}{16\xspace}
\newcommand{\EighthCellQL}{8\xspace}
\newcommand{\EighthCellQM}{5\xspace}
\newcommand{\EighthCellQN}{14\xspace}
\newcommand{\EighthCellQO}{12\xspace}
\newcommand{\EighthCellQP}{6\xspace}
\newcommand{\EighthCellQQ}{17\xspace}
\newcommand{\EighthCellQR}{13\xspace}
\newcommand{\EighthCellQS}{7\xspace}
\newcommand{\EighthCellQT}{19\xspace}
\newcommand{\EighthCellQU}{0.172799\xspace}
\newcommand{\EighthCellQV}{0.179010\xspace}
\newcommand{\EighthCellQW}{0.173633\xspace}
\newcommand{\EighthCellQX}{0.178085\xspace}
\newcommand{\EighthCellQY}{500\xspace}
\newcommand{\EighthCellQZ}{0.441748\xspace}
\newcommand{\EighthCellRA}{0.201588\xspace}
\newcommand{\EighthCellRB}{0.231254\xspace}
\newcommand{\EighthCellRC}{0.446113\xspace}
\newcommand{\EighthCellRD}{0.446113\xspace}
\newcommand{\EighthCellRE}{500\xspace}
\newcommand{\EighthCellRF}{0.417312\xspace}
\newcommand{\EighthCellRG}{0.227587\xspace}
\newcommand{\EighthCellRH}{0.283376\xspace}
\newcommand{\EighthCellRI}{0.425597\xspace}
\newcommand{\EighthCellRJ}{0.425597\xspace}
\newcommand{\EighthCellRK}{250\xspace}
\newcommand{\EighthCellRL}{0.450594\xspace}
\newcommand{\EighthCellRM}{0.451760\xspace}
\newcommand{\EighthCellRN}{0.460655\xspace}
\newcommand{\EighthCellRO}{0.446269\xspace}
\newcommand{\EighthCellRP}{0.446269\xspace}

\newcommand{\EighthCellRR}{221.2549\xspace}
\newcommand{\EighthCellRS}{668.3412\xspace}
\newcommand{\EighthCellRT}{0.0284\xspace}
\newcommand{\EighthCellRU}{0.0379\xspace}

\newcommand{\EighthCellRW}{203.3134\xspace}
\newcommand{\EighthCellRX}{570.7405\xspace}
\newcommand{\EighthCellRY}{0.0261\xspace}
\newcommand{\EighthCellRZ}{0.0324\xspace}

\newcommand{\EighthCellSB}{185.7291\xspace}
\newcommand{\EighthCellSC}{516.5718\xspace}
\newcommand{\EighthCellSD}{0.0239\xspace}
\newcommand{\EighthCellSE}{0.0293\xspace}

\newcommand{\EighthCellSG}{153.7272\xspace}
\newcommand{\EighthCellSH}{238.3692\xspace}
\newcommand{\EighthCellSI}{0.0198\xspace}
\newcommand{\EighthCellSJ}{0.0135\xspace}

\newcommand{\EighthCellSL}{143.1458\xspace}
\newcommand{\EighthCellSM}{429.7200\xspace}
\newcommand{\EighthCellSN}{0.0184\xspace}
\newcommand{\EighthCellSO}{0.0244\xspace}

\newcommand{\EighthCellSQ}{136.1392\xspace}
\newcommand{\EighthCellSR}{373.3296\xspace}
\newcommand{\EighthCellSS}{0.0175\xspace}
\newcommand{\EighthCellST}{0.0212\xspace}

\newcommand{\EighthCellSV}{134.8628\xspace}
\newcommand{\EighthCellSW}{269.0107\xspace}
\newcommand{\EighthCellSX}{0.0173\xspace}
\newcommand{\EighthCellSY}{0.0153\xspace}

\newcommand{\EighthCellTA}{130.4964\xspace}
\newcommand{\EighthCellTB}{342.5268\xspace}
\newcommand{\EighthCellTC}{0.0168\xspace}
\newcommand{\EighthCellTD}{0.0194\xspace}

\newcommand{\EighthCellTF}{130.4680\xspace}
\newcommand{\EighthCellTG}{222.3263\xspace}
\newcommand{\EighthCellTH}{0.0168\xspace}
\newcommand{\EighthCellTI}{0.0126\xspace}

\newcommand{\EighthCellTK}{129.7243\xspace}
\newcommand{\EighthCellTL}{443.1864\xspace}
\newcommand{\EighthCellTM}{0.0167\xspace}
\newcommand{\EighthCellTN}{0.0252\xspace}

\newcommand{\EighthCellTP}{359.6369\xspace}
\newcommand{\EighthCellTQ}{1193.6802\xspace}
\newcommand{\EighthCellTR}{0.0481\xspace}
\newcommand{\EighthCellTS}{0.0667\xspace}

\newcommand{\EighthCellTU}{243.5246\xspace}
\newcommand{\EighthCellTV}{809.9921\xspace}
\newcommand{\EighthCellTW}{0.0326\xspace}
\newcommand{\EighthCellTX}{0.0452\xspace}

\newcommand{\EighthCellTZ}{176.7439\xspace}
\newcommand{\EighthCellUA}{621.8869\xspace}
\newcommand{\EighthCellUB}{0.0236\xspace}
\newcommand{\EighthCellUC}{0.0347\xspace}

\newcommand{\EighthCellUE}{153.1525\xspace}
\newcommand{\EighthCellUF}{405.1327\xspace}
\newcommand{\EighthCellUG}{0.0205\xspace}
\newcommand{\EighthCellUH}{0.0226\xspace}

\newcommand{\EighthCellUJ}{150.3014\xspace}
\newcommand{\EighthCellUK}{554.6524\xspace}
\newcommand{\EighthCellUL}{0.0201\xspace}
\newcommand{\EighthCellUM}{0.0310\xspace}

\newcommand{\EighthCellUO}{135.9684\xspace}
\newcommand{\EighthCellUP}{360.4956\xspace}
\newcommand{\EighthCellUQ}{0.0182\xspace}
\newcommand{\EighthCellUR}{0.0201\xspace}

\newcommand{\EighthCellUT}{132.9938\xspace}
\newcommand{\EighthCellUU}{389.3682\xspace}
\newcommand{\EighthCellUV}{0.0178\xspace}
\newcommand{\EighthCellUW}{0.0217\xspace}

\newcommand{\EighthCellUY}{122.7852\xspace}
\newcommand{\EighthCellUZ}{350.7590\xspace}
\newcommand{\EighthCellVA}{0.0164\xspace}
\newcommand{\EighthCellVB}{0.0196\xspace}

\newcommand{\EighthCellVD}{115.0718\xspace}
\newcommand{\EighthCellVE}{152.8025\xspace}
\newcommand{\EighthCellVF}{0.0154\xspace}
\newcommand{\EighthCellVG}{0.0085\xspace}

\newcommand{\EighthCellVI}{110.7799\xspace}
\newcommand{\EighthCellVJ}{236.3962\xspace}
\newcommand{\EighthCellVK}{0.0148\xspace}
\newcommand{\EighthCellVL}{0.0132\xspace}

\newcommand{\EighthCellVN}{640.7925\xspace}
\newcommand{\EighthCellVO}{1313.4507\xspace}
\newcommand{\EighthCellVP}{0.1126\xspace}
\newcommand{\EighthCellVQ}{0.1040\xspace}

\newcommand{\EighthCellVS}{307.0712\xspace}
\newcommand{\EighthCellVT}{312.1440\xspace}
\newcommand{\EighthCellVU}{0.0539\xspace}
\newcommand{\EighthCellVV}{0.0247\xspace}

\newcommand{\EighthCellVX}{267.3303\xspace}
\newcommand{\EighthCellVY}{781.6648\xspace}
\newcommand{\EighthCellVZ}{0.0470\xspace}
\newcommand{\EighthCellWA}{0.0619\xspace}

\newcommand{\EighthCellWC}{266.5373\xspace}
\newcommand{\EighthCellWD}{899.7502\xspace}
\newcommand{\EighthCellWE}{0.0468\xspace}
\newcommand{\EighthCellWF}{0.0712\xspace}

\newcommand{\EighthCellWH}{215.0268\xspace}
\newcommand{\EighthCellWI}{914.9520\xspace}
\newcommand{\EighthCellWJ}{0.0378\xspace}
\newcommand{\EighthCellWK}{0.0724\xspace}

\newcommand{\EighthCellWM}{200.3951\xspace}
\newcommand{\EighthCellWN}{331.5416\xspace}
\newcommand{\EighthCellWO}{0.0352\xspace}
\newcommand{\EighthCellWP}{0.0262\xspace}

\newcommand{\EighthCellWR}{196.8857\xspace}
\newcommand{\EighthCellWS}{559.5544\xspace}
\newcommand{\EighthCellWT}{0.0346\xspace}
\newcommand{\EighthCellWU}{0.0443\xspace}

\newcommand{\EighthCellWW}{184.6641\xspace}
\newcommand{\EighthCellWX}{279.0496\xspace}
\newcommand{\EighthCellWY}{0.0324\xspace}
\newcommand{\EighthCellWZ}{0.0221\xspace}

\newcommand{\EighthCellXB}{167.4601\xspace}
\newcommand{\EighthCellXC}{268.9942\xspace}
\newcommand{\EighthCellXD}{0.0294\xspace}
\newcommand{\EighthCellXE}{0.0213\xspace}

\newcommand{\EighthCellXG}{125.7692\xspace}
\newcommand{\EighthCellXH}{159.1587\xspace}
\newcommand{\EighthCellXI}{0.0221\xspace}
\newcommand{\EighthCellXJ}{0.0126\xspace}
\newcommand{\EighthCellXK}{0.001415\xspace}
\newcommand{\EighthCellXL}{0.001324\xspace}
\newcommand{\EighthCellXM}{0.998585\xspace}
\newcommand{\EighthCellXN}{0.998676\xspace}
\newcommand{\EighthCellXO}{0.002174\xspace}
\newcommand{\EighthCellXP}{0.002680\xspace}
\newcommand{\EighthCellXQ}{0.997826\xspace}
\newcommand{\EighthCellXR}{0.997320\xspace}
\newcommand{\EighthCellXS}{0.001292\xspace}
\newcommand{\EighthCellXT}{0.001085\xspace}
\newcommand{\EighthCellXU}{0.998708\xspace}
\newcommand{\EighthCellXV}{0.998915\xspace}

\newcommand{\EighthCellXX}{11\xspace}
\newcommand{\EighthCellXY}{15\xspace}
\newcommand{\EighthCellXZ}{46.6589\xspace}
\newcommand{\EighthCellYA}{73.3217\xspace}
\newcommand{\EighthCellYB}{0.0060\xspace}
\newcommand{\EighthCellYC}{0.0042\xspace}

\newcommand{\EighthCellYE}{12\xspace}
\newcommand{\EighthCellYF}{0\xspace}
\newcommand{\EighthCellYG}{31.1495\xspace}
\newcommand{\EighthCellYH}{51.1882\xspace}
\newcommand{\EighthCellYI}{0.0040\xspace}
\newcommand{\EighthCellYJ}{0.0029\xspace}

\newcommand{\EighthCellYL}{7\xspace}
\newcommand{\EighthCellYM}{15\xspace}
\newcommand{\EighthCellYN}{30.5088\xspace}
\newcommand{\EighthCellYO}{91.1676\xspace}
\newcommand{\EighthCellYP}{0.0039\xspace}
\newcommand{\EighthCellYQ}{0.0052\xspace}

\newcommand{\EighthCellYS}{5\xspace}
\newcommand{\EighthCellYT}{15\xspace}
\newcommand{\EighthCellYU}{28.1634\xspace}
\newcommand{\EighthCellYV}{72.4099\xspace}
\newcommand{\EighthCellYW}{0.0036\xspace}
\newcommand{\EighthCellYX}{0.0041\xspace}

\newcommand{\EighthCellYZ}{7\xspace}
\newcommand{\EighthCellZA}{15\xspace}
\newcommand{\EighthCellZB}{27.6647\xspace}
\newcommand{\EighthCellZC}{65.5733\xspace}
\newcommand{\EighthCellZD}{0.0036\xspace}
\newcommand{\EighthCellZE}{0.0037\xspace}

\newcommand{\EighthCellZG}{2\xspace}
\newcommand{\EighthCellZH}{11\xspace}
\newcommand{\EighthCellZI}{27.5178\xspace}
\newcommand{\EighthCellZJ}{75.6526\xspace}
\newcommand{\EighthCellZK}{0.0035\xspace}
\newcommand{\EighthCellZL}{0.0043\xspace}

\newcommand{\EighthCellZN}{0\xspace}
\newcommand{\EighthCellZO}{3\xspace}
\newcommand{\EighthCellZP}{26.2774\xspace}
\newcommand{\EighthCellZQ}{53.3844\xspace}
\newcommand{\EighthCellZR}{0.0034\xspace}
\newcommand{\EighthCellZS}{0.0030\xspace}

\newcommand{\EighthCellZU}{0\xspace}
\newcommand{\EighthCellZV}{15\xspace}
\newcommand{\EighthCellZW}{26.2171\xspace}
\newcommand{\EighthCellZX}{58.3341\xspace}
\newcommand{\EighthCellZY}{0.0034\xspace}
\newcommand{\EighthCellZZ}{0.0033\xspace}

\newcommand{\EighthCellAAB}{13\xspace}
\newcommand{\EighthCellAAC}{15\xspace}
\newcommand{\EighthCellAAD}{25.4989\xspace}
\newcommand{\EighthCellAAE}{68.4147\xspace}
\newcommand{\EighthCellAAF}{0.0033\xspace}
\newcommand{\EighthCellAAG}{0.0039\xspace}

\newcommand{\EighthCellAAI}{3\xspace}
\newcommand{\EighthCellAAJ}{15\xspace}
\newcommand{\EighthCellAAK}{24.5615\xspace}
\newcommand{\EighthCellAAL}{72.6132\xspace}
\newcommand{\EighthCellAAM}{0.0032\xspace}
\newcommand{\EighthCellAAN}{0.0041\xspace}

\newcommand{\EighthCellAAP}{13\xspace}
\newcommand{\EighthCellAAQ}{15\xspace}
\newcommand{\EighthCellAAR}{30.7484\xspace}
\newcommand{\EighthCellAAS}{73.2226\xspace}
\newcommand{\EighthCellAAT}{0.0041\xspace}
\newcommand{\EighthCellAAU}{0.0041\xspace}

\newcommand{\EighthCellAAW}{6\xspace}
\newcommand{\EighthCellAAX}{12\xspace}
\newcommand{\EighthCellAAY}{26.7758\xspace}
\newcommand{\EighthCellAAZ}{65.0952\xspace}
\newcommand{\EighthCellABA}{0.0036\xspace}
\newcommand{\EighthCellABB}{0.0036\xspace}

\newcommand{\EighthCellABD}{7\xspace}
\newcommand{\EighthCellABE}{15\xspace}
\newcommand{\EighthCellABF}{25.6841\xspace}
\newcommand{\EighthCellABG}{62.9360\xspace}
\newcommand{\EighthCellABH}{0.0034\xspace}
\newcommand{\EighthCellABI}{0.0035\xspace}

\newcommand{\EighthCellABK}{1\xspace}
\newcommand{\EighthCellABL}{15\xspace}
\newcommand{\EighthCellABM}{25.0992\xspace}
\newcommand{\EighthCellABN}{89.1706\xspace}
\newcommand{\EighthCellABO}{0.0034\xspace}
\newcommand{\EighthCellABP}{0.0050\xspace}

\newcommand{\EighthCellABR}{15\xspace}
\newcommand{\EighthCellABS}{0\xspace}
\newcommand{\EighthCellABT}{22.6323\xspace}
\newcommand{\EighthCellABU}{44.5713\xspace}
\newcommand{\EighthCellABV}{0.0030\xspace}
\newcommand{\EighthCellABW}{0.0025\xspace}

\newcommand{\EighthCellABY}{2\xspace}
\newcommand{\EighthCellABZ}{9\xspace}
\newcommand{\EighthCellACA}{22.2180\xspace}
\newcommand{\EighthCellACB}{75.0060\xspace}
\newcommand{\EighthCellACC}{0.0030\xspace}
\newcommand{\EighthCellACD}{0.0042\xspace}

\newcommand{\EighthCellACF}{13\xspace}
\newcommand{\EighthCellACG}{15\xspace}
\newcommand{\EighthCellACH}{21.0935\xspace}
\newcommand{\EighthCellACI}{43.8139\xspace}
\newcommand{\EighthCellACJ}{0.0028\xspace}
\newcommand{\EighthCellACK}{0.0024\xspace}

\newcommand{\EighthCellACM}{9\xspace}
\newcommand{\EighthCellACN}{15\xspace}
\newcommand{\EighthCellACO}{21.0019\xspace}
\newcommand{\EighthCellACP}{68.0261\xspace}
\newcommand{\EighthCellACQ}{0.0028\xspace}
\newcommand{\EighthCellACR}{0.0038\xspace}

\newcommand{\EighthCellACT}{6\xspace}
\newcommand{\EighthCellACU}{15\xspace}
\newcommand{\EighthCellACV}{20.9989\xspace}
\newcommand{\EighthCellACW}{51.4659\xspace}
\newcommand{\EighthCellACX}{0.0028\xspace}
\newcommand{\EighthCellACY}{0.0029\xspace}

\newcommand{\EighthCellADA}{14\xspace}
\newcommand{\EighthCellADB}{15\xspace}
\newcommand{\EighthCellADC}{20.6871\xspace}
\newcommand{\EighthCellADD}{50.0633\xspace}
\newcommand{\EighthCellADE}{0.0028\xspace}
\newcommand{\EighthCellADF}{0.0028\xspace}

\newcommand{\EighthCellADH}{12\xspace}
\newcommand{\EighthCellADI}{15\xspace}
\newcommand{\EighthCellADJ}{28.4815\xspace}
\newcommand{\EighthCellADK}{49.6983\xspace}
\newcommand{\EighthCellADL}{0.0050\xspace}
\newcommand{\EighthCellADM}{0.0039\xspace}

\newcommand{\EighthCellADO}{2\xspace}
\newcommand{\EighthCellADP}{15\xspace}
\newcommand{\EighthCellADQ}{27.8701\xspace}
\newcommand{\EighthCellADR}{77.7529\xspace}
\newcommand{\EighthCellADS}{0.0049\xspace}
\newcommand{\EighthCellADT}{0.0062\xspace}

\newcommand{\EighthCellADV}{0\xspace}
\newcommand{\EighthCellADW}{12\xspace}
\newcommand{\EighthCellADX}{27.4957\xspace}
\newcommand{\EighthCellADY}{65.4243\xspace}
\newcommand{\EighthCellADZ}{0.0048\xspace}
\newcommand{\EighthCellAEA}{0.0052\xspace}

\newcommand{\EighthCellAEC}{2\xspace}
\newcommand{\EighthCellAED}{0\xspace}
\newcommand{\EighthCellAEE}{26.3361\xspace}
\newcommand{\EighthCellAEF}{30.3242\xspace}
\newcommand{\EighthCellAEG}{0.0046\xspace}
\newcommand{\EighthCellAEH}{0.0024\xspace}

\newcommand{\EighthCellAEJ}{13\xspace}
\newcommand{\EighthCellAEK}{15\xspace}
\newcommand{\EighthCellAEL}{26.2296\xspace}
\newcommand{\EighthCellAEM}{52.5436\xspace}
\newcommand{\EighthCellAEN}{0.0046\xspace}
\newcommand{\EighthCellAEO}{0.0042\xspace}

\newcommand{\EighthCellAEQ}{15\xspace}
\newcommand{\EighthCellAER}{1\xspace}
\newcommand{\EighthCellAES}{25.2496\xspace}
\newcommand{\EighthCellAET}{44.5582\xspace}
\newcommand{\EighthCellAEU}{0.0044\xspace}
\newcommand{\EighthCellAEV}{0.0035\xspace}

\newcommand{\EighthCellAEX}{14\xspace}
\newcommand{\EighthCellAEY}{0\xspace}
\newcommand{\EighthCellAEZ}{25.0273\xspace}
\newcommand{\EighthCellAFA}{34.9420\xspace}
\newcommand{\EighthCellAFB}{0.0044\xspace}
\newcommand{\EighthCellAFC}{0.0028\xspace}

\newcommand{\EighthCellAFE}{15\xspace}
\newcommand{\EighthCellAFF}{2\xspace}
\newcommand{\EighthCellAFG}{23.6098\xspace}
\newcommand{\EighthCellAFH}{37.0772\xspace}
\newcommand{\EighthCellAFI}{0.0041\xspace}
\newcommand{\EighthCellAFJ}{0.0029\xspace}

\newcommand{\EighthCellAFL}{1\xspace}
\newcommand{\EighthCellAFM}{15\xspace}
\newcommand{\EighthCellAFN}{23.0698\xspace}
\newcommand{\EighthCellAFO}{55.6904\xspace}
\newcommand{\EighthCellAFP}{0.0041\xspace}
\newcommand{\EighthCellAFQ}{0.0044\xspace}

\newcommand{\EighthCellAFS}{9\xspace}
\newcommand{\EighthCellAFT}{15\xspace}
\newcommand{\EighthCellAFU}{22.9631\xspace}
\newcommand{\EighthCellAFV}{60.7388\xspace}
\newcommand{\EighthCellAFW}{0.0040\xspace}
\newcommand{\EighthCellAFX}{0.0048\xspace}
\newcommand{\EighthCellAFY}{0\xspace}
\newcommand{\EighthCellAFZ}{0\xspace}
\newcommand{\EighthCellAGA}{0.0877\xspace}
\newcommand{\EighthCellAGB}{0.0663\xspace}
\newcommand{\EighthCellAGC}{0.0801\xspace}
\newcommand{\EighthCellAGD}{0.0627\xspace}
\newcommand{\EighthCellAGE}{0.0980\xspace}
\newcommand{\EighthCellAGF}{0.0740\xspace}
\newcommand{\EighthCellAGG}{0\xspace}
\newcommand{\EighthCellAGH}{1\xspace}
\newcommand{\EighthCellAGI}{0.0435\xspace}
\newcommand{\EighthCellAGJ}{0.0421\xspace}
\newcommand{\EighthCellAGK}{0.0358\xspace}
\newcommand{\EighthCellAGL}{0.0316\xspace}
\newcommand{\EighthCellAGM}{0.0392\xspace}
\newcommand{\EighthCellAGN}{0.0343\xspace}
\newcommand{\EighthCellAGO}{0\xspace}
\newcommand{\EighthCellAGP}{2\xspace}
\newcommand{\EighthCellAGQ}{0.0375\xspace}
\newcommand{\EighthCellAGR}{0.0428\xspace}
\newcommand{\EighthCellAGS}{0.0405\xspace}
\newcommand{\EighthCellAGT}{0.0476\xspace}
\newcommand{\EighthCellAGU}{0.0309\xspace}
\newcommand{\EighthCellAGV}{0.0343\xspace}
\newcommand{\EighthCellAGW}{0\xspace}
\newcommand{\EighthCellAGX}{3\xspace}
\newcommand{\EighthCellAGY}{0.1058\xspace}
\newcommand{\EighthCellAGZ}{0.1309\xspace}
\newcommand{\EighthCellAHA}{0.0826\xspace}
\newcommand{\EighthCellAHB}{0.0974\xspace}
\newcommand{\EighthCellAHC}{0.1028\xspace}
\newcommand{\EighthCellAHD}{0.1316\xspace}
\newcommand{\EighthCellAHE}{1\xspace}
\newcommand{\EighthCellAHF}{0\xspace}
\newcommand{\EighthCellAHG}{0.0784\xspace}
\newcommand{\EighthCellAHH}{0.0599\xspace}
\newcommand{\EighthCellAHI}{0.0762\xspace}
\newcommand{\EighthCellAHJ}{0.0565\xspace}
\newcommand{\EighthCellAHK}{0.0938\xspace}
\newcommand{\EighthCellAHL}{0.0729\xspace}
\newcommand{\EighthCellAHM}{1\xspace}
\newcommand{\EighthCellAHN}{1\xspace}
\newcommand{\EighthCellAHO}{0.0221\xspace}
\newcommand{\EighthCellAHP}{0.0189\xspace}
\newcommand{\EighthCellAHQ}{0.0222\xspace}
\newcommand{\EighthCellAHR}{0.0225\xspace}
\newcommand{\EighthCellAHS}{0.0265\xspace}
\newcommand{\EighthCellAHT}{0.0241\xspace}
\newcommand{\EighthCellAHU}{1\xspace}
\newcommand{\EighthCellAHV}{2\xspace}
\newcommand{\EighthCellAHW}{0.0332\xspace}
\newcommand{\EighthCellAHX}{0.0393\xspace}
\newcommand{\EighthCellAHY}{0.0420\xspace}
\newcommand{\EighthCellAHZ}{0.0517\xspace}
\newcommand{\EighthCellAIA}{0.0195\xspace}
\newcommand{\EighthCellAIB}{0.0235\xspace}
\newcommand{\EighthCellAIC}{1\xspace}
\newcommand{\EighthCellAID}{3\xspace}
\newcommand{\EighthCellAIE}{0.1123\xspace}
\newcommand{\EighthCellAIF}{0.1369\xspace}
\newcommand{\EighthCellAIG}{0.0902\xspace}
\newcommand{\EighthCellAIH}{0.1025\xspace}
\newcommand{\EighthCellAII}{0.0874\xspace}
\newcommand{\EighthCellAIJ}{0.1166\xspace}
\newcommand{\EighthCellAIK}{2\xspace}
\newcommand{\EighthCellAIL}{0\xspace}
\newcommand{\EighthCellAIM}{0.0796\xspace}
\newcommand{\EighthCellAIN}{0.0536\xspace}
\newcommand{\EighthCellAIO}{0.0770\xspace}
\newcommand{\EighthCellAIP}{0.0523\xspace}
\newcommand{\EighthCellAIQ}{0.0926\xspace}
\newcommand{\EighthCellAIR}{0.0688\xspace}
\newcommand{\EighthCellAIS}{2\xspace}
\newcommand{\EighthCellAIT}{1\xspace}
\newcommand{\EighthCellAIU}{0.0265\xspace}
\newcommand{\EighthCellAIV}{0.0233\xspace}
\newcommand{\EighthCellAIW}{0.0198\xspace}
\newcommand{\EighthCellAIX}{0.0204\xspace}
\newcommand{\EighthCellAIY}{0.0215\xspace}
\newcommand{\EighthCellAIZ}{0.0200\xspace}
\newcommand{\EighthCellAJA}{2\xspace}
\newcommand{\EighthCellAJB}{2\xspace}
\newcommand{\EighthCellAJC}{0.0286\xspace}
\newcommand{\EighthCellAJD}{0.0346\xspace}
\newcommand{\EighthCellAJE}{0.0338\xspace}
\newcommand{\EighthCellAJF}{0.0392\xspace}
\newcommand{\EighthCellAJG}{0.0196\xspace}
\newcommand{\EighthCellAJH}{0.0245\xspace}
\newcommand{\EighthCellAJI}{2\xspace}
\newcommand{\EighthCellAJJ}{3\xspace}
\newcommand{\EighthCellAJK}{0.0961\xspace}
\newcommand{\EighthCellAJL}{0.1122\xspace}
\newcommand{\EighthCellAJM}{0.1152\xspace}
\newcommand{\EighthCellAJN}{0.1409\xspace}
\newcommand{\EighthCellAJO}{0.0863\xspace}
\newcommand{\EighthCellAJP}{0.1141\xspace}
\newcommand{\EighthCellAJQ}{3\xspace}
\newcommand{\EighthCellAJR}{0\xspace}
\newcommand{\EighthCellAJS}{0.0994\xspace}
\newcommand{\EighthCellAJT}{0.0642\xspace}
\newcommand{\EighthCellAJU}{0.1055\xspace}
\newcommand{\EighthCellAJV}{0.0665\xspace}
\newcommand{\EighthCellAJW}{0.1372\xspace}
\newcommand{\EighthCellAJX}{0.0996\xspace}
\newcommand{\EighthCellAJY}{3\xspace}
\newcommand{\EighthCellAJZ}{1\xspace}
\newcommand{\EighthCellAKA}{0.0264\xspace}
\newcommand{\EighthCellAKB}{0.0229\xspace}
\newcommand{\EighthCellAKC}{0.0219\xspace}
\newcommand{\EighthCellAKD}{0.0214\xspace}
\newcommand{\EighthCellAKE}{0.0361\xspace}
\newcommand{\EighthCellAKF}{0.0325\xspace}
\newcommand{\EighthCellAKG}{3\xspace}
\newcommand{\EighthCellAKH}{2\xspace}
\newcommand{\EighthCellAKI}{0.0215\xspace}
\newcommand{\EighthCellAKJ}{0.0257\xspace}
\newcommand{\EighthCellAKK}{0.0294\xspace}
\newcommand{\EighthCellAKL}{0.0368\xspace}
\newcommand{\EighthCellAKM}{0.0186\xspace}
\newcommand{\EighthCellAKN}{0.0215\xspace}
\newcommand{\EighthCellAKO}{3\xspace}
\newcommand{\EighthCellAKP}{3\xspace}
\newcommand{\EighthCellAKQ}{0.1001\xspace}
\newcommand{\EighthCellAKR}{0.1250\xspace}
\newcommand{\EighthCellAKS}{0.1257\xspace}
\newcommand{\EighthCellAKT}{0.1474\xspace}
\newcommand{\EighthCellAKU}{0.0886\xspace}
\newcommand{\EighthCellAKV}{0.1066\xspace}
\newcommand{\IntervalLevelPct}{95\xspace}

\newcommand{\NinthFitJoint}{0.246\xspace}

\newcommand{\NinthFitCodes}{0.157\xspace}

\newcommand{\NinthFitOverlap}{0.137\xspace}

\newcommand{\NinthFitMagnitude}{0.00396\xspace}

\newcommand{\NinthErrJointA}{42\xspace}
\newcommand{\NinthErrJointB}{27\xspace}
\newcommand{\NinthErrJointC}{28\xspace}
\newcommand{\NinthErrCodesA}{33\xspace}
\newcommand{\NinthMoreCodesA}{-9\xspace}
\newcommand{\NinthMoreLowCodesA}{-17\xspace}
\newcommand{\NinthMoreHighCodesA}{-2\xspace}
\newcommand{\NinthErrCodesB}{54\xspace}
\newcommand{\NinthMoreCodesB}{27\xspace}
\newcommand{\NinthMoreLowCodesB}{13\xspace}
\newcommand{\NinthMoreHighCodesB}{44\xspace}
\newcommand{\NinthErrCodesC}{37\xspace}
\newcommand{\NinthMoreCodesC}{9\xspace}
\newcommand{\NinthMoreLowCodesC}{-1\xspace}
\newcommand{\NinthMoreHighCodesC}{21\xspace}
\newcommand{\NinthErrOverlapA}{41\xspace}

\newcommand{\NinthMoreLowOverlapA}{-9\xspace}
\newcommand{\NinthMoreHighOverlapA}{7\xspace}
\newcommand{\NinthErrOverlapB}{57\xspace}

\newcommand{\NinthMoreLowOverlapB}{15\xspace}
\newcommand{\NinthMoreHighOverlapB}{47\xspace}
\newcommand{\NinthErrOverlapC}{45\xspace}

\newcommand{\NinthMoreLowOverlapC}{3\xspace}
\newcommand{\NinthMoreHighOverlapC}{33\xspace}
\newcommand{\NinthErrMagnitudeA}{1,102\xspace}

\newcommand{\NinthMoreLowMagnitudeA}{992\xspace}
\newcommand{\NinthMoreHighMagnitudeA}{1,126\xspace}
\newcommand{\NinthErrMagnitudeB}{594\xspace}

\newcommand{\NinthMoreLowMagnitudeB}{514\xspace}
\newcommand{\NinthMoreHighMagnitudeB}{626\xspace}
\newcommand{\NinthErrMagnitudeC}{328\xspace}

\newcommand{\NinthMoreLowMagnitudeC}{259\xspace}
\newcommand{\NinthMoreHighMagnitudeC}{346\xspace}
\newcommand{\NinthErrCosineA}{65\xspace}
\newcommand{\NinthMoreCosineA}{23\xspace}
\newcommand{\NinthMoreLowCosineA}{12\xspace}
\newcommand{\NinthMoreHighCosineA}{34\xspace}
\newcommand{\NinthErrCosineB}{56\xspace}
\newcommand{\NinthMoreCosineB}{29\xspace}
\newcommand{\NinthMoreLowCosineB}{14\xspace}
\newcommand{\NinthMoreHighCosineB}{46\xspace}
\newcommand{\NinthErrCosineC}{81\xspace}
\newcommand{\NinthMoreCosineC}{53\xspace}
\newcommand{\NinthMoreLowCosineC}{35\xspace}
\newcommand{\NinthMoreHighCosineC}{75\xspace}
\newcommand{\NinthFreshErrJointA}{37\xspace}
\newcommand{\NinthFreshErrJointB}{17\xspace}
\newcommand{\NinthFreshErrJointC}{21\xspace}
\newcommand{\NinthFreshErrCodesA}{32\xspace}

\newcommand{\NinthFreshMoreLowCodesA}{-12\xspace}
\newcommand{\NinthFreshMoreHighCodesA}{1\xspace}
\newcommand{\NinthFreshErrCodesB}{36\xspace}

\newcommand{\NinthFreshMoreLowCodesB}{6\xspace}
\newcommand{\NinthFreshMoreHighCodesB}{34\xspace}
\newcommand{\NinthFreshErrCodesC}{28\xspace}

\newcommand{\NinthFreshMoreLowCodesC}{-3\xspace}
\newcommand{\NinthFreshMoreHighCodesC}{18\xspace}
\newcommand{\NinthFreshErrOverlapA}{37\xspace}

\newcommand{\NinthFreshMoreLowOverlapA}{-8\xspace}
\newcommand{\NinthFreshMoreHighOverlapA}{9\xspace}
\newcommand{\NinthFreshErrOverlapB}{51\xspace}

\newcommand{\NinthFreshMoreLowOverlapB}{18\xspace}
\newcommand{\NinthFreshMoreHighOverlapB}{51\xspace}
\newcommand{\NinthFreshErrOverlapC}{41\xspace}

\newcommand{\NinthFreshMoreLowOverlapC}{8\xspace}
\newcommand{\NinthFreshMoreHighOverlapC}{33\xspace}
\newcommand{\NinthFreshErrMagnitudeA}{1,088\xspace}

\newcommand{\NinthFreshMoreLowMagnitudeA}{982\xspace}
\newcommand{\NinthFreshMoreHighMagnitudeA}{1,119\xspace}
\newcommand{\NinthFreshErrMagnitudeB}{587\xspace}

\newcommand{\NinthFreshMoreLowMagnitudeB}{517\xspace}
\newcommand{\NinthFreshMoreHighMagnitudeB}{627\xspace}
\newcommand{\NinthFreshErrMagnitudeC}{350\xspace}

\newcommand{\NinthFreshMoreLowMagnitudeC}{287\xspace}
\newcommand{\NinthFreshMoreHighMagnitudeC}{375\xspace}
\newcommand{\NinthFreshErrCosineA}{65\xspace}

\newcommand{\NinthFreshMoreLowCosineA}{17\xspace}
\newcommand{\NinthFreshMoreHighCosineA}{40\xspace}
\newcommand{\NinthFreshErrCosineB}{56\xspace}

\newcommand{\NinthFreshMoreLowCosineB}{23\xspace}
\newcommand{\NinthFreshMoreHighCosineB}{57\xspace}
\newcommand{\NinthFreshErrCosineC}{81\xspace}

\newcommand{\NinthFreshMoreLowCosineC}{41\xspace}
\newcommand{\NinthFreshMoreHighCosineC}{81\xspace}
\newcommand{\NinthContSiteOne}{0.267\xspace}
\newcommand{\NinthContPreOne}{0.172\xspace}
\newcommand{\NinthContPostOne}{0.174\xspace}
\newcommand{\NinthContNormSiteOne}{1.076\xspace}
\newcommand{\NinthContNormPreOne}{1.055\xspace}
\newcommand{\NinthContNormPostOne}{1.048\xspace}
\newcommand{\NinthContSrcOne}{1\xspace}
\newcommand{\NinthContFeatOne}{1502\xspace}
\newcommand{\NinthContTgtOne}{5\xspace}
\newcommand{\NinthContPartOne}{1146\xspace}
\newcommand{\NinthContSiteTwo}{0.327\xspace}
\newcommand{\NinthContPreTwo}{0.224\xspace}
\newcommand{\NinthContPostTwo}{0.238\xspace}
\newcommand{\NinthContNormSiteTwo}{1.164\xspace}
\newcommand{\NinthContNormPreTwo}{1.121\xspace}
\newcommand{\NinthContNormPostTwo}{1.116\xspace}
\newcommand{\NinthContSrcTwo}{3\xspace}
\newcommand{\NinthContFeatTwo}{1443\xspace}
\newcommand{\NinthContTgtTwo}{1\xspace}
\newcommand{\NinthContPartTwo}{1479\xspace}
\newcommand{\NinthContSiteThree}{0.295\xspace}
\newcommand{\NinthContPreThree}{0.201\xspace}
\newcommand{\NinthContPostThree}{0.212\xspace}
\newcommand{\NinthContNormSiteThree}{1.165\xspace}
\newcommand{\NinthContNormPreThree}{1.124\xspace}
\newcommand{\NinthContNormPostThree}{1.118\xspace}
\newcommand{\NinthContSrcThree}{3\xspace}
\newcommand{\NinthContFeatThree}{1443\xspace}
\newcommand{\NinthContTgtThree}{4\xspace}
\newcommand{\NinthContPartThree}{1262\xspace}
\newcommand{\NinthContSiteFour}{0.306\xspace}
\newcommand{\NinthContPreFour}{0.214\xspace}
\newcommand{\NinthContPostFour}{0.217\xspace}
\newcommand{\NinthContNormSiteFour}{0.963\xspace}
\newcommand{\NinthContNormPreFour}{0.908\xspace}
\newcommand{\NinthContNormPostFour}{0.903\xspace}
\newcommand{\NinthContSrcFour}{4\xspace}
\newcommand{\NinthContFeatFour}{27\xspace}
\newcommand{\NinthContTgtFour}{5\xspace}
\newcommand{\NinthContPartFour}{5388\xspace}
\newcommand{\NinthContSiteFive}{0.260\xspace}
\newcommand{\NinthContPreFive}{0.210\xspace}
\newcommand{\NinthContPostFive}{0.215\xspace}
\newcommand{\NinthContNormSiteFive}{1.111\xspace}
\newcommand{\NinthContNormPreFive}{1.052\xspace}
\newcommand{\NinthContNormPostFive}{1.052\xspace}
\newcommand{\NinthContSrcFive}{5\xspace}
\newcommand{\NinthContFeatFive}{5381\xspace}
\newcommand{\NinthContTgtFive}{1\xspace}
\newcommand{\NinthContPartFive}{1721\xspace}
\newcommand{\NinthContSiteSix}{0.225\xspace}
\newcommand{\NinthContPreSix}{0.255\xspace}
\newcommand{\NinthContPostSix}{0.273\xspace}
\newcommand{\NinthContNormSiteSix}{0.911\xspace}
\newcommand{\NinthContNormPreSix}{0.894\xspace}
\newcommand{\NinthContNormPostSix}{0.875\xspace}
\newcommand{\NinthContSrcSix}{3\xspace}
\newcommand{\NinthContFeatSix}{1124\xspace}
\newcommand{\NinthContTgtSix}{2\xspace}
\newcommand{\NinthContPartSix}{1706\xspace}
\newcommand{\NinthContSiteSeven}{0.178\xspace}
\newcommand{\NinthContPreSeven}{0.408\xspace}
\newcommand{\NinthContPostSeven}{0.451\xspace}
\newcommand{\NinthContNormSiteSeven}{1.081\xspace}
\newcommand{\NinthContNormPreSeven}{1.093\xspace}
\newcommand{\NinthContNormPostSeven}{1.075\xspace}
\newcommand{\NinthContSrcSeven}{2\xspace}
\newcommand{\NinthContFeatSeven}{387\xspace}
\newcommand{\NinthContTgtSeven}{4\xspace}
\newcommand{\NinthContPartSeven}{744\xspace}
\newcommand{\NinthCasePre}{0.408\xspace}
\newcommand{\NinthCaseNormLow}{1.07\xspace}
\newcommand{\NinthCaseNormHigh}{1.09\xspace}
\newcommand{\NinthContSiteEight}{0.258\xspace}
\newcommand{\NinthContPreEight}{0.157\xspace}
\newcommand{\NinthContPostEight}{0.169\xspace}
\newcommand{\NinthContNormSiteEight}{0.923\xspace}
\newcommand{\NinthContNormPreEight}{0.953\xspace}
\newcommand{\NinthContNormPostEight}{0.955\xspace}
\newcommand{\NinthContSrcEight}{5\xspace}
\newcommand{\NinthContFeatEight}{442\xspace}
\newcommand{\NinthContTgtEight}{3\xspace}
\newcommand{\NinthContPartEight}{1738\xspace}

\newcommand{\NinthStableSelection}{24\xspace}

\newcommand{\TenthPermutations}{3\xspace}
\newcommand{\TenthSeedSubset}{2026111010\xspace}
\newcommand{\TenthSeedPermuted}{2026111060\xspace}
\newcommand{\TenthPermSourcesA}{128\xspace}
\newcommand{\TenthPermMatchedPairsA}{512\xspace}
\newcommand{\TenthPermMatchedA}{54\xspace}
\newcommand{\TenthPermMatchedLowA}{31\xspace}
\newcommand{\TenthPermMatchedHighA}{81\xspace}
\newcommand{\TenthPermOneA}{0\xspace}

\newcommand{\TenthPermTwoA}{0\xspace}

\newcommand{\TenthPermThreeA}{0\xspace}

\newcommand{\TenthPermAllA}{0\xspace}
\newcommand{\TenthPermAllPairsA}{1,536\xspace}
\newcommand{\TenthPermDiffA}{54.0\xspace}
\newcommand{\TenthPermDiffLowA}{31.0\xspace}
\newcommand{\TenthPermDiffHighA}{81.0\xspace}

\newcommand{\TenthPermMatchedB}{230\xspace}
\newcommand{\TenthPermMatchedLowB}{190\xspace}
\newcommand{\TenthPermMatchedHighB}{273\xspace}
\newcommand{\TenthPermOneB}{0\xspace}

\newcommand{\TenthPermTwoB}{0\xspace}

\newcommand{\TenthPermThreeB}{0\xspace}

\newcommand{\TenthPermAllB}{0\xspace}

\newcommand{\TenthPermDiffB}{230.0\xspace}
\newcommand{\TenthPermDiffLowB}{190.0\xspace}
\newcommand{\TenthPermDiffHighB}{273.0\xspace}

\newcommand{\TenthPermMatchedC}{356\xspace}
\newcommand{\TenthPermMatchedLowC}{321\xspace}
\newcommand{\TenthPermMatchedHighC}{389\xspace}
\newcommand{\TenthPermOneC}{0\xspace}

\newcommand{\TenthPermTwoC}{0\xspace}

\newcommand{\TenthPermThreeC}{0\xspace}

\newcommand{\TenthPermAllC}{0\xspace}

\newcommand{\TenthPermDiffC}{356.0\xspace}
\newcommand{\TenthPermDiffLowC}{321.0\xspace}
\newcommand{\TenthPermDiffHighC}{389.0\xspace}

\newcommand{\TenthPermPairsTotal}{4,608\xspace}

\newcommand{\TenthSecPairsA}{2,560\xspace}

\newcommand{\TenthSecPassA}{298\xspace}
\newcommand{\TenthSecPassLowA}{242\xspace}
\newcommand{\TenthSecPassHighA}{357\xspace}
\newcommand{\TenthSecPctA}{12\xspace}
\newcommand{\TenthFirstPctA}{12\xspace}
\newcommand{\TenthSecErrCosineA}{58\xspace}
\newcommand{\TenthSecErrCodesA}{30\xspace}
\newcommand{\TenthSecErrSiteA}{33\xspace}

\newcommand{\TenthSecCosLessA}{25\xspace}
\newcommand{\TenthSecCosLessLowA}{13\xspace}
\newcommand{\TenthSecCosLessHighA}{38\xspace}

\newcommand{\TenthSecCodesLessA}{-3\xspace}
\newcommand{\TenthSecCodesLessLowA}{-10\xspace}
\newcommand{\TenthSecCodesLessHighA}{3\xspace}

\newcommand{\TenthSecFreshErrCodesA}{25\xspace}
\newcommand{\TenthSecFreshErrSiteA}{28\xspace}

\newcommand{\TenthSecPairsB}{2,560\xspace}

\newcommand{\TenthSecPassB}{994\xspace}
\newcommand{\TenthSecPassLowB}{903\xspace}
\newcommand{\TenthSecPassHighB}{1,086\xspace}

\newcommand{\TenthSecErrCosineB}{63\xspace}
\newcommand{\TenthSecErrCodesB}{51\xspace}
\newcommand{\TenthSecErrSiteB}{32\xspace}

\newcommand{\TenthSecCosLessB}{31\xspace}
\newcommand{\TenthSecCosLessLowB}{14\xspace}
\newcommand{\TenthSecCosLessHighB}{47\xspace}

\newcommand{\TenthSecCodesLessB}{19\xspace}
\newcommand{\TenthSecCodesLessLowB}{9\xspace}
\newcommand{\TenthSecCodesLessHighB}{30\xspace}

\newcommand{\TenthSecFreshErrCodesB}{35\xspace}
\newcommand{\TenthSecFreshErrSiteB}{20\xspace}

\newcommand{\TenthSecPairsC}{2,560\xspace}

\newcommand{\TenthSecPassC}{1,812\xspace}
\newcommand{\TenthSecPassLowC}{1,734\xspace}
\newcommand{\TenthSecPassHighC}{1,893\xspace}
\newcommand{\TenthSecPctC}{71\xspace}
\newcommand{\TenthFirstPctC}{70\xspace}
\newcommand{\TenthSecErrCosineC}{63\xspace}
\newcommand{\TenthSecErrCodesC}{53\xspace}
\newcommand{\TenthSecErrSiteC}{38\xspace}

\newcommand{\TenthSecCosLessC}{25\xspace}
\newcommand{\TenthSecCosLessLowC}{9\xspace}
\newcommand{\TenthSecCosLessHighC}{42\xspace}

\newcommand{\TenthSecCodesLessC}{15\xspace}
\newcommand{\TenthSecCodesLessLowC}{5\xspace}
\newcommand{\TenthSecCodesLessHighC}{26\xspace}

\newcommand{\TenthSecFreshErrCodesC}{30\xspace}
\newcommand{\TenthSecFreshErrSiteC}{22\xspace}
\newcommand{\TenthSeedSecond}{20261110\xspace}
\newcommand{\TenthSeedSecondInterval}{2026111021\xspace}
\newcommand{\TenthSecErrSiteMin}{32\xspace}
\newcommand{\TenthSecErrSiteMax}{38\xspace}
\newcommand{\TenthSecErrCosineMin}{58\xspace}
\newcommand{\TenthSecErrCosineMax}{63\xspace}
\newcommand{\TenthSeconds}{7,691\xspace}
\newcommand{\TenthCalls}{30\xspace}
\newcommand{\TenthImageRQ}{578545\xspace}
\newcommand{\TenthImageRV}{41633\xspace}
\newcommand{\TenthImageSA}{60363\xspace}
\newcommand{\TenthImageSF}{345361\xspace}
\newcommand{\TenthImageSK}{351589\xspace}
\newcommand{\TenthImageSP}{389451\xspace}
\newcommand{\TenthImageSU}{109976\xspace}
\newcommand{\TenthImageSZ}{134722\xspace}
\newcommand{\TenthImageTE}{408830\xspace}
\newcommand{\TenthImageTJ}{402774\xspace}
\newcommand{\TenthImageTO}{531256\xspace}
\newcommand{\TenthImageTT}{71239\xspace}
\newcommand{\TenthImageTY}{161054\xspace}
\newcommand{\TenthImageUD}{369294\xspace}
\newcommand{\TenthImageUI}{565564\xspace}
\newcommand{\TenthImageUN}{383525\xspace}
\newcommand{\TenthImageUS}{534889\xspace}
\newcommand{\TenthImageUX}{314194\xspace}
\newcommand{\TenthImageVC}{144283\xspace}
\newcommand{\TenthImageVH}{421655\xspace}
\newcommand{\TenthImageVM}{67569\xspace}
\newcommand{\TenthImageVR}{251751\xspace}
\newcommand{\TenthImageVW}{133384\xspace}
\newcommand{\TenthImageWB}{242868\xspace}
\newcommand{\TenthImageWG}{346121\xspace}
\newcommand{\TenthImageWL}{369326\xspace}
\newcommand{\TenthImageWQ}{233872\xspace}
\newcommand{\TenthImageWV}{282631\xspace}
\newcommand{\TenthImageXA}{189805\xspace}
\newcommand{\TenthImageXF}{542934\xspace}
\newcommand{\TenthImageXW}{89556\xspace}
\newcommand{\TenthImageYD}{345361\xspace}
\newcommand{\TenthImageYK}{402774\xspace}
\newcommand{\TenthImageYR}{60363\xspace}
\newcommand{\TenthImageYY}{442161\xspace}
\newcommand{\TenthImageZF}{428867\xspace}
\newcommand{\TenthImageZM}{201646\xspace}
\newcommand{\TenthImageZT}{578545\xspace}
\newcommand{\TenthImageAAA}{60363\xspace}
\newcommand{\TenthImageAAH}{578545\xspace}
\newcommand{\TenthImageAAO}{531256\xspace}
\newcommand{\TenthImageAAV}{420658\xspace}
\newcommand{\TenthImageABC}{71972\xspace}
\newcommand{\TenthImageABJ}{565564\xspace}
\newcommand{\TenthImageABQ}{421655\xspace}
\newcommand{\TenthImageABX}{161054\xspace}
\newcommand{\TenthImageACE}{383525\xspace}
\newcommand{\TenthImageACL}{126265\xspace}
\newcommand{\TenthImageACS}{60434\xspace}
\newcommand{\TenthImageACZ}{531256\xspace}
\newcommand{\TenthImageADG}{68518\xspace}
\newcommand{\TenthImageADN}{242868\xspace}
\newcommand{\TenthImageADU}{133384\xspace}
\newcommand{\TenthImageAEB}{251751\xspace}
\newcommand{\TenthImageAEI}{427060\xspace}
\newcommand{\TenthImageAEP}{369326\xspace}
\newcommand{\TenthImageAEW}{189805\xspace}
\newcommand{\TenthImageAFD}{369326\xspace}
\newcommand{\TenthImageAFK}{457271\xspace}
\newcommand{\TenthImageAFR}{242868\xspace}

\newcommand{\EleventhTimedFeatures}{6,144\xspace}
\newcommand{\EleventhContainerProcessors}{20\xspace}
\newcommand{\EleventhContainerThreads}{4\xspace}

\newcommand{\EleventhZeroCodeSets}{16\xspace}

\newcommand{\EleventhOutPassTokensA}{316\xspace}
\newcommand{\EleventhOutPassImageA}{326\xspace}
\newcommand{\EleventhOutPctImageA}{13\xspace}
\newcommand{\EleventhOutLostA}{10\xspace}
\newcommand{\EleventhOutGainedA}{20\xspace}
\newcommand{\EleventhErrTokensCosineA}{65\xspace}
\newcommand{\EleventhErrImageCosineA}{73\xspace}
\newcommand{\EleventhRepImageCosineA}{62\xspace}
\newcommand{\EleventhErrTokensLayerFittedA}{42\xspace}
\newcommand{\EleventhErrImageLayerFittedA}{44\xspace}
\newcommand{\EleventhRepImageLayerFittedA}{33\xspace}
\newcommand{\EleventhErrTokensLayerAtBoundA}{42\xspace}
\newcommand{\EleventhErrImageLayerAtBoundA}{44\xspace}
\newcommand{\EleventhRepImageLayerAtBoundA}{33\xspace}
\newcommand{\EleventhErrTokensCodesA}{33\xspace}
\newcommand{\EleventhErrImageCodesA}{41\xspace}
\newcommand{\EleventhRepImageCodesA}{29\xspace}
\newcommand{\EleventhErrTokensOverlapA}{41\xspace}
\newcommand{\EleventhErrImageOverlapA}{43\xspace}
\newcommand{\EleventhRepImageOverlapA}{32\xspace}
\newcommand{\EleventhErrTokensMagnitudeA}{1,102\xspace}
\newcommand{\EleventhErrImageMagnitudeA}{1,108\xspace}
\newcommand{\EleventhRepImageMagnitudeA}{1,096\xspace}
\newcommand{\EleventhOutDiffA}{29\xspace}
\newcommand{\EleventhOutDiffLowA}{17\xspace}
\newcommand{\EleventhOutDiffHighA}{43\xspace}

\newcommand{\EleventhOutPassTokensB}{1,045\xspace}
\newcommand{\EleventhOutPassImageB}{1,094\xspace}

\newcommand{\EleventhOutLostB}{3\xspace}
\newcommand{\EleventhOutGainedB}{52\xspace}
\newcommand{\EleventhErrTokensCosineB}{56\xspace}
\newcommand{\EleventhErrImageCosineB}{105\xspace}
\newcommand{\EleventhRepImageCosineB}{79\xspace}
\newcommand{\EleventhErrTokensLayerFittedB}{27\xspace}
\newcommand{\EleventhErrImageLayerFittedB}{62\xspace}
\newcommand{\EleventhRepImageLayerFittedB}{39\xspace}
\newcommand{\EleventhErrTokensLayerAtBoundB}{27\xspace}
\newcommand{\EleventhErrImageLayerAtBoundB}{62\xspace}
\newcommand{\EleventhRepImageLayerAtBoundB}{39\xspace}
\newcommand{\EleventhErrTokensCodesB}{54\xspace}
\newcommand{\EleventhErrImageCodesB}{75\xspace}
\newcommand{\EleventhRepImageCodesB}{52\xspace}
\newcommand{\EleventhErrTokensOverlapB}{57\xspace}
\newcommand{\EleventhErrImageOverlapB}{90\xspace}
\newcommand{\EleventhRepImageOverlapB}{64\xspace}
\newcommand{\EleventhErrTokensMagnitudeB}{594\xspace}
\newcommand{\EleventhErrImageMagnitudeB}{599\xspace}
\newcommand{\EleventhRepImageMagnitudeB}{576\xspace}
\newcommand{\EleventhOutDiffB}{43\xspace}
\newcommand{\EleventhOutDiffLowB}{25\xspace}
\newcommand{\EleventhOutDiffHighB}{65\xspace}

\newcommand{\EleventhOutPassTokensC}{1,783\xspace}
\newcommand{\EleventhOutPassImageC}{1,770\xspace}
\newcommand{\EleventhOutPctImageC}{69\xspace}
\newcommand{\EleventhOutLostC}{47\xspace}
\newcommand{\EleventhOutGainedC}{34\xspace}
\newcommand{\EleventhErrTokensCosineC}{81\xspace}
\newcommand{\EleventhErrImageCosineC}{126\xspace}
\newcommand{\EleventhRepImageCosineC}{75\xspace}
\newcommand{\EleventhErrTokensLayerFittedC}{28\xspace}
\newcommand{\EleventhErrImageLayerFittedC}{93\xspace}
\newcommand{\EleventhRepImageLayerFittedC}{47\xspace}
\newcommand{\EleventhErrTokensLayerAtBoundC}{27\xspace}
\newcommand{\EleventhErrImageLayerAtBoundC}{92\xspace}
\newcommand{\EleventhRepImageLayerAtBoundC}{46\xspace}
\newcommand{\EleventhErrTokensCodesC}{37\xspace}
\newcommand{\EleventhErrImageCodesC}{98\xspace}
\newcommand{\EleventhRepImageCodesC}{51\xspace}
\newcommand{\EleventhErrTokensOverlapC}{45\xspace}
\newcommand{\EleventhErrImageOverlapC}{110\xspace}
\newcommand{\EleventhRepImageOverlapC}{63\xspace}
\newcommand{\EleventhErrTokensMagnitudeC}{328\xspace}
\newcommand{\EleventhErrImageMagnitudeC}{349\xspace}
\newcommand{\EleventhRepImageMagnitudeC}{297\xspace}
\newcommand{\EleventhOutDiffC}{33\xspace}
\newcommand{\EleventhOutDiffLowC}{14\xspace}
\newcommand{\EleventhOutDiffHighC}{53\xspace}
\newcommand{\EleventhCaseError}{0.339\xspace}
\newcommand{\EleventhCaseTokens}{0.417\xspace}
\newcommand{\EleventhCaseRepeats}{repeats\xspace}
\newcommand{\EleventhEightRepeat}{6\xspace}
\newcommand{\EleventhSeedOutputs}{2026111101\xspace}
\newcommand{\EleventhOutSeconds}{3,796\xspace}
\newcommand{\EleventhOutCalls}{15\xspace}
\newcommand{\EleventhOutDevices}{NVIDIA A100-SXM4-40GB and an NVIDIA A100-SXM4-80GB\xspace}

\newcommand{\AbstractCosineHigh}{3.4\xspace}

\newcommand{\AbstractLayerHigh}{1.3\xspace}

\newcommand{\AbstractControlZero}{0\xspace}

\newcommand{\VarianceSeedOne}{1\xspace}
\newcommand{\VarianceSeedTwo}{2\xspace}
\newcommand{\VarianceSeedThree}{3\xspace}
\newcommand{\VarianceSeedFour}{4\xspace}
\newcommand{\VarianceSeedFive}{5\xspace}

\date{}
\newcommand{\authoremail}{kl3685@columbia.edu}
\title{Carrying Feature Ablations Across Sparse Autoencoder Seeds}
\author{%
  Karl Li \\
  Columbia University \\
  {\normalfont\small\texttt{\authoremail}}
}
\hypersetup{pdfauthor={Karl Li},pdftitle={Carrying Feature Ablations Across Sparse Autoencoder Seeds}}
\begin{document}
\maketitle
\begin{abstract}
A vision encoder turns an image into one vector per patch, refined layer by layer. A sparse autoencoder (SAE) writes each vector of one layer as a sum of learned directions, most with a coefficient of zero. Researchers call each direction a feature and edit the encoder by removing one feature. An SAE trained from another random seed learns different features. Earlier work matches the two SAEs' features one to one by how closely their directions agree. Matched features can point alike yet respond to different image parts, so removing them may change the encoder differently. We remove a feature and, separately, its match, then compare the changes in the encoder's output on the same held out images. When the second change nearly reproduces the first, we say the removal carries across seeds. A setting fixes a \Dino or SigLIP layer and its SAE feature count. At our tolerance, the removal carries for \SharePctA{}\% to \SharePctC{}\% of matched pairs across three settings. The share is \AbstractControlZero{}\% in two controls: an SAE whose directions stay random and, post hoc, matches shuffled among trained features. Post hoc, predicting which removals carry is more accurate with layer contributions than with directions. Some incorrect predictions remain. One persists on independent images. So pairs can be screened at the layer. Measuring the output detects errors this screen misses.
\end{abstract}

\section{Introduction}
\label{sec:intro}

A vision encoder such as \Dino cuts an image into a grid of patches. At each layer it holds one vector, called a token, for each patch and one more for the whole image. A sparse autoencoder (SAE) trained on one layer writes each token as a sum of learned directions, each scaled by a nonnegative coefficient that is zero for most directions on any one token~\cite{cunningham2024sparse,stevens2025testable}. Researchers treat each learned direction, together with its coefficient on each token, as a feature of the encoder. Subtracting one feature's scaled direction from every token of the layer and running the later layers on what remains is called an ablation of that feature. Researchers name features from the images that activate them, ablate features to edit what the encoder computes~\cite{papadimitriou2025linear,klotz2026annotations} and align features across models~\cite{thasarathan2025universal}. Carrying an ablation across seeds assumes that a feature belongs to the encoder and not to one training run of the SAE, so that an SAE trained from another random seed would hold a feature whose ablation does the same.

\begin{figure}[!htbp]
  \centering
  \includegraphics{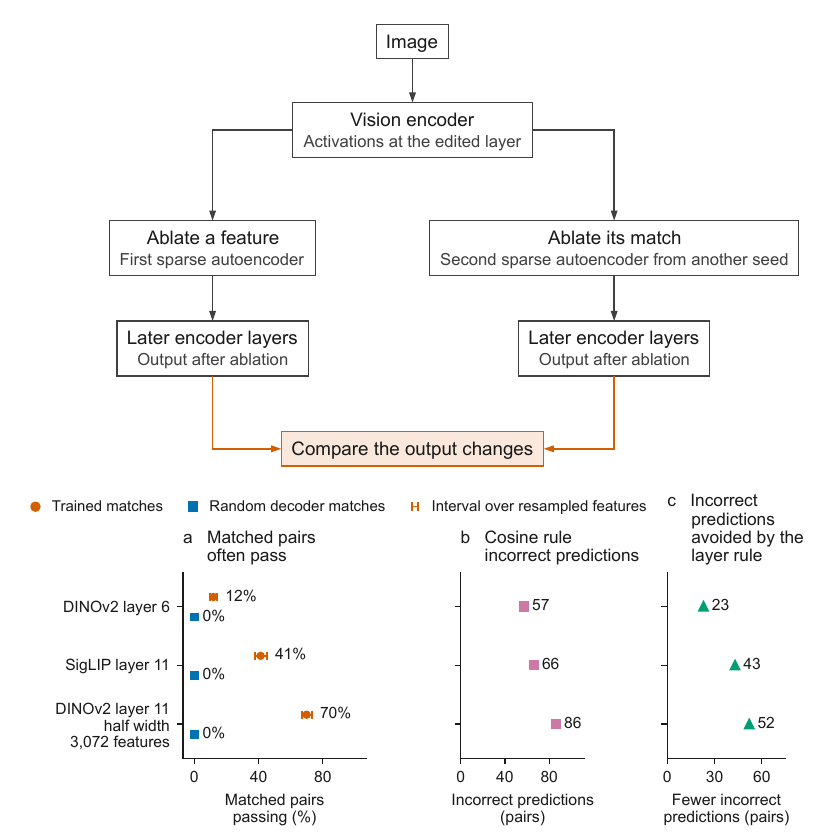}
  \caption{\figOverviewCaption}
  \label{fig:test}
  \tablenote{\noteFigOverview}
\end{figure}

SAEs trained on the same tokens from different seeds learn different features, and many features of one SAE have no close counterpart in the other~\cite{paulo2025different}. To compare two such SAEs, earlier work gives each feature of one exactly one partner in the other, its match, so that over all features the matched directions point as nearly the same way as possible~\cite{paulo2025different,papadimitriou2025linear}. The cosine of the angle between a feature's direction and its match's direction, the decoder cosine, is one when the two point the same way. Earlier work counts the matches with a high decoder cosine as features that recur across seeds. Two features can add the same direction and still be active on different tokens, for example on different parts of an image, so their ablations can change the encoder in different ways. A researcher who ablates a feature to edit an encoder needs to know whether ablating its match changes the encoder's output in the same way. The decoder cosine does not tell.

We judge a match by what its two features do, as \cref{fig:test} shows. We ablate a feature at one layer of the encoder, the edited layer, and separately ablate its match in an SAE trained on the same layer from another seed. We then compare the two changes in the encoder output, which is every token that the encoder returns after its last layer. We call this comparison the \emph{output test}. A pair passes the output test when the second change nearly reproduces the first. \Cref{sec:test} states how nearly. When a pair passes, we say that the ablation carries across seeds, as in our title.

We run the output test on \Dino~\cite{oquab2024dinov2} and on SigLIP~\cite{zhai2023siglip}, an image and text encoder trained with a sigmoid loss, in three settings of encoder, edited layer and SAE width, with \Seeds SAEs in each. From each SAE we sample \DrawnPerSource features and pair each with its match in every other SAE of the setting. We score the output test on held out images and again on fresh images that trained no SAE and fitted no prediction threshold. Two controls replace the second SAE by one whose decoder directions stay random or by one trained on Gaussian tokens. A third control, which we added post hoc, permutes the matches among trained features.

The output test runs the later layers for every ablation, which makes it the costly step. So we also ask whether two cheaper rules predict its outcome. Each feature also has an encoder row, the direction that sets how strongly the feature responds to a token. The cosine rule predicts a pass where both the decoder cosine and the cosine of the two encoder rows are high. The layer rule, which we added post hoc, predicts a pass where the two features add nearly the same vectors to the tokens of the edited layer. It therefore runs no later layer.

We make the following contributions.

\begin{enumerate}[leftmargin=*,nosep]
  \item \textbf{An output test for matched features.} \Cref{sec:test} defines the output test for two SAEs of equal width at one layer of one encoder, with two sanity checks and three controls.
  \item \textbf{Matched features often carry an ablation.} Up to \SharePctC{}\% of matched pairs pass. The share depends on the setting, as \Cref{sec:carrying} reports.
  \item \textbf{Passes are rare in the controls.} No pair into an SAE with random decoder directions passes. When we permute the matches post hoc, no pair passes either, as \Cref{sec:carrying} reports.
  \item \textbf{Two cosines predict most outcomes.} The cosine rule predicts incorrectly for at most \AbstractCosineHigh{}\% of a setting's matched pairs on the first scoring images, as \Cref{sec:cosine} reports.
  \item \textbf{A comparison at the edited layer predicts better.} The layer rule predicts incorrectly for at most \AbstractLayerHigh{}\% of a setting's matched pairs on the first scoring images, with fewer errors than the cosine rule there and on a second feature sample from the same SAEs, as \Cref{sec:cosine} reports.
  \item \textbf{One prediction error persists.} One incorrect prediction of the layer rule persists on a third, independent image set, as \Cref{sec:cosine} reports.
\end{enumerate}

\section{Related work}
\label{sec:related}

Earlier work compares SAEs by their directions or their codes, edits one feature and measures that edit, or predicts the effect of an edit before running it. We compare the effects of two edits, one in each of two SAEs. Each paragraph below closes on how that differs from one group of earlier work.

\paragraph{Similarity across training runs.}
Li \etal~\cite{li2016convergent} matched single units of two networks by correlation and found some units with a single partner and others shared only within subspaces. Kornblith \etal~\cite{kornblith2019similarity} compared whole layers by centred kernel alignment. For SAEs, Paulo and Belrose~\cite{paulo2025different} matched features across seeds with the Hungarian method~\cite{kuhn1955hungarian} on decoder cosine. Gerasimov \etal~\cite{gerasimov2026unstable} found that subspaces reproduce across seeds where single features do not. Brzozowski and Chung~\cite{brzozowski2026archetypes} traced the stability of archetypal SAEs to their initialization and to the metric that scores it. In contemporaneous work, Choi \etal~\cite{choi2026consistent} raise the consistency of concepts across seeds by optimizing each input's code and regularizing training. These studies compare SAEs by their codes or their directions. We compare two independently trained SAEs by what ablating their features does to the encoder output.

\paragraph{Matching and editing features.}
Cao \etal~\cite{cao2026semantic} match features across the layers of a language model by optimal transport and check each match by replacing one feature with the other. Thasarathan \etal~\cite{thasarathan2025universal} train one SAE jointly over several vision models, so a shared code aligns their concepts by construction. Papadimitriou \etal~\cite{papadimitriou2025linear} found the most active features of five SAEs on SigLIP embeddings stable under cosine matching. Stevens \etal~\cite{stevens2025testable} edit vision features and measure the edits' effects to test the features' explanations. We instead train every SAE on its own, edit by the ablation of one feature, and compare the effect of that ablation across seeds.

\paragraph{Predicting an intervention's effect.}
In contemporaneous work, Zhang \etal~\cite{zhang2026strength} show that how an intervention's strength is set changes how interventions compare across SAE codes. Duan~\cite{duan2026preintervention} predicts the side effects of steering a language model, that is, of adding a feature's direction to change its output, from statistics computed before the intervention. We predict whether a pair passes the output test from the two features' contributions at the edited layer of a vision encoder.

\paragraph{Naming features.}
Network Dissection~\cite{bau2017dissection} names a unit by the overlap of its activation with segmentation masks. Klotz \etal~\cite{klotz2026annotations} evaluate SAE features against concept annotations and controlled changes to images. In contemporaneous work, Mencattini \etal~\cite{mencattini2026fragments} show that the independence prior of SAEs splits one visual concept across several features. Shen \etal~\cite{shen2026graphs} learn graphs among features of mixed topology. Ro \etal~\cite{ro2026scope} separate the information scope of a contrastive image and text encoder's features from their semantics. Kravets \etal~\cite{kravets2026transfer} transfer interpretations from a language SAE to vision through a learned projector. These works name the features of one SAE. We asked whether a feature that passes keeps its name in the other seed's SAE. Object annotations name too few features to answer.

\section{Method}
\label{sec:test}

We define the output test in three steps: we ablate a feature in one SAE, we ablate its match in an SAE trained from another seed, and we compare the two changes in the encoder output. We then define two rules that predict the outcome of the output test without running the later layers, and the controls whose pass rates we compare.

\paragraph{Ablating a feature.}
Intuitively, an SAE splits each token of the edited layer into the parts that its features add. An ablation removes one part and lets the encoder run on. Let the vision encoder be $f = g \circ e$, where $e$ maps an image to its activations $h \in \mathbb{R}^{T \times n}$ at the edited layer, $T$ tokens of width $n$, and $g$ is the rest of the encoder, its later layers through the final normalization. An SAE takes each token $\eta$ of $h$ in scaled coordinates $x = (\eta - \mu)/\alpha$, where $\mu$ is the mean token over the training images and $\alpha$ is the root mean square of a token's coordinates about that mean. The SAE trained from seed $s$ reconstructs $x$ as
\begin{equation}
  \begin{split}
    \hat{x} &= b^{s} + \sum_{k} z^{s}_{k}(x)\, d^{s}_{k}, \\
    z^{s}_{k}(x) &= \max\bigl(0,\; w^{s}_{k} \cdot (x - b^{s}) + \beta^{s}_{k}\bigr),
  \end{split}
  \label{eq:dictionary}
\end{equation}
with unit decoder directions $d^{s}_{k}$, encoder rows $w^{s}_{k}$ and biases $b^{s}$ and $\beta^{s}_{k}$.

Feature $k$ has the code $z^{s}_{k}(x)$, the nonnegative coefficient of its decoder direction on a token. The feature is active on a token where its code is positive. Its contribution to a token of $h$ is $\alpha\, z^{s}_{k}(x)\, d^{s}_{k}$. Ablating feature $i$ subtracts exactly its contribution from every token of $h$, and the later layers then run on what remains. We use ablation because it has no strength to tune. Zhang \etal~\cite{zhang2026strength} show that how a strength is set changes how interventions compare. The ablation's effect on an image is the change in the encoder output on every token, $\delta^{s}_{i} = g(h - a^{s}_{i}) - g(h)$, where $a^{s}_{i}$ stacks the feature's contributions to the tokens of the image.

\paragraph{Comparing two ablations.}
We compare the ablation of a feature with the ablation of its match in a second SAE, trained on the same layer from seed $t$. The matching $\pi$ pairs each feature of the first SAE with exactly one feature of the second SAE. It is the one to one pairing that maximizes the summed cosine of matched decoder directions, which the Hungarian method~\cite{kuhn1955hungarian} finds, as in earlier work on seeds~\cite{paulo2025different}. The matched ablation subtracts the contribution of the matched feature $\pi(i)$, with the second SAE computing its own code. Its effect, the matched effect, is $\delta^{t}_{\pi(i)} = g(h - a^{t}_{\pi(i)}) - g(h)$. We call $\delta^{s}_{i}$ the first effect.

Intuitively, the matched ablation reproduces the first ablation when the matched effect lies close to the first effect, judged against the size of the first effect. We judge the two effects at the encoder output, since the output is what a user of the encoder receives. Measured as a distance, a matched effect that adds a change of its own counts against the pair as much as one that misses part of the first change.

The output error of the pair states this comparison exactly. It is the squared distance between the two effects divided by the squared size of the first effect, each summed over a set of images,
\begin{equation}
  \epsilon = \frac{\sum \lVert \delta^{t}_{\pi(i)} - \delta^{s}_{i} \rVert^{2}}{\sum \lVert \delta^{s}_{i} \rVert^{2}}.
  \label{eq:error}
\end{equation}
A pair passes the output test where $\epsilon$ is at most \Certify and fails otherwise. Equivalently, a pair passes where the norm of the difference between the two effects, taken over every token of every image in the set, is at most half the norm of the first effect. For example, a matched effect that points the same way as the first effect at half its size has an output error of exactly \Certify, so it just passes. We chose this bound. \Cref{sec:carrying} gives the shares at two other bounds.

\paragraph{Images.}
In the main comparisons, we score predictions on images that neither trained an SAE nor entered those predictions. Each setting uses the \ImagesAll images of the Common Objects in Context (COCO) 2017 validation set~\cite{lin2014coco} in one fixed random order. The first \ImagesTrain train the SAEs. The next \ImagesSelect are the selection images, on which we compute the rules' predictions and fit the replacements described below. The last \ImagesScore are the scoring images, on which we first measure the output errors. They split into two halves of \ImagesHalf. The fresh images, \FreshImages images from the COCO 2017 training split that no SAE, rule or fit has seen, score every setting again.

A feature enters the output test where it is active on at least \MinImages selection and \MinImages scoring images. It also enters only where its ablation's squared effect is at least \EffectFloor of the unedited output's squared size, which keeps the denominator of \cref{eq:error} positive.

\paragraph{Replacements built from several features.}
One feature of the first SAE may be split over several features of the second. A match of single features cannot follow such a split. So beside the match we score two replacements built from several features of the second SAE. Greedy selection adds up to \Budget features one at a time, each unscaled and each the feature whose contribution most lowers the squared distance to the first feature's contribution on the selection images. Ridge regression fits a weighted sum of the contributions of all features of the second SAE to the first feature's contribution on the same images. We ablate each replacement as we ablate the match and score it by \cref{eq:error}.

\paragraph{What we fixed in advance.}
Before measuring the three settings, we fixed the bound of \Certify, the first two controls and the cosine rule, which the paragraphs below define, and three comparisons. The first comparison asks that matched pairs pass more often than control pairs. The second asks that the output test agree with itself across the two halves of the scoring images more often than the cosine rule agrees with it. The third asks that features that pass keep their names. Intuitively, each comparison asks one share to beat a baseline by a clear margin that shrinks as the baseline nears one. Each comparison asks one share to exceed a baseline share $q$ by more than a required difference $m(q) = \min\{\MarginOne,\ (1-q)/2\}$. \Cref{sec:supp-halves} states each in full.

A development setting, \Dino layer~\SiteDev with \WidthDev features, shaped the output test and fitted the cosine rule beforehand. No comparison uses it. A dagger\later{} marks every analysis or measurement added after the comparisons were scored, which we call post hoc.

\paragraph{Controls and sanity checks.}
In each setting we train \Seeds SAEs from \Seeds seeds and sample \DrawnPerSource features from each before any output error is computed, \DrawnAll in all. We pair every sampled feature with its match in each of the other four SAEs, \UnitsTrained matched pairs a setting.

The controls show how often a pair passes when its second feature is not a learned counterpart of the first. Random decoder SAEs form the first control, \Seeds a setting. They use the same recipe with their decoder directions frozen at their random start, so only their encoder rows and biases learn. We match every sampled feature into each of them, \UnitsFrozen pairs. Surrogate SAEs form the second, also \Seeds, and train on Gaussian tokens with the covariance of the layer's training tokens. They give \UnitsSurrogate pairs where we trained them. The third control\later{} keeps the trained SAEs and permutes the matches among \TenthPermSourcesA sampled features a setting, so each feature takes the match of another sampled feature of its SAE, under \TenthPermutations permutations scored on the fresh images.

Two sanity checks alter copies of each SAE. We set their pass criteria before measurement. A rescaled copy permutes the features and divides each decoder direction by a positive factor while multiplying its encoder row and bias by that factor. That leaves every contribution unchanged, so the matching must recover the permutation and every pair must pass. A shuffled copy moves the encoder rows with their biases among the features and keeps every decoder direction. The matching then stays the same while the codes may change. So its pairs should fail. \Cref{sec:supp-recipe} gives both checks in full.

\paragraph{Predictors.}
The output test runs the later layers for every ablation, so we define two rules that predict its outcome from what can be computed at or before the edited layer. The cosine rule observes only the directions of the two features. The shuffled copy of the sanity checks shows how directions alone can mislead, since each of its pairs keeps a decoder cosine of one while the codes change. \Cref{sec:results} reports that none of those pairs passes. The layer rule observes what the two features add to the tokens of the edited layer on the selection images. A rule predicts incorrectly on a pair where its prediction differs from the outcome of the output test.

The cosine rule predicts a pass where the decoder cosine of a pair, the cosine of its two decoder directions, is at least \RuleDec and its encoder cosine, the cosine of its two encoder rows, is at least \RuleEnc. We fitted the two thresholds on the \DevUnits pairs of the development setting, by their agreement with the output test on its scoring images, and then froze them. Separately, a match is strong where both its cosines exceed \Weak and weak otherwise, a grouping that \cref{sec:carrying} uses.

Intuitively, two features add the same vectors at the edited layer when they point the same way, are active on the same tokens and are active as strongly. The layer error\later{} of a pair measures this agreement: it is \cref{eq:error} taken at the edited layer, with the two contributions in place of the two effects, summed over the selection images. Since every decoder direction has unit length, it equals
\begin{equation}
  1 + r^{2} - 2\,r\,c\,\rho,
  \label{eq:layer}
\end{equation}
where $c$ is the decoder cosine, $\rho$ the code cosine, which is the cosine of the two features' codes across the image tokens, and $r$ the code norm ratio, which is the norm of the matched feature's codes over the norm of the first feature's. So the layer error is small only where $c$, $\rho$ and $r$ are all near one. The decoder cosine measures only the first of the three. The layer rule predicts a pass where the layer error is at most \Certify. It needs one pass to the edited layer and none through the later layers.

\section{Experimental setup}
\label{sec:setup}

The vision encoders stay fixed throughout. Only the SAEs are trained, \Seeds times in each setting, each time from its own seed.

\paragraph{Settings.}
The development setting fitted the cosine rule. Three further settings score the output test, each changing one part of the development setting. We name each setting by its encoder and its edited layer, where the activations of layer $k$ are the output of block $k$ of the encoder's twelve. \cfgA takes \Dino base~\cite{oquab2024dinov2}, pretrained on \DinoCorpus, with SAEs of \WidthA features. It moves the edited layer of the development setting from block \SiteDev to block \SiteA. \cfgB uses the vision tower of SigLIP base~\cite{zhai2023siglip}, pretrained on the English image and text pairs of WebLI with a sigmoid loss, with SAEs of \WidthB features. It changes the encoder, and with it the training objective, the patches and the image processing. \cfgC keeps both \Dino base and the edited layer of the development setting and halves the SAE width to \WidthC features. Every edit passes through at least one later layer and the final normalization.

\paragraph{Images and training.}
The first \ImagesTrain images of each setting supply the tokens that train its SAEs. The selection and scoring images supply the first comparisons. \Dino resizes the shorter side of an image to \DinoShort pixels and crops its centre to \DinoCrop by \DinoCrop, which yields \TokensDino tokens with its class token. SigLIP resizes the image to \TowerSize by \TowerSize and yields \TokensSiglip tokens from patches of \TowerPatch pixels. Every SAE trains from its own seed for \Steps steps of Adam on batches of \Batch tokens. It minimizes the squared reconstruction error of each scaled token, averaged over its coordinates, plus \Coefficient times the summed code. \Cref{sec:supp-recipe} gives the recipe in full.

\paragraph{The SAEs.}
Every SAE, trained or control, has one construction and one training recipe. Each control changes one part of it. The trained SAEs reconstruct each layer better than the random decoder SAEs with fewer active features, as \cref{tab:dictionaries} shows. Its first column gives the share of the layer's variance that a reconstruction leaves unexplained, and its second the mean number of features active on a token. The trained SAEs of \cfgA are the densest, with \LzeroA features active on a token against \LzeroC at \cfgC. The random decoder SAEs have more active features and leave more of the variance unexplained in every setting. So that control measures how often an SAE with learned encoder rows and random directions passes a pair, at a density and a reconstruction of its own. We trained no surrogate SAEs at \cfgA, where an edit passes through the most later layers.

\begin{table}[!htbp]
  \caption{\capDictionaries}
  \label{tab:dictionaries}
  \centering
  \fontsize{9}{10.5}\selectfont
  \setlength{\tabcolsep}{1pt}
  \begin{tabular}{@{}lcc@{}}
    \toprule
    SAE class & \shortstack{Unexplained\\variance $\downarrow$} & \shortstack{Active\\features $\downarrow$} \\
    \midrule
    \multicolumn{3}{@{}l}{\cfgA, \WidthA features} \\
    \rowcolor{methodtint} Trained & \estimate{\MarkDictFvuASkFvuTrMeanA}{\SkFvuTrMinA to \SkFvuTrMaxA} & \estimate{\MarkDictLzeroASkLzeroTrMeanA}{\SkLzeroTrMinA to \SkLzeroTrMaxA} \\
    Random decoder & \estimate{\MarkDictFvuASkFvuFrMeanA}{\SkFvuFrMinA to \SkFvuFrMaxA} & \estimate{\MarkDictLzeroASkLzeroFrMeanA}{\SkLzeroFrMinA to \SkLzeroFrMaxA} \\
    Surrogate & none & none \\
    \midrule
    \multicolumn{3}{@{}l}{\cfgB, \WidthB features} \\
    \rowcolor{methodtint} Trained & \estimate{\MarkDictFvuBSkFvuTrMeanB}{\SkFvuTrMinB to \SkFvuTrMaxB} & \estimate{\MarkDictLzeroBSkLzeroTrMeanB}{\SkLzeroTrMinB to \SkLzeroTrMaxB} \\
    Random decoder & \estimate{\MarkDictFvuBSkFvuFrMeanB}{\SkFvuFrMinB to \SkFvuFrMaxB} & \estimate{\MarkDictLzeroBSkLzeroFrMeanB}{\SkLzeroFrMinB to \SkLzeroFrMaxB} \\
    Surrogate & \estimate{\MarkDictFvuBSkFvuSuMeanB}{\SkFvuSuMinB to \SkFvuSuMaxB} & \estimate{\MarkDictLzeroBSkLzeroSuMeanB}{\SkLzeroSuMinB to \SkLzeroSuMaxB} \\
    \midrule
    \multicolumn{3}{@{}l}{\cfgC, \WidthC features} \\
    \rowcolor{methodtint} Trained & \estimate{\MarkDictFvuCSkFvuTrMeanC}{\SkFvuTrMinC to \SkFvuTrMaxC} & \estimate{\MarkDictLzeroCSkLzeroTrMeanC}{\SkLzeroTrMinC to \SkLzeroTrMaxC} \\
    Random decoder & \estimate{\MarkDictFvuCSkFvuFrMeanC}{\SkFvuFrMinC to \SkFvuFrMaxC} & \estimate{\MarkDictLzeroCSkLzeroFrMeanC}{\SkLzeroFrMinC to \SkLzeroFrMaxC} \\
    Surrogate & \estimate{\MarkDictFvuCSkFvuSuMeanC}{\SkFvuSuMinC to \SkFvuSuMaxC} & \estimate{\MarkDictLzeroCSkLzeroSuMeanC}{\SkLzeroSuMinC to \SkLzeroSuMaxC} \\
    \bottomrule
  \end{tabular}
  \tablenote{\noteDictionaries}
\end{table}

\begin{table}[!htbp]
  \caption{\capCarrying}
  \label{tab:carrying}
  \centering
  \fontsize{9}{10.5}\selectfont
\renewcommand{\estimate}[2]{\ensuremath{\textnormal{#1}_{\publicspread#2\relax}}}
  \setlength{\tabcolsep}{1pt}
  \begin{tabular}{@{}lcccc@{}}
    \toprule
    & Development $\uparrow$ & \Dino $\uparrow$ & SigLIP $\uparrow$ & \Dino $\uparrow$ \\
    & \Dino layer \SiteDev & layer \SiteA & layer \SiteB & layer \SiteC \\
    & \WidthDev features & \WidthA features & \WidthB features & \WidthC features \\
    \midrule
    \multicolumn{5}{@{}l}{\emph{Passing pairs on the \ImagesScore scoring images of the COCO 2017 validation split}} \\
    \rowcolor{methodtint} Matched pairs, of \UnitsTrained & \MarkCarryCountDevDevCert & \MarkCarryCountACertA & \MarkCarryCountBCertB & \MarkCarryCountCCertC \\
    Into random decoder SAEs, of \UnitsFrozen & \MarkCarryCountDevDevFrozenCert & \MarkCarryCountAFrozenCertA & \MarkCarryCountBFrozenCertB & \MarkCarryCountCFrozenCertC \\
    Into surrogate SAEs, of \UnitsSurrogate & \MarkCarryCountDevDevSurrogateCert & none & \MarkCarryCountBSurrogateCertB & \MarkCarryCountCSurrogateCertC \\
    \midrule
    \multicolumn{5}{@{}l}{\emph{Share of pairs that pass}} \\
    \rowcolor{methodtint} Matched pairs & \inestimate{\MarkCarryShareDevTabDevTr}{\TabDevTrLow to \TabDevTrHigh} & \inestimate{\MarkCarryShareAShareA}{\TrLowA to \TrHighA} & \inestimate{\MarkCarryShareBShareB}{\TrLowB to \TrHighB} & \inestimate{\MarkCarryShareCShareC}{\TrLowC to \TrHighC} \\
    Into random decoder SAEs & \inestimate{\MarkCarryShareDevTabDevFr}{\TabDevFrLow to \TabDevFrHigh} & \inestimate{\MarkCarryShareAFrShareA}{\FrLowA to \FrHighA} & \inestimate{\MarkCarryShareBFrShareB}{\FrLowB to \FrHighB} & \inestimate{\MarkCarryShareCFrShareC}{\FrLowC to \FrHighC} \\
    Into surrogate SAEs & \inestimate{\MarkCarryShareDevTabDevSu}{\TabDevSuLow to \TabDevSuHigh} & none & \inestimate{\MarkCarryShareBSuShareB}{\SuLowB to \SuHighB} & \inestimate{\MarkCarryShareCSuShareC}{\SuLowC to \SuHighC} \\
    \midrule
    \multicolumn{5}{@{}l}{\emph{Matched pairs by the strength of their match}} \\
    Weak matches & \DevWeakUnits & \WeakUnitsA & \WeakUnitsB & \WeakUnitsC \\
    Weak matches that pass & \DevWeakMatch & \WeakMatchA & \WeakMatchB & \WeakMatchC \\
    Strong matches that pass & not computed & \MarkCarryCountACertA of \StrongPairsA & \MarkCarryCountBCertB of \StrongPairsB & \MarkCarryCountCCertC of \StrongPairsC \\
    \midrule
    \multicolumn{5}{@{}l}{\emph{Matched pairs that pass with a replacement built from several features}} \\
    Greedy selection of up to \Budget features & \DevClassGroup & \ClassGroupA & \ClassGroupB & \ClassGroupC \\
    Ridge regression on all features & \DevClassUpper & \ClassUpperA & \ClassUpperB & \ClassUpperC \\
    \midrule
    \multicolumn{5}{@{}l}{\emph{First SAEs and all other seeds}} \\
    Passing pairs of one first SAE & not computed & \SeedLeastA to \SeedMostA of \SeedPairs & \SeedLeastB to \SeedMostB of \SeedPairs & \SeedLeastC to \SeedMostC of \SeedPairs \\
    Features that pass into all other seeds & \DevEvery of \DrawnAll & \EveryA of \DrawnAll & \EveryB of \DrawnAll & \EveryC of \DrawnAll \\
    \bottomrule
  \end{tabular}
  \tablenote{\noteCarrying}
\end{table}

\paragraph{Intervals.}
Each interval shows how a result varies over samples of features drawn from the same trained SAEs. Unless stated otherwise, an interval spans the middle \IntervalLevelPct{}\% of a statistic's values over \Resamples resamples of the \DrawnAll sampled features, drawn with replacement without stratifying by seed, with every pair of a feature kept together. The trained SAEs stay fixed in every resample, so an interval measures the spread over sampled features alone. A share printed without an interval is a point estimate.

\section{Results}
\label{sec:results}

A close direction does not ensure a close effect: across the three settings, between \SharePctA{}\% and \SharePctC{}\% of matched pairs pass the output test. \Cref{sec:carrying} asks whether matched pairs pass more often than the controls. \Cref{sec:cosine} asks whether the cosine rule or the layer rule can predict them without running the later layers. \Cref{sec:every} asks which features pass into all other seeds and how much of the layer's variance they reconstruct. Before the main comparisons, every SAE passed both sanity checks. The matching returned the permutation of every rescaled copy, all \GaugeCertA pairs of rescaled copies passed, and none of the \UnitsKnown pairs of shuffled copies did. Every sampled feature also met the entry conditions of \cref{sec:test}.

\subsection{Matched pairs}
\label{sec:carrying}

The share of matched pairs that pass varies widely across the settings. Almost no pair passes in the controls. Of the \UnitsTrained matched pairs of each setting, \CertA pass at \cfgA, \CertB at \cfgB and \CertC at \cfgC, as \cref{tab:carrying} gives beside the development setting. The bound on the output error moves each share and leaves the order of the settings\later. At bounds of \BoundTight and \BoundLoose the shares are \BdTightA and \BdLooseA at \cfgA. At the same bounds they are \BdTightC and \BdLooseC at \cfgC. On the \FreshImages fresh images \FreshPassA, \FreshPassB and \FreshPassC pairs pass in the three settings. Again no pair into a random decoder SAE passes.

\paragraph{Almost no control pair passes.}
None of the \UnitsFrozen pairs into random decoder SAEs passes in any setting. Of the \UnitsSurrogate pairs into surrogate SAEs, none passes at \cfgB and one passes at \cfgC. So each observed matched share exceeds every observed control share by more than \MarginOne, the required difference of the first prespecified comparison. At \cfgA the matched share of \ShareA comes closest to that difference, since its interval reaches down to \TrLowA.

\paragraph{No permuted pair passes.}
When we keep every trained SAE and permute only the matches, none of the \TenthPermPairsTotal permuted pairs passes on the fresh images\later. Of the \TenthPermMatchedPairsA matched pairs of the same features, \TenthPermMatchedA, \TenthPermMatchedB and \TenthPermMatchedC pass in the three settings, as \cref{tab:repeat} counts. This control gives the rate of a random matching within the trained SAEs. Since it changes the matching alone, it isolates no effect of decoder learning.

\begin{table}[!htbp]
  \caption{\capRepeat}
  \label{tab:repeat}
  \centering
  \fontsize{9}{10.5}\selectfont
  \setlength{\tabcolsep}{3pt}
  \begin{tabular}{@{}>{\raggedright\arraybackslash}p{0.43\linewidth}ccc@{}}
    \toprule
    & \headA & \headB & \headCnarrow \\
    \midrule
    Features of an SAE & \WidthA & \WidthB & \WidthC \\
    \midrule
    \multicolumn{4}{@{}l}{\emph{Pairs that pass, of \UnitsTrained}} \\
    Scoring images & \CertA & \CertB & \CertC \\
    Fresh images & \FreshPassA & \FreshPassB & \FreshPassC \\
    Second sample\later & \TenthSecPassA & \TenthSecPassB & \TenthSecPassC \\
    \midrule
    \multicolumn{4}{@{}l}{\emph{Pairs that pass under the permuted control\later}} \\
    Matched pairs, of \TenthPermMatchedPairsA & \TenthPermMatchedA & \TenthPermMatchedB & \TenthPermMatchedC \\
    Permuted pairs, of \TenthPermAllPairsA & \TenthPermAllA & \TenthPermAllB & \TenthPermAllC \\
    \midrule
    \multicolumn{4}{@{}l}{\emph{Incorrect predictions on the fresh images $\downarrow$}} \\
    Cosine rule & \NinthErrCosineA & \NinthErrCosineB & \NinthErrCosineC \\
    Code norm ratio alone\later & \NinthErrMagnitudeA & \NinthErrMagnitudeB & \NinthErrMagnitudeC \\
    Code cosine alone\later & \NinthErrOverlapA & \NinthErrOverlapB & \NinthErrOverlapC \\
    Layer error with $c$ set to one\later & \textbf{\NinthErrCodesA} & \NinthErrCodesB & \NinthErrCodesC \\
    \rowcolor{methodtint} Layer rule\later & \NinthErrJointA & \textbf{\NinthErrJointB} & \textbf{\NinthErrJointC} \\
    \midrule
    \multicolumn{4}{@{}l}{\emph{Incorrect predictions on the second sample\later{} $\downarrow$}} \\
    Cosine rule & \TenthSecErrCosineA & \TenthSecErrCosineB & \TenthSecErrCosineC \\
    Layer error with $c$ set to one & \textbf{\TenthSecErrCodesA} & \TenthSecErrCodesB & \TenthSecErrCodesC \\
    \rowcolor{methodtint} Layer rule & \TenthSecErrSiteA & \textbf{\TenthSecErrSiteB} & \textbf{\TenthSecErrSiteC} \\
    \bottomrule
  \end{tabular}
  \tablenote{\noteRepeat}
\end{table}

\paragraph{Only strong matches pass in the three settings.}
No weak match passes in the three settings, against \DevWeakMatch of \DevWeakUnits weak matches in the development setting. A strong match is not enough, though, since \StrongNotA of the \StrongPairsA strong matches at \cfgA fail. In the settings we tested, the share that passes follows the number of strong matches. That number differs more from development in the changed layer and width settings than in the changed encoder setting. The densest SAEs, those of \cfgA, have the fewest strong matches. No single seed accounts for a share\later, since at \cfgA the pairs of each first SAE include \SeedLeastA to \SeedMostA passes of \SeedPairs.

\paragraph{The replacement pass totals are close.}
A replacement built from several features of the second SAE passes a similar number of pairs in total. At \cfgC greedy selection passes \ClassGroupC pairs and ridge regression \ClassUpperC, against \CertC for the match. The fourth block of \cref{tab:carrying} gives every setting. Among the \WeakUnitsA, \WeakUnitsB and \WeakUnitsC weak matches of the three settings, ridge regression passes at most \WeakUpperC. So even the fitted replacements rarely pass for weak matches.

\subsection{Predicting the outcome}
\label{sec:cosine}

\begin{figure}[!htbp]
  \centering
  \includegraphics[width=\linewidth]{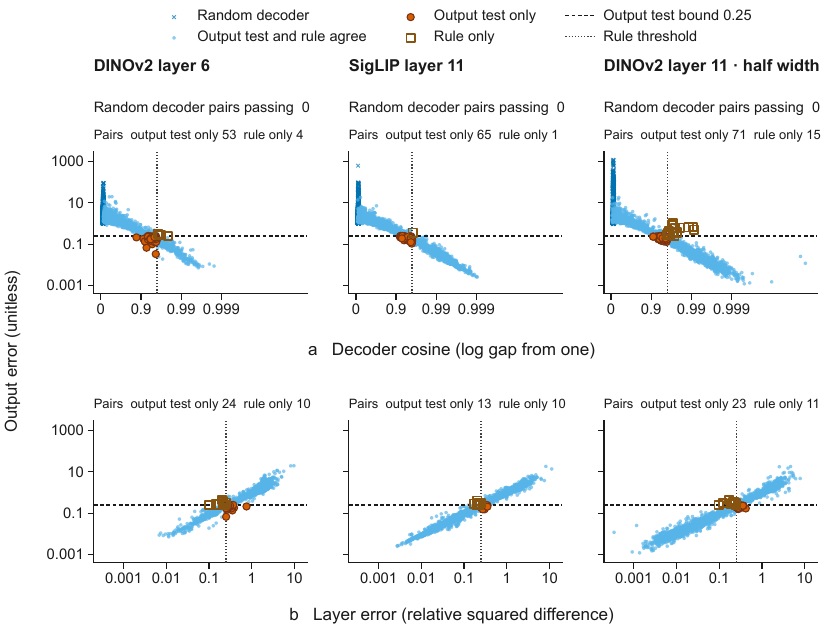}
  \caption{\figPredictorsCaption}
  \label{fig:cosine}
  \tablenote{\noteFigPredictors}
\end{figure}

\begin{table}[!htbp]
  \caption{\capCosine}
  \label{tab:cosine}
  \centering
  \fontsize{9}{10.5}\selectfont
\renewcommand{\estimate}[2]{\ensuremath{\textnormal{#1}_{\publicspread#2\relax}}}
  \setlength{\tabcolsep}{1pt}
  \begin{tabular}{@{}>{\raggedright\arraybackslash}p{135pt}ccc@{}}
    \toprule
    & \shortstack{\Dino layer \SiteA\\\WidthA features} & \shortstack{SigLIP layer \SiteB\\\WidthB features} & \shortstack{\Dino layer \SiteC\\\WidthC features} \\
    \midrule
    \multicolumn{4}{@{}l}{\emph{Agreement with the outcomes on the two halves of the scoring images $\uparrow$}} \\
    \rowcolor{methodtint} Split half agreement of the output test & \inestimate{\MarkAgreementASelfAgreeA}{\SelfLowA to \SelfHighA} & \inestimate{\MarkAgreementBSelfAgreeB}{\SelfLowB to \SelfHighB} & \inestimate{\MarkAgreementCSelfAgreeC}{\SelfLowC to \SelfHighC} \\
    Agreement of the cosine rule & \inestimate{\MarkAgreementARuleAgreeA}{\RuleLowA to \RuleHighA} & \inestimate{\MarkAgreementBRuleAgreeB}{\RuleLowB to \RuleHighB} & \inestimate{\MarkAgreementCRuleAgreeC}{\RuleLowC to \RuleHighC} \\
    Agreement of the layer rule & \inestimate{\MarkAgreementASiteHalvesA}{\SiteHalvesLowA to \SiteHalvesHighA} & \inestimate{\MarkAgreementBSiteHalvesB}{\SiteHalvesLowB to \SiteHalvesHighB} & \inestimate{\MarkAgreementCSiteHalvesC}{\SiteHalvesLowC to \SiteHalvesHighC} \\
    Output test less cosine rule & \inestimate{\DiffTwoA}{\DiffTwoLowA to \DiffTwoHighA} & \inestimate{\DiffTwoB}{\DiffTwoLowB to \DiffTwoHighB} & \inestimate{\DiffTwoC}{\DiffTwoLowC to \DiffTwoHighC} \\
    Required difference & \MarginTwoA & \MarginTwoB & \MarginTwoC \\
    \midrule
    \multicolumn{4}{@{}l}{\emph{Pairs of \UnitsTrained on all scoring images}} \\
    \rowcolor{methodtint} Passed by the output test & \CertA & \CertB & \CertC \\
    Predicted to pass by the cosine rule & \RulePredA & \RulePredB & \RulePredC \\
    Incorrect predictions of the cosine rule $\downarrow$ & \MarkDisagreementADisagreeA & \MarkDisagreementBDisagreeB & \MarkDisagreementCDisagreeC \\
    Passing pairs it predicts to fail & \RuleFalseNegA & \RuleFalseNegB & \RuleFalseNegC \\
    Failing pairs it predicts to pass & \RuleFalsePosA & \RuleFalsePosB & \RuleFalsePosC \\
    Incorrect predictions of the layer rule $\downarrow$ & \MarkDisagreementASiteDisagreeA & \MarkDisagreementBSiteDisagreeB & \MarkDisagreementCSiteDisagreeC \\
    Cosine rule less layer rule & \inestimate{\ErrDiffA}{\ErrDiffLowA to \ErrDiffHighA} & \inestimate{\ErrDiffB}{\ErrDiffLowB to \ErrDiffHighB} & \inestimate{\ErrDiffC}{\ErrDiffLowC to \ErrDiffHighC} \\
    Rank correlation of layer error and output error & \SelRhoA & \SelRhoB & \SelRhoC \\
    \bottomrule
  \end{tabular}
  \tablenote{\noteCosine}
\end{table}

Both rules predict the outcome of the output test for nearly every pair. The layer rule predicts incorrectly less often. These experiments ask whether the output test is needed once the two cosines are known. They also ask whether a comparison at the edited layer can stand in for it. The cosine rule predicts incorrectly on \DisagreeA to \DisagreeC of each setting's \UnitsTrained pairs, and the layer rule on \SiteDisagreeB to \SiteDisagreeA. The split half agreement of the output test is the share of pairs whose outcome is the same on the two halves of the scoring images. It shows how far the output test agrees with itself. \Cref{tab:cosine} gives agreement with the outcome on each half in its upper block and incorrect predictions on all scoring images in its lower block.

\paragraph{The cosine rule misses mostly passing pairs.}
On the development pairs that fitted it, the cosine rule agreed with the output test on \RuleDevAgree of \DevUnits. In the three settings most of its incorrect predictions are passing pairs that it predicts to fail, \RuleFalseNegA of the \CertA passing pairs at \cfgA. A higher decoder cosine goes with a lower output error in every setting, as the upper row of \cref{fig:cosine} shows. The pairs into random decoder SAEs lie apart there at low cosine and high output error. In each plot of that figure, a pair below the dashed line passes the output test. In the upper row a pair right of the dotted line meets the decoder threshold. The cosine rule predicts a pass only where the encoder cosine meets its threshold as well. In the lower row the layer rule predicts a pass left of the dotted line, so an output test only marker lies below the dashed line on the side where its rule predicts a failure.

\paragraph{The two agreement rates are close.}
The second comparison measures the gap between the two agreement rates. That comparison asks whether the split half agreement of the output test exceeds the cosine rule's agreement with the outcomes on the halves by more than the required difference. At \cfgA the split half agreement is \SelfAgreeA and the cosine rule's agreement is \RuleAgreeA, a gap that exceeds the required difference by one pair in \UnitsTrained. In the other two settings the gap is smaller than that difference. On the fresh images it is smaller in all three.

\paragraph{The layer rule mispredicts fewer pairs.}
The layer rule\later{} predicts incorrectly on \ErrDiffA to \ErrDiffC fewer pairs than the cosine rule. It agrees with the outcomes on the halves nearly as often as the output test agrees with itself. The layer error also ranks the pairs as the output error does, with rank correlations from \SelRhoA to \SelRhoC, so a pair with a larger layer error tends to have a larger output error, as the lower row of \cref{fig:cosine} shows.

The advantage repeats on the fresh images at the threshold of \NinthFitJoint that we fitted post hoc on the development pairs. There the layer rule predicts incorrectly on \NinthErrJointA, \NinthErrJointB and \NinthErrJointC pairs in the three settings, against the cosine rule's \NinthErrCosineA, \NinthErrCosineB and \NinthErrCosineC. No rule built from one part of \cref{eq:layer} and fitted the same way predicts incorrectly as rarely as the layer rule in every setting\later, as \cref{tab:repeat} counts. The layer error with the decoder cosine set to one predicts incorrectly on \NinthErrCodesA pairs at \cfgA, fewer than the layer rule's \NinthErrJointA. At \cfgB it predicts incorrectly on \NinthErrCodesB, more than the layer rule's \NinthErrJointB. So the data leave open which part of the layer error gives the layer rule its advantage.

\paragraph{A second sample repeats the gain.}
We sampled \DrawnPerSource further features from each SAE and scored their \TenthSecPairsA new pairs a setting on the fresh images with every rule frozen\later. The layer rule predicts incorrectly on \TenthSecErrSiteMin to \TenthSecErrSiteMax pairs a setting and the cosine rule on \TenthSecErrCosineMin to \TenthSecErrCosineMax. Each difference has an interval above zero. The advantage repeats across these two feature samples. At \cfgA the layer error with the decoder cosine set to one again predicts incorrectly on fewer pairs than the layer rule, though the interval of that difference now includes zero.

\paragraph{The image level output repeats the gain.}
A classifier or a retrieval rule takes one vector per image as its input, the class token for \Dino and the pooled image embedding for SigLIP, rather than every token. Measured at that vector on the fresh images\later, \EleventhOutPctImageA{}\% to \EleventhOutPctImageC{}\% of the \UnitsTrained matched pairs pass. Against that outcome the layer rule predicts incorrectly on \EleventhErrImageLayerFittedA to \EleventhErrImageLayerFittedC pairs and the cosine rule on \EleventhErrImageCosineA to \EleventhErrImageCosineC. Each difference has an interval above zero. So the advantage of the layer rule repeats at the image level output, the one vector per image. \Cref{sec:supp-outputs} gives the details.

\paragraph{One incorrect prediction persists on new images.}
Two features can agree at the edited layer and still differ at the output. To find such pairs, we took the \EightCount pairs on which the layer rule and the output test, computed on the same images, disagreed in one direction on every half of the scoring images and of the fresh images. We then scored them on a third image set of \SeventhThirdImages COCO training images that no earlier step had used\later. One pair, at \cfgC, keeps its disagreement there: the layer rule passes it at a layer error of \SeventhSevenThirdWholeSite and the output test fails it at an output error of \SeventhSevenThirdWholeOutput. Both values stay on their sides of \Certify in at least \SeventhFamilyShare of \Resamples paired resamples of the images, on the whole set and on each half. Because the third image set shares no image with the sets that selected the pair, its disagreement there does not come from those images.

The final normalization does not create this disagreement. Taken just before that normalization, the ratio of \cref{eq:error} already lies on the same side of \Certify for each of the \EightCount pairs as its output error does, so the normalization moves no pair across the bound. For the pair whose disagreement persists, the ratio is \NinthCasePre before the normalization and \SeventhSevenThirdWholeOutput after it. Its matched effect has a norm between \NinthCaseNormLow and \NinthCaseNormHigh times that of the first effect. \Cref{sec:supp-calibrated} follows the pair image by image and token by token.

\subsection{Features that pass into all other seeds}
\label{sec:every}

\begin{figure}[!htbp]
  \centering
  \includegraphics[width=\linewidth]{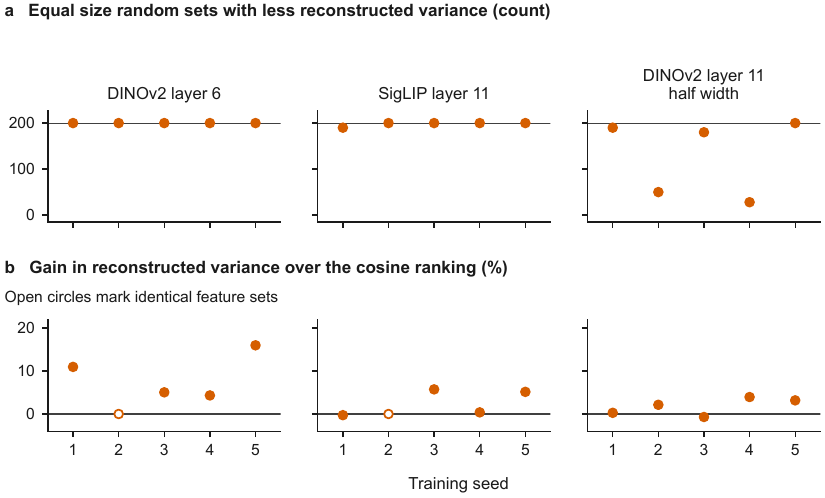}
  \begingroup
\fontsize{8}{10}\selectfont
\setlength{\tabcolsep}{0.8pt}
\begin{tabular*}{\linewidth}{@{}p{79pt}@{\extracolsep{\fill}}*{15}{>{\centering\arraybackslash}p{18pt}}@{}}
\toprule
Setting & \multicolumn{5}{c}{\Dino{} layer \SiteA} & \multicolumn{5}{c}{SigLIP layer \SiteB} & \multicolumn{5}{c}{\Dino{} layer \SiteC{} at half width} \\
Training seed & \VarianceSeedOne & \VarianceSeedTwo & \VarianceSeedThree & \VarianceSeedFour & \VarianceSeedFive & \VarianceSeedOne & \VarianceSeedTwo & \VarianceSeedThree & \VarianceSeedFour & \VarianceSeedFive & \VarianceSeedOne & \VarianceSeedTwo & \VarianceSeedThree & \VarianceSeedFour & \VarianceSeedFive \\
Features that pass & \SvAOneSize & \SvATwoSize & \SvAThreeSize & \SvAFourSize & \SvAFiveSize & \SvBOneSize & \SvBTwoSize & \SvBThreeSize & \SvBFourSize & \SvBFiveSize & \SvCOneSize & \SvCTwoSize & \SvCThreeSize & \SvCFourSize & \SvCFiveSize \\
Random sets with less variance & \SvAOneBelow & \SvATwoBelow & \SvAThreeBelow & \SvAFourBelow & \SvAFiveBelow & \SvBOneBelow & \SvBTwoBelow & \SvBThreeBelow & \SvBFourBelow & \SvBFiveBelow & \SvCOneBelow & \SvCTwoBelow & \SvCThreeBelow & \SvCFourBelow & \SvCFiveBelow \\
Not ranked highest by cosine\later{} & \SaAOne & \SaATwo & \SaAThree & \SaAFour & \SaAFive & \SaBOne & \SaBTwo & \SaBThree & \SaBFour & \SaBFive & \SaCOne & \SaCTwo & \SaCThree & \SaCFour & \SaCFive \\
\bottomrule
\end{tabular*}
\endgroup

  \caption{\figEveryCaption}
  \label{fig:variance}
  \tablenote{\noteFigEvery}
\end{figure}

The number of features that pass into all other seeds varies with the setting as the share of passing pairs does. These features are nearly the ones that the cosines rank highest\later. These are the features whose ablation their matches reproduce in every other sampled SAE of the setting. Of the \DrawnAll sampled features, \EveryA, \EveryB and \EveryC pass into all four other seeds in the three settings. To compare them with the cosines, we rank each feature by the smaller of its match's two cosines in each other seed and take the least of these over the four seeds. On each SAE we then keep as many features from the top of this ranking as pass into all other seeds. The two sets differ in at most \AtomsApartMost features on every SAE, as the last row of the table in \cref{fig:variance} counts. They are the same set on one SAE of \cfgA and one of \cfgB.

\paragraph{Those features usually reconstruct more variance than equally many cosine ranked features.}
We compare each set with sets of equal size drawn from the same SAE. On all five SAEs of \cfgA and four of \cfgB, the features that pass into all other seeds reconstruct more of the layer's variance on the scoring images than each of \RandomSets random sets of the same size, as \cref{fig:variance} shows. At \cfgC they do so on one SAE. On two others, \RandomBelowCSecond and \RandomBelowCFourth of the random sets reconstruct less\later. Against as many features ranked highest by cosine, they reconstruct more on most SAEs and less on one SAE of \cfgB and one of \cfgC. On the fresh images they keep a mean advantage in all three settings. So on most SAEs the features that pass reconstruct more of the layer's variance than the sampled random sets of equal size.

\paragraph{Object annotations name too few features.}
COCO's object annotations name too few features to say whether a feature that passes keeps its meaning in another seed. Each annotation outlines one object of a named category with a mask, the set of pixels that the object covers. We name a feature after the object category whose masks cover at least half of its positive codes over patches. Only \NameFixedA to \NameFixedC of the \DrawnAll sampled features reach that share on the selection images. A feature keeps its name only where the same category covers that share on the scoring images as well. After that check, more than nine in ten take no name in every setting. For the median feature the best category covers \NameMedianB to \NameMedianA of its codes\later. \Cref{sec:supp-recipe} gives the counts.

\section{Discussion}
\label{sec:discussion}

In the timed settings, the layer rule adds a small fraction of the output test's time per pair. One of its incorrect predictions persists, though. The output test detects that error. We timed the layer rule and the output test on one accelerator at \cfgA\later. Beyond a pass to the edited layer that both share, the layer rule adds \TimeSitePairA seconds a pair, while the output test adds \TimeFullPairA seconds a pair to run the later layers. At \cfgB, where an edit passes through one later layer, the two times are \TimeSitePairB and \TimeFullPairB seconds a pair. The two steps of the cosine rule quoted next ran on two different processors and not on that accelerator, so their times compare neither with each other nor with the accelerator times. Matching two SAEs took a median of \SeventhMacA seconds on an \SeventhProcessor. The cosine rule's predictions for their \WidthA matched features took \SeventhScreenCallsA seconds in a container whose log names no processor model, as \cref{sec:supp-sources} details.

The two kinds of incorrect prediction differ in what they cost a researcher. A pair that a rule rejects although it passes costs only a match left unused. A pair that a rule accepts although it fails lets an edit stand whose effect differs in the other seed. The pair whose incorrect prediction persists is of this kind.

So a researcher who needs features that recur across seeds can match the SAEs, screen the pairs by layer error and keep the output test for the edits whose final effect matters. Combining the two rules\later{} shows how many output tests such screening avoids. If we accept a prediction wherever the cosine rule and the layer rule agree and run the output test wherever they disagree, \ScreenSentA to \ScreenSentC of each setting's \UnitsTrained pairs go to the output test. At \cfgA, for example, the output test would run on \ScreenSentA of the \UnitsTrained pairs instead of on all of them. Among the accepted predictions, \ScreenWrongB to \ScreenWrongA are still incorrect, so agreement between these rules can miss a failing pair.

\section{Conclusion}
\label{sec:conclusion}

A study that matches SAE features across seeds can report how many matched pairs pass the output test beside how many reach a high decoder cosine. Where an edit's final effect matters, the study can screen pairs by the layer error before it runs the output test. The same comparison of two effects applies to steering and to other edits of a feature's contribution. On \Dino the share that passes rises from layer~\SiteA to layer~\SiteDev. Whether it keeps rising toward the output is open. Whether it rises on encoders trained with other objectives is open as well. Denser annotations of object parts and textures would let us ask whether a feature that passes keeps its name in the other seed.

\FloatBarrier
\bibliographystyle{plainnat}
\bibliography{references}
\clearpage
\appendix

\section{Training and measurement details}
\label{sec:supp-recipe}

Every SAE is trained by one recipe from a stated seed, and two sanity checks on altered copies precede the main comparisons.

\paragraph{The recipe.}
Each SAE trains on the tokens of its setting's edited layer from the first \ImagesTrain images, scaled as in \cref{sec:test} by the mean and the root mean square of those training tokens. On batches of \Batch tokens an SAE minimizes the squared reconstruction error of each scaled token, summed over the token's coordinates, divided by the width $n$ and averaged over the batch, plus \Coefficient times the summed code averaged over the batch. The penalty rises linearly over the first \WarmSteps steps. Adam runs \Steps steps at a rate of \Rate with PyTorch's default coefficients and no weight decay, the rate rising linearly over the first \WarmSteps steps and falling linearly to zero over the last \DecaySteps. Each decoder direction starts as a draw from a standard normal distribution scaled to unit length, each encoder row starts as its decoder direction, and both biases start at zero. Before each step we remove the part of each decoder direction's gradient along the direction, and after it we scale the direction back to unit length.

Both encoders resize an image by bicubic interpolation, divide its pixel values by \Rescale and normalize each channel by the mean and standard deviation that the encoder's release gives, \DinoMeanOne, \DinoMeanTwo and \DinoMeanThree and \DinoStdOne, \DinoStdTwo and \DinoStdThree for \Dino and \TowerMean and \TowerStd for SigLIP. Arithmetic runs in single precision with reduced precision matrix products switched off, and both encoders run in evaluation mode. We compute the matching in double precision with the linear sum assignment of SciPy~\cite{virtanen2020scipy} \ScipyVersion.

We fit the cosine rule's thresholds over a grid of hundredths from zero to one on each cosine, with one value beyond each end that allows a rule predicting a pass for every pair or for none. On a tie, which \RuleTies pairs of thresholds reached, we keep the lower decoder threshold and then the lower encoder threshold. An SAE's unexplained variance in \cref{tab:dictionaries} is its summed squared reconstruction error over the summed squared distance of the scaled held out tokens from their own mean, and its active features are the mean count of positive codes on a token. We compute each pair's layer error from the sums of squared codes and cross products of codes over the image tokens, multiplied by the inner products of the corresponding decoder directions. Every rank correlation is Spearman's, with average ranks on ties.

\paragraph{Seeds.}
The training seed of each SAE is what the main text calls its seed, and separate sampling seeds fix every random draw of a measurement. The images take their order from seed \SeedImages over their sorted file names, and we sample an SAE's features without replacement from seed \SeedDraw plus the SAE's own seed. The default generator of NumPy~\cite{harris2020numpy} \NumpyVersion draws the image order, the sampled features, the copies' permutations and factors, the surrogate tokens, the random sets and the resamples, each from the seed stated for it. PyTorch's generator on the processor, seeded by an SAE's own seed, draws its initialization and its token order, and where a pass through the training tokens cannot fill a last batch it leaves those tokens out and starts a fresh order.

The trained SAEs take seeds \SeedTrainedFirst to \SeedTrainedLast and the random decoder SAEs seeds \SeedFrozenFirst to \SeedFrozenLast. The surrogate tokens of a setting are Gaussian draws of mean zero with the covariance of the layer's scaled training tokens, then centred and scaled on their own statistics as the real tokens are, from seed \SeedSurrogate, as many as the training tokens and the tokens held out of training. The surrogate SAEs take seeds \SeedSurrogateFirst to \SeedSurrogateLast, and a feature of one seed is matched into the surrogate SAEs of the other four.

The halves of the scoring images are their first and last \ImagesHalf images. The fresh images take their order from seed \SeedFreshSet over the sorted file names of the \FreshList images of the COCO 2017 training split, and the first \FreshImages in that order form the set, whose first and last \FreshHalf form its halves. The third image set uses seed \ThirdSeed to permute the COCO 2017 training images, sorted by file name, after all validation images and the fresh images are excluded. The first \SeventhThirdImages images in that order form the set, with no candidate skipped, and its two halves have one hundred twenty-five images each. We keep the ordered image identifiers with the measurements.

The random sets of \cref{sec:every} come from seed \SeedRandomSets plus the SAE's seed. The resamples behind the intervals of the output test come from seed \SeedBoot, and the bound for the fresh images reuses those resamples. Among the post hoc quantities, the resamples of the agreement difference over the sampled features come from seed \SeedContrast, those of the repeated incorrect predictions on the selection images from seed \SeedDepartures and those of the repeated incorrect predictions on the same halves from seed \SeedSame. The resamples of the scoring images within each half come from seed \SeedImageDraws, those of the fresh images within each half from seed \SeedFreshDraws, and the intervals of the layer rule's agreements with the halves, of the difference of the two rules' incorrect predictions and of the match's shares at the bounds of \BoundTight and \BoundLoose from seed \SeedAfterDraws. The three settings add zero, one and two to each seed, and each interval is the middle \IntervalLevelPct{}\% of its resamples under NumPy's linear rule of interpolation.

The paired image resamples of the eight pairs of \cref{sec:supp-halves} use seed \PairImageSeed plus one hundred times the setting's index, ten times the set's index and the part's index. Those indices start at zero in the orders \cfgA, \cfgB and \cfgC, then scoring, fresh and third, then whole, earlier half and later half. The counts at the fitted threshold use seeds \FittedSeedA, \FittedSeedB and \FittedSeedC in the three settings, with \Resamples paired resamples within each fresh half. The summaries by image and by token use all images without resampling.

\paragraph{The two sanity checks.}
Each sanity check builds a copy of an SAE whose correct outcome we know before measuring. A rescaled copy of an SAE permutes its features, divides each decoder direction by a positive factor and multiplies its encoder row and bias by the same factor, which leaves every contribution unchanged. The matching returns the permutation where each copied decoder direction has a strict cosine maximum at its original, and we required at least nineteen pairs in twenty to pass. A shuffled copy moves the encoder rows with their biases among the features under a permutation that leaves no feature in place, and it keeps every decoder direction. Where no two encoder rows with their biases are equal, the shuffled copy can change each feature's code, which alone does not ensure a failure, so we required at most one pair in twenty to pass. A setting whose checks failed on any SAE would have been excluded.

The rescaled copy takes its permutation and its factors from seed \SeedGauge plus the SAE's seed, each factor the exponential of a draw uniform between minus \LogScale and \LogScale, and the shuffled copy draws its permutation from seed \SeedShuffle plus the SAE's seed, drawing again until no feature stays in place. In every setting, the development setting among them, each copied decoder direction has its strict cosine maximum at its own original, and no two encoder rows with their biases are equal\later. The smallest gap between the cosine of a copied direction with its original and its cosine with any other direction is \CopyGapA at \cfgA, \CopyGapB at \cfgB, \CopyGapC at \cfgC and \CopyGapDev on development, and each setting has \CopyDuplicates duplicate encoder rows with biases. So the rescaled copy has one permutation to recover and the shuffled copy permutes the encoding rows, and the failure of its pairs is what we measured.

The encoder's later layers applied to the unedited activations return the encoder's own output exactly. A feature is active on an image where any token of the image, the class token among them, has a positive code. Every sampled feature's ablation was at least \EnergyFloorTimesA times the minimum squared effect of \cref{sec:test} at \cfgA, \EnergyFloorTimesB at \cfgB and \EnergyFloorTimesC at \cfgC.

\paragraph{Replacements built from several features.}
Both replacements stand in for the first feature with several features of the second SAE, where the match uses one. Greedy selection, as \cref{sec:test} defines it, adds each feature unscaled. Its search stops after \Budget features or when no feature lowers the distance to the first contribution by more than \GreedyTol of that contribution's squared size, and on a tie it takes the feature with the lower index. Ridge regression chooses the weights that minimize the squared distance between the first contribution and the weighted sum, summed over the selection tokens, plus a penalty on the summed squared weights, with no intercept. The penalty's coefficient is \Ridge times the mean, over the second SAE's features, of each feature's squared contribution summed over the same tokens. We score both on every pair, as we score the match.

\paragraph{The naming rule.}
A feature takes as its name the COCO instance category whose masks cover the largest share of its summed positive codes over image patches on the selection images, where that share is at least \NameHold and the feature is active on at least \MinImages of those images. It keeps the name only where the same category covers at least \NameHold of its summed positive codes over patches on the scoring images. A feature that is active on the class token alone has no positive code at a patch and takes no name, and equal shares select the lowest COCO category identifier. A named pair is one whose two features both take a name, and \cref{tab:names} counts the features by the share their best category covers and the named pairs that agree.

We fill each instance's polygons with the polygon drawing of Pillow~\cite{pillow2024release} \PillowVersion and decode each crowd region from its run lengths, at the image's own size. A category's mask is the union of its instances, crowd regions among them, resized to the encoder's input size by area averaging and cropped as the image is cropped. Averaged over each patch, it gives the share of the patch's pixels the category covers. Each category covers a patch on its own, so the shares of overlapping categories are computed independently.

\begin{table}[H]
  \caption{\capNames}
  \label{tab:names}
  \centering
  \fontsize{9}{10.5}\selectfont
  \setlength{\tabcolsep}{2pt}
  \begin{tabular}{@{}>{\raggedright\arraybackslash}p{0.36\linewidth}ccc@{}}
    \toprule
    & \headA & \headB & \headCnarrow \\
    \midrule
    Largest share at or above \NameHold & \NameFixedA & \NameFixedB & \NameFixedC \\
    Largest share below \NameHold & \NameShareLowA & \NameShareLowB & \NameShareLowC \\
    Name lost on the scoring images & \NameLostA & \NameLostB & \NameLostC \\
    Median largest share & \NameMedianA & \NameMedianB & \NameMedianC \\
    Features without a name & \UnnamedA & \UnnamedB & \UnnamedC \\
    \midrule
    Passing pairs & \AgreeCertA{} of \NamedCertA & \AgreeCertB{} of \NamedCertB & \AgreeCertC{} of \NamedCertC \\
    Failing pairs & \AgreeNotCertA{} of \NamedNotCertA & \AgreeNotCertB{} of \NamedNotCertB & \AgreeNotCertC{} of \NamedNotCertC \\
    Pairs the cosine rule predicts to pass & \AgreeRuleA{} of \NamedRuleA & \AgreeRuleB{} of \NamedRuleB & \AgreeRuleC{} of \NamedRuleC \\
    Pairs of the shuffled copy & \AgreeShufA{} of \NamedShufA & \AgreeShufB{} of \NamedShufB & \AgreeShufC{} of \NamedShufC \\
    \bottomrule
  \end{tabular}
  \tablenote{\noteNames}
\end{table}

\clearpage
\section{Comparisons fixed in advance and the cosine rule in detail}
\label{sec:supp-halves}

We fixed three comparisons before we measured the three settings. The first is met in every setting, the second in one setting by one pair that the fresh images do not repeat, and the third cannot be made.

\paragraph{The three criteria.}
The first comparison could establish that matched pairs pass more often than the controls, the second that the output test agrees better across halves than with the cosine rule, and the third that a feature that passes keeps its meaning. The required difference $m(q)$ of \cref{sec:test} shrinks to zero as the baseline share $q$ nears one, and a tie does not exceed it. The first comparison asks that the share of matched pairs that pass exceed the share of pairs into random decoder SAEs, and the share of pairs into surrogate SAEs where we trained them, by more than this difference. The second sets the output test against the cosine rule. Write $v^{1}_{p}$ and $v^{2}_{p}$ for pair $p$'s outcomes on the two halves of the scoring images and $r_{p}$ for the rule's prediction, each one for a pass and zero otherwise, over the $N$ pairs eligible on both halves. The split half agreement of the output test and the agreement of the rule are
\begin{equation}
  \begin{split}
    A_{\mathrm{self}} &= \frac{1}{N} \sum_{p} \mathbf{1}\bigl[v^{1}_{p} = v^{2}_{p}\bigr], \\
    A_{\mathrm{rule}} &= \frac{1}{2N} \sum_{p} \sum_{j} \mathbf{1}\bigl[r_{p} = v^{j}_{p}\bigr].
  \end{split}
  \label{eq:agreement}
\end{equation}

A pair's outcome on a half enters only where the ablation's squared effect on that half alone meets the minimum squared effect of \cref{sec:test}. The comparison asks that $A_{\mathrm{self}} - A_{\mathrm{rule}} > m(A_{\mathrm{rule}})$, and we fixed in advance that one setting of three meeting it would suffice. The third asks whether a feature that passes keeps its name. It compares how often passing pairs agree in name with how often three other groups do: the failing pairs, the pairs the cosine rule predicts to pass and the pairs of the shuffled copy, which set each sampled feature with its matched feature in that copy. Each difference must exceed $m$ in at least two settings. It is made only where at most nine sampled features in ten lack a name and each group has at least fifty named pairs, which no setting meets.

\paragraph{The second criterion is met by one pair.}
At \cfgA the difference between the split half agreement of the output test and the cosine rule's agreement is \DiffTwoA, which exceeds the required \MarginTwoA by the agreement of one pair in \UnitsTrained. At \cfgB and \cfgC the differences lie below the required difference, the latter by the same one pair. Resampling the sampled features keeps the difference at \cfgA above its required difference on \ContrastPassCountA of \Resamples resamples, and resampling the scoring images within each half keeps it above on \ImageContrastA\later. On the fresh images the difference lies below its required difference in every setting, at \cfgA by the agreement of one pair, as \cref{tab:supp-fresh} gives. So neither resampling nor the fresh images separate the agreement of the output test from the cosine rule's by the difference we required.

\renewcommand{\dbltopfraction}{0.95}
\begin{table}[!htbp]
  \caption{\capHalves}
  \label{tab:supp-halves}
  \centering
  \fontsize{9}{10.5}\selectfont
\renewcommand{\estimate}[2]{\ensuremath{\textnormal{#1}_{\publicspread#2\relax}}}
  \setlength{\tabcolsep}{1pt}
  \begin{tabular}{@{}>{\raggedright\arraybackslash}p{175pt}ccc@{}}
    \toprule
    & \headA & \headB & \headC \\
    \midrule
    Failing pairs the cosine rule predicts to pass, earlier half & \ShFirstPosA & \ShFirstPosB & \ShFirstPosC \\
    Passing pairs the cosine rule predicts to fail, earlier half & \ShFirstNegA & \ShFirstNegB & \ShFirstNegC \\
    Failing pairs the cosine rule predicts to pass, later half & \ShSecondPosA & \ShSecondPosB & \ShSecondPosC \\
    Passing pairs the cosine rule predicts to fail, later half & \ShSecondNegA & \ShSecondNegB & \ShSecondNegC \\
    \midrule
    Agreement difference less required difference & $\ContrastA$ & $\ContrastB$ & $\ContrastC$ \\
    Its interval & $\ContrastLowA$ to $\ContrastHighA$ & $\ContrastLowB$ to $\ContrastHighB$ & $\ContrastLowC$ to $\ContrastHighC$ \\
    Resamples above zero & \ContrastPassCountA & \ContrastPassCountB & \ContrastPassCountC \\
    Resamples of the images above zero & \ImageContrastA & \ImageContrastB & \ImageContrastC \\
    First SAE seeds left out, still above zero & \LeaveSourceA & \LeaveSourceB & \LeaveSourceC \\
    SAEs left out, still above zero & \LeaveDictA & \LeaveDictB & \LeaveDictC \\
    \midrule
    Pairs whose halves disagree & \ExHalfA & \ExHalfB & \ExHalfC \\
    Repeated incorrect predictions, cosine rule & \ExCosA & \ExCosB & \ExCosC \\
    Cosine rule less disagreeing pairs, interval & $\ExCosLowA$ to $\ExCosHighA$ & $\ExCosLowB$ to $\ExCosHighB$ & $\ExCosLowC$ to $\ExCosHighC$ \\
    SAEs left out, difference still above zero & \ExCosLeaveA & \ExCosLeaveB & \ExCosLeaveC \\
    Repeated incorrect predictions, layer rule & \ExSiteA & \ExSiteB & \ExSiteC \\
    Layer rule less disagreeing pairs, interval & $\ExSiteLowA$ to $\ExSiteHighA$ & $\ExSiteLowB$ to $\ExSiteHighB$ & $\ExSiteLowC$ to $\ExSiteHighC$ \\
    SAEs left out, difference still above zero & \ExSiteLeaveA & \ExSiteLeaveB & \ExSiteLeaveC \\
    Repeated incorrect predictions of both rules & \ExBothA & \ExBothB & \ExBothC \\
    \midrule
    Repeated incorrect predictions, layer rule, same halves & \SameSiteA & \SameSiteB & \SameSiteC \\
    Layer rule less disagreeing pairs, same halves, interval & $\SameSiteLowA$ to $\SameSiteHighA$ & $\SameSiteLowB$ to $\SameSiteHighB$ & $\SameSiteLowC$ to $\SameSiteHighC$ \\
    SAEs left out, difference still above zero & \SameSiteLeaveA & \SameSiteLeaveB & \SameSiteLeaveC \\
    Repeated incorrect predictions of both rules, same halves & \SameBothA & \SameBothB & \SameBothC \\
    Random decoder pairs the layer rule predicts to pass & \FrozenSiteA & \FrozenSiteB & \FrozenSiteC \\
    \midrule
    Pairs on which the two rules disagree & \ScreenSentA & \ScreenSentB & \ScreenSentC \\
    Incorrect predictions where the two rules agree & \ScreenWrongA & \ScreenWrongB & \ScreenWrongC \\
    \bottomrule
  \end{tabular}
  \tablenote{\noteHalves}
\end{table}

\Cref{tab:supp-halves} gives this difference less the required difference, so that a positive value meets the second comparison. The table also gives the cosine rule's incorrect predictions on each half, the response of that value to resampling and to leaving a seed out, and the pairs whose outcomes on both halves differ from each rule's prediction. On each half most of the cosine rule's incorrect predictions are passing pairs that it predicts to fail, \ShFirstNegA on the earlier half at \cfgA and \ShSecondNegA on its later half.

The difference less the required difference stays above zero at \cfgA in \LeaveSourceA of the \Seeds cases that leave out the pairs of one first SAE in turn, and in \LeaveDictA of the \Seeds cases that leave out one SAE as first and as second SAE. It does so at \cfgB in \LeaveSourceB and \LeaveDictB cases and at \cfgC in \LeaveSourceC and \LeaveDictC\later. It stays above zero on \ContrastPassCountB resamples of the sampled features at \cfgB and \ContrastPassCountC at \cfgC, and in at least one setting on \ContrastAnyCount of the \Resamples resamples.

Resampling the images within each half repeats some images and drops others, so the two resampled halves split more pairs than the own halves of the scoring images do. They split \ImgSplitMeanA pairs on average at \cfgA against \ExHalfA on the own halves, \ImgSplitMeanB against \ExHalfB at \cfgB and \ImgSplitMeanC against \ExHalfC at \cfgC. So at \cfgA the split half agreement of the output test falls on average from \SelfAgreeA to \ImgSelfMeanA, while the cosine rule's agreement falls from \RuleAgreeA to \ImgRuleMeanA, and the difference falls with them. Resampling the scoring images within each half \Resamples times keeps the difference less the required difference above zero on \ImageContrastA resamples at \cfgA, \ImageContrastB at \cfgB and \ImageContrastC at \cfgC.

\paragraph{Repeated incorrect predictions.}
A rule's incorrect prediction on a pair repeats where it occurs on both halves of the scoring images, and repeated incorrect predictions show what the output test adds to that rule. The cosine rule's repeated incorrect predictions outnumber the pairs whose halves disagree in every setting, while the layer rule's are fewer, on the scoring images and on the fresh images alike\later. The layer rule's repeated incorrect predictions exceed the pairs whose halves disagree with \ExSiteLeaveA of the \Seeds SAEs of \cfgA left out and with none of those of \cfgB or \cfgC, while the cosine rule's exceed them with every SAE left out in every setting. With the layer error computed on all \ImagesScore scoring images, the layer rule predicts incorrectly on \SameDisagreeA, \SameDisagreeB and \SameDisagreeC pairs in the three settings. Against the halves it agrees at \SameHalvesA, \SameHalvesB and \SameHalvesC, with intervals from \SameHalvesLowA to \SameHalvesHighA, \SameHalvesLowB to \SameHalvesHighB and \SameHalvesLowC to \SameHalvesHighC, beside the split half agreement of the output test of \SelfAgreeA, \SelfAgreeB and \SelfAgreeC. The layer rule's prediction changes between the selection and the scoring images on \MovesSelA, \MovesSelB and \MovesSelC pairs and between the two halves on \MovesHalfA, \MovesHalfB and \MovesHalfC, and the layer error on all scoring images has a rank correlation with the output error of \SameRhoA, \SameRhoB and \SameRhoC.

\paragraph{Combining the two rules.}
We combine the two rules\later: we accept a prediction where the cosine rule and the layer rule at \Certify agree, and we run the output test on the pairs where they disagree. On the scoring images the output test then runs on \ScreenSentA to \ScreenSentC of each setting's \UnitsTrained pairs. Among the predictions we accept, \ScreenWrongB to \ScreenWrongA are incorrect, against \SiteDisagreeB to \SiteDisagreeA for the layer rule alone, and most of them are passing pairs that both rules predict to fail, as the last rows of \cref{tab:supp-halves} count.

\paragraph{Fresh images.}
The fresh images check whether the second comparison and the two rules behave the same on images that no SAE, rule or fit has seen. \Cref{tab:supp-fresh} repeats the second comparison on the fresh images\later. Each resample of the fresh images takes the required difference at the cosine rule's agreement in that resample, and the last rows of the table set the layer rule's incorrect predictions on the fresh images beside the cosine rule's.
\begin{table}[!htb]
  \caption{\capFresh}
  \label{tab:supp-fresh}
  \centering
  \fontsize{9}{10.5}\selectfont
\renewcommand{\estimate}[2]{\ensuremath{\textnormal{#1}_{\publicspread#2\relax}}}
  \setlength{\tabcolsep}{1pt}
  \begin{tabular}{@{}lccc@{}}
    \toprule
    & \headA & \headB & \headC \\
    \midrule
    Pairs the output test passes on the scoring images & \CertA & \CertB & \CertC \\
    Pairs the output test passes on the fresh images & \FreshPassA & \FreshPassB & \FreshPassC \\
    Pairs with one outcome on all four halves & \FreshFourA & \FreshFourB & \FreshFourC \\
    Their share of the matched pairs & \FreshFourShareA & \FreshFourShareB & \FreshFourShareC \\
    \midrule
    \rowcolor{methodtint} Split half agreement of the output test $\uparrow$ & \MarkFreshAgreementAFreshSelfA & \MarkFreshAgreementBFreshSelfB & \MarkFreshAgreementCFreshSelfC \\
    Agreement of the cosine rule $\uparrow$ & \MarkFreshAgreementAFreshRuleA & \MarkFreshAgreementBFreshRuleB & \MarkFreshAgreementCFreshRuleC \\
    Difference of the two & \inestimate{\FreshDiffA}{\FreshDiffLowA to \FreshDiffHighA} & \inestimate{\FreshDiffB}{\FreshDiffLowB to \FreshDiffHighB} & \inestimate{\FreshDiffC}{\FreshDiffLowC to \FreshDiffHighC} \\
    Fifth percentile of the difference & \FreshBoundA & \FreshBoundB & \FreshBoundC \\
    Required difference & \FreshFnMarginA & \FreshFnMarginB & \FreshFnMarginC \\
    Resamples above the required difference & \FreshImageAboveA & \FreshImageAboveB & \FreshImageAboveC \\
    \midrule
    Pairs whose fresh halves disagree & \FreshSplitA & \FreshSplitB & \FreshSplitC \\
    Repeated incorrect predictions, cosine rule & \FreshCosRepeatA & \FreshCosRepeatB & \FreshCosRepeatC \\
    Repeated incorrect predictions, layer rule & \FreshSiteRepeatA & \FreshSiteRepeatB & \FreshSiteRepeatC \\
    Pairs into random decoder SAEs that pass & \FreshFrozenA & \FreshFrozenB & \FreshFrozenC \\
    \midrule
    Incorrect predictions of the cosine rule $\downarrow$ & \MarkFreshErrorsAFreshCosErrA & \MarkFreshErrorsBFreshCosErrB & \MarkFreshErrorsCFreshCosErrC \\
    Incorrect predictions of the layer rule $\downarrow$ & \MarkFreshErrorsAFreshSiteErrA & \MarkFreshErrorsBFreshSiteErrB & \MarkFreshErrorsCFreshSiteErrC \\
    Incorrect predictions of the layer rule at \NinthFitJoint & \MarkFreshErrorsAFittedCountA & \MarkFreshErrorsBFittedCountB & \MarkFreshErrorsCFittedCountC \\
    \bottomrule
  \end{tabular}
  \tablenote{\noteFresh}
\end{table}

\paragraph{The parts of the layer error.}
We score each part of the layer error as a rule of its own, with the full layer rule as the reference, and \cref{tab:supp-predictors} scores each against the output test on the fresh images\later. With $r$, $\rho$ and $c$ as in \cref{eq:layer}, the rule on the code norm ratio alone scores $(1-r)^{2}$, the rule on the code cosine alone $2(1-\rho)$, and the layer error with the decoder cosine set to one scores $1 + r^{2} - 2r\rho$. Where a matched feature's code is zero on every token of an image set the code cosine is undefined, and we set it to zero, so the code cosine alone scores two and the other three score one. That case arises for no pair of the first sample, of the second sample or of the development pairs, on any of the \EleventhZeroCodeSets combinations of those pairs with an image set that a rule uses. Each rule predicts a pass at or below a threshold fitted on the \DevUnits development pairs by the objective and the tie order of \cref{sec:supp-calibrated}, and for the whole layer error the fit returns the threshold of that section.

With every statistic on the selection images, the layer error with the decoder cosine set to one predicts incorrectly on \NinthErrCodesA pairs at \cfgA against the layer rule's \NinthErrJointA, fewer by a difference whose interval lies below zero. It predicts incorrectly on \NinthErrCodesC at \cfgC against \NinthErrJointC, with an interval that includes zero, and on \NinthErrCodesB at \cfgB against \NinthErrJointB, more by a difference whose interval lies above zero. The code cosine alone predicts incorrectly on \NinthErrOverlapA pairs at \cfgA, and the interval of its difference from the layer rule, from $\NinthMoreLowOverlapA$ to $\NinthMoreHighOverlapA$, includes zero, so the data do not separate the two rules there. The code norm ratio alone predicts incorrectly on \NinthErrMagnitudeC to \NinthErrMagnitudeA pairs. With every statistic recomputed on the fresh images, the interval for the layer error with the decoder cosine set to one includes zero at \cfgA and \cfgC and lies above zero at \cfgB.
\begin{table}[!htb]
  \caption{\capPredictors}
  \label{tab:supp-predictors}
  \centering
  \fontsize{9}{10.5}\selectfont
\renewcommand{\estimate}[2]{\ensuremath{\textnormal{#1}_{\publicspread#2\relax}}}
  \setlength{\tabcolsep}{1pt}
  \begin{tabular}{@{}>{\raggedright\arraybackslash}p{175pt}lccc@{}}
    \toprule
    Rule & Threshold & \headA{} $\downarrow$ & \headB{} $\downarrow$ & \headC{} $\downarrow$ \\
    \midrule
    \multicolumn{5}{@{}l}{\emph{Statistics on the selection images}} \\
    \rowcolor{methodtint} Layer error, $1 + r\squared - 2rc\rho$ & \NinthFitJoint & \NinthErrJointA & \textbf{\NinthErrJointB} & \textbf{\NinthErrJointC} \\
    Layer error with $c$ set to one, $1 + r\squared - 2r\rho$ & \NinthFitCodes & \inestimate{\textbf{\NinthErrCodesA}}{$\NinthMoreLowCodesA$ to $\NinthMoreHighCodesA$} & \inestimate{\NinthErrCodesB}{$\NinthMoreLowCodesB$ to $\NinthMoreHighCodesB$} & \inestimate{\NinthErrCodesC}{$\NinthMoreLowCodesC$ to $\NinthMoreHighCodesC$} \\
    Code cosine alone, $2(1 - \rho)$ & \NinthFitOverlap & \inestimate{\NinthErrOverlapA}{$\NinthMoreLowOverlapA$ to $\NinthMoreHighOverlapA$} & \inestimate{\NinthErrOverlapB}{$\NinthMoreLowOverlapB$ to $\NinthMoreHighOverlapB$} & \inestimate{\NinthErrOverlapC}{$\NinthMoreLowOverlapC$ to $\NinthMoreHighOverlapC$} \\
    Code norm ratio alone, $(1 - r)\squared$ & \NinthFitMagnitude & \inestimate{\NinthErrMagnitudeA}{$\NinthMoreLowMagnitudeA$ to $\NinthMoreHighMagnitudeA$} & \inestimate{\NinthErrMagnitudeB}{$\NinthMoreLowMagnitudeB$ to $\NinthMoreHighMagnitudeB$} & \inestimate{\NinthErrMagnitudeC}{$\NinthMoreLowMagnitudeC$ to $\NinthMoreHighMagnitudeC$} \\
    Cosine rule & \RuleDec and \RuleEnc & \inestimate{\NinthErrCosineA}{$\NinthMoreLowCosineA$ to $\NinthMoreHighCosineA$} & \inestimate{\NinthErrCosineB}{$\NinthMoreLowCosineB$ to $\NinthMoreHighCosineB$} & \inestimate{\NinthErrCosineC}{$\NinthMoreLowCosineC$ to $\NinthMoreHighCosineC$} \\
    \midrule
    \multicolumn{5}{@{}l}{\emph{Statistics on the fresh images}} \\
    \rowcolor{methodtint} Layer error, $1 + r\squared - 2rc\rho$ & \NinthFitJoint & \NinthFreshErrJointA & \textbf{\NinthFreshErrJointB} & \textbf{\NinthFreshErrJointC} \\
    Layer error with $c$ set to one, $1 + r\squared - 2r\rho$ & \NinthFitCodes & \inestimate{\textbf{\NinthFreshErrCodesA}}{$\NinthFreshMoreLowCodesA$ to $\NinthFreshMoreHighCodesA$} & \inestimate{\NinthFreshErrCodesB}{$\NinthFreshMoreLowCodesB$ to $\NinthFreshMoreHighCodesB$} & \inestimate{\NinthFreshErrCodesC}{$\NinthFreshMoreLowCodesC$ to $\NinthFreshMoreHighCodesC$} \\
    Code cosine alone, $2(1 - \rho)$ & \NinthFitOverlap & \inestimate{\NinthFreshErrOverlapA}{$\NinthFreshMoreLowOverlapA$ to $\NinthFreshMoreHighOverlapA$} & \inestimate{\NinthFreshErrOverlapB}{$\NinthFreshMoreLowOverlapB$ to $\NinthFreshMoreHighOverlapB$} & \inestimate{\NinthFreshErrOverlapC}{$\NinthFreshMoreLowOverlapC$ to $\NinthFreshMoreHighOverlapC$} \\
    Code norm ratio alone, $(1 - r)\squared$ & \NinthFitMagnitude & \inestimate{\NinthFreshErrMagnitudeA}{$\NinthFreshMoreLowMagnitudeA$ to $\NinthFreshMoreHighMagnitudeA$} & \inestimate{\NinthFreshErrMagnitudeB}{$\NinthFreshMoreLowMagnitudeB$ to $\NinthFreshMoreHighMagnitudeB$} & \inestimate{\NinthFreshErrMagnitudeC}{$\NinthFreshMoreLowMagnitudeC$ to $\NinthFreshMoreHighMagnitudeC$} \\
    Cosine rule & \RuleDec and \RuleEnc & \inestimate{\NinthFreshErrCosineA}{$\NinthFreshMoreLowCosineA$ to $\NinthFreshMoreHighCosineA$} & \inestimate{\NinthFreshErrCosineB}{$\NinthFreshMoreLowCosineB$ to $\NinthFreshMoreHighCosineB$} & \inestimate{\NinthFreshErrCosineC}{$\NinthFreshMoreLowCosineC$ to $\NinthFreshMoreHighCosineC$} \\
    \bottomrule
  \end{tabular}
  \tablenote{\notePredictors}
\end{table}

\paragraph{Eight pairs on a third image set.}
The scoring and fresh images selected eight pairs, on each of which the layer rule and the output test disagree in one direction on all four halves of the scoring and fresh images, with the layer error computed on the same halves as the output error. With the layer rule's prediction fixed on the selection images instead, \NinthStableSelection pairs disagree so. \Cref{tab:supp-six} gives the output error and the layer error of the eight pairs on a third image set, with both computed on the same halves\later. The third image set excludes every image used for training, selection, scoring, development or the fresh images. A disagreement persists where both the output error and the layer error keep their stored opposite sides of \Certify on at least \SeventhFamilyShare of \Resamples paired image resamples of the whole third image set and of each of its halves, a tie counting against. The share divides an allowance of one twentieth over the eight pairs, and the intervals are pointwise, with the trained SAEs and the eight pairs fixed.

One disagreement persists, at \cfgC, for feature \EightSevenFeature of the SAE of seed \EightSevenSource matched into the SAE of seed \EightSevenTarget. The output test fails the pair at an output error of \SeventhSevenThirdWholeOutput, and the layer rule passes it at a layer error of \SeventhSevenThirdWholeSite.

\Cref{tab:supp-continuation} follows the same \EightCount pairs on the third image set through the encoder's later layers\later. These passes ran on one \DeviceName under the versions of the earlier passes and reproduce each pair's stored output error on the third image set within a relative tolerance of \ReplayTolerance. A pass on a processor under later versions did not meet that tolerance at \cfgA, whose edits cross six later layers.

A recomputation checks the stored values in every setting\later. It takes the first fifty images of each scoring half, \PlantedImages in all, through the image preprocessing, the encoder up to the edited layer, the ablations and the later layers used for the fresh images, and it reproduces their stored activations at the edited layer bit for bit. Its squared differences and first effects on every image meet the relative tolerance of \ReplayTolerance, with a lower limit of \ReplayFloor times the unedited output's squared size on the denominator. The largest absolute difference in a finite output error of one image is \ReplayMaximumA at \cfgA, \ReplayMaximumB at \cfgB and \ReplayMaximumC at \cfgC, and in the output error pooled over these images it is \ReplayWholeA, \ReplayWholeB and \ReplayWholeC. On \ReplayUndefinedA, \ReplayUndefinedB and \ReplayUndefinedC combinations of an image and a pair the first effect is zero, so the ratio is undefined, and the comparisons of the squared differences and of the first effects still cover them.

\begin{table}[p]
  \caption{\capEight}
  \label{tab:supp-six}
  \centering
  \fontsize{9}{10.5}\selectfont
  \setlength{\tabcolsep}{2pt}
  \begin{tabular}{@{}llcccccc@{}}
    \toprule
Pair and set & Part & \shortstack{Output\\error $\epsilon$} & \shortstack{Layer\\error $d$} & \shortstack{$\epsilon$ less\\\Certify} & \shortstack{$d$ less\\\Certify} & \shortstack{Share\\opposite} & \shortstack{Share in stored\\direction} \\
    \midrule
    \multicolumn{8}{l}{\cfgA, seed \EightOneSource feature \EightOneFeature into seed \EightOneTarget, passes, does not persist} \\
    Selection & whole & & $\EightOneSiteSelection$ & & $\EightOneSiteOffsetSelection$ & & \\
    S & whole & $\SeventhOneScoringWholeOutput_{\text{\fontsize{9}{10}\selectfont [\SeventhOneScoringWholeOutputLow,\SeventhOneScoringWholeOutputHigh]}}$ & $\SeventhOneScoringWholeSite_{\text{\fontsize{9}{10}\selectfont [\SeventhOneScoringWholeSiteLow,\SeventhOneScoringWholeSiteHigh]}}$ & $\SeventhOneScoringWholeOutputDistance$ & $\SeventhOneScoringWholeSiteDistance$ & $\SeventhOneScoringWholeOpposite$ & $\SeventhOneScoringWholeFixed$ \\
    S & first & $\SeventhOneScoringEarlierOutput_{\text{\fontsize{9}{10}\selectfont [\SeventhOneScoringEarlierOutputLow,\SeventhOneScoringEarlierOutputHigh]}}$ & $\SeventhOneScoringEarlierSite_{\text{\fontsize{9}{10}\selectfont [\SeventhOneScoringEarlierSiteLow,\SeventhOneScoringEarlierSiteHigh]}}$ & $\SeventhOneScoringEarlierOutputDistance$ & $\SeventhOneScoringEarlierSiteDistance$ & $\SeventhOneScoringEarlierOpposite$ & $\SeventhOneScoringEarlierFixed$ \\
    S & last & $\SeventhOneScoringLaterOutput_{\text{\fontsize{9}{10}\selectfont [\SeventhOneScoringLaterOutputLow,\SeventhOneScoringLaterOutputHigh]}}$ & $\SeventhOneScoringLaterSite_{\text{\fontsize{9}{10}\selectfont [\SeventhOneScoringLaterSiteLow,\SeventhOneScoringLaterSiteHigh]}}$ & $\SeventhOneScoringLaterOutputDistance$ & $\SeventhOneScoringLaterSiteDistance$ & $\SeventhOneScoringLaterOpposite$ & $\SeventhOneScoringLaterFixed$ \\
    F & whole & $\SeventhOneFreshWholeOutput_{\text{\fontsize{9}{10}\selectfont [\SeventhOneFreshWholeOutputLow,\SeventhOneFreshWholeOutputHigh]}}$ & $\SeventhOneFreshWholeSite_{\text{\fontsize{9}{10}\selectfont [\SeventhOneFreshWholeSiteLow,\SeventhOneFreshWholeSiteHigh]}}$ & $\SeventhOneFreshWholeOutputDistance$ & $\SeventhOneFreshWholeSiteDistance$ & $\SeventhOneFreshWholeOpposite$ & $\SeventhOneFreshWholeFixed$ \\
    F & first & $\SeventhOneFreshEarlierOutput_{\text{\fontsize{9}{10}\selectfont [\SeventhOneFreshEarlierOutputLow,\SeventhOneFreshEarlierOutputHigh]}}$ & $\SeventhOneFreshEarlierSite_{\text{\fontsize{9}{10}\selectfont [\SeventhOneFreshEarlierSiteLow,\SeventhOneFreshEarlierSiteHigh]}}$ & $\SeventhOneFreshEarlierOutputDistance$ & $\SeventhOneFreshEarlierSiteDistance$ & $\SeventhOneFreshEarlierOpposite$ & $\SeventhOneFreshEarlierFixed$ \\
    F & last & $\SeventhOneFreshLaterOutput_{\text{\fontsize{9}{10}\selectfont [\SeventhOneFreshLaterOutputLow,\SeventhOneFreshLaterOutputHigh]}}$ & $\SeventhOneFreshLaterSite_{\text{\fontsize{9}{10}\selectfont [\SeventhOneFreshLaterSiteLow,\SeventhOneFreshLaterSiteHigh]}}$ & $\SeventhOneFreshLaterOutputDistance$ & $\SeventhOneFreshLaterSiteDistance$ & $\SeventhOneFreshLaterOpposite$ & $\SeventhOneFreshLaterFixed$ \\
    U & whole & $\SeventhOneThirdWholeOutput_{\text{\fontsize{9}{10}\selectfont [\SeventhOneThirdWholeOutputLow,\SeventhOneThirdWholeOutputHigh]}}$ & $\SeventhOneThirdWholeSite_{\text{\fontsize{9}{10}\selectfont [\SeventhOneThirdWholeSiteLow,\SeventhOneThirdWholeSiteHigh]}}$ & $\SeventhOneThirdWholeOutputDistance$ & $\SeventhOneThirdWholeSiteDistance$ & $\SeventhOneThirdWholeOpposite$ & $\SeventhOneThirdWholeFixed$ \\
    U & first & $\SeventhOneThirdEarlierOutput_{\text{\fontsize{9}{10}\selectfont [\SeventhOneThirdEarlierOutputLow,\SeventhOneThirdEarlierOutputHigh]}}$ & $\SeventhOneThirdEarlierSite_{\text{\fontsize{9}{10}\selectfont [\SeventhOneThirdEarlierSiteLow,\SeventhOneThirdEarlierSiteHigh]}}$ & $\SeventhOneThirdEarlierOutputDistance$ & $\SeventhOneThirdEarlierSiteDistance$ & $\SeventhOneThirdEarlierOpposite$ & $\SeventhOneThirdEarlierFixed$ \\
    U & last & $\SeventhOneThirdLaterOutput_{\text{\fontsize{9}{10}\selectfont [\SeventhOneThirdLaterOutputLow,\SeventhOneThirdLaterOutputHigh]}}$ & $\SeventhOneThirdLaterSite_{\text{\fontsize{9}{10}\selectfont [\SeventhOneThirdLaterSiteLow,\SeventhOneThirdLaterSiteHigh]}}$ & $\SeventhOneThirdLaterOutputDistance$ & $\SeventhOneThirdLaterSiteDistance$ & $\SeventhOneThirdLaterOpposite$ & $\SeventhOneThirdLaterFixed$ \\
    \midrule
    \multicolumn{8}{l}{\cfgA, seed \EightTwoSource feature \EightTwoFeature into seed \EightTwoTarget, passes, does not persist} \\
    Selection & whole & & $\EightTwoSiteSelection$ & & $\EightTwoSiteOffsetSelection$ & & \\
    S & whole & $\SeventhTwoScoringWholeOutput_{\text{\fontsize{9}{10}\selectfont [\SeventhTwoScoringWholeOutputLow,\SeventhTwoScoringWholeOutputHigh]}}$ & $\SeventhTwoScoringWholeSite_{\text{\fontsize{9}{10}\selectfont [\SeventhTwoScoringWholeSiteLow,\SeventhTwoScoringWholeSiteHigh]}}$ & $\SeventhTwoScoringWholeOutputDistance$ & $\SeventhTwoScoringWholeSiteDistance$ & $\SeventhTwoScoringWholeOpposite$ & $\SeventhTwoScoringWholeFixed$ \\
    S & first & $\SeventhTwoScoringEarlierOutput_{\text{\fontsize{9}{10}\selectfont [\SeventhTwoScoringEarlierOutputLow,\SeventhTwoScoringEarlierOutputHigh]}}$ & $\SeventhTwoScoringEarlierSite_{\text{\fontsize{9}{10}\selectfont [\SeventhTwoScoringEarlierSiteLow,\SeventhTwoScoringEarlierSiteHigh]}}$ & $\SeventhTwoScoringEarlierOutputDistance$ & $\SeventhTwoScoringEarlierSiteDistance$ & $\SeventhTwoScoringEarlierOpposite$ & $\SeventhTwoScoringEarlierFixed$ \\
    S & last & $\SeventhTwoScoringLaterOutput_{\text{\fontsize{9}{10}\selectfont [\SeventhTwoScoringLaterOutputLow,\SeventhTwoScoringLaterOutputHigh]}}$ & $\SeventhTwoScoringLaterSite_{\text{\fontsize{9}{10}\selectfont [\SeventhTwoScoringLaterSiteLow,\SeventhTwoScoringLaterSiteHigh]}}$ & $\SeventhTwoScoringLaterOutputDistance$ & $\SeventhTwoScoringLaterSiteDistance$ & $\SeventhTwoScoringLaterOpposite$ & $\SeventhTwoScoringLaterFixed$ \\
    F & whole & $\SeventhTwoFreshWholeOutput_{\text{\fontsize{9}{10}\selectfont [\SeventhTwoFreshWholeOutputLow,\SeventhTwoFreshWholeOutputHigh]}}$ & $\SeventhTwoFreshWholeSite_{\text{\fontsize{9}{10}\selectfont [\SeventhTwoFreshWholeSiteLow,\SeventhTwoFreshWholeSiteHigh]}}$ & $\SeventhTwoFreshWholeOutputDistance$ & $\SeventhTwoFreshWholeSiteDistance$ & $\SeventhTwoFreshWholeOpposite$ & $\SeventhTwoFreshWholeFixed$ \\
    F & first & $\SeventhTwoFreshEarlierOutput_{\text{\fontsize{9}{10}\selectfont [\SeventhTwoFreshEarlierOutputLow,\SeventhTwoFreshEarlierOutputHigh]}}$ & $\SeventhTwoFreshEarlierSite_{\text{\fontsize{9}{10}\selectfont [\SeventhTwoFreshEarlierSiteLow,\SeventhTwoFreshEarlierSiteHigh]}}$ & $\SeventhTwoFreshEarlierOutputDistance$ & $\SeventhTwoFreshEarlierSiteDistance$ & $\SeventhTwoFreshEarlierOpposite$ & $\SeventhTwoFreshEarlierFixed$ \\
    F & last & $\SeventhTwoFreshLaterOutput_{\text{\fontsize{9}{10}\selectfont [\SeventhTwoFreshLaterOutputLow,\SeventhTwoFreshLaterOutputHigh]}}$ & $\SeventhTwoFreshLaterSite_{\text{\fontsize{9}{10}\selectfont [\SeventhTwoFreshLaterSiteLow,\SeventhTwoFreshLaterSiteHigh]}}$ & $\SeventhTwoFreshLaterOutputDistance$ & $\SeventhTwoFreshLaterSiteDistance$ & $\SeventhTwoFreshLaterOpposite$ & $\SeventhTwoFreshLaterFixed$ \\
    U & whole & $\SeventhTwoThirdWholeOutput_{\text{\fontsize{9}{10}\selectfont [\SeventhTwoThirdWholeOutputLow,\SeventhTwoThirdWholeOutputHigh]}}$ & $\SeventhTwoThirdWholeSite_{\text{\fontsize{9}{10}\selectfont [\SeventhTwoThirdWholeSiteLow,\SeventhTwoThirdWholeSiteHigh]}}$ & $\SeventhTwoThirdWholeOutputDistance$ & $\SeventhTwoThirdWholeSiteDistance$ & $\SeventhTwoThirdWholeOpposite$ & $\SeventhTwoThirdWholeFixed$ \\
    U & first & $\SeventhTwoThirdEarlierOutput_{\text{\fontsize{9}{10}\selectfont [\SeventhTwoThirdEarlierOutputLow,\SeventhTwoThirdEarlierOutputHigh]}}$ & $\SeventhTwoThirdEarlierSite_{\text{\fontsize{9}{10}\selectfont [\SeventhTwoThirdEarlierSiteLow,\SeventhTwoThirdEarlierSiteHigh]}}$ & $\SeventhTwoThirdEarlierOutputDistance$ & $\SeventhTwoThirdEarlierSiteDistance$ & $\SeventhTwoThirdEarlierOpposite$ & $\SeventhTwoThirdEarlierFixed$ \\
    U & last & $\SeventhTwoThirdLaterOutput_{\text{\fontsize{9}{10}\selectfont [\SeventhTwoThirdLaterOutputLow,\SeventhTwoThirdLaterOutputHigh]}}$ & $\SeventhTwoThirdLaterSite_{\text{\fontsize{9}{10}\selectfont [\SeventhTwoThirdLaterSiteLow,\SeventhTwoThirdLaterSiteHigh]}}$ & $\SeventhTwoThirdLaterOutputDistance$ & $\SeventhTwoThirdLaterSiteDistance$ & $\SeventhTwoThirdLaterOpposite$ & $\SeventhTwoThirdLaterFixed$ \\
    \midrule
    \multicolumn{8}{l}{\cfgA, seed \EightThreeSource feature \EightThreeFeature into seed \EightThreeTarget, passes, does not persist} \\
    Selection & whole & & $\EightThreeSiteSelection$ & & $\EightThreeSiteOffsetSelection$ & & \\
    S & whole & $\SeventhThreeScoringWholeOutput_{\text{\fontsize{9}{10}\selectfont [\SeventhThreeScoringWholeOutputLow,\SeventhThreeScoringWholeOutputHigh]}}$ & $\SeventhThreeScoringWholeSite_{\text{\fontsize{9}{10}\selectfont [\SeventhThreeScoringWholeSiteLow,\SeventhThreeScoringWholeSiteHigh]}}$ & $\SeventhThreeScoringWholeOutputDistance$ & $\SeventhThreeScoringWholeSiteDistance$ & $\SeventhThreeScoringWholeOpposite$ & $\SeventhThreeScoringWholeFixed$ \\
    S & first & $\SeventhThreeScoringEarlierOutput_{\text{\fontsize{9}{10}\selectfont [\SeventhThreeScoringEarlierOutputLow,\SeventhThreeScoringEarlierOutputHigh]}}$ & $\SeventhThreeScoringEarlierSite_{\text{\fontsize{9}{10}\selectfont [\SeventhThreeScoringEarlierSiteLow,\SeventhThreeScoringEarlierSiteHigh]}}$ & $\SeventhThreeScoringEarlierOutputDistance$ & $\SeventhThreeScoringEarlierSiteDistance$ & $\SeventhThreeScoringEarlierOpposite$ & $\SeventhThreeScoringEarlierFixed$ \\
    S & last & $\SeventhThreeScoringLaterOutput_{\text{\fontsize{9}{10}\selectfont [\SeventhThreeScoringLaterOutputLow,\SeventhThreeScoringLaterOutputHigh]}}$ & $\SeventhThreeScoringLaterSite_{\text{\fontsize{9}{10}\selectfont [\SeventhThreeScoringLaterSiteLow,\SeventhThreeScoringLaterSiteHigh]}}$ & $\SeventhThreeScoringLaterOutputDistance$ & $\SeventhThreeScoringLaterSiteDistance$ & $\SeventhThreeScoringLaterOpposite$ & $\SeventhThreeScoringLaterFixed$ \\
    F & whole & $\SeventhThreeFreshWholeOutput_{\text{\fontsize{9}{10}\selectfont [\SeventhThreeFreshWholeOutputLow,\SeventhThreeFreshWholeOutputHigh]}}$ & $\SeventhThreeFreshWholeSite_{\text{\fontsize{9}{10}\selectfont [\SeventhThreeFreshWholeSiteLow,\SeventhThreeFreshWholeSiteHigh]}}$ & $\SeventhThreeFreshWholeOutputDistance$ & $\SeventhThreeFreshWholeSiteDistance$ & $\SeventhThreeFreshWholeOpposite$ & $\SeventhThreeFreshWholeFixed$ \\
    F & first & $\SeventhThreeFreshEarlierOutput_{\text{\fontsize{9}{10}\selectfont [\SeventhThreeFreshEarlierOutputLow,\SeventhThreeFreshEarlierOutputHigh]}}$ & $\SeventhThreeFreshEarlierSite_{\text{\fontsize{9}{10}\selectfont [\SeventhThreeFreshEarlierSiteLow,\SeventhThreeFreshEarlierSiteHigh]}}$ & $\SeventhThreeFreshEarlierOutputDistance$ & $\SeventhThreeFreshEarlierSiteDistance$ & $\SeventhThreeFreshEarlierOpposite$ & $\SeventhThreeFreshEarlierFixed$ \\
    F & last & $\SeventhThreeFreshLaterOutput_{\text{\fontsize{9}{10}\selectfont [\SeventhThreeFreshLaterOutputLow,\SeventhThreeFreshLaterOutputHigh]}}$ & $\SeventhThreeFreshLaterSite_{\text{\fontsize{9}{10}\selectfont [\SeventhThreeFreshLaterSiteLow,\SeventhThreeFreshLaterSiteHigh]}}$ & $\SeventhThreeFreshLaterOutputDistance$ & $\SeventhThreeFreshLaterSiteDistance$ & $\SeventhThreeFreshLaterOpposite$ & $\SeventhThreeFreshLaterFixed$ \\
    U & whole & $\SeventhThreeThirdWholeOutput_{\text{\fontsize{9}{10}\selectfont [\SeventhThreeThirdWholeOutputLow,\SeventhThreeThirdWholeOutputHigh]}}$ & $\SeventhThreeThirdWholeSite_{\text{\fontsize{9}{10}\selectfont [\SeventhThreeThirdWholeSiteLow,\SeventhThreeThirdWholeSiteHigh]}}$ & $\SeventhThreeThirdWholeOutputDistance$ & $\SeventhThreeThirdWholeSiteDistance$ & $\SeventhThreeThirdWholeOpposite$ & $\SeventhThreeThirdWholeFixed$ \\
    U & first & $\SeventhThreeThirdEarlierOutput_{\text{\fontsize{9}{10}\selectfont [\SeventhThreeThirdEarlierOutputLow,\SeventhThreeThirdEarlierOutputHigh]}}$ & $\SeventhThreeThirdEarlierSite_{\text{\fontsize{9}{10}\selectfont [\SeventhThreeThirdEarlierSiteLow,\SeventhThreeThirdEarlierSiteHigh]}}$ & $\SeventhThreeThirdEarlierOutputDistance$ & $\SeventhThreeThirdEarlierSiteDistance$ & $\SeventhThreeThirdEarlierOpposite$ & $\SeventhThreeThirdEarlierFixed$ \\
    U & last & $\SeventhThreeThirdLaterOutput_{\text{\fontsize{9}{10}\selectfont [\SeventhThreeThirdLaterOutputLow,\SeventhThreeThirdLaterOutputHigh]}}$ & $\SeventhThreeThirdLaterSite_{\text{\fontsize{9}{10}\selectfont [\SeventhThreeThirdLaterSiteLow,\SeventhThreeThirdLaterSiteHigh]}}$ & $\SeventhThreeThirdLaterOutputDistance$ & $\SeventhThreeThirdLaterSiteDistance$ & $\SeventhThreeThirdLaterOpposite$ & $\SeventhThreeThirdLaterFixed$ \\
    \midrule
    \multicolumn{8}{l}{\cfgA, seed \EightFourSource feature \EightFourFeature into seed \EightFourTarget, passes, does not persist} \\
    Selection & whole & & $\EightFourSiteSelection$ & & $\EightFourSiteOffsetSelection$ & & \\
    S & whole & $\SeventhFourScoringWholeOutput_{\text{\fontsize{9}{10}\selectfont [\SeventhFourScoringWholeOutputLow,\SeventhFourScoringWholeOutputHigh]}}$ & $\SeventhFourScoringWholeSite_{\text{\fontsize{9}{10}\selectfont [\SeventhFourScoringWholeSiteLow,\SeventhFourScoringWholeSiteHigh]}}$ & $\SeventhFourScoringWholeOutputDistance$ & $\SeventhFourScoringWholeSiteDistance$ & $\SeventhFourScoringWholeOpposite$ & $\SeventhFourScoringWholeFixed$ \\
    S & first & $\SeventhFourScoringEarlierOutput_{\text{\fontsize{9}{10}\selectfont [\SeventhFourScoringEarlierOutputLow,\SeventhFourScoringEarlierOutputHigh]}}$ & $\SeventhFourScoringEarlierSite_{\text{\fontsize{9}{10}\selectfont [\SeventhFourScoringEarlierSiteLow,\SeventhFourScoringEarlierSiteHigh]}}$ & $\SeventhFourScoringEarlierOutputDistance$ & $\SeventhFourScoringEarlierSiteDistance$ & $\SeventhFourScoringEarlierOpposite$ & $\SeventhFourScoringEarlierFixed$ \\
    S & last & $\SeventhFourScoringLaterOutput_{\text{\fontsize{9}{10}\selectfont [\SeventhFourScoringLaterOutputLow,\SeventhFourScoringLaterOutputHigh]}}$ & $\SeventhFourScoringLaterSite_{\text{\fontsize{9}{10}\selectfont [\SeventhFourScoringLaterSiteLow,\SeventhFourScoringLaterSiteHigh]}}$ & $\SeventhFourScoringLaterOutputDistance$ & $\SeventhFourScoringLaterSiteDistance$ & $\SeventhFourScoringLaterOpposite$ & $\SeventhFourScoringLaterFixed$ \\
    F & whole & $\SeventhFourFreshWholeOutput_{\text{\fontsize{9}{10}\selectfont [\SeventhFourFreshWholeOutputLow,\SeventhFourFreshWholeOutputHigh]}}$ & $\SeventhFourFreshWholeSite_{\text{\fontsize{9}{10}\selectfont [\SeventhFourFreshWholeSiteLow,\SeventhFourFreshWholeSiteHigh]}}$ & $\SeventhFourFreshWholeOutputDistance$ & $\SeventhFourFreshWholeSiteDistance$ & $\SeventhFourFreshWholeOpposite$ & $\SeventhFourFreshWholeFixed$ \\
    F & first & $\SeventhFourFreshEarlierOutput_{\text{\fontsize{9}{10}\selectfont [\SeventhFourFreshEarlierOutputLow,\SeventhFourFreshEarlierOutputHigh]}}$ & $\SeventhFourFreshEarlierSite_{\text{\fontsize{9}{10}\selectfont [\SeventhFourFreshEarlierSiteLow,\SeventhFourFreshEarlierSiteHigh]}}$ & $\SeventhFourFreshEarlierOutputDistance$ & $\SeventhFourFreshEarlierSiteDistance$ & $\SeventhFourFreshEarlierOpposite$ & $\SeventhFourFreshEarlierFixed$ \\
    F & last & $\SeventhFourFreshLaterOutput_{\text{\fontsize{9}{10}\selectfont [\SeventhFourFreshLaterOutputLow,\SeventhFourFreshLaterOutputHigh]}}$ & $\SeventhFourFreshLaterSite_{\text{\fontsize{9}{10}\selectfont [\SeventhFourFreshLaterSiteLow,\SeventhFourFreshLaterSiteHigh]}}$ & $\SeventhFourFreshLaterOutputDistance$ & $\SeventhFourFreshLaterSiteDistance$ & $\SeventhFourFreshLaterOpposite$ & $\SeventhFourFreshLaterFixed$ \\
    U & whole & $\SeventhFourThirdWholeOutput_{\text{\fontsize{9}{10}\selectfont [\SeventhFourThirdWholeOutputLow,\SeventhFourThirdWholeOutputHigh]}}$ & $\SeventhFourThirdWholeSite_{\text{\fontsize{9}{10}\selectfont [\SeventhFourThirdWholeSiteLow,\SeventhFourThirdWholeSiteHigh]}}$ & $\SeventhFourThirdWholeOutputDistance$ & $\SeventhFourThirdWholeSiteDistance$ & $\SeventhFourThirdWholeOpposite$ & $\SeventhFourThirdWholeFixed$ \\
    U & first & $\SeventhFourThirdEarlierOutput_{\text{\fontsize{9}{10}\selectfont [\SeventhFourThirdEarlierOutputLow,\SeventhFourThirdEarlierOutputHigh]}}$ & $\SeventhFourThirdEarlierSite_{\text{\fontsize{9}{10}\selectfont [\SeventhFourThirdEarlierSiteLow,\SeventhFourThirdEarlierSiteHigh]}}$ & $\SeventhFourThirdEarlierOutputDistance$ & $\SeventhFourThirdEarlierSiteDistance$ & $\SeventhFourThirdEarlierOpposite$ & $\SeventhFourThirdEarlierFixed$ \\
    U & last & $\SeventhFourThirdLaterOutput_{\text{\fontsize{9}{10}\selectfont [\SeventhFourThirdLaterOutputLow,\SeventhFourThirdLaterOutputHigh]}}$ & $\SeventhFourThirdLaterSite_{\text{\fontsize{9}{10}\selectfont [\SeventhFourThirdLaterSiteLow,\SeventhFourThirdLaterSiteHigh]}}$ & $\SeventhFourThirdLaterOutputDistance$ & $\SeventhFourThirdLaterSiteDistance$ & $\SeventhFourThirdLaterOpposite$ & $\SeventhFourThirdLaterFixed$ \\
    \bottomrule
  \end{tabular}
  \tablenote{\noteEight{} 
Each signed difference subtracts the bound \Certify from $\epsilon$ or $d$.}
\end{table}
\begin{table}[p]
\ContinuedFloat
  \caption{\capEightContinued}
  \centering
  \fontsize{9}{10.5}\selectfont
  \setlength{\tabcolsep}{2pt}
  \begin{tabular}{@{}llcccccc@{}}
    \toprule
Pair and set & Part & \shortstack{Output\\error $\epsilon$} & \shortstack{Layer\\error $d$} & \shortstack{$\epsilon$ less\\\Certify} & \shortstack{$d$ less\\\Certify} & \shortstack{Share\\opposite} & \shortstack{Share in stored\\direction} \\
    \midrule
    \multicolumn{8}{l}{\cfgA, seed \EightFiveSource feature \EightFiveFeature into seed \EightFiveTarget, passes, does not persist} \\
    Selection & whole & & $\EightFiveSiteSelection$ & & $\EightFiveSiteOffsetSelection$ & & \\
    S & whole & $\SeventhFiveScoringWholeOutput_{\text{\fontsize{9}{10}\selectfont [\SeventhFiveScoringWholeOutputLow,\SeventhFiveScoringWholeOutputHigh]}}$ & $\SeventhFiveScoringWholeSite_{\text{\fontsize{9}{10}\selectfont [\SeventhFiveScoringWholeSiteLow,\SeventhFiveScoringWholeSiteHigh]}}$ & $\SeventhFiveScoringWholeOutputDistance$ & $\SeventhFiveScoringWholeSiteDistance$ & $\SeventhFiveScoringWholeOpposite$ & $\SeventhFiveScoringWholeFixed$ \\
    S & first & $\SeventhFiveScoringEarlierOutput_{\text{\fontsize{9}{10}\selectfont [\SeventhFiveScoringEarlierOutputLow,\SeventhFiveScoringEarlierOutputHigh]}}$ & $\SeventhFiveScoringEarlierSite_{\text{\fontsize{9}{10}\selectfont [\SeventhFiveScoringEarlierSiteLow,\SeventhFiveScoringEarlierSiteHigh]}}$ & $\SeventhFiveScoringEarlierOutputDistance$ & $\SeventhFiveScoringEarlierSiteDistance$ & $\SeventhFiveScoringEarlierOpposite$ & $\SeventhFiveScoringEarlierFixed$ \\
    S & last & $\SeventhFiveScoringLaterOutput_{\text{\fontsize{9}{10}\selectfont [\SeventhFiveScoringLaterOutputLow,\SeventhFiveScoringLaterOutputHigh]}}$ & $\SeventhFiveScoringLaterSite_{\text{\fontsize{9}{10}\selectfont [\SeventhFiveScoringLaterSiteLow,\SeventhFiveScoringLaterSiteHigh]}}$ & $\SeventhFiveScoringLaterOutputDistance$ & $\SeventhFiveScoringLaterSiteDistance$ & $\SeventhFiveScoringLaterOpposite$ & $\SeventhFiveScoringLaterFixed$ \\
    F & whole & $\SeventhFiveFreshWholeOutput_{\text{\fontsize{9}{10}\selectfont [\SeventhFiveFreshWholeOutputLow,\SeventhFiveFreshWholeOutputHigh]}}$ & $\SeventhFiveFreshWholeSite_{\text{\fontsize{9}{10}\selectfont [\SeventhFiveFreshWholeSiteLow,\SeventhFiveFreshWholeSiteHigh]}}$ & $\SeventhFiveFreshWholeOutputDistance$ & $\SeventhFiveFreshWholeSiteDistance$ & $\SeventhFiveFreshWholeOpposite$ & $\SeventhFiveFreshWholeFixed$ \\
    F & first & $\SeventhFiveFreshEarlierOutput_{\text{\fontsize{9}{10}\selectfont [\SeventhFiveFreshEarlierOutputLow,\SeventhFiveFreshEarlierOutputHigh]}}$ & $\SeventhFiveFreshEarlierSite_{\text{\fontsize{9}{10}\selectfont [\SeventhFiveFreshEarlierSiteLow,\SeventhFiveFreshEarlierSiteHigh]}}$ & $\SeventhFiveFreshEarlierOutputDistance$ & $\SeventhFiveFreshEarlierSiteDistance$ & $\SeventhFiveFreshEarlierOpposite$ & $\SeventhFiveFreshEarlierFixed$ \\
    F & last & $\SeventhFiveFreshLaterOutput_{\text{\fontsize{9}{10}\selectfont [\SeventhFiveFreshLaterOutputLow,\SeventhFiveFreshLaterOutputHigh]}}$ & $\SeventhFiveFreshLaterSite_{\text{\fontsize{9}{10}\selectfont [\SeventhFiveFreshLaterSiteLow,\SeventhFiveFreshLaterSiteHigh]}}$ & $\SeventhFiveFreshLaterOutputDistance$ & $\SeventhFiveFreshLaterSiteDistance$ & $\SeventhFiveFreshLaterOpposite$ & $\SeventhFiveFreshLaterFixed$ \\
    U & whole & $\SeventhFiveThirdWholeOutput_{\text{\fontsize{9}{10}\selectfont [\SeventhFiveThirdWholeOutputLow,\SeventhFiveThirdWholeOutputHigh]}}$ & $\SeventhFiveThirdWholeSite_{\text{\fontsize{9}{10}\selectfont [\SeventhFiveThirdWholeSiteLow,\SeventhFiveThirdWholeSiteHigh]}}$ & $\SeventhFiveThirdWholeOutputDistance$ & $\SeventhFiveThirdWholeSiteDistance$ & $\SeventhFiveThirdWholeOpposite$ & $\SeventhFiveThirdWholeFixed$ \\
    U & first & $\SeventhFiveThirdEarlierOutput_{\text{\fontsize{9}{10}\selectfont [\SeventhFiveThirdEarlierOutputLow,\SeventhFiveThirdEarlierOutputHigh]}}$ & $\SeventhFiveThirdEarlierSite_{\text{\fontsize{9}{10}\selectfont [\SeventhFiveThirdEarlierSiteLow,\SeventhFiveThirdEarlierSiteHigh]}}$ & $\SeventhFiveThirdEarlierOutputDistance$ & $\SeventhFiveThirdEarlierSiteDistance$ & $\SeventhFiveThirdEarlierOpposite$ & $\SeventhFiveThirdEarlierFixed$ \\
    U & last & $\SeventhFiveThirdLaterOutput_{\text{\fontsize{9}{10}\selectfont [\SeventhFiveThirdLaterOutputLow,\SeventhFiveThirdLaterOutputHigh]}}$ & $\SeventhFiveThirdLaterSite_{\text{\fontsize{9}{10}\selectfont [\SeventhFiveThirdLaterSiteLow,\SeventhFiveThirdLaterSiteHigh]}}$ & $\SeventhFiveThirdLaterOutputDistance$ & $\SeventhFiveThirdLaterSiteDistance$ & $\SeventhFiveThirdLaterOpposite$ & $\SeventhFiveThirdLaterFixed$ \\
    \midrule
    \multicolumn{8}{l}{\cfgB, seed \EightSixSource feature \EightSixFeature into seed \EightSixTarget, fails, does not persist} \\
    Selection & whole & & $\EightSixSiteSelection$ & & $\EightSixSiteOffsetSelection$ & & \\
    S & whole & $\SeventhSixScoringWholeOutput_{\text{\fontsize{9}{10}\selectfont [\SeventhSixScoringWholeOutputLow,\SeventhSixScoringWholeOutputHigh]}}$ & $\SeventhSixScoringWholeSite_{\text{\fontsize{9}{10}\selectfont [\SeventhSixScoringWholeSiteLow,\SeventhSixScoringWholeSiteHigh]}}$ & $\SeventhSixScoringWholeOutputDistance$ & $\SeventhSixScoringWholeSiteDistance$ & $\SeventhSixScoringWholeOpposite$ & $\SeventhSixScoringWholeFixed$ \\
    S & first & $\SeventhSixScoringEarlierOutput_{\text{\fontsize{9}{10}\selectfont [\SeventhSixScoringEarlierOutputLow,\SeventhSixScoringEarlierOutputHigh]}}$ & $\SeventhSixScoringEarlierSite_{\text{\fontsize{9}{10}\selectfont [\SeventhSixScoringEarlierSiteLow,\SeventhSixScoringEarlierSiteHigh]}}$ & $\SeventhSixScoringEarlierOutputDistance$ & $\SeventhSixScoringEarlierSiteDistance$ & $\SeventhSixScoringEarlierOpposite$ & $\SeventhSixScoringEarlierFixed$ \\
    S & last & $\SeventhSixScoringLaterOutput_{\text{\fontsize{9}{10}\selectfont [\SeventhSixScoringLaterOutputLow,\SeventhSixScoringLaterOutputHigh]}}$ & $\SeventhSixScoringLaterSite_{\text{\fontsize{9}{10}\selectfont [\SeventhSixScoringLaterSiteLow,\SeventhSixScoringLaterSiteHigh]}}$ & $\SeventhSixScoringLaterOutputDistance$ & $\SeventhSixScoringLaterSiteDistance$ & $\SeventhSixScoringLaterOpposite$ & $\SeventhSixScoringLaterFixed$ \\
    F & whole & $\SeventhSixFreshWholeOutput_{\text{\fontsize{9}{10}\selectfont [\SeventhSixFreshWholeOutputLow,\SeventhSixFreshWholeOutputHigh]}}$ & $\SeventhSixFreshWholeSite_{\text{\fontsize{9}{10}\selectfont [\SeventhSixFreshWholeSiteLow,\SeventhSixFreshWholeSiteHigh]}}$ & $\SeventhSixFreshWholeOutputDistance$ & $\SeventhSixFreshWholeSiteDistance$ & $\SeventhSixFreshWholeOpposite$ & $\SeventhSixFreshWholeFixed$ \\
    F & first & $\SeventhSixFreshEarlierOutput_{\text{\fontsize{9}{10}\selectfont [\SeventhSixFreshEarlierOutputLow,\SeventhSixFreshEarlierOutputHigh]}}$ & $\SeventhSixFreshEarlierSite_{\text{\fontsize{9}{10}\selectfont [\SeventhSixFreshEarlierSiteLow,\SeventhSixFreshEarlierSiteHigh]}}$ & $\SeventhSixFreshEarlierOutputDistance$ & $\SeventhSixFreshEarlierSiteDistance$ & $\SeventhSixFreshEarlierOpposite$ & $\SeventhSixFreshEarlierFixed$ \\
    F & last & $\SeventhSixFreshLaterOutput_{\text{\fontsize{9}{10}\selectfont [\SeventhSixFreshLaterOutputLow,\SeventhSixFreshLaterOutputHigh]}}$ & $\SeventhSixFreshLaterSite_{\text{\fontsize{9}{10}\selectfont [\SeventhSixFreshLaterSiteLow,\SeventhSixFreshLaterSiteHigh]}}$ & $\SeventhSixFreshLaterOutputDistance$ & $\SeventhSixFreshLaterSiteDistance$ & $\SeventhSixFreshLaterOpposite$ & $\SeventhSixFreshLaterFixed$ \\
    U & whole & $\SeventhSixThirdWholeOutput_{\text{\fontsize{9}{10}\selectfont [\SeventhSixThirdWholeOutputLow,\SeventhSixThirdWholeOutputHigh]}}$ & $\SeventhSixThirdWholeSite_{\text{\fontsize{9}{10}\selectfont [\SeventhSixThirdWholeSiteLow,\SeventhSixThirdWholeSiteHigh]}}$ & $\SeventhSixThirdWholeOutputDistance$ & $\SeventhSixThirdWholeSiteDistance$ & $\SeventhSixThirdWholeOpposite$ & $\SeventhSixThirdWholeFixed$ \\
    U & first & $\SeventhSixThirdEarlierOutput_{\text{\fontsize{9}{10}\selectfont [\SeventhSixThirdEarlierOutputLow,\SeventhSixThirdEarlierOutputHigh]}}$ & $\SeventhSixThirdEarlierSite_{\text{\fontsize{9}{10}\selectfont [\SeventhSixThirdEarlierSiteLow,\SeventhSixThirdEarlierSiteHigh]}}$ & $\SeventhSixThirdEarlierOutputDistance$ & $\SeventhSixThirdEarlierSiteDistance$ & $\SeventhSixThirdEarlierOpposite$ & $\SeventhSixThirdEarlierFixed$ \\
    U & last & $\SeventhSixThirdLaterOutput_{\text{\fontsize{9}{10}\selectfont [\SeventhSixThirdLaterOutputLow,\SeventhSixThirdLaterOutputHigh]}}$ & $\SeventhSixThirdLaterSite_{\text{\fontsize{9}{10}\selectfont [\SeventhSixThirdLaterSiteLow,\SeventhSixThirdLaterSiteHigh]}}$ & $\SeventhSixThirdLaterOutputDistance$ & $\SeventhSixThirdLaterSiteDistance$ & $\SeventhSixThirdLaterOpposite$ & $\SeventhSixThirdLaterFixed$ \\
    \midrule
    \multicolumn{8}{l}{\cfgC, seed \EightSevenSource feature \EightSevenFeature into seed \EightSevenTarget, fails, persists} \\
    Selection & whole & & $\EightSevenSiteSelection$ & & $\EightSevenSiteOffsetSelection$ & & \\
    S & whole & $\SeventhSevenScoringWholeOutput_{\text{\fontsize{9}{10}\selectfont [\SeventhSevenScoringWholeOutputLow,\SeventhSevenScoringWholeOutputHigh]}}$ & $\SeventhSevenScoringWholeSite_{\text{\fontsize{9}{10}\selectfont [\SeventhSevenScoringWholeSiteLow,\SeventhSevenScoringWholeSiteHigh]}}$ & $\SeventhSevenScoringWholeOutputDistance$ & $\SeventhSevenScoringWholeSiteDistance$ & $\SeventhSevenScoringWholeOpposite$ & $\SeventhSevenScoringWholeFixed$ \\
    S & first & $\SeventhSevenScoringEarlierOutput_{\text{\fontsize{9}{10}\selectfont [\SeventhSevenScoringEarlierOutputLow,\SeventhSevenScoringEarlierOutputHigh]}}$ & $\SeventhSevenScoringEarlierSite_{\text{\fontsize{9}{10}\selectfont [\SeventhSevenScoringEarlierSiteLow,\SeventhSevenScoringEarlierSiteHigh]}}$ & $\SeventhSevenScoringEarlierOutputDistance$ & $\SeventhSevenScoringEarlierSiteDistance$ & $\SeventhSevenScoringEarlierOpposite$ & $\SeventhSevenScoringEarlierFixed$ \\
    S & last & $\SeventhSevenScoringLaterOutput_{\text{\fontsize{9}{10}\selectfont [\SeventhSevenScoringLaterOutputLow,\SeventhSevenScoringLaterOutputHigh]}}$ & $\SeventhSevenScoringLaterSite_{\text{\fontsize{9}{10}\selectfont [\SeventhSevenScoringLaterSiteLow,\SeventhSevenScoringLaterSiteHigh]}}$ & $\SeventhSevenScoringLaterOutputDistance$ & $\SeventhSevenScoringLaterSiteDistance$ & $\SeventhSevenScoringLaterOpposite$ & $\SeventhSevenScoringLaterFixed$ \\
    F & whole & $\SeventhSevenFreshWholeOutput_{\text{\fontsize{9}{10}\selectfont [\SeventhSevenFreshWholeOutputLow,\SeventhSevenFreshWholeOutputHigh]}}$ & $\SeventhSevenFreshWholeSite_{\text{\fontsize{9}{10}\selectfont [\SeventhSevenFreshWholeSiteLow,\SeventhSevenFreshWholeSiteHigh]}}$ & $\SeventhSevenFreshWholeOutputDistance$ & $\SeventhSevenFreshWholeSiteDistance$ & $\SeventhSevenFreshWholeOpposite$ & $\SeventhSevenFreshWholeFixed$ \\
    F & first & $\SeventhSevenFreshEarlierOutput_{\text{\fontsize{9}{10}\selectfont [\SeventhSevenFreshEarlierOutputLow,\SeventhSevenFreshEarlierOutputHigh]}}$ & $\SeventhSevenFreshEarlierSite_{\text{\fontsize{9}{10}\selectfont [\SeventhSevenFreshEarlierSiteLow,\SeventhSevenFreshEarlierSiteHigh]}}$ & $\SeventhSevenFreshEarlierOutputDistance$ & $\SeventhSevenFreshEarlierSiteDistance$ & $\SeventhSevenFreshEarlierOpposite$ & $\SeventhSevenFreshEarlierFixed$ \\
    F & last & $\SeventhSevenFreshLaterOutput_{\text{\fontsize{9}{10}\selectfont [\SeventhSevenFreshLaterOutputLow,\SeventhSevenFreshLaterOutputHigh]}}$ & $\SeventhSevenFreshLaterSite_{\text{\fontsize{9}{10}\selectfont [\SeventhSevenFreshLaterSiteLow,\SeventhSevenFreshLaterSiteHigh]}}$ & $\SeventhSevenFreshLaterOutputDistance$ & $\SeventhSevenFreshLaterSiteDistance$ & $\SeventhSevenFreshLaterOpposite$ & $\SeventhSevenFreshLaterFixed$ \\
    U & whole & $\SeventhSevenThirdWholeOutput_{\text{\fontsize{9}{10}\selectfont [\SeventhSevenThirdWholeOutputLow,\SeventhSevenThirdWholeOutputHigh]}}$ & $\SeventhSevenThirdWholeSite_{\text{\fontsize{9}{10}\selectfont [\SeventhSevenThirdWholeSiteLow,\SeventhSevenThirdWholeSiteHigh]}}$ & $\SeventhSevenThirdWholeOutputDistance$ & $\SeventhSevenThirdWholeSiteDistance$ & $\SeventhSevenThirdWholeOpposite$ & $\SeventhSevenThirdWholeFixed$ \\
    U & first & $\SeventhSevenThirdEarlierOutput_{\text{\fontsize{9}{10}\selectfont [\SeventhSevenThirdEarlierOutputLow,\SeventhSevenThirdEarlierOutputHigh]}}$ & $\SeventhSevenThirdEarlierSite_{\text{\fontsize{9}{10}\selectfont [\SeventhSevenThirdEarlierSiteLow,\SeventhSevenThirdEarlierSiteHigh]}}$ & $\SeventhSevenThirdEarlierOutputDistance$ & $\SeventhSevenThirdEarlierSiteDistance$ & $\SeventhSevenThirdEarlierOpposite$ & $\SeventhSevenThirdEarlierFixed$ \\
    U & last & $\SeventhSevenThirdLaterOutput_{\text{\fontsize{9}{10}\selectfont [\SeventhSevenThirdLaterOutputLow,\SeventhSevenThirdLaterOutputHigh]}}$ & $\SeventhSevenThirdLaterSite_{\text{\fontsize{9}{10}\selectfont [\SeventhSevenThirdLaterSiteLow,\SeventhSevenThirdLaterSiteHigh]}}$ & $\SeventhSevenThirdLaterOutputDistance$ & $\SeventhSevenThirdLaterSiteDistance$ & $\SeventhSevenThirdLaterOpposite$ & $\SeventhSevenThirdLaterFixed$ \\
    \midrule
    \multicolumn{8}{l}{\cfgC, seed \EightEightSource feature \EightEightFeature into seed \EightEightTarget, passes, does not persist} \\
    Selection & whole & & $\EightEightSiteSelection$ & & $\EightEightSiteOffsetSelection$ & & \\
    S & whole & $\SeventhEightScoringWholeOutput_{\text{\fontsize{9}{10}\selectfont [\SeventhEightScoringWholeOutputLow,\SeventhEightScoringWholeOutputHigh]}}$ & $\SeventhEightScoringWholeSite_{\text{\fontsize{9}{10}\selectfont [\SeventhEightScoringWholeSiteLow,\SeventhEightScoringWholeSiteHigh]}}$ & $\SeventhEightScoringWholeOutputDistance$ & $\SeventhEightScoringWholeSiteDistance$ & $\SeventhEightScoringWholeOpposite$ & $\SeventhEightScoringWholeFixed$ \\
    S & first & $\SeventhEightScoringEarlierOutput_{\text{\fontsize{9}{10}\selectfont [\SeventhEightScoringEarlierOutputLow,\SeventhEightScoringEarlierOutputHigh]}}$ & $\SeventhEightScoringEarlierSite_{\text{\fontsize{9}{10}\selectfont [\SeventhEightScoringEarlierSiteLow,\SeventhEightScoringEarlierSiteHigh]}}$ & $\SeventhEightScoringEarlierOutputDistance$ & $\SeventhEightScoringEarlierSiteDistance$ & $\SeventhEightScoringEarlierOpposite$ & $\SeventhEightScoringEarlierFixed$ \\
    S & last & $\SeventhEightScoringLaterOutput_{\text{\fontsize{9}{10}\selectfont [\SeventhEightScoringLaterOutputLow,\SeventhEightScoringLaterOutputHigh]}}$ & $\SeventhEightScoringLaterSite_{\text{\fontsize{9}{10}\selectfont [\SeventhEightScoringLaterSiteLow,\SeventhEightScoringLaterSiteHigh]}}$ & $\SeventhEightScoringLaterOutputDistance$ & $\SeventhEightScoringLaterSiteDistance$ & $\SeventhEightScoringLaterOpposite$ & $\SeventhEightScoringLaterFixed$ \\
    F & whole & $\SeventhEightFreshWholeOutput_{\text{\fontsize{9}{10}\selectfont [\SeventhEightFreshWholeOutputLow,\SeventhEightFreshWholeOutputHigh]}}$ & $\SeventhEightFreshWholeSite_{\text{\fontsize{9}{10}\selectfont [\SeventhEightFreshWholeSiteLow,\SeventhEightFreshWholeSiteHigh]}}$ & $\SeventhEightFreshWholeOutputDistance$ & $\SeventhEightFreshWholeSiteDistance$ & $\SeventhEightFreshWholeOpposite$ & $\SeventhEightFreshWholeFixed$ \\
    F & first & $\SeventhEightFreshEarlierOutput_{\text{\fontsize{9}{10}\selectfont [\SeventhEightFreshEarlierOutputLow,\SeventhEightFreshEarlierOutputHigh]}}$ & $\SeventhEightFreshEarlierSite_{\text{\fontsize{9}{10}\selectfont [\SeventhEightFreshEarlierSiteLow,\SeventhEightFreshEarlierSiteHigh]}}$ & $\SeventhEightFreshEarlierOutputDistance$ & $\SeventhEightFreshEarlierSiteDistance$ & $\SeventhEightFreshEarlierOpposite$ & $\SeventhEightFreshEarlierFixed$ \\
    F & last & $\SeventhEightFreshLaterOutput_{\text{\fontsize{9}{10}\selectfont [\SeventhEightFreshLaterOutputLow,\SeventhEightFreshLaterOutputHigh]}}$ & $\SeventhEightFreshLaterSite_{\text{\fontsize{9}{10}\selectfont [\SeventhEightFreshLaterSiteLow,\SeventhEightFreshLaterSiteHigh]}}$ & $\SeventhEightFreshLaterOutputDistance$ & $\SeventhEightFreshLaterSiteDistance$ & $\SeventhEightFreshLaterOpposite$ & $\SeventhEightFreshLaterFixed$ \\
    U & whole & $\SeventhEightThirdWholeOutput_{\text{\fontsize{9}{10}\selectfont [\SeventhEightThirdWholeOutputLow,\SeventhEightThirdWholeOutputHigh]}}$ & $\SeventhEightThirdWholeSite_{\text{\fontsize{9}{10}\selectfont [\SeventhEightThirdWholeSiteLow,\SeventhEightThirdWholeSiteHigh]}}$ & $\SeventhEightThirdWholeOutputDistance$ & $\SeventhEightThirdWholeSiteDistance$ & $\SeventhEightThirdWholeOpposite$ & $\SeventhEightThirdWholeFixed$ \\
    U & first & $\SeventhEightThirdEarlierOutput_{\text{\fontsize{9}{10}\selectfont [\SeventhEightThirdEarlierOutputLow,\SeventhEightThirdEarlierOutputHigh]}}$ & $\SeventhEightThirdEarlierSite_{\text{\fontsize{9}{10}\selectfont [\SeventhEightThirdEarlierSiteLow,\SeventhEightThirdEarlierSiteHigh]}}$ & $\SeventhEightThirdEarlierOutputDistance$ & $\SeventhEightThirdEarlierSiteDistance$ & $\SeventhEightThirdEarlierOpposite$ & $\SeventhEightThirdEarlierFixed$ \\
    U & last & $\SeventhEightThirdLaterOutput_{\text{\fontsize{9}{10}\selectfont [\SeventhEightThirdLaterOutputLow,\SeventhEightThirdLaterOutputHigh]}}$ & $\SeventhEightThirdLaterSite_{\text{\fontsize{9}{10}\selectfont [\SeventhEightThirdLaterSiteLow,\SeventhEightThirdLaterSiteHigh]}}$ & $\SeventhEightThirdLaterOutputDistance$ & $\SeventhEightThirdLaterSiteDistance$ & $\SeventhEightThirdLaterOpposite$ & $\SeventhEightThirdLaterFixed$ \\
    \bottomrule
  \end{tabular}
  \tablenote{\noteEightContinued}
\end{table}

\begin{table}[!htb]
  \caption{\capContinuation}
  \label{tab:supp-continuation}
  \centering
  \fontsize{9}{10.5}\selectfont
  \setlength{\tabcolsep}{1pt}
  \begin{tabular}{@{}lllcccccc@{}}
    \toprule
    & & & \multicolumn{3}{c}{\shortstack{Squared difference over\\squared first effect}} & \multicolumn{3}{c}{\shortstack{Norm of matched effect\\over norm of first}} \\
    \cmidrule(lr){4-6}\cmidrule(l){7-9}
    & Pair & \shortstack{Output\\test} & \shortstack{Edited\\layer} & \shortstack{Before\\normalization} & Output & \shortstack{Edited\\layer} & \shortstack{Before\\normalization} & Output \\
    \midrule
    \rowAone & \NinthContSrcOne/\NinthContFeatOne to \NinthContTgtOne/\NinthContPartOne & passes & \NinthContSiteOne & \NinthContPreOne & \NinthContPostOne & \NinthContNormSiteOne & \NinthContNormPreOne & \NinthContNormPostOne \\
    \rowAtwo & \NinthContSrcTwo/\NinthContFeatTwo to \NinthContTgtTwo/\NinthContPartTwo & passes & \NinthContSiteTwo & \NinthContPreTwo & \NinthContPostTwo & \NinthContNormSiteTwo & \NinthContNormPreTwo & \NinthContNormPostTwo \\
     & \NinthContSrcThree/\NinthContFeatThree to \NinthContTgtThree/\NinthContPartThree & passes & \NinthContSiteThree & \NinthContPreThree & \NinthContPostThree & \NinthContNormSiteThree & \NinthContNormPreThree & \NinthContNormPostThree \\
     & \NinthContSrcFour/\NinthContFeatFour to \NinthContTgtFour/\NinthContPartFour & passes & \NinthContSiteFour & \NinthContPreFour & \NinthContPostFour & \NinthContNormSiteFour & \NinthContNormPreFour & \NinthContNormPostFour \\
     & \NinthContSrcFive/\NinthContFeatFive to \NinthContTgtFive/\NinthContPartFive & passes & \NinthContSiteFive & \NinthContPreFive & \NinthContPostFive & \NinthContNormSiteFive & \NinthContNormPreFive & \NinthContNormPostFive \\
    \cfgB & \NinthContSrcSix/\NinthContFeatSix to \NinthContTgtSix/\NinthContPartSix & fails & \NinthContSiteSix & \NinthContPreSix & \NinthContPostSix & \NinthContNormSiteSix & \NinthContNormPreSix & \NinthContNormPostSix \\
    \rowCone & \NinthContSrcSeven/\NinthContFeatSeven to \NinthContTgtSeven/\NinthContPartSeven & fails & \NinthContSiteSeven & \NinthContPreSeven & \NinthContPostSeven & \NinthContNormSiteSeven & \NinthContNormPreSeven & \NinthContNormPostSeven \\
    \rowCtwo & \NinthContSrcEight/\NinthContFeatEight to \NinthContTgtEight/\NinthContPartEight & passes & \NinthContSiteEight & \NinthContPreEight & \NinthContPostEight & \NinthContNormSiteEight & \NinthContNormPreEight & \NinthContNormPostEight \\
    \bottomrule
  \end{tabular}
  \tablenote{\noteContinuation{} 
Each column of the second ratio divides the norm of the matched effect by the norm of the first effect.}
\end{table}

\clearpage
\section{The development setting}
\label{sec:supp-development}

On the development setting the match passes \TabDevTr of the \DevUnits matched pairs, and the replacements built from several features have similar pass totals. We designed the output test and fitted the cosine rule on this setting, \Dino base at layer~\SiteDev with SAEs of \WidthDev features, before we froze the output test. It enters none of the three comparisons, and \cref{sec:carrying} compares layer and width against it. Its pairs are scored on the same \ImagesScore scoring images under every definition of \cref{sec:test}.

\begin{figure}[!htb]
  \centering
  \includegraphics[width=\linewidth]{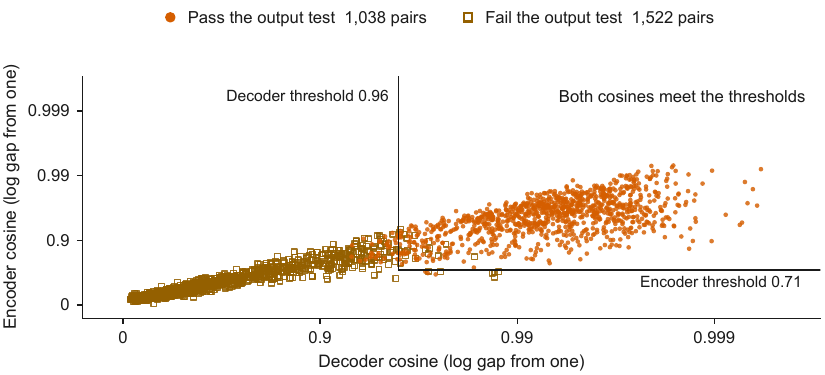}
  \caption{\figRuleCaption}
  \label{fig:rule}
  \tablenote{\noteFigRule}
\end{figure}

\Cref{tab:supp-development} gives the share each replacement passes in each kind of pair, where the match passes \TabDevTr of the matched pairs and ridge regression \SdUpperTr. At bounds of \BoundTight and \BoundLoose on the output error the match passes \SdMatchTight and \SdMatchLoose of those pairs, as \cref{tab:supp-bounds} gives for every replacement beside the match's shares in the three settings. \Cref{tab:supp-weak-grid} gives the weak matches at five cutoffs. \Cref{fig:rule} plots the pairs on which we fitted the cosine rule, where the boundary between passing pairs and the rest runs close to the decoder threshold.

\begin{table}[!htbp]
  \caption{\capDevelopment}
  \label{tab:supp-development}
  \centering
  \fontsize{9}{10.5}\selectfont
\renewcommand{\estimate}[2]{\ensuremath{\textnormal{#1}_{\publicspread#2\relax}}}
  \setlength{\tabcolsep}{1pt}
  \begin{tabular}{@{}lccccc@{}}
    \toprule
    Replacement & Matched $\uparrow$ & Random decoder $\downarrow$ & Surrogate $\downarrow$ & Rescaled $\uparrow$ & Shuffled $\downarrow$ \\
    \midrule
    \rowcolor{methodtint} The match & \estimate{\MarkDevelopmentTrSdMatchTr}{\SdMatchTrLow to \SdMatchTrHigh} & \estimate{\MarkDevelopmentFrSdMatchFr}{\SdMatchFrLow to \SdMatchFrHigh} & \estimate{\MarkDevelopmentSuSdMatchSu}{\SdMatchSuLow to \SdMatchSuHigh} & \estimate{\MarkDevelopmentGaSdMatchGa}{\SdMatchGaLow to \SdMatchGaHigh} & \estimate{\MarkDevelopmentShSdMatchSh}{\SdMatchShLow to \SdMatchShHigh} \\
    Greedy, one feature & \estimate{\MarkDevelopmentTrSdGroupOneTr}{\SdGroupOneTrLow to \SdGroupOneTrHigh} & \estimate{\MarkDevelopmentFrSdGroupOneFr}{\SdGroupOneFrLow to \SdGroupOneFrHigh} & \estimate{\MarkDevelopmentSuSdGroupOneSu}{\SdGroupOneSuLow to \SdGroupOneSuHigh} & \estimate{\MarkDevelopmentGaSdGroupOneGa}{\SdGroupOneGaLow to \SdGroupOneGaHigh} & \estimate{\MarkDevelopmentShSdGroupOneSh}{\SdGroupOneShLow to \SdGroupOneShHigh} \\
    Greedy, two features & \estimate{\MarkDevelopmentTrSdGroupTwoTr}{\SdGroupTwoTrLow to \SdGroupTwoTrHigh} & \estimate{\MarkDevelopmentFrSdGroupTwoFr}{\SdGroupTwoFrLow to \SdGroupTwoFrHigh} & \estimate{\MarkDevelopmentSuSdGroupTwoSu}{\SdGroupTwoSuLow to \SdGroupTwoSuHigh} & \estimate{\MarkDevelopmentGaSdGroupTwoGa}{\SdGroupTwoGaLow to \SdGroupTwoGaHigh} & \estimate{\MarkDevelopmentShSdGroupTwoSh}{\SdGroupTwoShLow to \SdGroupTwoShHigh} \\
    Greedy, four features & \estimate{\MarkDevelopmentTrSdGroupFourTr}{\SdGroupFourTrLow to \SdGroupFourTrHigh} & \estimate{\MarkDevelopmentFrSdGroupFourFr}{\SdGroupFourFrLow to \SdGroupFourFrHigh} & \estimate{\MarkDevelopmentSuSdGroupFourSu}{\SdGroupFourSuLow to \SdGroupFourSuHigh} & \estimate{\MarkDevelopmentGaSdGroupFourGa}{\SdGroupFourGaLow to \SdGroupFourGaHigh} & \estimate{\MarkDevelopmentShSdGroupFourSh}{\SdGroupFourShLow to \SdGroupFourShHigh} \\
    Greedy, eight features & \estimate{\MarkDevelopmentTrSdGroupEightTr}{\SdGroupEightTrLow to \SdGroupEightTrHigh} & \estimate{\MarkDevelopmentFrSdGroupEightFr}{\SdGroupEightFrLow to \SdGroupEightFrHigh} & \estimate{\MarkDevelopmentSuSdGroupEightSu}{\SdGroupEightSuLow to \SdGroupEightSuHigh} & \estimate{\MarkDevelopmentGaSdGroupEightGa}{\SdGroupEightGaLow to \SdGroupEightGaHigh} & \estimate{\MarkDevelopmentShSdGroupEightSh}{\SdGroupEightShLow to \SdGroupEightShHigh} \\
    Ridge regression & \estimate{\MarkDevelopmentTrSdUpperTr}{\SdUpperTrLow to \SdUpperTrHigh} & \estimate{\MarkDevelopmentFrSdUpperFr}{\SdUpperFrLow to \SdUpperFrHigh} & \estimate{\MarkDevelopmentSuSdUpperSu}{\SdUpperSuLow to \SdUpperSuHigh} & \estimate{\MarkDevelopmentGaSdUpperGa}{\SdUpperGaLow to \SdUpperGaHigh} & \estimate{\MarkDevelopmentShSdUpperSh}{\SdUpperShLow to \SdUpperShHigh} \\
    \bottomrule
  \end{tabular}
  \tablenote{\noteDevelopment}
\end{table}

\begin{table}[!htbp]
  \caption{\capBounds}
  \label{tab:supp-bounds}
  \centering
  \fontsize{9}{10.5}\selectfont
\renewcommand{\estimate}[2]{\ensuremath{\textnormal{#1}_{\publicspread#2\relax}}}
  \begin{tabular}{@{}lccc@{}}
    \toprule
    Replacement & $\epsilon \leq$ \BoundTight $\uparrow$ & $\epsilon \leq$ \Certify $\uparrow$ & $\epsilon \leq$ \BoundLoose $\uparrow$ \\
    \midrule
    \rowcolor{methodtint} The match & \inestimate{\MarkDevelopmentTightSdMatchTight}{\SdMatchTightLow to \SdMatchTightHigh} & \inestimate{\MarkDevelopmentTrSdMatchTr}{\SdMatchTrLow to \SdMatchTrHigh} & \inestimate{\MarkDevelopmentLooseSdMatchLoose}{\SdMatchLooseLow to \SdMatchLooseHigh} \\
    Greedy selection of at most one feature & \inestimate{\MarkDevelopmentTightSdGroupOneTight}{\SdGroupOneTightLow to \SdGroupOneTightHigh} & \inestimate{\MarkDevelopmentTrSdGroupOneTr}{\SdGroupOneTrLow to \SdGroupOneTrHigh} & \inestimate{\MarkDevelopmentLooseSdGroupOneLoose}{\SdGroupOneLooseLow to \SdGroupOneLooseHigh} \\
    Greedy selection of at most two features & \inestimate{\MarkDevelopmentTightSdGroupTwoTight}{\SdGroupTwoTightLow to \SdGroupTwoTightHigh} & \inestimate{\MarkDevelopmentTrSdGroupTwoTr}{\SdGroupTwoTrLow to \SdGroupTwoTrHigh} & \inestimate{\MarkDevelopmentLooseSdGroupTwoLoose}{\SdGroupTwoLooseLow to \SdGroupTwoLooseHigh} \\
    Greedy selection of at most four features & \inestimate{\MarkDevelopmentTightSdGroupFourTight}{\SdGroupFourTightLow to \SdGroupFourTightHigh} & \inestimate{\MarkDevelopmentTrSdGroupFourTr}{\SdGroupFourTrLow to \SdGroupFourTrHigh} & \inestimate{\MarkDevelopmentLooseSdGroupFourLoose}{\SdGroupFourLooseLow to \SdGroupFourLooseHigh} \\
    Greedy selection of at most eight features & \inestimate{\MarkDevelopmentTightSdGroupEightTight}{\SdGroupEightTightLow to \SdGroupEightTightHigh} & \inestimate{\MarkDevelopmentTrSdGroupEightTr}{\SdGroupEightTrLow to \SdGroupEightTrHigh} & \inestimate{\MarkDevelopmentLooseSdGroupEightLoose}{\SdGroupEightLooseLow to \SdGroupEightLooseHigh} \\
    Ridge regression on all features & \inestimate{\MarkDevelopmentTightSdUpperTight}{\SdUpperTightLow to \SdUpperTightHigh} & \inestimate{\MarkDevelopmentTrSdUpperTr}{\SdUpperTrLow to \SdUpperTrHigh} & \inestimate{\MarkDevelopmentLooseSdUpperLoose}{\SdUpperLooseLow to \SdUpperLooseHigh} \\
    \midrule
    \rowcolor{methodtint} The match at \cfgA & \inestimate{\BdTightA}{\BdTightLowA to \BdTightHighA} & \inestimate{\ShareA}{\TrLowA to \TrHighA} & \inestimate{\BdLooseA}{\BdLooseLowA to \BdLooseHighA} \\
    \rowcolor{methodtint} The match at \cfgB & \inestimate{\BdTightB}{\BdTightLowB to \BdTightHighB} & \inestimate{\ShareB}{\TrLowB to \TrHighB} & \inestimate{\BdLooseB}{\BdLooseLowB to \BdLooseHighB} \\
    \rowcolor{methodtint} The match at \cfgC & \inestimate{\BdTightC}{\BdTightLowC to \BdTightHighC} & \inestimate{\ShareC}{\TrLowC to \TrHighC} & \inestimate{\BdLooseC}{\BdLooseLowC to \BdLooseHighC} \\
    \bottomrule
  \end{tabular}
  \tablenote{\noteBounds}
\end{table}

\begin{table}[!htb]
  \caption{\capWeakGrid}
  \label{tab:supp-weak-grid}
  \centering
  \fontsize{9}{10.5}\selectfont
  \setlength{\tabcolsep}{4pt}
  \begin{tabular}{@{}cccccc@{}}
    \toprule
    Cutoff on either cosine & \SgAtFive $\uparrow$ & \SgAtSix $\uparrow$ & \SgAtSeven $\uparrow$ & \SgAtEight $\uparrow$ & \SgAtNine $\uparrow$ \\
    \midrule
    Weak matches & \SgUnitsFive & \SgUnitsSix & \SgUnitsSeven & \SgUnitsEight & \SgUnitsNine \\
    \rowcolor{methodtint} Passed by the match & \MarkWeakGridFiveSgMatchFive & \MarkWeakGridSixSgMatchSix & \MarkWeakGridSevenSgMatchSeven & \MarkWeakGridEightSgMatchEight & \MarkWeakGridNineSgMatchNine \\
    Passed by greedy selection & \MarkWeakGridFiveSgGroupFive & \MarkWeakGridSixSgGroupSix & \MarkWeakGridSevenSgGroupSeven & \MarkWeakGridEightSgGroupEight & \MarkWeakGridNineSgGroupNine \\
    Passed by ridge regression & \MarkWeakGridFiveSgUpperFive & \MarkWeakGridSixSgUpperSix & \MarkWeakGridSevenSgUpperSeven & \MarkWeakGridEightSgUpperEight & \MarkWeakGridNineSgUpperNine \\
    Empty greedy selections & \SgEmptyFive & \SgEmptySix & \SgEmptySeven & \SgEmptyEight & \SgEmptyNine \\
    \bottomrule
  \end{tabular}
  \tablenote{\noteWeakGrid{} 
A match is weak at a cutoff where its decoder cosine or its encoder cosine is at or below that cutoff.}
\end{table}

\clearpage
\section{Weak matches and features that pass into all other seeds}
\label{sec:supp-classes}

A weak match has a decoder cosine or an encoder cosine at or below \Weak, and a feature passes into all other seeds where it passes with its match in all four other SAEs. No replacement makes more than a few weak matches pass, and on most SAEs the features that pass into all other seeds reconstruct more variance than as many features ranked highest by cosine.

\paragraph{Weak matches.}
\Cref{tab:weak} gives the passes of each replacement over every matched pair and over the weak matches of every setting. Greedy selection passes \WeakGroupA, \WeakGroupB and \WeakGroupC of the weak matches in the three settings and stays empty on \WeakEmptyA, \WeakEmptyB and \WeakEmptyC of them, where no feature of the second SAE lowers the distance to the first contribution by more than the search's tolerance. Ridge regression passes two weak matches at \cfgA, \WeakUpperB at \cfgB and \WeakUpperC at \cfgC. In the three settings every replacement passes all \UnitsKnown pairs of the rescaled copies and none of the shuffled copies or of the pairs into random decoder SAEs. Of the \UnitsSurrogate pairs into surrogate SAEs at \cfgB, the match, greedy selection and ridge regression pass \SurrogateBMatch, \SurrogateBGreedy and \SurrogateBDense, and at \cfgC the match and greedy selection pass one and ridge regression none.
\begin{table}[!htb]
  \caption{\capWeak}
  \label{tab:weak}
  \centering
  \fontsize{9}{10.5}\selectfont
  \setlength{\tabcolsep}{6pt}
  \begin{tabular}{@{}lcccc@{}}
    \toprule
    & \headDev{} $\uparrow$ & \headA{} $\uparrow$ & \headB{} $\uparrow$ & \headC{} $\uparrow$ \\
    \midrule
    \multicolumn{5}{@{}l}{\emph{All matched pairs}} \\
    \rowcolor{methodtint} Passed by the match & \MarkWeakAllDevDevCert & \MarkWeakAllACertA & \MarkWeakAllBCertB & \MarkWeakAllCCertC \\
    Passed by greedy selection & \MarkWeakAllDevDevClassGroup & \MarkWeakAllAClassGroupA & \MarkWeakAllBClassGroupB & \MarkWeakAllCClassGroupC \\
    Passed by ridge regression & \MarkWeakAllDevDevClassUpper & \MarkWeakAllAClassUpperA & \MarkWeakAllBClassUpperB & \MarkWeakAllCClassUpperC \\
    \midrule
    \multicolumn{5}{@{}l}{\emph{Weak matches}} \\
    Weak matches & \DevWeakUnits & \WeakUnitsA & \WeakUnitsB & \WeakUnitsC \\
    \rowcolor{methodtint} Passed by the match & \MarkWeakOnlyDevDevWeakMatch & \MarkWeakOnlyAWeakMatchA & \MarkWeakOnlyBWeakMatchB & \MarkWeakOnlyCWeakMatchC \\
    Passed by greedy selection & \MarkWeakOnlyDevDevWeakGroup & \MarkWeakOnlyAWeakGroupA & \MarkWeakOnlyBWeakGroupB & \MarkWeakOnlyCWeakGroupC \\
    Passed by ridge regression & \MarkWeakOnlyDevDevWeakUpper & \MarkWeakOnlyAWeakUpperA & \MarkWeakOnlyBWeakUpperB & \MarkWeakOnlyCWeakUpperC \\
    Empty greedy selections & \DevWeakEmpty & \WeakEmptyA & \WeakEmptyB & \WeakEmptyC \\
    \bottomrule
  \end{tabular}
  \tablenote{\noteWeak}
\end{table}

\paragraph{Features that pass into all other seeds.}
The output test alone selects these features, while the comparison below measures how much of the layer's variance they reconstruct. The share of the layer's centred variance a set reconstructs is one minus the squared distance between the tokens and the sum of the set's contributions, each centred on its own mean over the scoring tokens, over the squared size of the centred tokens. The codes come from the SAE's encoder rows, and the decoder bias stays outside the sum.

For each SAE, \cref{tab:supp-variance} gives the features that pass into all other seeds, the share of variance they and as many features ranked highest by cosine reconstruct, and their rank among the random sets. On one SAE of \cfgA and one of \cfgB the two sets are the same set, and the features that pass reconstruct less than the features ranked highest by cosine on one SAE of \cfgB and one of \cfgC. \Cref{tab:supp-fresh-variance} repeats the comparison on the fresh images, where every set keeps its members and its reconstruction and the layer's tokens are centred on their own fresh means\later.

\begin{table}[!htb]
  \caption{\capVariance}
  \label{tab:supp-variance}
  \centering
  \fontsize{9}{10.5}\selectfont
  \setlength{\tabcolsep}{4pt}
  \begin{tabular}{@{}llccccc@{}}
    \toprule
    & \shortstack{Training\\seed} & \shortstack{Features\\that pass} & \shortstack{Not top\\by cosine} & \shortstack{Variance,\\passing $\uparrow$} & \shortstack{Variance,\\top cosine $\uparrow$} & \shortstack{Random sets\\with less} \\
    \midrule
    \multirow{5}{*}{\headA} & first & \SvAOneSize & \SaAOne & \cellcolor{methodtint}\MarkVarianceAOneSvAOneCarried & \MarkVarianceAOneSvAOneCosine & \SvAOneBelow \\
     & second & \SvATwoSize & \SaATwo & \cellcolor{methodtint}\MarkVarianceATwoSvATwoCarried & \MarkVarianceATwoSvATwoCosine & \SvATwoBelow \\
     & third & \SvAThreeSize & \SaAThree & \cellcolor{methodtint}\MarkVarianceAThreeSvAThreeCarried & \MarkVarianceAThreeSvAThreeCosine & \SvAThreeBelow \\
     & fourth & \SvAFourSize & \SaAFour & \cellcolor{methodtint}\MarkVarianceAFourSvAFourCarried & \MarkVarianceAFourSvAFourCosine & \SvAFourBelow \\
     & fifth & \SvAFiveSize & \SaAFive & \cellcolor{methodtint}\MarkVarianceAFiveSvAFiveCarried & \MarkVarianceAFiveSvAFiveCosine & \SvAFiveBelow \\
    \midrule
    \multirow{5}{*}{\headB} & first & \SvBOneSize & \SaBOne & \cellcolor{methodtint}\MarkVarianceBOneSvBOneCarried & \MarkVarianceBOneSvBOneCosine & \SvBOneBelow \\
     & second & \SvBTwoSize & \SaBTwo & \cellcolor{methodtint}\MarkVarianceBTwoSvBTwoCarried & \MarkVarianceBTwoSvBTwoCosine & \SvBTwoBelow \\
     & third & \SvBThreeSize & \SaBThree & \cellcolor{methodtint}\MarkVarianceBThreeSvBThreeCarried & \MarkVarianceBThreeSvBThreeCosine & \SvBThreeBelow \\
     & fourth & \SvBFourSize & \SaBFour & \cellcolor{methodtint}\MarkVarianceBFourSvBFourCarried & \MarkVarianceBFourSvBFourCosine & \SvBFourBelow \\
     & fifth & \SvBFiveSize & \SaBFive & \cellcolor{methodtint}\MarkVarianceBFiveSvBFiveCarried & \MarkVarianceBFiveSvBFiveCosine & \SvBFiveBelow \\
    \midrule
    \multirow{5}{*}{\headC} & first & \SvCOneSize & \SaCOne & \cellcolor{methodtint}\MarkVarianceCOneSvCOneCarried & \MarkVarianceCOneSvCOneCosine & \SvCOneBelow \\
     & second & \SvCTwoSize & \SaCTwo & \cellcolor{methodtint}\MarkVarianceCTwoSvCTwoCarried & \MarkVarianceCTwoSvCTwoCosine & \SvCTwoBelow \\
     & third & \SvCThreeSize & \SaCThree & \cellcolor{methodtint}\MarkVarianceCThreeSvCThreeCarried & \MarkVarianceCThreeSvCThreeCosine & \SvCThreeBelow \\
     & fourth & \SvCFourSize & \SaCFour & \cellcolor{methodtint}\MarkVarianceCFourSvCFourCarried & \MarkVarianceCFourSvCFourCosine & \SvCFourBelow \\
     & fifth & \SvCFiveSize & \SaCFive & \cellcolor{methodtint}\MarkVarianceCFiveSvCFiveCarried & \MarkVarianceCFiveSvCFiveCosine & \SvCFiveBelow \\
    \bottomrule
  \end{tabular}
  \tablenote{\noteVariance{} 
The last column counts the \RandomSets random sets of the same size, drawn from the SAE's \DrawnPerSource sampled features, that reconstruct less.}
\end{table}

\begin{table}[!htb]
  \caption{\capFreshVariance}
  \label{tab:supp-fresh-variance}
  \centering
  \fontsize{9}{10.5}\selectfont
  \begin{tabular}{@{}llccccc@{}}
    \toprule
    & \shortstack{Training\\seed} & \shortstack{Features\\that pass} & \shortstack{Variance,\\passing $\uparrow$} & \shortstack{Variance,\\top cosine $\uparrow$} & Difference & \shortstack{Random sets\\with less} \\
    \midrule
    \rowAone & 1 & $\FreshVarianceAOneSize$ & $\cellcolor{methodtint}\MarkFreshVarianceAOneFreshVarianceAOnePassing$ & $\MarkFreshVarianceAOneFreshVarianceAOneCosine$ & $\FreshVarianceAOneDifference$ & $\FreshVarianceAOneBelow$ \\
    \rowAtwo & 2 & $\FreshVarianceATwoSize$ & $\cellcolor{methodtint}\MarkFreshVarianceATwoFreshVarianceATwoPassing$ & $\MarkFreshVarianceATwoFreshVarianceATwoCosine$ & $\FreshVarianceATwoDifference$ & $\FreshVarianceATwoBelow$ \\
     & 3 & $\FreshVarianceAThreeSize$ & $\cellcolor{methodtint}\MarkFreshVarianceAThreeFreshVarianceAThreePassing$ & $\MarkFreshVarianceAThreeFreshVarianceAThreeCosine$ & $\FreshVarianceAThreeDifference$ & $\FreshVarianceAThreeBelow$ \\
     & 4 & $\FreshVarianceAFourSize$ & $\cellcolor{methodtint}\MarkFreshVarianceAFourFreshVarianceAFourPassing$ & $\MarkFreshVarianceAFourFreshVarianceAFourCosine$ & $\FreshVarianceAFourDifference$ & $\FreshVarianceAFourBelow$ \\
     & 5 & $\FreshVarianceAFiveSize$ & $\cellcolor{methodtint}\MarkFreshVarianceAFiveFreshVarianceAFivePassing$ & $\MarkFreshVarianceAFiveFreshVarianceAFiveCosine$ & $\FreshVarianceAFiveDifference$ & $\FreshVarianceAFiveBelow$ \\
    Mean & & & & & $\FreshVarianceDifferenceA$ &  \\
    \midrule
    \rowBone & 1 & $\FreshVarianceBOneSize$ & $\cellcolor{methodtint}\MarkFreshVarianceBOneFreshVarianceBOnePassing$ & $\MarkFreshVarianceBOneFreshVarianceBOneCosine$ & $\FreshVarianceBOneDifference$ & $\FreshVarianceBOneBelow$ \\
    \rowBtwo & 2 & $\FreshVarianceBTwoSize$ & $\cellcolor{methodtint}\MarkFreshVarianceBTwoFreshVarianceBTwoPassing$ & $\MarkFreshVarianceBTwoFreshVarianceBTwoCosine$ & $\FreshVarianceBTwoDifference$ & $\FreshVarianceBTwoBelow$ \\
     & 3 & $\FreshVarianceBThreeSize$ & $\cellcolor{methodtint}\MarkFreshVarianceBThreeFreshVarianceBThreePassing$ & $\MarkFreshVarianceBThreeFreshVarianceBThreeCosine$ & $\FreshVarianceBThreeDifference$ & $\FreshVarianceBThreeBelow$ \\
     & 4 & $\FreshVarianceBFourSize$ & $\cellcolor{methodtint}\MarkFreshVarianceBFourFreshVarianceBFourPassing$ & $\MarkFreshVarianceBFourFreshVarianceBFourCosine$ & $\FreshVarianceBFourDifference$ & $\FreshVarianceBFourBelow$ \\
     & 5 & $\FreshVarianceBFiveSize$ & $\cellcolor{methodtint}\MarkFreshVarianceBFiveFreshVarianceBFivePassing$ & $\MarkFreshVarianceBFiveFreshVarianceBFiveCosine$ & $\FreshVarianceBFiveDifference$ & $\FreshVarianceBFiveBelow$ \\
    Mean & & & & & $\FreshVarianceDifferenceB$ &  \\
    \midrule
    \rowCone & 1 & $\FreshVarianceCOneSize$ & $\cellcolor{methodtint}\MarkFreshVarianceCOneFreshVarianceCOnePassing$ & $\MarkFreshVarianceCOneFreshVarianceCOneCosine$ & $\FreshVarianceCOneDifference$ & $\FreshVarianceCOneBelow$ \\
    \rowCtwo & 2 & $\FreshVarianceCTwoSize$ & $\cellcolor{methodtint}\MarkFreshVarianceCTwoFreshVarianceCTwoPassing$ & $\MarkFreshVarianceCTwoFreshVarianceCTwoCosine$ & $\FreshVarianceCTwoDifference$ & $\FreshVarianceCTwoBelow$ \\
     & 3 & $\FreshVarianceCThreeSize$ & $\cellcolor{methodtint}\MarkFreshVarianceCThreeFreshVarianceCThreePassing$ & $\MarkFreshVarianceCThreeFreshVarianceCThreeCosine$ & $\FreshVarianceCThreeDifference$ & $\FreshVarianceCThreeBelow$ \\
     & 4 & $\FreshVarianceCFourSize$ & $\cellcolor{methodtint}\MarkFreshVarianceCFourFreshVarianceCFourPassing$ & $\MarkFreshVarianceCFourFreshVarianceCFourCosine$ & $\FreshVarianceCFourDifference$ & $\FreshVarianceCFourBelow$ \\
     & 5 & $\FreshVarianceCFiveSize$ & $\cellcolor{methodtint}\MarkFreshVarianceCFiveFreshVarianceCFivePassing$ & $\MarkFreshVarianceCFiveFreshVarianceCFiveCosine$ & $\FreshVarianceCFiveDifference$ & $\FreshVarianceCFiveBelow$ \\
    Mean & & & & & $\FreshVarianceDifferenceC$ &  \\
    \bottomrule
  \end{tabular}
  \tablenote{\noteFreshVariance{} 
The last column counts the \RandomSets random sets of the same size that reconstruct less.}
\end{table}

\clearpage
\section{Assets and compute}
\label{sec:supp-sources}

Both encoders and the images are public assets, and every timing ran on the hardware its log names.

\paragraph{Assets.}
On the Hugging Face Hub the encoders' weights are the checkpoints \texttt{\DinoCheckpoint} and \texttt{\TowerCheckpoint}. Both are released under the Apache licence their model cards state. The images are the COCO 2017 validation set with its instance annotations, and the fresh and third images come from its training split. The annotations are released under the Creative Commons Attribution licence, while each image's rights stay with its owner under the collection's terms of use.

\paragraph{Accelerator time.}
Every training and scoring run of the three settings used one accelerator, an \DeviceName or an \DevicePcie, under PyTorch~\cite{paszke2019pytorch} \TorchVersion and Transformers~\cite{wolf2020transformers} \TransformersVersion. All \ComputeCalls runs used an accelerator for \ComputeSeconds seconds, \SecondsA at \cfgA, \SecondsB at \cfgB, \SecondsC at \cfgC and \SecondsNone on a run that names no setting\later. Scoring the sampled features took \ComputeSources of those seconds and training the SAEs \ComputeTrain, and the rest went to activations, names, the sanity checks and the recomputations. The development setting's runs used one for \SecondsDev seconds.

Measuring the layer error on each half of the scoring images and each pair's output error image by image took \SitePassCalls further runs on the same accelerators and versions, which used one for \SitePassSeconds seconds. The fresh images of \cfgA and \cfgB and the timing of the three procedures took \FreshCalls further runs, which used one for \FreshSeconds seconds, and the fresh images of \cfgC took \FifthCalls more, which used one for \FifthSeconds seconds. Recomputing the coordinate means and root mean squares of \cfgA and \cfgB from their stored training images in a separate encoder pass, then repeating the timing, took \SixthCalls more runs, which used one for \SixthSeconds seconds. The permuted control and the second sample of features took \TenthCalls further runs on the same accelerators and versions, which used one for \TenthSeconds seconds\later. The measurement at the image level output of \cref{sec:supp-outputs} took \EleventhOutCalls further runs under the same versions, on an \EleventhOutDevices, which used one for \EleventhOutSeconds seconds\later. Without further training, additional passes computed the third image set, the layer errors on the scoring and fresh images and the matching of whole SAEs\later, and with the token computation of \cref{sec:supp-calibrated} they add \AdditionalSeconds measured accelerator seconds. The stated durations are a lower bound, since no duration was logged for the run that computed the layer errors of \cfgA on the scoring and fresh images or for the run that followed the eight pairs through the later layers\later.

\paragraph{Timing.}
\Cref{tab:supp-timing} times the steps of the three procedures, each on the device it ran on, so that each comparison rests on one measurement basis. The timing uses the first eight sampled features of the first SAE and their four matches at \cfgA and at \cfgB, thirty-two pairs, on the \FreshImages fresh images\later. Each measured step runs \TimingRepeats times, and the table gives each step's median and full range, printed to three decimal places. The image and pair steps follow one batch and one pair of warm up that no clock counts and use batches of \TimingBatch images, with the accelerator synchronized before each clock starts and stops, on one \TimingDevice. The activations at the edited layer are shared by the layer rule and the output test, the unedited outputs belong to the output test, and each first ablation serves its four matches.

The matching and the cosine rule's predictions each follow one identical warm up and run on a processor with \TimingThreads threads, each over all \EleventhTimedFeatures matched features of the trained SAEs of seeds one and two. Their sum, which the table gives from the unrounded medians, is the cost of the cosine rule for one pair of SAEs. It includes no timing of the layer error and none of an ablation, so it is no cost of combining the two rules as \cref{sec:supp-halves} does, which adds the layer error for every pair and the output test for the pairs on which the rules disagree. Those two steps ran in a container whose log gives \EleventhContainerProcessors processors and \EleventhContainerThreads threads and names no processor model, so their seconds describe that container alone. The matching also ran on an \SeventhProcessor with the same repetitions, warm up and thread count, on a machine shared with other jobs, where its median at \cfgA is \SeventhMacA seconds.

\begin{table}[!htb]
  \caption{\capTiming}
  \label{tab:supp-timing}
  \centering
  \fontsize{9}{10.5}\selectfont
  \begin{tabular}{@{}lcc@{}}
    \toprule
Step & \headA{} $\downarrow$ & \headB{} $\downarrow$ \\
    \midrule
Matching two SAEs, in the container & $\SeventhScreenAssignmentA_{\text{\fontsize{9}{10}\selectfont [\SeventhScreenAssignmentALow,\SeventhScreenAssignmentAHigh]}}$ & $\SeventhScreenAssignmentB_{\text{\fontsize{9}{10}\selectfont [\SeventhScreenAssignmentBLow,\SeventhScreenAssignmentBHigh]}}$ \\
Cosine predictions for two SAEs, in the container & $\SeventhScreenCallsA_{\text{\fontsize{9}{10}\selectfont [\SeventhScreenCallsALow,\SeventhScreenCallsAHigh]}}$ & $\SeventhScreenCallsB_{\text{\fontsize{9}{10}\selectfont [\SeventhScreenCallsBLow,\SeventhScreenCallsBHigh]}}$ \\
Cosine rule for two SAEs, the sum of the two medians & $\SeventhScreenTotalA$ & $\SeventhScreenTotalB$ \\
Matching two SAEs, on the \SeventhProcessor & $\SeventhMacA_{\text{\fontsize{9}{10}\selectfont [\SeventhMacALow,\SeventhMacAHigh]}}$ & $\SeventhMacB_{\text{\fontsize{9}{10}\selectfont [\SeventhMacBLow,\SeventhMacBHigh]}}$ \\
    \midrule
Cosine predictions for thirty-two pairs & $\TimeCosineA_{\text{\fontsize{9}{10}\selectfont [\TimeCosineLowA,\TimeCosineHighA]}}$ & $\TimeCosineB_{\text{\fontsize{9}{10}\selectfont [\TimeCosineLowB,\TimeCosineHighB]}}$ \\
Pass to the edited layer for all images & $\TimeSiteBatchA_{\text{\fontsize{9}{10}\selectfont [\TimeSiteBatchLowA,\TimeSiteBatchHighA]}}$ & $\TimeSiteBatchB_{\text{\fontsize{9}{10}\selectfont [\TimeSiteBatchLowB,\TimeSiteBatchHighB]}}$ \\
Codes at the edited layer for thirty-two pairs & $\TimeSiteCodesA_{\text{\fontsize{9}{10}\selectfont [\TimeSiteCodesLowA,\TimeSiteCodesHighA]}}$ & $\TimeSiteCodesB_{\text{\fontsize{9}{10}\selectfont [\TimeSiteCodesLowB,\TimeSiteCodesHighB]}}$ \\
Unedited later layers for all images & $\TimeFullBatchA_{\text{\fontsize{9}{10}\selectfont [\TimeFullBatchLowA,\TimeFullBatchHighA]}}$ & $\TimeFullBatchB_{\text{\fontsize{9}{10}\selectfont [\TimeFullBatchLowB,\TimeFullBatchHighB]}}$ \\
Ablations for thirty-two pairs & $\TimeRemovalsA_{\text{\fontsize{9}{10}\selectfont [\TimeRemovalsLowA,\TimeRemovalsHighA]}}$ & $\TimeRemovalsB_{\text{\fontsize{9}{10}\selectfont [\TimeRemovalsLowB,\TimeRemovalsHighB]}}$ \\
    \midrule
Pass to the edited layer for one image & $\TimeSiteImageA$ & $\TimeSiteImageB$ \\
Layer rule for one pair on all images & $\MarkTimingPairATimeSitePairA$ & $\MarkTimingPairBTimeSitePairB$ \\
    \rowcolor{methodtint} Output test for one pair on all images & $\MarkTimingPairATimeFullPairA$ & $\MarkTimingPairBTimeFullPairB$ \\
    \bottomrule
  \end{tabular}
  \tablenote{\noteTiming}
\end{table}

\clearpage
\section{Permuted matches and a second sample of features}
\label{sec:supp-added}

No pair passes when the matches are permuted, and a second sample of features repeats the layer rule's advantage over the cosine rule. We added both measurements after every outcome above was known\later. Both use the trained SAEs, the stored matchings and the stored activations of the fresh images as they are, and neither trains an SAE.

\paragraph{Permuted matches.}
\Cref{tab:supp-permuted} gives the permuted control of \cref{sec:test}, which shuffles the matches among trained features and which we chose after the comparisons were scored. Its \TenthPermSourcesA features a setting are sampled from the \DrawnAll sampled features by the generator of \cref{sec:supp-recipe} from seed \TenthSeedSubset plus the setting's index, which starts at zero at \cfgA. For each first SAE and each second SAE, a permutation of the first SAE's sampled features, drawn again until no feature keeps its place, gives each feature the stored match of another, and the matching is not computed again. So every permuted pair lies within the two trained SAEs of a matched pair, at their own density and reconstruction. No permuted pair passes in any setting under any of the \TenthPermutations permutations, \TenthPermAllPairsA pairs a setting, and the matched pairs of the same features reproduce their stored output errors on the fresh images exactly. Each interval of the matched pairs less the mean of the permuted pairs lies above zero, over \Resamples resamples of the features from seed \TenthSeedPermuted plus the setting's index.
\begin{table}[!htb]
  \caption{\capPermuted}
  \label{tab:supp-permuted}
  \centering
  \fontsize{9}{10.5}\selectfont
  \setlength{\tabcolsep}{4pt}
  \begin{tabular}{@{}lccc@{}}
    \toprule
    Matching & \headA{} $\uparrow$ & \headB{} $\uparrow$ & \headC{} $\uparrow$ \\
    \midrule
    \rowcolor{methodtint} Matched pairs & \inestimate{\textbf{\TenthPermMatchedA}}{\TenthPermMatchedLowA to \TenthPermMatchedHighA} & \inestimate{\textbf{\TenthPermMatchedB}}{\TenthPermMatchedLowB to \TenthPermMatchedHighB} & \inestimate{\textbf{\TenthPermMatchedC}}{\TenthPermMatchedLowC to \TenthPermMatchedHighC} \\
    Permuted, first & \TenthPermOneA & \TenthPermOneB & \TenthPermOneC \\
    Permuted, second & \TenthPermTwoA & \TenthPermTwoB & \TenthPermTwoC \\
    Permuted, third & \TenthPermThreeA & \TenthPermThreeB & \TenthPermThreeC \\
    \midrule
    Matched pairs less mean permuted & \inestimate{\TenthPermDiffA}{\TenthPermDiffLowA to \TenthPermDiffHighA} & \inestimate{\TenthPermDiffB}{\TenthPermDiffLowB to \TenthPermDiffHighB} & \inestimate{\TenthPermDiffC}{\TenthPermDiffLowC to \TenthPermDiffHighC} \\
    \bottomrule
  \end{tabular}
  \tablenote{\notePermuted}
\end{table}

\paragraph{A second sample of features.}
For each SAE the second sample takes \DrawnPerSource of the features the first sample left, from seed \TenthSeedSecond plus the SAE's own seed, and \cref{tab:supp-second} gives its counts. Every sampled feature meets the entry conditions of \cref{sec:test}, so each setting has \TenthSecPairsA new pairs, of which \TenthSecPctA{}\% to \TenthSecPctC{}\% pass, against \TenthFirstPctA{}\% to \TenthFirstPctC{}\% of the first sample's on the same images. The cosine rule, the layer error with the decoder cosine set to one and the layer rule keep the thresholds of \cref{tab:supp-predictors}, with every statistic on the selection images. The cosine rule's incorrect predictions less the layer rule's have an interval above zero in every setting, so the layer rule's advantage over the cosine rule repeats.

At \cfgA the layer error with the decoder cosine set to one again predicts incorrectly on fewer pairs than the layer rule, \TenthSecErrCodesA against \TenthSecErrSiteA, as on the first sample. The interval of that difference, from $\TenthSecCodesLessLowA$ to $\TenthSecCodesLessHighA$, now includes zero, so the order of the two counts repeats and its separation from zero does not. At \cfgB and \cfgC that rule predicts incorrectly on more pairs than the layer rule, with intervals above zero. With every statistic recomputed on the fresh images, the layer rule predicts incorrectly on \TenthSecFreshErrSiteA, \TenthSecFreshErrSiteB and \TenthSecFreshErrSiteC pairs and the layer error with the decoder cosine set to one on \TenthSecFreshErrCodesA, \TenthSecFreshErrCodesB and \TenthSecFreshErrCodesC, and the cosine rule's incorrect predictions less the layer rule's keep an interval above zero in every setting. The intervals resample the sampled features from seed \TenthSeedSecondInterval plus the setting's index.

\begin{table}[!htb]
  \caption{\capSecond}
  \label{tab:supp-second}
  \centering
  \fontsize{9}{10.5}\selectfont
  \setlength{\tabcolsep}{4pt}
  \begin{tabular}{@{}lccc@{}}
    \toprule
    & \headA & \headB & \headC \\
    \midrule
    \multicolumn{4}{@{}l}{\emph{First sample}} \\
    Pairs & \UnitsTrained & \UnitsTrained & \UnitsTrained \\
    Passing pairs & \FreshPassA & \FreshPassB & \FreshPassC \\
    Incorrect predictions, cosine rule $\downarrow$ & \NinthErrCosineA & \NinthErrCosineB & \NinthErrCosineC \\
    Incorrect predictions, $c$ set to one $\downarrow$ & \textbf{\NinthErrCodesA} & \NinthErrCodesB & \NinthErrCodesC \\
    \rowcolor{methodtint} Incorrect predictions, layer rule $\downarrow$ & \NinthErrJointA & \textbf{\NinthErrJointB} & \textbf{\NinthErrJointC} \\
    Cosine rule less layer rule & \inestimate{$\NinthMoreCosineA$}{$\NinthMoreLowCosineA$ to $\NinthMoreHighCosineA$} & \inestimate{$\NinthMoreCosineB$}{$\NinthMoreLowCosineB$ to $\NinthMoreHighCosineB$} & \inestimate{$\NinthMoreCosineC$}{$\NinthMoreLowCosineC$ to $\NinthMoreHighCosineC$} \\
    Codes less layer rule & \inestimate{$\NinthMoreCodesA$}{$\NinthMoreLowCodesA$ to $\NinthMoreHighCodesA$} & \inestimate{$\NinthMoreCodesB$}{$\NinthMoreLowCodesB$ to $\NinthMoreHighCodesB$} & \inestimate{$\NinthMoreCodesC$}{$\NinthMoreLowCodesC$ to $\NinthMoreHighCodesC$} \\
    \midrule
    \multicolumn{4}{@{}l}{\emph{Second sample}} \\
    Pairs & \TenthSecPairsA & \TenthSecPairsB & \TenthSecPairsC \\
    Passing pairs & \inestimate{\TenthSecPassA}{\TenthSecPassLowA to \TenthSecPassHighA} & \inestimate{\TenthSecPassB}{\TenthSecPassLowB to \TenthSecPassHighB} & \inestimate{\TenthSecPassC}{\TenthSecPassLowC to \TenthSecPassHighC} \\
    Incorrect predictions, cosine rule $\downarrow$ & \TenthSecErrCosineA & \TenthSecErrCosineB & \TenthSecErrCosineC \\
    Incorrect predictions, $c$ set to one $\downarrow$ & \textbf{\TenthSecErrCodesA} & \TenthSecErrCodesB & \TenthSecErrCodesC \\
    \rowcolor{methodtint} Incorrect predictions, layer rule $\downarrow$ & \TenthSecErrSiteA & \textbf{\TenthSecErrSiteB} & \textbf{\TenthSecErrSiteC} \\
    Cosine rule less layer rule & \inestimate{$\TenthSecCosLessA$}{$\TenthSecCosLessLowA$ to $\TenthSecCosLessHighA$} & \inestimate{$\TenthSecCosLessB$}{$\TenthSecCosLessLowB$ to $\TenthSecCosLessHighB$} & \inestimate{$\TenthSecCosLessC$}{$\TenthSecCosLessLowC$ to $\TenthSecCosLessHighC$} \\
    Codes less layer rule & \inestimate{$\TenthSecCodesLessA$}{$\TenthSecCodesLessLowA$ to $\TenthSecCodesLessHighA$} & \inestimate{$\TenthSecCodesLessB$}{$\TenthSecCodesLessLowB$ to $\TenthSecCodesLessHighB$} & \inestimate{$\TenthSecCodesLessC$}{$\TenthSecCodesLessLowC$ to $\TenthSecCodesLessHighC$} \\
    \bottomrule
  \end{tabular}
  \tablenote{\noteSecond}
\end{table}

\clearpage
\section{The image level output}
\label{sec:supp-outputs}

At the one vector an encoder returns for an image, \EleventhOutPctImageA{}\% to \EleventhOutPctImageC{}\% of the matched pairs pass, and the layer rule predicts incorrectly on fewer pairs than the cosine rule in every setting\later. We added this measurement after every outcome above was known. It uses the trained SAEs, the stored matchings, the ablations of the fresh images and each rule's stored prediction as they are, and it trains nothing.

\paragraph{The output and the output error.}
The output test of \cref{sec:test} compares two ablations on every token of the encoder output. \Cref{tab:supp-outputs} scores the same \UnitsTrained matched pairs of each setting on the \FreshImages fresh images at the vector a classifier or a retrieval rule would take as its input, and we apply neither. For \Dino that vector is the class token of the encoder output, taken after the final normalization. For SigLIP it is the pooled image embedding, the output of the encoder's attention pooling head on the encoder output. The output error is \cref{eq:error} with this vector in place of all tokens, a pair passes at an output error of at most \Certify, and the minimum squared effect of \cref{sec:test} is taken at this vector. Every pair meets that minimum, so every pair has an outcome.

\paragraph{The shares.}
At the image level output \EleventhOutPassImageA, \EleventhOutPassImageB and \EleventhOutPassImageC of the \UnitsTrained matched pairs pass in the three settings, against \EleventhOutPassTokensA, \EleventhOutPassTokensB and \EleventhOutPassTokensC on all tokens of the same images. In the three settings \EleventhOutLostA, \EleventhOutLostB and \EleventhOutLostC pairs pass on all tokens and fail at the image level output, and \EleventhOutGainedA, \EleventhOutGainedB and \EleventhOutGainedC fail on all tokens and pass there.

\paragraph{The rules.}
Each rule keeps its threshold and the prediction it made from the selection images. Against the outcome at the image level output the cosine rule predicts incorrectly on \EleventhErrImageCosineA, \EleventhErrImageCosineB and \EleventhErrImageCosineC pairs and the layer rule at its fitted threshold on \EleventhErrImageLayerFittedA, \EleventhErrImageLayerFittedB and \EleventhErrImageLayerFittedC. The difference is \EleventhOutDiffA, \EleventhOutDiffB and \EleventhOutDiffC pairs, and each interval lies above zero, over \Resamples resamples of the sampled features from seed \EleventhSeedOutputs plus the setting's index. Both rules predict incorrectly on more pairs than against the outcome on all tokens, where the cosine rule predicts incorrectly on \EleventhErrTokensCosineA, \EleventhErrTokensCosineB and \EleventhErrTokensCosineC and the layer rule on \EleventhErrTokensLayerFittedA, \EleventhErrTokensLayerFittedB and \EleventhErrTokensLayerFittedC. Of the layer rule's incorrect predictions at the image level output, \EleventhRepImageLayerFittedA, \EleventhRepImageLayerFittedB and \EleventhRepImageLayerFittedC repeat on both halves of the fresh images. At \cfgA the layer error with the decoder cosine set to one predicts incorrectly on \EleventhErrImageCodesA pairs and the code cosine alone on \EleventhErrImageOverlapA, beside the layer rule's \EleventhErrImageLayerFittedA.

\paragraph{The eight pairs.}
On \EleventhEightRepeat of the eight pairs of \cref{tab:supp-six} the layer rule's incorrect prediction repeats at the image level output. The pair whose incorrect prediction persists on the third image set fails at the class token at an output error of \EleventhCaseError, against \EleventhCaseTokens on all tokens, and the layer rule predicts a pass. That incorrect prediction \EleventhCaseRepeats on both halves of the fresh images.

\begin{table}[!htb]
  \caption{\capOutputs}
  \label{tab:supp-outputs}
  \centering
  \fontsize{9}{10.5}\selectfont
  \setlength{\tabcolsep}{2pt}
  \begin{tabular}{@{}lccc@{}}
    \toprule
    & \headA & \headB & \headC \\
    \midrule
    \multicolumn{4}{@{}l}{\emph{Pairs that pass on the \FreshImages fresh images, of \UnitsTrained}} \\
    On all tokens & \EleventhOutPassTokensA & \EleventhOutPassTokensB & \EleventhOutPassTokensC \\
    At the image level output & \EleventhOutPassImageA & \EleventhOutPassImageB & \EleventhOutPassImageC \\
    Passing on all tokens alone and at the image level output alone & \EleventhOutLostA and \EleventhOutGainedA & \EleventhOutLostB and \EleventhOutGainedB & \EleventhOutLostC and \EleventhOutGainedC \\
    \midrule
    \multicolumn{4}{@{}l}{\emph{Incorrect predictions against the outcome on all tokens $\downarrow$}} \\
    Cosine rule & \EleventhErrTokensCosineA & \EleventhErrTokensCosineB & \EleventhErrTokensCosineC \\
    \rowcolor{methodtint} Layer rule at its fitted threshold & \EleventhErrTokensLayerFittedA & \textbf{\EleventhErrTokensLayerFittedB} & \EleventhErrTokensLayerFittedC \\
    Layer rule at \Certify & \EleventhErrTokensLayerAtBoundA & \textbf{\EleventhErrTokensLayerAtBoundB} & \textbf{\EleventhErrTokensLayerAtBoundC} \\
    Layer error with $c$ set to one & \textbf{\EleventhErrTokensCodesA} & \EleventhErrTokensCodesB & \EleventhErrTokensCodesC \\
    Code cosine alone & \EleventhErrTokensOverlapA & \EleventhErrTokensOverlapB & \EleventhErrTokensOverlapC \\
    Code norm ratio alone & \EleventhErrTokensMagnitudeA & \EleventhErrTokensMagnitudeB & \EleventhErrTokensMagnitudeC \\
    \midrule
    \multicolumn{4}{@{}l}{\emph{Incorrect predictions against the outcome at the image level output $\downarrow$}} \\
    Cosine rule & \inestimate{\EleventhErrImageCosineA}{\EleventhRepImageCosineA} & \inestimate{\EleventhErrImageCosineB}{\EleventhRepImageCosineB} & \inestimate{\EleventhErrImageCosineC}{\EleventhRepImageCosineC} \\
    \rowcolor{methodtint} Layer rule at its fitted threshold & \inestimate{\EleventhErrImageLayerFittedA}{\EleventhRepImageLayerFittedA} & \inestimate{\textbf{\EleventhErrImageLayerFittedB}}{\EleventhRepImageLayerFittedB} & \inestimate{\EleventhErrImageLayerFittedC}{\EleventhRepImageLayerFittedC} \\
    Layer rule at \Certify & \inestimate{\EleventhErrImageLayerAtBoundA}{\EleventhRepImageLayerAtBoundA} & \inestimate{\textbf{\EleventhErrImageLayerAtBoundB}}{\EleventhRepImageLayerAtBoundB} & \inestimate{\textbf{\EleventhErrImageLayerAtBoundC}}{\EleventhRepImageLayerAtBoundC} \\
    Layer error with $c$ set to one & \inestimate{\textbf{\EleventhErrImageCodesA}}{\EleventhRepImageCodesA} & \inestimate{\EleventhErrImageCodesB}{\EleventhRepImageCodesB} & \inestimate{\EleventhErrImageCodesC}{\EleventhRepImageCodesC} \\
    Code cosine alone & \inestimate{\EleventhErrImageOverlapA}{\EleventhRepImageOverlapA} & \inestimate{\EleventhErrImageOverlapB}{\EleventhRepImageOverlapB} & \inestimate{\EleventhErrImageOverlapC}{\EleventhRepImageOverlapC} \\
    Code norm ratio alone & \inestimate{\EleventhErrImageMagnitudeA}{\EleventhRepImageMagnitudeA} & \inestimate{\EleventhErrImageMagnitudeB}{\EleventhRepImageMagnitudeB} & \inestimate{\EleventhErrImageMagnitudeC}{\EleventhRepImageMagnitudeC} \\
    Cosine rule less layer rule at its fitted threshold & \inestimate{$\EleventhOutDiffA$}{$\EleventhOutDiffLowA$ to $\EleventhOutDiffHighA$} & \inestimate{$\EleventhOutDiffB$}{$\EleventhOutDiffLowB$ to $\EleventhOutDiffHighB$} & \inestimate{$\EleventhOutDiffC$}{$\EleventhOutDiffLowC$ to $\EleventhOutDiffHighC$} \\
    \bottomrule
  \end{tabular}
  \tablenote{\noteOutputs}
\end{table}

\clearpage
\section{The fitted threshold and the pair whose incorrect prediction persists}
\label{sec:supp-calibrated}

Fitting the layer rule's threshold moves it from \Certify to \NinthFitJoint and leaves incorrect predictions on the fresh images, and one pair's incorrect prediction persists on the third image set. We added the fit and these measurements after the outcomes were known\later.

\paragraph{The fit.}
The fit chooses the threshold on the development pairs alone, and the fresh images then measure its incorrect predictions. It maximizes agreement with the outcomes of the output test on all scoring images of the development pairs, the objective that fitted the cosine rule. It searches every distinct layer error on the selection images, predicts a pass at or below the candidate, and on a tie takes the value nearest \Certify and then the smaller value. Its one optimum is \NinthFitJoint, which the note of \cref{tab:supp-predictors} gives in full, with an agreement of \FittedAgreement. We froze this value before we evaluated any pair on the fresh images at it, and no threshold is fitted again during evaluation or resampling. \Cref{tab:fitted-selection} gives both directions of incorrect prediction by first SAE with the layer rule's prediction fixed on the selection images, and \cref{tab:fitted-fresh} gives them with the layer error computed on the same resampled fresh images as the output error.
\begin{table}[!htb]
  \caption{\capFittedSelection}
  \label{tab:fitted-selection}
  \centering\fontsize{9}{10.5}\selectfont
  \setlength{\tabcolsep}{4pt}
  \begin{tabular}{@{}llcccc@{}}
  \toprule
 & Seed & \multicolumn{2}{c}{Layer rule passes, output test fails} & \multicolumn{2}{c}{Layer rule fails, output test passes} \\  & & At \Certify & At \NinthFitJoint & At \Certify & At \NinthFitJoint \\
  \midrule
\rowAone & 1 & \inestimate{\EighthCellAE}{\EighthCellAF, \EighthCellAG} & \inestimate{\EighthCellAH}{\EighthCellAI, \EighthCellAJ} & \inestimate{\EighthCellAK}{\EighthCellAL, \EighthCellAM} & \inestimate{\EighthCellAN}{\EighthCellAO, \EighthCellAP} \\
\rowAtwo & 2 & \inestimate{\EighthCellAQ}{\EighthCellAR, \EighthCellAS} & \inestimate{\EighthCellAT}{\EighthCellAU, \EighthCellAV} & \inestimate{\EighthCellAW}{\EighthCellAX, \EighthCellAY} & \inestimate{\EighthCellAZ}{\EighthCellBA, \EighthCellBB} \\
 & 3 & \inestimate{\EighthCellBC}{\EighthCellBD, \EighthCellBE} & \inestimate{\EighthCellBF}{\EighthCellBG, \EighthCellBH} & \inestimate{\EighthCellBI}{\EighthCellBJ, \EighthCellBK} & \inestimate{\EighthCellBL}{\EighthCellBM, \EighthCellBN} \\
 & 4 & \inestimate{\EighthCellBO}{\EighthCellBP, \EighthCellBQ} & \inestimate{\EighthCellBR}{\EighthCellBS, \EighthCellBT} & \inestimate{\EighthCellBU}{\EighthCellBV, \EighthCellBW} & \inestimate{\EighthCellBX}{\EighthCellBY, \EighthCellBZ} \\
 & 5 & \inestimate{\EighthCellCA}{\EighthCellCB, \EighthCellCC} & \inestimate{\EighthCellCD}{\EighthCellCE, \EighthCellCF} & \inestimate{\EighthCellCG}{\EighthCellCH, \EighthCellCI} & \inestimate{\EighthCellCJ}{\EighthCellCK, \EighthCellCL} \\
 & all & \inestimate{\EighthCellCM}{\EighthCellCN, \EighthCellCO} & \inestimate{\EighthCellCP}{\EighthCellCQ, \EighthCellCR} & \inestimate{\EighthCellCS}{\EighthCellCT, \EighthCellCU} & \inestimate{\EighthCellCV}{\EighthCellCW, \EighthCellCX} \\
\rowBone & 1 & \inestimate{\EighthCellCY}{\EighthCellCZ, \EighthCellDA} & \inestimate{\EighthCellDB}{\EighthCellDC, \EighthCellDD} & \inestimate{\EighthCellDE}{\EighthCellDF, \EighthCellDG} & \inestimate{\EighthCellDH}{\EighthCellDI, \EighthCellDJ} \\
\rowBtwo & 2 & \inestimate{\EighthCellDK}{\EighthCellDL, \EighthCellDM} & \inestimate{\EighthCellDN}{\EighthCellDO, \EighthCellDP} & \inestimate{\EighthCellDQ}{\EighthCellDR, \EighthCellDS} & \inestimate{\EighthCellDT}{\EighthCellDU, \EighthCellDV} \\
 & 3 & \inestimate{\EighthCellDW}{\EighthCellDX, \EighthCellDY} & \inestimate{\EighthCellDZ}{\EighthCellEA, \EighthCellEB} & \inestimate{\EighthCellEC}{\EighthCellED, \EighthCellEE} & \inestimate{\EighthCellEF}{\EighthCellEG, \EighthCellEH} \\
 & 4 & \inestimate{\EighthCellEI}{\EighthCellEJ, \EighthCellEK} & \inestimate{\EighthCellEL}{\EighthCellEM, \EighthCellEN} & \inestimate{\EighthCellEO}{\EighthCellEP, \EighthCellEQ} & \inestimate{\EighthCellER}{\EighthCellES, \EighthCellET} \\
 & 5 & \inestimate{\EighthCellEU}{\EighthCellEV, \EighthCellEW} & \inestimate{\EighthCellEX}{\EighthCellEY, \EighthCellEZ} & \inestimate{\EighthCellFA}{\EighthCellFB, \EighthCellFC} & \inestimate{\EighthCellFD}{\EighthCellFE, \EighthCellFF} \\
 & all & \inestimate{\EighthCellFG}{\EighthCellFH, \EighthCellFI} & \inestimate{\EighthCellFJ}{\EighthCellFK, \EighthCellFL} & \inestimate{\EighthCellFM}{\EighthCellFN, \EighthCellFO} & \inestimate{\EighthCellFP}{\EighthCellFQ, \EighthCellFR} \\
\rowCone & 1 & \inestimate{\EighthCellFS}{\EighthCellFT, \EighthCellFU} & \inestimate{\EighthCellFV}{\EighthCellFW, \EighthCellFX} & \inestimate{\EighthCellFY}{\EighthCellFZ, \EighthCellGA} & \inestimate{\EighthCellGB}{\EighthCellGC, \EighthCellGD} \\
\rowCtwo & 2 & \inestimate{\EighthCellGE}{\EighthCellGF, \EighthCellGG} & \inestimate{\EighthCellGH}{\EighthCellGI, \EighthCellGJ} & \inestimate{\EighthCellGK}{\EighthCellGL, \EighthCellGM} & \inestimate{\EighthCellGN}{\EighthCellGO, \EighthCellGP} \\
 & 3 & \inestimate{\EighthCellGQ}{\EighthCellGR, \EighthCellGS} & \inestimate{\EighthCellGT}{\EighthCellGU, \EighthCellGV} & \inestimate{\EighthCellGW}{\EighthCellGX, \EighthCellGY} & \inestimate{\EighthCellGZ}{\EighthCellHA, \EighthCellHB} \\
 & 4 & \inestimate{\EighthCellHC}{\EighthCellHD, \EighthCellHE} & \inestimate{\EighthCellHF}{\EighthCellHG, \EighthCellHH} & \inestimate{\EighthCellHI}{\EighthCellHJ, \EighthCellHK} & \inestimate{\EighthCellHL}{\EighthCellHM, \EighthCellHN} \\
 & 5 & \inestimate{\EighthCellHO}{\EighthCellHP, \EighthCellHQ} & \inestimate{\EighthCellHR}{\EighthCellHS, \EighthCellHT} & \inestimate{\EighthCellHU}{\EighthCellHV, \EighthCellHW} & \inestimate{\EighthCellHX}{\EighthCellHY, \EighthCellHZ} \\
 & all & \inestimate{\EighthCellIA}{\EighthCellIB, \EighthCellIC} & \inestimate{\EighthCellID}{\EighthCellIE, \EighthCellIF} & \inestimate{\EighthCellIG}{\EighthCellIH, \EighthCellII} & \inestimate{\EighthCellIJ}{\EighthCellIK, \EighthCellIL} \\
  \bottomrule
  \end{tabular}
  \tablenote{\noteFittedSelection}
\end{table}

\begin{table}[!htb]
  \caption{\capFittedFresh}
  \label{tab:fitted-fresh}
  \centering\fontsize{9}{10.5}\selectfont
  \setlength{\tabcolsep}{4pt}
  \begin{tabular}{@{}llcccc@{}}
  \toprule
 & Seed & \multicolumn{2}{c}{Layer rule passes, output test fails} & \multicolumn{2}{c}{Layer rule fails, output test passes} \\  & & At \Certify & At \NinthFitJoint & At \Certify & At \NinthFitJoint \\
  \midrule
\rowAone & 1 & \inestimate{\EighthCellIM}{\EighthCellIN, \EighthCellIO} & \inestimate{\EighthCellIP}{\EighthCellIQ, \EighthCellIR} & \inestimate{\EighthCellIS}{\EighthCellIT, \EighthCellIU} & \inestimate{\EighthCellIV}{\EighthCellIW, \EighthCellIX} \\
\rowAtwo & 2 & \inestimate{\EighthCellIY}{\EighthCellIZ, \EighthCellJA} & \inestimate{\EighthCellJB}{\EighthCellJC, \EighthCellJD} & \inestimate{\EighthCellJE}{\EighthCellJF, \EighthCellJG} & \inestimate{\EighthCellJH}{\EighthCellJI, \EighthCellJJ} \\
 & 3 & \inestimate{\EighthCellJK}{\EighthCellJL, \EighthCellJM} & \inestimate{\EighthCellJN}{\EighthCellJO, \EighthCellJP} & \inestimate{\EighthCellJQ}{\EighthCellJR, \EighthCellJS} & \inestimate{\EighthCellJT}{\EighthCellJU, \EighthCellJV} \\
 & 4 & \inestimate{\EighthCellJW}{\EighthCellJX, \EighthCellJY} & \inestimate{\EighthCellJZ}{\EighthCellKA, \EighthCellKB} & \inestimate{\EighthCellKC}{\EighthCellKD, \EighthCellKE} & \inestimate{\EighthCellKF}{\EighthCellKG, \EighthCellKH} \\
 & 5 & \inestimate{\EighthCellKI}{\EighthCellKJ, \EighthCellKK} & \inestimate{\EighthCellKL}{\EighthCellKM, \EighthCellKN} & \inestimate{\EighthCellKO}{\EighthCellKP, \EighthCellKQ} & \inestimate{\EighthCellKR}{\EighthCellKS, \EighthCellKT} \\
 & all & \inestimate{\EighthCellKU}{\EighthCellKV, \EighthCellKW} & \inestimate{\EighthCellKX}{\EighthCellKY, \EighthCellKZ} & \inestimate{\EighthCellLA}{\EighthCellLB, \EighthCellLC} & \inestimate{\EighthCellLD}{\EighthCellLE, \EighthCellLF} \\
\rowBone & 1 & \inestimate{\EighthCellLG}{\EighthCellLH, \EighthCellLI} & \inestimate{\EighthCellLJ}{\EighthCellLK, \EighthCellLL} & \inestimate{\EighthCellLM}{\EighthCellLN, \EighthCellLO} & \inestimate{\EighthCellLP}{\EighthCellLQ, \EighthCellLR} \\
\rowBtwo & 2 & \inestimate{\EighthCellLS}{\EighthCellLT, \EighthCellLU} & \inestimate{\EighthCellLV}{\EighthCellLW, \EighthCellLX} & \inestimate{\EighthCellLY}{\EighthCellLZ, \EighthCellMA} & \inestimate{\EighthCellMB}{\EighthCellMC, \EighthCellMD} \\
 & 3 & \inestimate{\EighthCellME}{\EighthCellMF, \EighthCellMG} & \inestimate{\EighthCellMH}{\EighthCellMI, \EighthCellMJ} & \inestimate{\EighthCellMK}{\EighthCellML, \EighthCellMM} & \inestimate{\EighthCellMN}{\EighthCellMO, \EighthCellMP} \\
 & 4 & \inestimate{\EighthCellMQ}{\EighthCellMR, \EighthCellMS} & \inestimate{\EighthCellMT}{\EighthCellMU, \EighthCellMV} & \inestimate{\EighthCellMW}{\EighthCellMX, \EighthCellMY} & \inestimate{\EighthCellMZ}{\EighthCellNA, \EighthCellNB} \\
 & 5 & \inestimate{\EighthCellNC}{\EighthCellND, \EighthCellNE} & \inestimate{\EighthCellNF}{\EighthCellNG, \EighthCellNH} & \inestimate{\EighthCellNI}{\EighthCellNJ, \EighthCellNK} & \inestimate{\EighthCellNL}{\EighthCellNM, \EighthCellNN} \\
 & all & \inestimate{\EighthCellNO}{\EighthCellNP, \EighthCellNQ} & \inestimate{\EighthCellNR}{\EighthCellNS, \EighthCellNT} & \inestimate{\EighthCellNU}{\EighthCellNV, \EighthCellNW} & \inestimate{\EighthCellNX}{\EighthCellNY, \EighthCellNZ} \\
\rowCone & 1 & \inestimate{\EighthCellOA}{\EighthCellOB, \EighthCellOC} & \inestimate{\EighthCellOD}{\EighthCellOE, \EighthCellOF} & \inestimate{\EighthCellOG}{\EighthCellOH, \EighthCellOI} & \inestimate{\EighthCellOJ}{\EighthCellOK, \EighthCellOL} \\
\rowCtwo & 2 & \inestimate{\EighthCellOM}{\EighthCellON, \EighthCellOO} & \inestimate{\EighthCellOP}{\EighthCellOQ, \EighthCellOR} & \inestimate{\EighthCellOS}{\EighthCellOT, \EighthCellOU} & \inestimate{\EighthCellOV}{\EighthCellOW, \EighthCellOX} \\
 & 3 & \inestimate{\EighthCellOY}{\EighthCellOZ, \EighthCellPA} & \inestimate{\EighthCellPB}{\EighthCellPC, \EighthCellPD} & \inestimate{\EighthCellPE}{\EighthCellPF, \EighthCellPG} & \inestimate{\EighthCellPH}{\EighthCellPI, \EighthCellPJ} \\
 & 4 & \inestimate{\EighthCellPK}{\EighthCellPL, \EighthCellPM} & \inestimate{\EighthCellPN}{\EighthCellPO, \EighthCellPP} & \inestimate{\EighthCellPQ}{\EighthCellPR, \EighthCellPS} & \inestimate{\EighthCellPT}{\EighthCellPU, \EighthCellPV} \\
 & 5 & \inestimate{\EighthCellPW}{\EighthCellPX, \EighthCellPY} & \inestimate{\EighthCellPZ}{\EighthCellQA, \EighthCellQB} & \inestimate{\EighthCellQC}{\EighthCellQD, \EighthCellQE} & \inestimate{\EighthCellQF}{\EighthCellQG, \EighthCellQH} \\
 & all & \inestimate{\EighthCellQI}{\EighthCellQJ, \EighthCellQK} & \inestimate{\EighthCellQL}{\EighthCellQM, \EighthCellQN} & \inestimate{\EighthCellQO}{\EighthCellQP, \EighthCellQQ} & \inestimate{\EighthCellQR}{\EighthCellQS, \EighthCellQT} \\
  \bottomrule
  \end{tabular}
  \tablenote{\noteFittedFresh}
\end{table}

\paragraph{Incorrect predictions by first SAE at three bounds.}
At bounds of \BoundTight, \Certify and \BoundLoose, with the layer rule computed on the selection images, some of its incorrect predictions persist across both halves of both sets in every setting. Yet the upper pointwise image bound on its count of incorrect predictions on the fresh images stays at or below \SeventhFreshUpper of \UnitsTrained pairs. \Cref{tab:supp-site-bounds} counts the layer rule's incorrect predictions by first SAE, one bound to a part, with its prediction fixed on the selection images and with the layer error computed on the same images as the output error\later.

\begin{table}[p]
  \caption{\capSiteBoundsOne}
  \label{tab:supp-site-bounds}
  \centering
  \fontsize{9}{10.5}\selectfont
  \setlength{\tabcolsep}{0.2pt}
  \begin{tabular}{@{}llcccccc@{}}
    \toprule
 & & \multicolumn{3}{c}{Prediction from selection images} & \multicolumn{3}{c}{Prediction from the same images} \\
 & Seed & Scoring & Fresh & All halves & Scoring & Fresh & All halves \\
    \midrule
    \rowAone & 1 & $\SeventhBoundOneAOneSelectionScoring_{\text{\fontsize{9}{10}\selectfont [\SeventhBoundOneAOneSelectionScoringLow,\SeventhBoundOneAOneSelectionScoringHigh]}}$ & $\SeventhBoundOneAOneSelectionFresh_{\text{\fontsize{9}{10}\selectfont [\SeventhBoundOneAOneSelectionFreshLow,\SeventhBoundOneAOneSelectionFreshHigh]}}$ & $\SeventhBoundOneAOneSelectionStable_{\text{\fontsize{9}{10}\selectfont [\SeventhBoundOneAOneSelectionStableLow,\SeventhBoundOneAOneSelectionStableHigh]}}$ & $\SeventhBoundOneAOneSameScoring_{\text{\fontsize{9}{10}\selectfont [\SeventhBoundOneAOneSameScoringLow,\SeventhBoundOneAOneSameScoringHigh]}}$ & $\SeventhBoundOneAOneSameFresh_{\text{\fontsize{9}{10}\selectfont [\SeventhBoundOneAOneSameFreshLow,\SeventhBoundOneAOneSameFreshHigh]}}$ & $\SeventhBoundOneAOneSameStable_{\text{\fontsize{9}{10}\selectfont [\SeventhBoundOneAOneSameStableLow,\SeventhBoundOneAOneSameStableHigh]}}$ \\
    \rowAtwo & 2 & $\SeventhBoundOneATwoSelectionScoring_{\text{\fontsize{9}{10}\selectfont [\SeventhBoundOneATwoSelectionScoringLow,\SeventhBoundOneATwoSelectionScoringHigh]}}$ & $\SeventhBoundOneATwoSelectionFresh_{\text{\fontsize{9}{10}\selectfont [\SeventhBoundOneATwoSelectionFreshLow,\SeventhBoundOneATwoSelectionFreshHigh]}}$ & $\SeventhBoundOneATwoSelectionStable_{\text{\fontsize{9}{10}\selectfont [\SeventhBoundOneATwoSelectionStableLow,\SeventhBoundOneATwoSelectionStableHigh]}}$ & $\SeventhBoundOneATwoSameScoring_{\text{\fontsize{9}{10}\selectfont [\SeventhBoundOneATwoSameScoringLow,\SeventhBoundOneATwoSameScoringHigh]}}$ & $\SeventhBoundOneATwoSameFresh_{\text{\fontsize{9}{10}\selectfont [\SeventhBoundOneATwoSameFreshLow,\SeventhBoundOneATwoSameFreshHigh]}}$ & $\SeventhBoundOneATwoSameStable_{\text{\fontsize{9}{10}\selectfont [\SeventhBoundOneATwoSameStableLow,\SeventhBoundOneATwoSameStableHigh]}}$ \\
     & 3 & $\SeventhBoundOneAThreeSelectionScoring_{\text{\fontsize{9}{10}\selectfont [\SeventhBoundOneAThreeSelectionScoringLow,\SeventhBoundOneAThreeSelectionScoringHigh]}}$ & $\SeventhBoundOneAThreeSelectionFresh_{\text{\fontsize{9}{10}\selectfont [\SeventhBoundOneAThreeSelectionFreshLow,\SeventhBoundOneAThreeSelectionFreshHigh]}}$ & $\SeventhBoundOneAThreeSelectionStable_{\text{\fontsize{9}{10}\selectfont [\SeventhBoundOneAThreeSelectionStableLow,\SeventhBoundOneAThreeSelectionStableHigh]}}$ & $\SeventhBoundOneAThreeSameScoring_{\text{\fontsize{9}{10}\selectfont [\SeventhBoundOneAThreeSameScoringLow,\SeventhBoundOneAThreeSameScoringHigh]}}$ & $\SeventhBoundOneAThreeSameFresh_{\text{\fontsize{9}{10}\selectfont [\SeventhBoundOneAThreeSameFreshLow,\SeventhBoundOneAThreeSameFreshHigh]}}$ & $\SeventhBoundOneAThreeSameStable_{\text{\fontsize{9}{10}\selectfont [\SeventhBoundOneAThreeSameStableLow,\SeventhBoundOneAThreeSameStableHigh]}}$ \\
     & 4 & $\SeventhBoundOneAFourSelectionScoring_{\text{\fontsize{9}{10}\selectfont [\SeventhBoundOneAFourSelectionScoringLow,\SeventhBoundOneAFourSelectionScoringHigh]}}$ & $\SeventhBoundOneAFourSelectionFresh_{\text{\fontsize{9}{10}\selectfont [\SeventhBoundOneAFourSelectionFreshLow,\SeventhBoundOneAFourSelectionFreshHigh]}}$ & $\SeventhBoundOneAFourSelectionStable_{\text{\fontsize{9}{10}\selectfont [\SeventhBoundOneAFourSelectionStableLow,\SeventhBoundOneAFourSelectionStableHigh]}}$ & $\SeventhBoundOneAFourSameScoring_{\text{\fontsize{9}{10}\selectfont [\SeventhBoundOneAFourSameScoringLow,\SeventhBoundOneAFourSameScoringHigh]}}$ & $\SeventhBoundOneAFourSameFresh_{\text{\fontsize{9}{10}\selectfont [\SeventhBoundOneAFourSameFreshLow,\SeventhBoundOneAFourSameFreshHigh]}}$ & $\SeventhBoundOneAFourSameStable_{\text{\fontsize{9}{10}\selectfont [\SeventhBoundOneAFourSameStableLow,\SeventhBoundOneAFourSameStableHigh]}}$ \\
     & 5 & $\SeventhBoundOneAFiveSelectionScoring_{\text{\fontsize{9}{10}\selectfont [\SeventhBoundOneAFiveSelectionScoringLow,\SeventhBoundOneAFiveSelectionScoringHigh]}}$ & $\SeventhBoundOneAFiveSelectionFresh_{\text{\fontsize{9}{10}\selectfont [\SeventhBoundOneAFiveSelectionFreshLow,\SeventhBoundOneAFiveSelectionFreshHigh]}}$ & $\SeventhBoundOneAFiveSelectionStable_{\text{\fontsize{9}{10}\selectfont [\SeventhBoundOneAFiveSelectionStableLow,\SeventhBoundOneAFiveSelectionStableHigh]}}$ & $\SeventhBoundOneAFiveSameScoring_{\text{\fontsize{9}{10}\selectfont [\SeventhBoundOneAFiveSameScoringLow,\SeventhBoundOneAFiveSameScoringHigh]}}$ & $\SeventhBoundOneAFiveSameFresh_{\text{\fontsize{9}{10}\selectfont [\SeventhBoundOneAFiveSameFreshLow,\SeventhBoundOneAFiveSameFreshHigh]}}$ & $\SeventhBoundOneAFiveSameStable_{\text{\fontsize{9}{10}\selectfont [\SeventhBoundOneAFiveSameStableLow,\SeventhBoundOneAFiveSameStableHigh]}}$ \\
     & all & $\SeventhBoundOneAAllSelectionScoring_{\text{\fontsize{9}{10}\selectfont [\SeventhBoundOneAAllSelectionScoringLow,\SeventhBoundOneAAllSelectionScoringHigh]}}$ & $\SeventhBoundOneAAllSelectionFresh_{\text{\fontsize{9}{10}\selectfont [\SeventhBoundOneAAllSelectionFreshLow,\SeventhBoundOneAAllSelectionFreshHigh]}}$ & $\SeventhBoundOneAAllSelectionStable_{\text{\fontsize{9}{10}\selectfont [\SeventhBoundOneAAllSelectionStableLow,\SeventhBoundOneAAllSelectionStableHigh]}}$ & $\SeventhBoundOneAAllSameScoring_{\text{\fontsize{9}{10}\selectfont [\SeventhBoundOneAAllSameScoringLow,\SeventhBoundOneAAllSameScoringHigh]}}$ & $\SeventhBoundOneAAllSameFresh_{\text{\fontsize{9}{10}\selectfont [\SeventhBoundOneAAllSameFreshLow,\SeventhBoundOneAAllSameFreshHigh]}}$ & $\SeventhBoundOneAAllSameStable_{\text{\fontsize{9}{10}\selectfont [\SeventhBoundOneAAllSameStableLow,\SeventhBoundOneAAllSameStableHigh]}}$ \\
    \midrule
    \rowBone & 1 & $\SeventhBoundOneBOneSelectionScoring_{\text{\fontsize{9}{10}\selectfont [\SeventhBoundOneBOneSelectionScoringLow,\SeventhBoundOneBOneSelectionScoringHigh]}}$ & $\SeventhBoundOneBOneSelectionFresh_{\text{\fontsize{9}{10}\selectfont [\SeventhBoundOneBOneSelectionFreshLow,\SeventhBoundOneBOneSelectionFreshHigh]}}$ & $\SeventhBoundOneBOneSelectionStable_{\text{\fontsize{9}{10}\selectfont [\SeventhBoundOneBOneSelectionStableLow,\SeventhBoundOneBOneSelectionStableHigh]}}$ & $\SeventhBoundOneBOneSameScoring_{\text{\fontsize{9}{10}\selectfont [\SeventhBoundOneBOneSameScoringLow,\SeventhBoundOneBOneSameScoringHigh]}}$ & $\SeventhBoundOneBOneSameFresh_{\text{\fontsize{9}{10}\selectfont [\SeventhBoundOneBOneSameFreshLow,\SeventhBoundOneBOneSameFreshHigh]}}$ & $\SeventhBoundOneBOneSameStable_{\text{\fontsize{9}{10}\selectfont [\SeventhBoundOneBOneSameStableLow,\SeventhBoundOneBOneSameStableHigh]}}$ \\
    \rowBtwo & 2 & $\SeventhBoundOneBTwoSelectionScoring_{\text{\fontsize{9}{10}\selectfont [\SeventhBoundOneBTwoSelectionScoringLow,\SeventhBoundOneBTwoSelectionScoringHigh]}}$ & $\SeventhBoundOneBTwoSelectionFresh_{\text{\fontsize{9}{10}\selectfont [\SeventhBoundOneBTwoSelectionFreshLow,\SeventhBoundOneBTwoSelectionFreshHigh]}}$ & $\SeventhBoundOneBTwoSelectionStable_{\text{\fontsize{9}{10}\selectfont [\SeventhBoundOneBTwoSelectionStableLow,\SeventhBoundOneBTwoSelectionStableHigh]}}$ & $\SeventhBoundOneBTwoSameScoring_{\text{\fontsize{9}{10}\selectfont [\SeventhBoundOneBTwoSameScoringLow,\SeventhBoundOneBTwoSameScoringHigh]}}$ & $\SeventhBoundOneBTwoSameFresh_{\text{\fontsize{9}{10}\selectfont [\SeventhBoundOneBTwoSameFreshLow,\SeventhBoundOneBTwoSameFreshHigh]}}$ & $\SeventhBoundOneBTwoSameStable_{\text{\fontsize{9}{10}\selectfont [\SeventhBoundOneBTwoSameStableLow,\SeventhBoundOneBTwoSameStableHigh]}}$ \\
     & 3 & $\SeventhBoundOneBThreeSelectionScoring_{\text{\fontsize{9}{10}\selectfont [\SeventhBoundOneBThreeSelectionScoringLow,\SeventhBoundOneBThreeSelectionScoringHigh]}}$ & $\SeventhBoundOneBThreeSelectionFresh_{\text{\fontsize{9}{10}\selectfont [\SeventhBoundOneBThreeSelectionFreshLow,\SeventhBoundOneBThreeSelectionFreshHigh]}}$ & $\SeventhBoundOneBThreeSelectionStable_{\text{\fontsize{9}{10}\selectfont [\SeventhBoundOneBThreeSelectionStableLow,\SeventhBoundOneBThreeSelectionStableHigh]}}$ & $\SeventhBoundOneBThreeSameScoring_{\text{\fontsize{9}{10}\selectfont [\SeventhBoundOneBThreeSameScoringLow,\SeventhBoundOneBThreeSameScoringHigh]}}$ & $\SeventhBoundOneBThreeSameFresh_{\text{\fontsize{9}{10}\selectfont [\SeventhBoundOneBThreeSameFreshLow,\SeventhBoundOneBThreeSameFreshHigh]}}$ & $\SeventhBoundOneBThreeSameStable_{\text{\fontsize{9}{10}\selectfont [\SeventhBoundOneBThreeSameStableLow,\SeventhBoundOneBThreeSameStableHigh]}}$ \\
     & 4 & $\SeventhBoundOneBFourSelectionScoring_{\text{\fontsize{9}{10}\selectfont [\SeventhBoundOneBFourSelectionScoringLow,\SeventhBoundOneBFourSelectionScoringHigh]}}$ & $\SeventhBoundOneBFourSelectionFresh_{\text{\fontsize{9}{10}\selectfont [\SeventhBoundOneBFourSelectionFreshLow,\SeventhBoundOneBFourSelectionFreshHigh]}}$ & $\SeventhBoundOneBFourSelectionStable_{\text{\fontsize{9}{10}\selectfont [\SeventhBoundOneBFourSelectionStableLow,\SeventhBoundOneBFourSelectionStableHigh]}}$ & $\SeventhBoundOneBFourSameScoring_{\text{\fontsize{9}{10}\selectfont [\SeventhBoundOneBFourSameScoringLow,\SeventhBoundOneBFourSameScoringHigh]}}$ & $\SeventhBoundOneBFourSameFresh_{\text{\fontsize{9}{10}\selectfont [\SeventhBoundOneBFourSameFreshLow,\SeventhBoundOneBFourSameFreshHigh]}}$ & $\SeventhBoundOneBFourSameStable_{\text{\fontsize{9}{10}\selectfont [\SeventhBoundOneBFourSameStableLow,\SeventhBoundOneBFourSameStableHigh]}}$ \\
     & 5 & $\SeventhBoundOneBFiveSelectionScoring_{\text{\fontsize{9}{10}\selectfont [\SeventhBoundOneBFiveSelectionScoringLow,\SeventhBoundOneBFiveSelectionScoringHigh]}}$ & $\SeventhBoundOneBFiveSelectionFresh_{\text{\fontsize{9}{10}\selectfont [\SeventhBoundOneBFiveSelectionFreshLow,\SeventhBoundOneBFiveSelectionFreshHigh]}}$ & $\SeventhBoundOneBFiveSelectionStable_{\text{\fontsize{9}{10}\selectfont [\SeventhBoundOneBFiveSelectionStableLow,\SeventhBoundOneBFiveSelectionStableHigh]}}$ & $\SeventhBoundOneBFiveSameScoring_{\text{\fontsize{9}{10}\selectfont [\SeventhBoundOneBFiveSameScoringLow,\SeventhBoundOneBFiveSameScoringHigh]}}$ & $\SeventhBoundOneBFiveSameFresh_{\text{\fontsize{9}{10}\selectfont [\SeventhBoundOneBFiveSameFreshLow,\SeventhBoundOneBFiveSameFreshHigh]}}$ & $\SeventhBoundOneBFiveSameStable_{\text{\fontsize{9}{10}\selectfont [\SeventhBoundOneBFiveSameStableLow,\SeventhBoundOneBFiveSameStableHigh]}}$ \\
     & all & $\SeventhBoundOneBAllSelectionScoring_{\text{\fontsize{9}{10}\selectfont [\SeventhBoundOneBAllSelectionScoringLow,\SeventhBoundOneBAllSelectionScoringHigh]}}$ & $\SeventhBoundOneBAllSelectionFresh_{\text{\fontsize{9}{10}\selectfont [\SeventhBoundOneBAllSelectionFreshLow,\SeventhBoundOneBAllSelectionFreshHigh]}}$ & $\SeventhBoundOneBAllSelectionStable_{\text{\fontsize{9}{10}\selectfont [\SeventhBoundOneBAllSelectionStableLow,\SeventhBoundOneBAllSelectionStableHigh]}}$ & $\SeventhBoundOneBAllSameScoring_{\text{\fontsize{9}{10}\selectfont [\SeventhBoundOneBAllSameScoringLow,\SeventhBoundOneBAllSameScoringHigh]}}$ & $\SeventhBoundOneBAllSameFresh_{\text{\fontsize{9}{10}\selectfont [\SeventhBoundOneBAllSameFreshLow,\SeventhBoundOneBAllSameFreshHigh]}}$ & $\SeventhBoundOneBAllSameStable_{\text{\fontsize{9}{10}\selectfont [\SeventhBoundOneBAllSameStableLow,\SeventhBoundOneBAllSameStableHigh]}}$ \\
    \midrule
    \rowCone & 1 & $\SeventhBoundOneCOneSelectionScoring_{\text{\fontsize{9}{10}\selectfont [\SeventhBoundOneCOneSelectionScoringLow,\SeventhBoundOneCOneSelectionScoringHigh]}}$ & $\SeventhBoundOneCOneSelectionFresh_{\text{\fontsize{9}{10}\selectfont [\SeventhBoundOneCOneSelectionFreshLow,\SeventhBoundOneCOneSelectionFreshHigh]}}$ & $\SeventhBoundOneCOneSelectionStable_{\text{\fontsize{9}{10}\selectfont [\SeventhBoundOneCOneSelectionStableLow,\SeventhBoundOneCOneSelectionStableHigh]}}$ & $\SeventhBoundOneCOneSameScoring_{\text{\fontsize{9}{10}\selectfont [\SeventhBoundOneCOneSameScoringLow,\SeventhBoundOneCOneSameScoringHigh]}}$ & $\SeventhBoundOneCOneSameFresh_{\text{\fontsize{9}{10}\selectfont [\SeventhBoundOneCOneSameFreshLow,\SeventhBoundOneCOneSameFreshHigh]}}$ & $\SeventhBoundOneCOneSameStable_{\text{\fontsize{9}{10}\selectfont [\SeventhBoundOneCOneSameStableLow,\SeventhBoundOneCOneSameStableHigh]}}$ \\
    \rowCtwo & 2 & $\SeventhBoundOneCTwoSelectionScoring_{\text{\fontsize{9}{10}\selectfont [\SeventhBoundOneCTwoSelectionScoringLow,\SeventhBoundOneCTwoSelectionScoringHigh]}}$ & $\SeventhBoundOneCTwoSelectionFresh_{\text{\fontsize{9}{10}\selectfont [\SeventhBoundOneCTwoSelectionFreshLow,\SeventhBoundOneCTwoSelectionFreshHigh]}}$ & $\SeventhBoundOneCTwoSelectionStable_{\text{\fontsize{9}{10}\selectfont [\SeventhBoundOneCTwoSelectionStableLow,\SeventhBoundOneCTwoSelectionStableHigh]}}$ & $\SeventhBoundOneCTwoSameScoring_{\text{\fontsize{9}{10}\selectfont [\SeventhBoundOneCTwoSameScoringLow,\SeventhBoundOneCTwoSameScoringHigh]}}$ & $\SeventhBoundOneCTwoSameFresh_{\text{\fontsize{9}{10}\selectfont [\SeventhBoundOneCTwoSameFreshLow,\SeventhBoundOneCTwoSameFreshHigh]}}$ & $\SeventhBoundOneCTwoSameStable_{\text{\fontsize{9}{10}\selectfont [\SeventhBoundOneCTwoSameStableLow,\SeventhBoundOneCTwoSameStableHigh]}}$ \\
     & 3 & $\SeventhBoundOneCThreeSelectionScoring_{\text{\fontsize{9}{10}\selectfont [\SeventhBoundOneCThreeSelectionScoringLow,\SeventhBoundOneCThreeSelectionScoringHigh]}}$ & $\SeventhBoundOneCThreeSelectionFresh_{\text{\fontsize{9}{10}\selectfont [\SeventhBoundOneCThreeSelectionFreshLow,\SeventhBoundOneCThreeSelectionFreshHigh]}}$ & $\SeventhBoundOneCThreeSelectionStable_{\text{\fontsize{9}{10}\selectfont [\SeventhBoundOneCThreeSelectionStableLow,\SeventhBoundOneCThreeSelectionStableHigh]}}$ & $\SeventhBoundOneCThreeSameScoring_{\text{\fontsize{9}{10}\selectfont [\SeventhBoundOneCThreeSameScoringLow,\SeventhBoundOneCThreeSameScoringHigh]}}$ & $\SeventhBoundOneCThreeSameFresh_{\text{\fontsize{9}{10}\selectfont [\SeventhBoundOneCThreeSameFreshLow,\SeventhBoundOneCThreeSameFreshHigh]}}$ & $\SeventhBoundOneCThreeSameStable_{\text{\fontsize{9}{10}\selectfont [\SeventhBoundOneCThreeSameStableLow,\SeventhBoundOneCThreeSameStableHigh]}}$ \\
     & 4 & $\SeventhBoundOneCFourSelectionScoring_{\text{\fontsize{9}{10}\selectfont [\SeventhBoundOneCFourSelectionScoringLow,\SeventhBoundOneCFourSelectionScoringHigh]}}$ & $\SeventhBoundOneCFourSelectionFresh_{\text{\fontsize{9}{10}\selectfont [\SeventhBoundOneCFourSelectionFreshLow,\SeventhBoundOneCFourSelectionFreshHigh]}}$ & $\SeventhBoundOneCFourSelectionStable_{\text{\fontsize{9}{10}\selectfont [\SeventhBoundOneCFourSelectionStableLow,\SeventhBoundOneCFourSelectionStableHigh]}}$ & $\SeventhBoundOneCFourSameScoring_{\text{\fontsize{9}{10}\selectfont [\SeventhBoundOneCFourSameScoringLow,\SeventhBoundOneCFourSameScoringHigh]}}$ & $\SeventhBoundOneCFourSameFresh_{\text{\fontsize{9}{10}\selectfont [\SeventhBoundOneCFourSameFreshLow,\SeventhBoundOneCFourSameFreshHigh]}}$ & $\SeventhBoundOneCFourSameStable_{\text{\fontsize{9}{10}\selectfont [\SeventhBoundOneCFourSameStableLow,\SeventhBoundOneCFourSameStableHigh]}}$ \\
     & 5 & $\SeventhBoundOneCFiveSelectionScoring_{\text{\fontsize{9}{10}\selectfont [\SeventhBoundOneCFiveSelectionScoringLow,\SeventhBoundOneCFiveSelectionScoringHigh]}}$ & $\SeventhBoundOneCFiveSelectionFresh_{\text{\fontsize{9}{10}\selectfont [\SeventhBoundOneCFiveSelectionFreshLow,\SeventhBoundOneCFiveSelectionFreshHigh]}}$ & $\SeventhBoundOneCFiveSelectionStable_{\text{\fontsize{9}{10}\selectfont [\SeventhBoundOneCFiveSelectionStableLow,\SeventhBoundOneCFiveSelectionStableHigh]}}$ & $\SeventhBoundOneCFiveSameScoring_{\text{\fontsize{9}{10}\selectfont [\SeventhBoundOneCFiveSameScoringLow,\SeventhBoundOneCFiveSameScoringHigh]}}$ & $\SeventhBoundOneCFiveSameFresh_{\text{\fontsize{9}{10}\selectfont [\SeventhBoundOneCFiveSameFreshLow,\SeventhBoundOneCFiveSameFreshHigh]}}$ & $\SeventhBoundOneCFiveSameStable_{\text{\fontsize{9}{10}\selectfont [\SeventhBoundOneCFiveSameStableLow,\SeventhBoundOneCFiveSameStableHigh]}}$ \\
     & all & $\SeventhBoundOneCAllSelectionScoring_{\text{\fontsize{9}{10}\selectfont [\SeventhBoundOneCAllSelectionScoringLow,\SeventhBoundOneCAllSelectionScoringHigh]}}$ & $\SeventhBoundOneCAllSelectionFresh_{\text{\fontsize{9}{10}\selectfont [\SeventhBoundOneCAllSelectionFreshLow,\SeventhBoundOneCAllSelectionFreshHigh]}}$ & $\SeventhBoundOneCAllSelectionStable_{\text{\fontsize{9}{10}\selectfont [\SeventhBoundOneCAllSelectionStableLow,\SeventhBoundOneCAllSelectionStableHigh]}}$ & $\SeventhBoundOneCAllSameScoring_{\text{\fontsize{9}{10}\selectfont [\SeventhBoundOneCAllSameScoringLow,\SeventhBoundOneCAllSameScoringHigh]}}$ & $\SeventhBoundOneCAllSameFresh_{\text{\fontsize{9}{10}\selectfont [\SeventhBoundOneCAllSameFreshLow,\SeventhBoundOneCAllSameFreshHigh]}}$ & $\SeventhBoundOneCAllSameStable_{\text{\fontsize{9}{10}\selectfont [\SeventhBoundOneCAllSameStableLow,\SeventhBoundOneCAllSameStableHigh]}}$ \\
    \midrule
    \bottomrule
  \end{tabular}
  \tablenote{\noteSiteBounds}
\end{table}
\begin{table}[p]
\ContinuedFloat
  \caption{\capSiteBoundsTwo}
  \centering
  \fontsize{9}{10.5}\selectfont
  \setlength{\tabcolsep}{0.2pt}
  \begin{tabular}{@{}llcccccc@{}}
    \toprule
 & & \multicolumn{3}{c}{Prediction from selection images} & \multicolumn{3}{c}{Prediction from the same images} \\
 & Seed & Scoring & Fresh & All halves & Scoring & Fresh & All halves \\
    \midrule
    \rowAone & 1 & $\SeventhBoundTwoAOneSelectionScoring_{\text{\fontsize{9}{10}\selectfont [\SeventhBoundTwoAOneSelectionScoringLow,\SeventhBoundTwoAOneSelectionScoringHigh]}}$ & $\SeventhBoundTwoAOneSelectionFresh_{\text{\fontsize{9}{10}\selectfont [\SeventhBoundTwoAOneSelectionFreshLow,\SeventhBoundTwoAOneSelectionFreshHigh]}}$ & $\SeventhBoundTwoAOneSelectionStable_{\text{\fontsize{9}{10}\selectfont [\SeventhBoundTwoAOneSelectionStableLow,\SeventhBoundTwoAOneSelectionStableHigh]}}$ & $\SeventhBoundTwoAOneSameScoring_{\text{\fontsize{9}{10}\selectfont [\SeventhBoundTwoAOneSameScoringLow,\SeventhBoundTwoAOneSameScoringHigh]}}$ & $\SeventhBoundTwoAOneSameFresh_{\text{\fontsize{9}{10}\selectfont [\SeventhBoundTwoAOneSameFreshLow,\SeventhBoundTwoAOneSameFreshHigh]}}$ & $\SeventhBoundTwoAOneSameStable_{\text{\fontsize{9}{10}\selectfont [\SeventhBoundTwoAOneSameStableLow,\SeventhBoundTwoAOneSameStableHigh]}}$ \\
    \rowAtwo & 2 & $\SeventhBoundTwoATwoSelectionScoring_{\text{\fontsize{9}{10}\selectfont [\SeventhBoundTwoATwoSelectionScoringLow,\SeventhBoundTwoATwoSelectionScoringHigh]}}$ & $\SeventhBoundTwoATwoSelectionFresh_{\text{\fontsize{9}{10}\selectfont [\SeventhBoundTwoATwoSelectionFreshLow,\SeventhBoundTwoATwoSelectionFreshHigh]}}$ & $\SeventhBoundTwoATwoSelectionStable_{\text{\fontsize{9}{10}\selectfont [\SeventhBoundTwoATwoSelectionStableLow,\SeventhBoundTwoATwoSelectionStableHigh]}}$ & $\SeventhBoundTwoATwoSameScoring_{\text{\fontsize{9}{10}\selectfont [\SeventhBoundTwoATwoSameScoringLow,\SeventhBoundTwoATwoSameScoringHigh]}}$ & $\SeventhBoundTwoATwoSameFresh_{\text{\fontsize{9}{10}\selectfont [\SeventhBoundTwoATwoSameFreshLow,\SeventhBoundTwoATwoSameFreshHigh]}}$ & $\SeventhBoundTwoATwoSameStable_{\text{\fontsize{9}{10}\selectfont [\SeventhBoundTwoATwoSameStableLow,\SeventhBoundTwoATwoSameStableHigh]}}$ \\
     & 3 & $\SeventhBoundTwoAThreeSelectionScoring_{\text{\fontsize{9}{10}\selectfont [\SeventhBoundTwoAThreeSelectionScoringLow,\SeventhBoundTwoAThreeSelectionScoringHigh]}}$ & $\SeventhBoundTwoAThreeSelectionFresh_{\text{\fontsize{9}{10}\selectfont [\SeventhBoundTwoAThreeSelectionFreshLow,\SeventhBoundTwoAThreeSelectionFreshHigh]}}$ & $\SeventhBoundTwoAThreeSelectionStable_{\text{\fontsize{9}{10}\selectfont [\SeventhBoundTwoAThreeSelectionStableLow,\SeventhBoundTwoAThreeSelectionStableHigh]}}$ & $\SeventhBoundTwoAThreeSameScoring_{\text{\fontsize{9}{10}\selectfont [\SeventhBoundTwoAThreeSameScoringLow,\SeventhBoundTwoAThreeSameScoringHigh]}}$ & $\SeventhBoundTwoAThreeSameFresh_{\text{\fontsize{9}{10}\selectfont [\SeventhBoundTwoAThreeSameFreshLow,\SeventhBoundTwoAThreeSameFreshHigh]}}$ & $\SeventhBoundTwoAThreeSameStable_{\text{\fontsize{9}{10}\selectfont [\SeventhBoundTwoAThreeSameStableLow,\SeventhBoundTwoAThreeSameStableHigh]}}$ \\
     & 4 & $\SeventhBoundTwoAFourSelectionScoring_{\text{\fontsize{9}{10}\selectfont [\SeventhBoundTwoAFourSelectionScoringLow,\SeventhBoundTwoAFourSelectionScoringHigh]}}$ & $\SeventhBoundTwoAFourSelectionFresh_{\text{\fontsize{9}{10}\selectfont [\SeventhBoundTwoAFourSelectionFreshLow,\SeventhBoundTwoAFourSelectionFreshHigh]}}$ & $\SeventhBoundTwoAFourSelectionStable_{\text{\fontsize{9}{10}\selectfont [\SeventhBoundTwoAFourSelectionStableLow,\SeventhBoundTwoAFourSelectionStableHigh]}}$ & $\SeventhBoundTwoAFourSameScoring_{\text{\fontsize{9}{10}\selectfont [\SeventhBoundTwoAFourSameScoringLow,\SeventhBoundTwoAFourSameScoringHigh]}}$ & $\SeventhBoundTwoAFourSameFresh_{\text{\fontsize{9}{10}\selectfont [\SeventhBoundTwoAFourSameFreshLow,\SeventhBoundTwoAFourSameFreshHigh]}}$ & $\SeventhBoundTwoAFourSameStable_{\text{\fontsize{9}{10}\selectfont [\SeventhBoundTwoAFourSameStableLow,\SeventhBoundTwoAFourSameStableHigh]}}$ \\
     & 5 & $\SeventhBoundTwoAFiveSelectionScoring_{\text{\fontsize{9}{10}\selectfont [\SeventhBoundTwoAFiveSelectionScoringLow,\SeventhBoundTwoAFiveSelectionScoringHigh]}}$ & $\SeventhBoundTwoAFiveSelectionFresh_{\text{\fontsize{9}{10}\selectfont [\SeventhBoundTwoAFiveSelectionFreshLow,\SeventhBoundTwoAFiveSelectionFreshHigh]}}$ & $\SeventhBoundTwoAFiveSelectionStable_{\text{\fontsize{9}{10}\selectfont [\SeventhBoundTwoAFiveSelectionStableLow,\SeventhBoundTwoAFiveSelectionStableHigh]}}$ & $\SeventhBoundTwoAFiveSameScoring_{\text{\fontsize{9}{10}\selectfont [\SeventhBoundTwoAFiveSameScoringLow,\SeventhBoundTwoAFiveSameScoringHigh]}}$ & $\SeventhBoundTwoAFiveSameFresh_{\text{\fontsize{9}{10}\selectfont [\SeventhBoundTwoAFiveSameFreshLow,\SeventhBoundTwoAFiveSameFreshHigh]}}$ & $\SeventhBoundTwoAFiveSameStable_{\text{\fontsize{9}{10}\selectfont [\SeventhBoundTwoAFiveSameStableLow,\SeventhBoundTwoAFiveSameStableHigh]}}$ \\
     & all & $\SeventhBoundTwoAAllSelectionScoring_{\text{\fontsize{9}{10}\selectfont [\SeventhBoundTwoAAllSelectionScoringLow,\SeventhBoundTwoAAllSelectionScoringHigh]}}$ & $\SeventhBoundTwoAAllSelectionFresh_{\text{\fontsize{9}{10}\selectfont [\SeventhBoundTwoAAllSelectionFreshLow,\SeventhBoundTwoAAllSelectionFreshHigh]}}$ & $\SeventhBoundTwoAAllSelectionStable_{\text{\fontsize{9}{10}\selectfont [\SeventhBoundTwoAAllSelectionStableLow,\SeventhBoundTwoAAllSelectionStableHigh]}}$ & $\SeventhBoundTwoAAllSameScoring_{\text{\fontsize{9}{10}\selectfont [\SeventhBoundTwoAAllSameScoringLow,\SeventhBoundTwoAAllSameScoringHigh]}}$ & $\SeventhBoundTwoAAllSameFresh_{\text{\fontsize{9}{10}\selectfont [\SeventhBoundTwoAAllSameFreshLow,\SeventhBoundTwoAAllSameFreshHigh]}}$ & $\SeventhBoundTwoAAllSameStable_{\text{\fontsize{9}{10}\selectfont [\SeventhBoundTwoAAllSameStableLow,\SeventhBoundTwoAAllSameStableHigh]}}$ \\
    \midrule
    \rowBone & 1 & $\SeventhBoundTwoBOneSelectionScoring_{\text{\fontsize{9}{10}\selectfont [\SeventhBoundTwoBOneSelectionScoringLow,\SeventhBoundTwoBOneSelectionScoringHigh]}}$ & $\SeventhBoundTwoBOneSelectionFresh_{\text{\fontsize{9}{10}\selectfont [\SeventhBoundTwoBOneSelectionFreshLow,\SeventhBoundTwoBOneSelectionFreshHigh]}}$ & $\SeventhBoundTwoBOneSelectionStable_{\text{\fontsize{9}{10}\selectfont [\SeventhBoundTwoBOneSelectionStableLow,\SeventhBoundTwoBOneSelectionStableHigh]}}$ & $\SeventhBoundTwoBOneSameScoring_{\text{\fontsize{9}{10}\selectfont [\SeventhBoundTwoBOneSameScoringLow,\SeventhBoundTwoBOneSameScoringHigh]}}$ & $\SeventhBoundTwoBOneSameFresh_{\text{\fontsize{9}{10}\selectfont [\SeventhBoundTwoBOneSameFreshLow,\SeventhBoundTwoBOneSameFreshHigh]}}$ & $\SeventhBoundTwoBOneSameStable_{\text{\fontsize{9}{10}\selectfont [\SeventhBoundTwoBOneSameStableLow,\SeventhBoundTwoBOneSameStableHigh]}}$ \\
    \rowBtwo & 2 & $\SeventhBoundTwoBTwoSelectionScoring_{\text{\fontsize{9}{10}\selectfont [\SeventhBoundTwoBTwoSelectionScoringLow,\SeventhBoundTwoBTwoSelectionScoringHigh]}}$ & $\SeventhBoundTwoBTwoSelectionFresh_{\text{\fontsize{9}{10}\selectfont [\SeventhBoundTwoBTwoSelectionFreshLow,\SeventhBoundTwoBTwoSelectionFreshHigh]}}$ & $\SeventhBoundTwoBTwoSelectionStable_{\text{\fontsize{9}{10}\selectfont [\SeventhBoundTwoBTwoSelectionStableLow,\SeventhBoundTwoBTwoSelectionStableHigh]}}$ & $\SeventhBoundTwoBTwoSameScoring_{\text{\fontsize{9}{10}\selectfont [\SeventhBoundTwoBTwoSameScoringLow,\SeventhBoundTwoBTwoSameScoringHigh]}}$ & $\SeventhBoundTwoBTwoSameFresh_{\text{\fontsize{9}{10}\selectfont [\SeventhBoundTwoBTwoSameFreshLow,\SeventhBoundTwoBTwoSameFreshHigh]}}$ & $\SeventhBoundTwoBTwoSameStable_{\text{\fontsize{9}{10}\selectfont [\SeventhBoundTwoBTwoSameStableLow,\SeventhBoundTwoBTwoSameStableHigh]}}$ \\
     & 3 & $\SeventhBoundTwoBThreeSelectionScoring_{\text{\fontsize{9}{10}\selectfont [\SeventhBoundTwoBThreeSelectionScoringLow,\SeventhBoundTwoBThreeSelectionScoringHigh]}}$ & $\SeventhBoundTwoBThreeSelectionFresh_{\text{\fontsize{9}{10}\selectfont [\SeventhBoundTwoBThreeSelectionFreshLow,\SeventhBoundTwoBThreeSelectionFreshHigh]}}$ & $\SeventhBoundTwoBThreeSelectionStable_{\text{\fontsize{9}{10}\selectfont [\SeventhBoundTwoBThreeSelectionStableLow,\SeventhBoundTwoBThreeSelectionStableHigh]}}$ & $\SeventhBoundTwoBThreeSameScoring_{\text{\fontsize{9}{10}\selectfont [\SeventhBoundTwoBThreeSameScoringLow,\SeventhBoundTwoBThreeSameScoringHigh]}}$ & $\SeventhBoundTwoBThreeSameFresh_{\text{\fontsize{9}{10}\selectfont [\SeventhBoundTwoBThreeSameFreshLow,\SeventhBoundTwoBThreeSameFreshHigh]}}$ & $\SeventhBoundTwoBThreeSameStable_{\text{\fontsize{9}{10}\selectfont [\SeventhBoundTwoBThreeSameStableLow,\SeventhBoundTwoBThreeSameStableHigh]}}$ \\
     & 4 & $\SeventhBoundTwoBFourSelectionScoring_{\text{\fontsize{9}{10}\selectfont [\SeventhBoundTwoBFourSelectionScoringLow,\SeventhBoundTwoBFourSelectionScoringHigh]}}$ & $\SeventhBoundTwoBFourSelectionFresh_{\text{\fontsize{9}{10}\selectfont [\SeventhBoundTwoBFourSelectionFreshLow,\SeventhBoundTwoBFourSelectionFreshHigh]}}$ & $\SeventhBoundTwoBFourSelectionStable_{\text{\fontsize{9}{10}\selectfont [\SeventhBoundTwoBFourSelectionStableLow,\SeventhBoundTwoBFourSelectionStableHigh]}}$ & $\SeventhBoundTwoBFourSameScoring_{\text{\fontsize{9}{10}\selectfont [\SeventhBoundTwoBFourSameScoringLow,\SeventhBoundTwoBFourSameScoringHigh]}}$ & $\SeventhBoundTwoBFourSameFresh_{\text{\fontsize{9}{10}\selectfont [\SeventhBoundTwoBFourSameFreshLow,\SeventhBoundTwoBFourSameFreshHigh]}}$ & $\SeventhBoundTwoBFourSameStable_{\text{\fontsize{9}{10}\selectfont [\SeventhBoundTwoBFourSameStableLow,\SeventhBoundTwoBFourSameStableHigh]}}$ \\
     & 5 & $\SeventhBoundTwoBFiveSelectionScoring_{\text{\fontsize{9}{10}\selectfont [\SeventhBoundTwoBFiveSelectionScoringLow,\SeventhBoundTwoBFiveSelectionScoringHigh]}}$ & $\SeventhBoundTwoBFiveSelectionFresh_{\text{\fontsize{9}{10}\selectfont [\SeventhBoundTwoBFiveSelectionFreshLow,\SeventhBoundTwoBFiveSelectionFreshHigh]}}$ & $\SeventhBoundTwoBFiveSelectionStable_{\text{\fontsize{9}{10}\selectfont [\SeventhBoundTwoBFiveSelectionStableLow,\SeventhBoundTwoBFiveSelectionStableHigh]}}$ & $\SeventhBoundTwoBFiveSameScoring_{\text{\fontsize{9}{10}\selectfont [\SeventhBoundTwoBFiveSameScoringLow,\SeventhBoundTwoBFiveSameScoringHigh]}}$ & $\SeventhBoundTwoBFiveSameFresh_{\text{\fontsize{9}{10}\selectfont [\SeventhBoundTwoBFiveSameFreshLow,\SeventhBoundTwoBFiveSameFreshHigh]}}$ & $\SeventhBoundTwoBFiveSameStable_{\text{\fontsize{9}{10}\selectfont [\SeventhBoundTwoBFiveSameStableLow,\SeventhBoundTwoBFiveSameStableHigh]}}$ \\
     & all & $\SeventhBoundTwoBAllSelectionScoring_{\text{\fontsize{9}{10}\selectfont [\SeventhBoundTwoBAllSelectionScoringLow,\SeventhBoundTwoBAllSelectionScoringHigh]}}$ & $\SeventhBoundTwoBAllSelectionFresh_{\text{\fontsize{9}{10}\selectfont [\SeventhBoundTwoBAllSelectionFreshLow,\SeventhBoundTwoBAllSelectionFreshHigh]}}$ & $\SeventhBoundTwoBAllSelectionStable_{\text{\fontsize{9}{10}\selectfont [\SeventhBoundTwoBAllSelectionStableLow,\SeventhBoundTwoBAllSelectionStableHigh]}}$ & $\SeventhBoundTwoBAllSameScoring_{\text{\fontsize{9}{10}\selectfont [\SeventhBoundTwoBAllSameScoringLow,\SeventhBoundTwoBAllSameScoringHigh]}}$ & $\SeventhBoundTwoBAllSameFresh_{\text{\fontsize{9}{10}\selectfont [\SeventhBoundTwoBAllSameFreshLow,\SeventhBoundTwoBAllSameFreshHigh]}}$ & $\SeventhBoundTwoBAllSameStable_{\text{\fontsize{9}{10}\selectfont [\SeventhBoundTwoBAllSameStableLow,\SeventhBoundTwoBAllSameStableHigh]}}$ \\
    \midrule
    \rowCone & 1 & $\SeventhBoundTwoCOneSelectionScoring_{\text{\fontsize{9}{10}\selectfont [\SeventhBoundTwoCOneSelectionScoringLow,\SeventhBoundTwoCOneSelectionScoringHigh]}}$ & $\SeventhBoundTwoCOneSelectionFresh_{\text{\fontsize{9}{10}\selectfont [\SeventhBoundTwoCOneSelectionFreshLow,\SeventhBoundTwoCOneSelectionFreshHigh]}}$ & $\SeventhBoundTwoCOneSelectionStable_{\text{\fontsize{9}{10}\selectfont [\SeventhBoundTwoCOneSelectionStableLow,\SeventhBoundTwoCOneSelectionStableHigh]}}$ & $\SeventhBoundTwoCOneSameScoring_{\text{\fontsize{9}{10}\selectfont [\SeventhBoundTwoCOneSameScoringLow,\SeventhBoundTwoCOneSameScoringHigh]}}$ & $\SeventhBoundTwoCOneSameFresh_{\text{\fontsize{9}{10}\selectfont [\SeventhBoundTwoCOneSameFreshLow,\SeventhBoundTwoCOneSameFreshHigh]}}$ & $\SeventhBoundTwoCOneSameStable_{\text{\fontsize{9}{10}\selectfont [\SeventhBoundTwoCOneSameStableLow,\SeventhBoundTwoCOneSameStableHigh]}}$ \\
    \rowCtwo & 2 & $\SeventhBoundTwoCTwoSelectionScoring_{\text{\fontsize{9}{10}\selectfont [\SeventhBoundTwoCTwoSelectionScoringLow,\SeventhBoundTwoCTwoSelectionScoringHigh]}}$ & $\SeventhBoundTwoCTwoSelectionFresh_{\text{\fontsize{9}{10}\selectfont [\SeventhBoundTwoCTwoSelectionFreshLow,\SeventhBoundTwoCTwoSelectionFreshHigh]}}$ & $\SeventhBoundTwoCTwoSelectionStable_{\text{\fontsize{9}{10}\selectfont [\SeventhBoundTwoCTwoSelectionStableLow,\SeventhBoundTwoCTwoSelectionStableHigh]}}$ & $\SeventhBoundTwoCTwoSameScoring_{\text{\fontsize{9}{10}\selectfont [\SeventhBoundTwoCTwoSameScoringLow,\SeventhBoundTwoCTwoSameScoringHigh]}}$ & $\SeventhBoundTwoCTwoSameFresh_{\text{\fontsize{9}{10}\selectfont [\SeventhBoundTwoCTwoSameFreshLow,\SeventhBoundTwoCTwoSameFreshHigh]}}$ & $\SeventhBoundTwoCTwoSameStable_{\text{\fontsize{9}{10}\selectfont [\SeventhBoundTwoCTwoSameStableLow,\SeventhBoundTwoCTwoSameStableHigh]}}$ \\
     & 3 & $\SeventhBoundTwoCThreeSelectionScoring_{\text{\fontsize{9}{10}\selectfont [\SeventhBoundTwoCThreeSelectionScoringLow,\SeventhBoundTwoCThreeSelectionScoringHigh]}}$ & $\SeventhBoundTwoCThreeSelectionFresh_{\text{\fontsize{9}{10}\selectfont [\SeventhBoundTwoCThreeSelectionFreshLow,\SeventhBoundTwoCThreeSelectionFreshHigh]}}$ & $\SeventhBoundTwoCThreeSelectionStable_{\text{\fontsize{9}{10}\selectfont [\SeventhBoundTwoCThreeSelectionStableLow,\SeventhBoundTwoCThreeSelectionStableHigh]}}$ & $\SeventhBoundTwoCThreeSameScoring_{\text{\fontsize{9}{10}\selectfont [\SeventhBoundTwoCThreeSameScoringLow,\SeventhBoundTwoCThreeSameScoringHigh]}}$ & $\SeventhBoundTwoCThreeSameFresh_{\text{\fontsize{9}{10}\selectfont [\SeventhBoundTwoCThreeSameFreshLow,\SeventhBoundTwoCThreeSameFreshHigh]}}$ & $\SeventhBoundTwoCThreeSameStable_{\text{\fontsize{9}{10}\selectfont [\SeventhBoundTwoCThreeSameStableLow,\SeventhBoundTwoCThreeSameStableHigh]}}$ \\
     & 4 & $\SeventhBoundTwoCFourSelectionScoring_{\text{\fontsize{9}{10}\selectfont [\SeventhBoundTwoCFourSelectionScoringLow,\SeventhBoundTwoCFourSelectionScoringHigh]}}$ & $\SeventhBoundTwoCFourSelectionFresh_{\text{\fontsize{9}{10}\selectfont [\SeventhBoundTwoCFourSelectionFreshLow,\SeventhBoundTwoCFourSelectionFreshHigh]}}$ & $\SeventhBoundTwoCFourSelectionStable_{\text{\fontsize{9}{10}\selectfont [\SeventhBoundTwoCFourSelectionStableLow,\SeventhBoundTwoCFourSelectionStableHigh]}}$ & $\SeventhBoundTwoCFourSameScoring_{\text{\fontsize{9}{10}\selectfont [\SeventhBoundTwoCFourSameScoringLow,\SeventhBoundTwoCFourSameScoringHigh]}}$ & $\SeventhBoundTwoCFourSameFresh_{\text{\fontsize{9}{10}\selectfont [\SeventhBoundTwoCFourSameFreshLow,\SeventhBoundTwoCFourSameFreshHigh]}}$ & $\SeventhBoundTwoCFourSameStable_{\text{\fontsize{9}{10}\selectfont [\SeventhBoundTwoCFourSameStableLow,\SeventhBoundTwoCFourSameStableHigh]}}$ \\
     & 5 & $\SeventhBoundTwoCFiveSelectionScoring_{\text{\fontsize{9}{10}\selectfont [\SeventhBoundTwoCFiveSelectionScoringLow,\SeventhBoundTwoCFiveSelectionScoringHigh]}}$ & $\SeventhBoundTwoCFiveSelectionFresh_{\text{\fontsize{9}{10}\selectfont [\SeventhBoundTwoCFiveSelectionFreshLow,\SeventhBoundTwoCFiveSelectionFreshHigh]}}$ & $\SeventhBoundTwoCFiveSelectionStable_{\text{\fontsize{9}{10}\selectfont [\SeventhBoundTwoCFiveSelectionStableLow,\SeventhBoundTwoCFiveSelectionStableHigh]}}$ & $\SeventhBoundTwoCFiveSameScoring_{\text{\fontsize{9}{10}\selectfont [\SeventhBoundTwoCFiveSameScoringLow,\SeventhBoundTwoCFiveSameScoringHigh]}}$ & $\SeventhBoundTwoCFiveSameFresh_{\text{\fontsize{9}{10}\selectfont [\SeventhBoundTwoCFiveSameFreshLow,\SeventhBoundTwoCFiveSameFreshHigh]}}$ & $\SeventhBoundTwoCFiveSameStable_{\text{\fontsize{9}{10}\selectfont [\SeventhBoundTwoCFiveSameStableLow,\SeventhBoundTwoCFiveSameStableHigh]}}$ \\
     & all & $\SeventhBoundTwoCAllSelectionScoring_{\text{\fontsize{9}{10}\selectfont [\SeventhBoundTwoCAllSelectionScoringLow,\SeventhBoundTwoCAllSelectionScoringHigh]}}$ & $\SeventhBoundTwoCAllSelectionFresh_{\text{\fontsize{9}{10}\selectfont [\SeventhBoundTwoCAllSelectionFreshLow,\SeventhBoundTwoCAllSelectionFreshHigh]}}$ & $\SeventhBoundTwoCAllSelectionStable_{\text{\fontsize{9}{10}\selectfont [\SeventhBoundTwoCAllSelectionStableLow,\SeventhBoundTwoCAllSelectionStableHigh]}}$ & $\SeventhBoundTwoCAllSameScoring_{\text{\fontsize{9}{10}\selectfont [\SeventhBoundTwoCAllSameScoringLow,\SeventhBoundTwoCAllSameScoringHigh]}}$ & $\SeventhBoundTwoCAllSameFresh_{\text{\fontsize{9}{10}\selectfont [\SeventhBoundTwoCAllSameFreshLow,\SeventhBoundTwoCAllSameFreshHigh]}}$ & $\SeventhBoundTwoCAllSameStable_{\text{\fontsize{9}{10}\selectfont [\SeventhBoundTwoCAllSameStableLow,\SeventhBoundTwoCAllSameStableHigh]}}$ \\
    \midrule
    \bottomrule
  \end{tabular}
\end{table}
\begin{table}[p]
\ContinuedFloat
  \caption{\capSiteBoundsThree}
  \centering
  \fontsize{9}{10.5}\selectfont
  \setlength{\tabcolsep}{0.2pt}
  \begin{tabular}{@{}llcccccc@{}}
    \toprule
 & & \multicolumn{3}{c}{Prediction from selection images} & \multicolumn{3}{c}{Prediction from the same images} \\
 & Seed & Scoring & Fresh & All halves & Scoring & Fresh & All halves \\
    \midrule
    \rowAone & 1 & $\SeventhBoundThreeAOneSelectionScoring_{\text{\fontsize{9}{10}\selectfont [\SeventhBoundThreeAOneSelectionScoringLow,\SeventhBoundThreeAOneSelectionScoringHigh]}}$ & $\SeventhBoundThreeAOneSelectionFresh_{\text{\fontsize{9}{10}\selectfont [\SeventhBoundThreeAOneSelectionFreshLow,\SeventhBoundThreeAOneSelectionFreshHigh]}}$ & $\SeventhBoundThreeAOneSelectionStable_{\text{\fontsize{9}{10}\selectfont [\SeventhBoundThreeAOneSelectionStableLow,\SeventhBoundThreeAOneSelectionStableHigh]}}$ & $\SeventhBoundThreeAOneSameScoring_{\text{\fontsize{9}{10}\selectfont [\SeventhBoundThreeAOneSameScoringLow,\SeventhBoundThreeAOneSameScoringHigh]}}$ & $\SeventhBoundThreeAOneSameFresh_{\text{\fontsize{9}{10}\selectfont [\SeventhBoundThreeAOneSameFreshLow,\SeventhBoundThreeAOneSameFreshHigh]}}$ & $\SeventhBoundThreeAOneSameStable_{\text{\fontsize{9}{10}\selectfont [\SeventhBoundThreeAOneSameStableLow,\SeventhBoundThreeAOneSameStableHigh]}}$ \\
    \rowAtwo & 2 & $\SeventhBoundThreeATwoSelectionScoring_{\text{\fontsize{9}{10}\selectfont [\SeventhBoundThreeATwoSelectionScoringLow,\SeventhBoundThreeATwoSelectionScoringHigh]}}$ & $\SeventhBoundThreeATwoSelectionFresh_{\text{\fontsize{9}{10}\selectfont [\SeventhBoundThreeATwoSelectionFreshLow,\SeventhBoundThreeATwoSelectionFreshHigh]}}$ & $\SeventhBoundThreeATwoSelectionStable_{\text{\fontsize{9}{10}\selectfont [\SeventhBoundThreeATwoSelectionStableLow,\SeventhBoundThreeATwoSelectionStableHigh]}}$ & $\SeventhBoundThreeATwoSameScoring_{\text{\fontsize{9}{10}\selectfont [\SeventhBoundThreeATwoSameScoringLow,\SeventhBoundThreeATwoSameScoringHigh]}}$ & $\SeventhBoundThreeATwoSameFresh_{\text{\fontsize{9}{10}\selectfont [\SeventhBoundThreeATwoSameFreshLow,\SeventhBoundThreeATwoSameFreshHigh]}}$ & $\SeventhBoundThreeATwoSameStable_{\text{\fontsize{9}{10}\selectfont [\SeventhBoundThreeATwoSameStableLow,\SeventhBoundThreeATwoSameStableHigh]}}$ \\
     & 3 & $\SeventhBoundThreeAThreeSelectionScoring_{\text{\fontsize{9}{10}\selectfont [\SeventhBoundThreeAThreeSelectionScoringLow,\SeventhBoundThreeAThreeSelectionScoringHigh]}}$ & $\SeventhBoundThreeAThreeSelectionFresh_{\text{\fontsize{9}{10}\selectfont [\SeventhBoundThreeAThreeSelectionFreshLow,\SeventhBoundThreeAThreeSelectionFreshHigh]}}$ & $\SeventhBoundThreeAThreeSelectionStable_{\text{\fontsize{9}{10}\selectfont [\SeventhBoundThreeAThreeSelectionStableLow,\SeventhBoundThreeAThreeSelectionStableHigh]}}$ & $\SeventhBoundThreeAThreeSameScoring_{\text{\fontsize{9}{10}\selectfont [\SeventhBoundThreeAThreeSameScoringLow,\SeventhBoundThreeAThreeSameScoringHigh]}}$ & $\SeventhBoundThreeAThreeSameFresh_{\text{\fontsize{9}{10}\selectfont [\SeventhBoundThreeAThreeSameFreshLow,\SeventhBoundThreeAThreeSameFreshHigh]}}$ & $\SeventhBoundThreeAThreeSameStable_{\text{\fontsize{9}{10}\selectfont [\SeventhBoundThreeAThreeSameStableLow,\SeventhBoundThreeAThreeSameStableHigh]}}$ \\
     & 4 & $\SeventhBoundThreeAFourSelectionScoring_{\text{\fontsize{9}{10}\selectfont [\SeventhBoundThreeAFourSelectionScoringLow,\SeventhBoundThreeAFourSelectionScoringHigh]}}$ & $\SeventhBoundThreeAFourSelectionFresh_{\text{\fontsize{9}{10}\selectfont [\SeventhBoundThreeAFourSelectionFreshLow,\SeventhBoundThreeAFourSelectionFreshHigh]}}$ & $\SeventhBoundThreeAFourSelectionStable_{\text{\fontsize{9}{10}\selectfont [\SeventhBoundThreeAFourSelectionStableLow,\SeventhBoundThreeAFourSelectionStableHigh]}}$ & $\SeventhBoundThreeAFourSameScoring_{\text{\fontsize{9}{10}\selectfont [\SeventhBoundThreeAFourSameScoringLow,\SeventhBoundThreeAFourSameScoringHigh]}}$ & $\SeventhBoundThreeAFourSameFresh_{\text{\fontsize{9}{10}\selectfont [\SeventhBoundThreeAFourSameFreshLow,\SeventhBoundThreeAFourSameFreshHigh]}}$ & $\SeventhBoundThreeAFourSameStable_{\text{\fontsize{9}{10}\selectfont [\SeventhBoundThreeAFourSameStableLow,\SeventhBoundThreeAFourSameStableHigh]}}$ \\
     & 5 & $\SeventhBoundThreeAFiveSelectionScoring_{\text{\fontsize{9}{10}\selectfont [\SeventhBoundThreeAFiveSelectionScoringLow,\SeventhBoundThreeAFiveSelectionScoringHigh]}}$ & $\SeventhBoundThreeAFiveSelectionFresh_{\text{\fontsize{9}{10}\selectfont [\SeventhBoundThreeAFiveSelectionFreshLow,\SeventhBoundThreeAFiveSelectionFreshHigh]}}$ & $\SeventhBoundThreeAFiveSelectionStable_{\text{\fontsize{9}{10}\selectfont [\SeventhBoundThreeAFiveSelectionStableLow,\SeventhBoundThreeAFiveSelectionStableHigh]}}$ & $\SeventhBoundThreeAFiveSameScoring_{\text{\fontsize{9}{10}\selectfont [\SeventhBoundThreeAFiveSameScoringLow,\SeventhBoundThreeAFiveSameScoringHigh]}}$ & $\SeventhBoundThreeAFiveSameFresh_{\text{\fontsize{9}{10}\selectfont [\SeventhBoundThreeAFiveSameFreshLow,\SeventhBoundThreeAFiveSameFreshHigh]}}$ & $\SeventhBoundThreeAFiveSameStable_{\text{\fontsize{9}{10}\selectfont [\SeventhBoundThreeAFiveSameStableLow,\SeventhBoundThreeAFiveSameStableHigh]}}$ \\
     & all & $\SeventhBoundThreeAAllSelectionScoring_{\text{\fontsize{9}{10}\selectfont [\SeventhBoundThreeAAllSelectionScoringLow,\SeventhBoundThreeAAllSelectionScoringHigh]}}$ & $\SeventhBoundThreeAAllSelectionFresh_{\text{\fontsize{9}{10}\selectfont [\SeventhBoundThreeAAllSelectionFreshLow,\SeventhBoundThreeAAllSelectionFreshHigh]}}$ & $\SeventhBoundThreeAAllSelectionStable_{\text{\fontsize{9}{10}\selectfont [\SeventhBoundThreeAAllSelectionStableLow,\SeventhBoundThreeAAllSelectionStableHigh]}}$ & $\SeventhBoundThreeAAllSameScoring_{\text{\fontsize{9}{10}\selectfont [\SeventhBoundThreeAAllSameScoringLow,\SeventhBoundThreeAAllSameScoringHigh]}}$ & $\SeventhBoundThreeAAllSameFresh_{\text{\fontsize{9}{10}\selectfont [\SeventhBoundThreeAAllSameFreshLow,\SeventhBoundThreeAAllSameFreshHigh]}}$ & $\SeventhBoundThreeAAllSameStable_{\text{\fontsize{9}{10}\selectfont [\SeventhBoundThreeAAllSameStableLow,\SeventhBoundThreeAAllSameStableHigh]}}$ \\
    \midrule
    \rowBone & 1 & $\SeventhBoundThreeBOneSelectionScoring_{\text{\fontsize{9}{10}\selectfont [\SeventhBoundThreeBOneSelectionScoringLow,\SeventhBoundThreeBOneSelectionScoringHigh]}}$ & $\SeventhBoundThreeBOneSelectionFresh_{\text{\fontsize{9}{10}\selectfont [\SeventhBoundThreeBOneSelectionFreshLow,\SeventhBoundThreeBOneSelectionFreshHigh]}}$ & $\SeventhBoundThreeBOneSelectionStable_{\text{\fontsize{9}{10}\selectfont [\SeventhBoundThreeBOneSelectionStableLow,\SeventhBoundThreeBOneSelectionStableHigh]}}$ & $\SeventhBoundThreeBOneSameScoring_{\text{\fontsize{9}{10}\selectfont [\SeventhBoundThreeBOneSameScoringLow,\SeventhBoundThreeBOneSameScoringHigh]}}$ & $\SeventhBoundThreeBOneSameFresh_{\text{\fontsize{9}{10}\selectfont [\SeventhBoundThreeBOneSameFreshLow,\SeventhBoundThreeBOneSameFreshHigh]}}$ & $\SeventhBoundThreeBOneSameStable_{\text{\fontsize{9}{10}\selectfont [\SeventhBoundThreeBOneSameStableLow,\SeventhBoundThreeBOneSameStableHigh]}}$ \\
    \rowBtwo & 2 & $\SeventhBoundThreeBTwoSelectionScoring_{\text{\fontsize{9}{10}\selectfont [\SeventhBoundThreeBTwoSelectionScoringLow,\SeventhBoundThreeBTwoSelectionScoringHigh]}}$ & $\SeventhBoundThreeBTwoSelectionFresh_{\text{\fontsize{9}{10}\selectfont [\SeventhBoundThreeBTwoSelectionFreshLow,\SeventhBoundThreeBTwoSelectionFreshHigh]}}$ & $\SeventhBoundThreeBTwoSelectionStable_{\text{\fontsize{9}{10}\selectfont [\SeventhBoundThreeBTwoSelectionStableLow,\SeventhBoundThreeBTwoSelectionStableHigh]}}$ & $\SeventhBoundThreeBTwoSameScoring_{\text{\fontsize{9}{10}\selectfont [\SeventhBoundThreeBTwoSameScoringLow,\SeventhBoundThreeBTwoSameScoringHigh]}}$ & $\SeventhBoundThreeBTwoSameFresh_{\text{\fontsize{9}{10}\selectfont [\SeventhBoundThreeBTwoSameFreshLow,\SeventhBoundThreeBTwoSameFreshHigh]}}$ & $\SeventhBoundThreeBTwoSameStable_{\text{\fontsize{9}{10}\selectfont [\SeventhBoundThreeBTwoSameStableLow,\SeventhBoundThreeBTwoSameStableHigh]}}$ \\
     & 3 & $\SeventhBoundThreeBThreeSelectionScoring_{\text{\fontsize{9}{10}\selectfont [\SeventhBoundThreeBThreeSelectionScoringLow,\SeventhBoundThreeBThreeSelectionScoringHigh]}}$ & $\SeventhBoundThreeBThreeSelectionFresh_{\text{\fontsize{9}{10}\selectfont [\SeventhBoundThreeBThreeSelectionFreshLow,\SeventhBoundThreeBThreeSelectionFreshHigh]}}$ & $\SeventhBoundThreeBThreeSelectionStable_{\text{\fontsize{9}{10}\selectfont [\SeventhBoundThreeBThreeSelectionStableLow,\SeventhBoundThreeBThreeSelectionStableHigh]}}$ & $\SeventhBoundThreeBThreeSameScoring_{\text{\fontsize{9}{10}\selectfont [\SeventhBoundThreeBThreeSameScoringLow,\SeventhBoundThreeBThreeSameScoringHigh]}}$ & $\SeventhBoundThreeBThreeSameFresh_{\text{\fontsize{9}{10}\selectfont [\SeventhBoundThreeBThreeSameFreshLow,\SeventhBoundThreeBThreeSameFreshHigh]}}$ & $\SeventhBoundThreeBThreeSameStable_{\text{\fontsize{9}{10}\selectfont [\SeventhBoundThreeBThreeSameStableLow,\SeventhBoundThreeBThreeSameStableHigh]}}$ \\
     & 4 & $\SeventhBoundThreeBFourSelectionScoring_{\text{\fontsize{9}{10}\selectfont [\SeventhBoundThreeBFourSelectionScoringLow,\SeventhBoundThreeBFourSelectionScoringHigh]}}$ & $\SeventhBoundThreeBFourSelectionFresh_{\text{\fontsize{9}{10}\selectfont [\SeventhBoundThreeBFourSelectionFreshLow,\SeventhBoundThreeBFourSelectionFreshHigh]}}$ & $\SeventhBoundThreeBFourSelectionStable_{\text{\fontsize{9}{10}\selectfont [\SeventhBoundThreeBFourSelectionStableLow,\SeventhBoundThreeBFourSelectionStableHigh]}}$ & $\SeventhBoundThreeBFourSameScoring_{\text{\fontsize{9}{10}\selectfont [\SeventhBoundThreeBFourSameScoringLow,\SeventhBoundThreeBFourSameScoringHigh]}}$ & $\SeventhBoundThreeBFourSameFresh_{\text{\fontsize{9}{10}\selectfont [\SeventhBoundThreeBFourSameFreshLow,\SeventhBoundThreeBFourSameFreshHigh]}}$ & $\SeventhBoundThreeBFourSameStable_{\text{\fontsize{9}{10}\selectfont [\SeventhBoundThreeBFourSameStableLow,\SeventhBoundThreeBFourSameStableHigh]}}$ \\
     & 5 & $\SeventhBoundThreeBFiveSelectionScoring_{\text{\fontsize{9}{10}\selectfont [\SeventhBoundThreeBFiveSelectionScoringLow,\SeventhBoundThreeBFiveSelectionScoringHigh]}}$ & $\SeventhBoundThreeBFiveSelectionFresh_{\text{\fontsize{9}{10}\selectfont [\SeventhBoundThreeBFiveSelectionFreshLow,\SeventhBoundThreeBFiveSelectionFreshHigh]}}$ & $\SeventhBoundThreeBFiveSelectionStable_{\text{\fontsize{9}{10}\selectfont [\SeventhBoundThreeBFiveSelectionStableLow,\SeventhBoundThreeBFiveSelectionStableHigh]}}$ & $\SeventhBoundThreeBFiveSameScoring_{\text{\fontsize{9}{10}\selectfont [\SeventhBoundThreeBFiveSameScoringLow,\SeventhBoundThreeBFiveSameScoringHigh]}}$ & $\SeventhBoundThreeBFiveSameFresh_{\text{\fontsize{9}{10}\selectfont [\SeventhBoundThreeBFiveSameFreshLow,\SeventhBoundThreeBFiveSameFreshHigh]}}$ & $\SeventhBoundThreeBFiveSameStable_{\text{\fontsize{9}{10}\selectfont [\SeventhBoundThreeBFiveSameStableLow,\SeventhBoundThreeBFiveSameStableHigh]}}$ \\
     & all & $\SeventhBoundThreeBAllSelectionScoring_{\text{\fontsize{9}{10}\selectfont [\SeventhBoundThreeBAllSelectionScoringLow,\SeventhBoundThreeBAllSelectionScoringHigh]}}$ & $\SeventhBoundThreeBAllSelectionFresh_{\text{\fontsize{9}{10}\selectfont [\SeventhBoundThreeBAllSelectionFreshLow,\SeventhBoundThreeBAllSelectionFreshHigh]}}$ & $\SeventhBoundThreeBAllSelectionStable_{\text{\fontsize{9}{10}\selectfont [\SeventhBoundThreeBAllSelectionStableLow,\SeventhBoundThreeBAllSelectionStableHigh]}}$ & $\SeventhBoundThreeBAllSameScoring_{\text{\fontsize{9}{10}\selectfont [\SeventhBoundThreeBAllSameScoringLow,\SeventhBoundThreeBAllSameScoringHigh]}}$ & $\SeventhBoundThreeBAllSameFresh_{\text{\fontsize{9}{10}\selectfont [\SeventhBoundThreeBAllSameFreshLow,\SeventhBoundThreeBAllSameFreshHigh]}}$ & $\SeventhBoundThreeBAllSameStable_{\text{\fontsize{9}{10}\selectfont [\SeventhBoundThreeBAllSameStableLow,\SeventhBoundThreeBAllSameStableHigh]}}$ \\
    \midrule
    \rowCone & 1 & $\SeventhBoundThreeCOneSelectionScoring_{\text{\fontsize{9}{10}\selectfont [\SeventhBoundThreeCOneSelectionScoringLow,\SeventhBoundThreeCOneSelectionScoringHigh]}}$ & $\SeventhBoundThreeCOneSelectionFresh_{\text{\fontsize{9}{10}\selectfont [\SeventhBoundThreeCOneSelectionFreshLow,\SeventhBoundThreeCOneSelectionFreshHigh]}}$ & $\SeventhBoundThreeCOneSelectionStable_{\text{\fontsize{9}{10}\selectfont [\SeventhBoundThreeCOneSelectionStableLow,\SeventhBoundThreeCOneSelectionStableHigh]}}$ & $\SeventhBoundThreeCOneSameScoring_{\text{\fontsize{9}{10}\selectfont [\SeventhBoundThreeCOneSameScoringLow,\SeventhBoundThreeCOneSameScoringHigh]}}$ & $\SeventhBoundThreeCOneSameFresh_{\text{\fontsize{9}{10}\selectfont [\SeventhBoundThreeCOneSameFreshLow,\SeventhBoundThreeCOneSameFreshHigh]}}$ & $\SeventhBoundThreeCOneSameStable_{\text{\fontsize{9}{10}\selectfont [\SeventhBoundThreeCOneSameStableLow,\SeventhBoundThreeCOneSameStableHigh]}}$ \\
    \rowCtwo & 2 & $\SeventhBoundThreeCTwoSelectionScoring_{\text{\fontsize{9}{10}\selectfont [\SeventhBoundThreeCTwoSelectionScoringLow,\SeventhBoundThreeCTwoSelectionScoringHigh]}}$ & $\SeventhBoundThreeCTwoSelectionFresh_{\text{\fontsize{9}{10}\selectfont [\SeventhBoundThreeCTwoSelectionFreshLow,\SeventhBoundThreeCTwoSelectionFreshHigh]}}$ & $\SeventhBoundThreeCTwoSelectionStable_{\text{\fontsize{9}{10}\selectfont [\SeventhBoundThreeCTwoSelectionStableLow,\SeventhBoundThreeCTwoSelectionStableHigh]}}$ & $\SeventhBoundThreeCTwoSameScoring_{\text{\fontsize{9}{10}\selectfont [\SeventhBoundThreeCTwoSameScoringLow,\SeventhBoundThreeCTwoSameScoringHigh]}}$ & $\SeventhBoundThreeCTwoSameFresh_{\text{\fontsize{9}{10}\selectfont [\SeventhBoundThreeCTwoSameFreshLow,\SeventhBoundThreeCTwoSameFreshHigh]}}$ & $\SeventhBoundThreeCTwoSameStable_{\text{\fontsize{9}{10}\selectfont [\SeventhBoundThreeCTwoSameStableLow,\SeventhBoundThreeCTwoSameStableHigh]}}$ \\
     & 3 & $\SeventhBoundThreeCThreeSelectionScoring_{\text{\fontsize{9}{10}\selectfont [\SeventhBoundThreeCThreeSelectionScoringLow,\SeventhBoundThreeCThreeSelectionScoringHigh]}}$ & $\SeventhBoundThreeCThreeSelectionFresh_{\text{\fontsize{9}{10}\selectfont [\SeventhBoundThreeCThreeSelectionFreshLow,\SeventhBoundThreeCThreeSelectionFreshHigh]}}$ & $\SeventhBoundThreeCThreeSelectionStable_{\text{\fontsize{9}{10}\selectfont [\SeventhBoundThreeCThreeSelectionStableLow,\SeventhBoundThreeCThreeSelectionStableHigh]}}$ & $\SeventhBoundThreeCThreeSameScoring_{\text{\fontsize{9}{10}\selectfont [\SeventhBoundThreeCThreeSameScoringLow,\SeventhBoundThreeCThreeSameScoringHigh]}}$ & $\SeventhBoundThreeCThreeSameFresh_{\text{\fontsize{9}{10}\selectfont [\SeventhBoundThreeCThreeSameFreshLow,\SeventhBoundThreeCThreeSameFreshHigh]}}$ & $\SeventhBoundThreeCThreeSameStable_{\text{\fontsize{9}{10}\selectfont [\SeventhBoundThreeCThreeSameStableLow,\SeventhBoundThreeCThreeSameStableHigh]}}$ \\
     & 4 & $\SeventhBoundThreeCFourSelectionScoring_{\text{\fontsize{9}{10}\selectfont [\SeventhBoundThreeCFourSelectionScoringLow,\SeventhBoundThreeCFourSelectionScoringHigh]}}$ & $\SeventhBoundThreeCFourSelectionFresh_{\text{\fontsize{9}{10}\selectfont [\SeventhBoundThreeCFourSelectionFreshLow,\SeventhBoundThreeCFourSelectionFreshHigh]}}$ & $\SeventhBoundThreeCFourSelectionStable_{\text{\fontsize{9}{10}\selectfont [\SeventhBoundThreeCFourSelectionStableLow,\SeventhBoundThreeCFourSelectionStableHigh]}}$ & $\SeventhBoundThreeCFourSameScoring_{\text{\fontsize{9}{10}\selectfont [\SeventhBoundThreeCFourSameScoringLow,\SeventhBoundThreeCFourSameScoringHigh]}}$ & $\SeventhBoundThreeCFourSameFresh_{\text{\fontsize{9}{10}\selectfont [\SeventhBoundThreeCFourSameFreshLow,\SeventhBoundThreeCFourSameFreshHigh]}}$ & $\SeventhBoundThreeCFourSameStable_{\text{\fontsize{9}{10}\selectfont [\SeventhBoundThreeCFourSameStableLow,\SeventhBoundThreeCFourSameStableHigh]}}$ \\
     & 5 & $\SeventhBoundThreeCFiveSelectionScoring_{\text{\fontsize{9}{10}\selectfont [\SeventhBoundThreeCFiveSelectionScoringLow,\SeventhBoundThreeCFiveSelectionScoringHigh]}}$ & $\SeventhBoundThreeCFiveSelectionFresh_{\text{\fontsize{9}{10}\selectfont [\SeventhBoundThreeCFiveSelectionFreshLow,\SeventhBoundThreeCFiveSelectionFreshHigh]}}$ & $\SeventhBoundThreeCFiveSelectionStable_{\text{\fontsize{9}{10}\selectfont [\SeventhBoundThreeCFiveSelectionStableLow,\SeventhBoundThreeCFiveSelectionStableHigh]}}$ & $\SeventhBoundThreeCFiveSameScoring_{\text{\fontsize{9}{10}\selectfont [\SeventhBoundThreeCFiveSameScoringLow,\SeventhBoundThreeCFiveSameScoringHigh]}}$ & $\SeventhBoundThreeCFiveSameFresh_{\text{\fontsize{9}{10}\selectfont [\SeventhBoundThreeCFiveSameFreshLow,\SeventhBoundThreeCFiveSameFreshHigh]}}$ & $\SeventhBoundThreeCFiveSameStable_{\text{\fontsize{9}{10}\selectfont [\SeventhBoundThreeCFiveSameStableLow,\SeventhBoundThreeCFiveSameStableHigh]}}$ \\
     & all & $\SeventhBoundThreeCAllSelectionScoring_{\text{\fontsize{9}{10}\selectfont [\SeventhBoundThreeCAllSelectionScoringLow,\SeventhBoundThreeCAllSelectionScoringHigh]}}$ & $\SeventhBoundThreeCAllSelectionFresh_{\text{\fontsize{9}{10}\selectfont [\SeventhBoundThreeCAllSelectionFreshLow,\SeventhBoundThreeCAllSelectionFreshHigh]}}$ & $\SeventhBoundThreeCAllSelectionStable_{\text{\fontsize{9}{10}\selectfont [\SeventhBoundThreeCAllSelectionStableLow,\SeventhBoundThreeCAllSelectionStableHigh]}}$ & $\SeventhBoundThreeCAllSameScoring_{\text{\fontsize{9}{10}\selectfont [\SeventhBoundThreeCAllSameScoringLow,\SeventhBoundThreeCAllSameScoringHigh]}}$ & $\SeventhBoundThreeCAllSameFresh_{\text{\fontsize{9}{10}\selectfont [\SeventhBoundThreeCAllSameFreshLow,\SeventhBoundThreeCAllSameFreshHigh]}}$ & $\SeventhBoundThreeCAllSameStable_{\text{\fontsize{9}{10}\selectfont [\SeventhBoundThreeCAllSameStableLow,\SeventhBoundThreeCAllSameStableHigh]}}$ \\
    \midrule
    \bottomrule
  \end{tabular}
\end{table}

\paragraph{The pair whose incorrect prediction persists.}
The pair's layer errors stay below both thresholds on every image set, as \cref{tab:case-verdict} shows\later, so it stays an incorrect prediction of the layer rule at the fitted threshold. Its first feature takes no category name under the rule of \cref{sec:supp-recipe}, and before the final normalization its ratio already lies above \Certify, as \cref{sec:cosine} gives. On the fresh images the two ablations' effects differ by \CaseFreshError of the first effect's squared size, with \CaseFreshPatch of that difference at patch tokens and \CaseFreshTopShare in the ten images of largest squared difference. Dropping the one image of largest first effect leaves output errors of \CaseScoreRemoved, \CaseFreshRemoved and \CaseThirdRemoved on the scoring, fresh and third image sets, each above \Certify, and dropping the image of largest squared difference gives the same result on each set, as \cref{tab:case-removals} gives. The ten largest images account for \CaseScoreTopShare, \CaseFreshTopShare and \CaseThirdTopShare of their sets' squared difference and \CaseScoreTopEnergy, \CaseFreshTopEnergy and \CaseThirdTopEnergy of the first effect's squared size, and \cref{tab:case-images} lists them.

\begin{table}[!htb]
  \caption{\capCaseVerdict}
  \label{tab:case-verdict}
  \centering\fontsize{9}{10.5}\selectfont
  \setlength{\tabcolsep}{4pt}
  \begin{tabular}{@{}lccc@{}}
  \toprule
Images and statistic & Value & At \Certify & At \NinthFitJoint \\
  \midrule
Selection, layer error & \EighthCellQU & Pass & Pass \\
Scoring, layer error & \EighthCellQV & Pass & Pass \\
Fresh, layer error & \EighthCellQW & Pass & Pass \\
Third, layer error & \EighthCellQX & Pass & Pass \\
Scoring, output error & \CaseScoreError & Fail & Fail \\
Fresh, output error & \CaseFreshError & Fail & Fail \\
Third, output error & \CaseThirdError & Fail & Fail \\
  \bottomrule
  \end{tabular}
  \tablenote{\noteCaseVerdict}
\end{table}

\begin{table}[!htb]
  \caption{\capCaseRemovals}
  \label{tab:case-removals}
  \centering\fontsize{9}{10.5}\selectfont
  \setlength{\tabcolsep}{4pt}
  \begin{tabular}{@{}lcccccc@{}}
  \toprule
Image set & Images & \shortstack{Output\\error} & \shortstack{Top ten\\share of squared\\difference} & \shortstack{Top ten\\share of squared\\first effect} & \shortstack{Output error\\without largest\\first effect} & \shortstack{Output error\\without largest\\difference} \\
  \midrule
Scoring & \EighthCellQY & \EighthCellQZ & \EighthCellRA & \EighthCellRB & \EighthCellRC & \EighthCellRD \\
Fresh & \EighthCellRE & \EighthCellRF & \EighthCellRG & \EighthCellRH & \EighthCellRI & \EighthCellRJ \\
Third & \EighthCellRK & \EighthCellRL & \EighthCellRM & \EighthCellRN & \EighthCellRO & \EighthCellRP \\
  \bottomrule
  \end{tabular}
  \tablenote{\noteCaseRemovals}
\end{table}

\begin{table}[!htb]
  \caption{\capCaseImages}
  \label{tab:case-images}
  \centering\fontsize{9}{10.5}\selectfont
  \setlength{\tabcolsep}{4pt}
  \begin{tabular}{@{}llrrrr@{}}
  \toprule
Image set & \shortstack{COCO image\\number} & \shortstack{Squared\\difference} & \shortstack{Squared\\first effect} & \shortstack{Share of squared\\difference} & \shortstack{Share of squared\\first effect} \\
  \midrule
Scoring & \texttt{\TenthImageRQ} & \EighthCellRR & \EighthCellRS & \EighthCellRT & \EighthCellRU \\
Scoring & \texttt{\TenthImageRV} & \EighthCellRW & \EighthCellRX & \EighthCellRY & \EighthCellRZ \\
Scoring & \texttt{\TenthImageSA} & \EighthCellSB & \EighthCellSC & \EighthCellSD & \EighthCellSE \\
Scoring & \texttt{\TenthImageSF} & \EighthCellSG & \EighthCellSH & \EighthCellSI & \EighthCellSJ \\
Scoring & \texttt{\TenthImageSK} & \EighthCellSL & \EighthCellSM & \EighthCellSN & \EighthCellSO \\
Scoring & \texttt{\TenthImageSP} & \EighthCellSQ & \EighthCellSR & \EighthCellSS & \EighthCellST \\
Scoring & \texttt{\TenthImageSU} & \EighthCellSV & \EighthCellSW & \EighthCellSX & \EighthCellSY \\
Scoring & \texttt{\TenthImageSZ} & \EighthCellTA & \EighthCellTB & \EighthCellTC & \EighthCellTD \\
Scoring & \texttt{\TenthImageTE} & \EighthCellTF & \EighthCellTG & \EighthCellTH & \EighthCellTI \\
Scoring & \texttt{\TenthImageTJ} & \EighthCellTK & \EighthCellTL & \EighthCellTM & \EighthCellTN \\
Fresh & \texttt{\TenthImageTO} & \EighthCellTP & \EighthCellTQ & \EighthCellTR & \EighthCellTS \\
Fresh & \texttt{\TenthImageTT} & \EighthCellTU & \EighthCellTV & \EighthCellTW & \EighthCellTX \\
Fresh & \texttt{\TenthImageTY} & \EighthCellTZ & \EighthCellUA & \EighthCellUB & \EighthCellUC \\
Fresh & \texttt{\TenthImageUD} & \EighthCellUE & \EighthCellUF & \EighthCellUG & \EighthCellUH \\
Fresh & \texttt{\TenthImageUI} & \EighthCellUJ & \EighthCellUK & \EighthCellUL & \EighthCellUM \\
Fresh & \texttt{\TenthImageUN} & \EighthCellUO & \EighthCellUP & \EighthCellUQ & \EighthCellUR \\
Fresh & \texttt{\TenthImageUS} & \EighthCellUT & \EighthCellUU & \EighthCellUV & \EighthCellUW \\
Fresh & \texttt{\TenthImageUX} & \EighthCellUY & \EighthCellUZ & \EighthCellVA & \EighthCellVB \\
Fresh & \texttt{\TenthImageVC} & \EighthCellVD & \EighthCellVE & \EighthCellVF & \EighthCellVG \\
Fresh & \texttt{\TenthImageVH} & \EighthCellVI & \EighthCellVJ & \EighthCellVK & \EighthCellVL \\
Third & \texttt{\TenthImageVM} & \EighthCellVN & \EighthCellVO & \EighthCellVP & \EighthCellVQ \\
Third & \texttt{\TenthImageVR} & \EighthCellVS & \EighthCellVT & \EighthCellVU & \EighthCellVV \\
Third & \texttt{\TenthImageVW} & \EighthCellVX & \EighthCellVY & \EighthCellVZ & \EighthCellWA \\
Third & \texttt{\TenthImageWB} & \EighthCellWC & \EighthCellWD & \EighthCellWE & \EighthCellWF \\
Third & \texttt{\TenthImageWG} & \EighthCellWH & \EighthCellWI & \EighthCellWJ & \EighthCellWK \\
Third & \texttt{\TenthImageWL} & \EighthCellWM & \EighthCellWN & \EighthCellWO & \EighthCellWP \\
Third & \texttt{\TenthImageWQ} & \EighthCellWR & \EighthCellWS & \EighthCellWT & \EighthCellWU \\
Third & \texttt{\TenthImageWV} & \EighthCellWW & \EighthCellWX & \EighthCellWY & \EighthCellWZ \\
Third & \texttt{\TenthImageXA} & \EighthCellXB & \EighthCellXC & \EighthCellXD & \EighthCellXE \\
Third & \texttt{\TenthImageXF} & \EighthCellXG & \EighthCellXH & \EighthCellXI & \EighthCellXJ \\
  \bottomrule
  \end{tabular}
  \tablenote{\noteCaseImages}
\end{table}

Almost all of the squared difference and of the first effect's squared size lie at patch tokens, as \cref{tab:case-tokens} shows by splitting both between the class token and the patch tokens. \Cref{tab:case-top-tokens} lists the ten largest tokens on each set, and \cref{tab:case-grid} groups the patch tokens into sixteen equal regions, across which the difference is spread. Before it computes anything by token, the token computation reproduces the stored activations of the fresh images at the edited layer bit for bit and the stored effects on the scoring images within the recomputation's tolerance\later. It keeps every output token and channel.

\begin{table}[!htb]
  \caption{\capCaseTokens}
  \label{tab:case-tokens}
  \centering\fontsize{9}{10.5}\selectfont
  \setlength{\tabcolsep}{4pt}
  \begin{tabular}{@{}lrrrr@{}}
  \toprule
Image set & \shortstack{Class token share of\\squared difference} & \shortstack{Class token share of\\squared first effect} & \shortstack{Patch token share of\\squared difference} & \shortstack{Patch token share of\\squared first effect} \\
  \midrule
Scoring & \EighthCellXK & \EighthCellXL & \EighthCellXM & \EighthCellXN \\
Fresh & \EighthCellXO & \EighthCellXP & \EighthCellXQ & \EighthCellXR \\
Third & \EighthCellXS & \EighthCellXT & \EighthCellXU & \EighthCellXV \\
  \bottomrule
  \end{tabular}
  \tablenote{\noteCaseTokens}
\end{table}

\begin{table}[!htb]
  \caption{\capCaseTopTokens}
  \label{tab:case-top-tokens}
  \centering\fontsize{9}{10.5}\selectfont
  \setlength{\tabcolsep}{4pt}
  \begin{tabular}{@{}llrrrrrr@{}}
  \toprule
Image set & \shortstack{COCO image\\number} & \shortstack{Patch\\row} & \shortstack{Patch\\column} & \shortstack{Squared\\difference} & \shortstack{Squared\\first effect} & \shortstack{Share of squared\\difference} & \shortstack{Share of squared\\first effect} \\
  \midrule
Scoring & \texttt{\TenthImageXW} & \EighthCellXX & \EighthCellXY & \EighthCellXZ & \EighthCellYA & \EighthCellYB & \EighthCellYC \\
Scoring & \texttt{\TenthImageYD} & \EighthCellYE & \EighthCellYF & \EighthCellYG & \EighthCellYH & \EighthCellYI & \EighthCellYJ \\
Scoring & \texttt{\TenthImageYK} & \EighthCellYL & \EighthCellYM & \EighthCellYN & \EighthCellYO & \EighthCellYP & \EighthCellYQ \\
Scoring & \texttt{\TenthImageYR} & \EighthCellYS & \EighthCellYT & \EighthCellYU & \EighthCellYV & \EighthCellYW & \EighthCellYX \\
Scoring & \texttt{\TenthImageYY} & \EighthCellYZ & \EighthCellZA & \EighthCellZB & \EighthCellZC & \EighthCellZD & \EighthCellZE \\
Scoring & \texttt{\TenthImageZF} & \EighthCellZG & \EighthCellZH & \EighthCellZI & \EighthCellZJ & \EighthCellZK & \EighthCellZL \\
Scoring & \texttt{\TenthImageZM} & \EighthCellZN & \EighthCellZO & \EighthCellZP & \EighthCellZQ & \EighthCellZR & \EighthCellZS \\
Scoring & \texttt{\TenthImageZT} & \EighthCellZU & \EighthCellZV & \EighthCellZW & \EighthCellZX & \EighthCellZY & \EighthCellZZ \\
Scoring & \texttt{\TenthImageAAA} & \EighthCellAAB & \EighthCellAAC & \EighthCellAAD & \EighthCellAAE & \EighthCellAAF & \EighthCellAAG \\
Scoring & \texttt{\TenthImageAAH} & \EighthCellAAI & \EighthCellAAJ & \EighthCellAAK & \EighthCellAAL & \EighthCellAAM & \EighthCellAAN \\
Fresh & \texttt{\TenthImageAAO} & \EighthCellAAP & \EighthCellAAQ & \EighthCellAAR & \EighthCellAAS & \EighthCellAAT & \EighthCellAAU \\
Fresh & \texttt{\TenthImageAAV} & \EighthCellAAW & \EighthCellAAX & \EighthCellAAY & \EighthCellAAZ & \EighthCellABA & \EighthCellABB \\
Fresh & \texttt{\TenthImageABC} & \EighthCellABD & \EighthCellABE & \EighthCellABF & \EighthCellABG & \EighthCellABH & \EighthCellABI \\
Fresh & \texttt{\TenthImageABJ} & \EighthCellABK & \EighthCellABL & \EighthCellABM & \EighthCellABN & \EighthCellABO & \EighthCellABP \\
Fresh & \texttt{\TenthImageABQ} & \EighthCellABR & \EighthCellABS & \EighthCellABT & \EighthCellABU & \EighthCellABV & \EighthCellABW \\
Fresh & \texttt{\TenthImageABX} & \EighthCellABY & \EighthCellABZ & \EighthCellACA & \EighthCellACB & \EighthCellACC & \EighthCellACD \\
Fresh & \texttt{\TenthImageACE} & \EighthCellACF & \EighthCellACG & \EighthCellACH & \EighthCellACI & \EighthCellACJ & \EighthCellACK \\
Fresh & \texttt{\TenthImageACL} & \EighthCellACM & \EighthCellACN & \EighthCellACO & \EighthCellACP & \EighthCellACQ & \EighthCellACR \\
Fresh & \texttt{\TenthImageACS} & \EighthCellACT & \EighthCellACU & \EighthCellACV & \EighthCellACW & \EighthCellACX & \EighthCellACY \\
Fresh & \texttt{\TenthImageACZ} & \EighthCellADA & \EighthCellADB & \EighthCellADC & \EighthCellADD & \EighthCellADE & \EighthCellADF \\
Third & \texttt{\TenthImageADG} & \EighthCellADH & \EighthCellADI & \EighthCellADJ & \EighthCellADK & \EighthCellADL & \EighthCellADM \\
Third & \texttt{\TenthImageADN} & \EighthCellADO & \EighthCellADP & \EighthCellADQ & \EighthCellADR & \EighthCellADS & \EighthCellADT \\
Third & \texttt{\TenthImageADU} & \EighthCellADV & \EighthCellADW & \EighthCellADX & \EighthCellADY & \EighthCellADZ & \EighthCellAEA \\
Third & \texttt{\TenthImageAEB} & \EighthCellAEC & \EighthCellAED & \EighthCellAEE & \EighthCellAEF & \EighthCellAEG & \EighthCellAEH \\
Third & \texttt{\TenthImageAEI} & \EighthCellAEJ & \EighthCellAEK & \EighthCellAEL & \EighthCellAEM & \EighthCellAEN & \EighthCellAEO \\
Third & \texttt{\TenthImageAEP} & \EighthCellAEQ & \EighthCellAER & \EighthCellAES & \EighthCellAET & \EighthCellAEU & \EighthCellAEV \\
Third & \texttt{\TenthImageAEW} & \EighthCellAEX & \EighthCellAEY & \EighthCellAEZ & \EighthCellAFA & \EighthCellAFB & \EighthCellAFC \\
Third & \texttt{\TenthImageAFD} & \EighthCellAFE & \EighthCellAFF & \EighthCellAFG & \EighthCellAFH & \EighthCellAFI & \EighthCellAFJ \\
Third & \texttt{\TenthImageAFK} & \EighthCellAFL & \EighthCellAFM & \EighthCellAFN & \EighthCellAFO & \EighthCellAFP & \EighthCellAFQ \\
Third & \texttt{\TenthImageAFR} & \EighthCellAFS & \EighthCellAFT & \EighthCellAFU & \EighthCellAFV & \EighthCellAFW & \EighthCellAFX \\
  \bottomrule
  \end{tabular}
  \tablenote{\noteCaseTopTokens}
\end{table}

\begin{table}[!htb]
  \caption{\capCaseGrid}
  \label{tab:case-grid}
  \centering\fontsize{9}{10.5}\selectfont
  \setlength{\tabcolsep}{4pt}
  \begin{tabular}{@{}rrrrrrrr@{}}
  \toprule
Region row & Region column & \multicolumn{2}{c}{Shares on scoring} & \multicolumn{2}{c}{Shares on fresh} & \multicolumn{2}{c}{Shares on third} \\  & & \shortstack{Squared\\difference} & \shortstack{Squared\\first effect} & \shortstack{Squared\\difference} & \shortstack{Squared\\first effect} & \shortstack{Squared\\difference} & \shortstack{Squared\\first effect} \\
  \midrule
\EighthCellAFY & \EighthCellAFZ & \EighthCellAGA & \EighthCellAGB & \EighthCellAGC & \EighthCellAGD & \EighthCellAGE & \EighthCellAGF \\
\EighthCellAGG & \EighthCellAGH & \EighthCellAGI & \EighthCellAGJ & \EighthCellAGK & \EighthCellAGL & \EighthCellAGM & \EighthCellAGN \\
\EighthCellAGO & \EighthCellAGP & \EighthCellAGQ & \EighthCellAGR & \EighthCellAGS & \EighthCellAGT & \EighthCellAGU & \EighthCellAGV \\
\EighthCellAGW & \EighthCellAGX & \EighthCellAGY & \EighthCellAGZ & \EighthCellAHA & \EighthCellAHB & \EighthCellAHC & \EighthCellAHD \\
\EighthCellAHE & \EighthCellAHF & \EighthCellAHG & \EighthCellAHH & \EighthCellAHI & \EighthCellAHJ & \EighthCellAHK & \EighthCellAHL \\
\EighthCellAHM & \EighthCellAHN & \EighthCellAHO & \EighthCellAHP & \EighthCellAHQ & \EighthCellAHR & \EighthCellAHS & \EighthCellAHT \\
\EighthCellAHU & \EighthCellAHV & \EighthCellAHW & \EighthCellAHX & \EighthCellAHY & \EighthCellAHZ & \EighthCellAIA & \EighthCellAIB \\
\EighthCellAIC & \EighthCellAID & \EighthCellAIE & \EighthCellAIF & \EighthCellAIG & \EighthCellAIH & \EighthCellAII & \EighthCellAIJ \\
\EighthCellAIK & \EighthCellAIL & \EighthCellAIM & \EighthCellAIN & \EighthCellAIO & \EighthCellAIP & \EighthCellAIQ & \EighthCellAIR \\
\EighthCellAIS & \EighthCellAIT & \EighthCellAIU & \EighthCellAIV & \EighthCellAIW & \EighthCellAIX & \EighthCellAIY & \EighthCellAIZ \\
\EighthCellAJA & \EighthCellAJB & \EighthCellAJC & \EighthCellAJD & \EighthCellAJE & \EighthCellAJF & \EighthCellAJG & \EighthCellAJH \\
\EighthCellAJI & \EighthCellAJJ & \EighthCellAJK & \EighthCellAJL & \EighthCellAJM & \EighthCellAJN & \EighthCellAJO & \EighthCellAJP \\
\EighthCellAJQ & \EighthCellAJR & \EighthCellAJS & \EighthCellAJT & \EighthCellAJU & \EighthCellAJV & \EighthCellAJW & \EighthCellAJX \\
\EighthCellAJY & \EighthCellAJZ & \EighthCellAKA & \EighthCellAKB & \EighthCellAKC & \EighthCellAKD & \EighthCellAKE & \EighthCellAKF \\
\EighthCellAKG & \EighthCellAKH & \EighthCellAKI & \EighthCellAKJ & \EighthCellAKK & \EighthCellAKL & \EighthCellAKM & \EighthCellAKN \\
\EighthCellAKO & \EighthCellAKP & \EighthCellAKQ & \EighthCellAKR & \EighthCellAKS & \EighthCellAKT & \EighthCellAKU & \EighthCellAKV \\
  \bottomrule
  \end{tabular}
  \tablenote{\noteCaseGrid}
\end{table}

\end{document}